\documentclass[10pt]{article}

\usepackage{epsfig}
\usepackage{amsmath}
\usepackage{amssymb}
\usepackage{color}
\usepackage{url}

\usepackage{verbatim} 
\usepackage{epstopdf} 
\usepackage{graphicx} 
\usepackage[font=small]{caption} 

\usepackage{afterpage}

\usepackage{caption}
\usepackage{subcaption}
\usepackage{multicol}

\usepackage{multirow}

\usepackage{setspace}

\usepackage{afterpage}

\usepackage{algorithm}
\usepackage{algorithmic}
\usepackage{mathrsfs}

\usepackage{slashbox}

\usepackage{pifont}

\newcommand{\commentout}[1]{}

\newcommand{\xmark}{\ding{55}}
\newcommand{\checkm}{\pmb\checkmark}

\renewcommand{\arraystretch}{1.2}

\begin{document}

\title {\Large \textbf{Intelligent Sampling for Surrogate Modeling, Hyperparameter Optimization, and Data Analysis}}

\author{%
Chandrika Kamath\\[0.5em]
{\small\begin{minipage}{\linewidth}\begin{center}
\begin{tabular}{c}
Lawrence Livermore National Laboratory \\
7000 East Avenue, Livermore, CA 94551, USA\\
\url{kamath2@llnl.gov}\\
\hspace*{0.8in}
\end{tabular}
\end{center}\end{minipage}}
}

\date{December 6, 2021}
\maketitle

\thispagestyle{empty}

\begin{abstract}
 
  Sampling techniques are used in many fields, including design of
  experiments, image processing, and graphics. The techniques in each
  field are designed to meet the constraints specific to that field
  such as uniform coverage of the range of each dimension or random
  samples that are at least a certain distance apart from each other.
  When an application imposes new constraints, for example, by
  requiring samples in a non-rectangular domain or the addition of new
  samples to an existing set, a common solution is to modify the
  algorithm currently in use, often with less than satisfactory
  results.  As an alternative, we propose the concept of intelligent
  sampling, where we devise algorithms specifically tailored to meet
  our sampling needs, either by creating new algorithms or by
  modifying suitable algorithms from other fields.  Surprisingly, both
  qualitative and quantitative comparisons indicate that some
  relatively simple algorithms can be easily modified to meet the many
  sampling requirements of surrogate modeling, hyperparameter
  optimization, and data analysis; these algorithms outperform their
  more sophisticated counterparts currently in use, resulting in
  better use of time and computer resources.

\end{abstract}

\clearpage

\begin{spacing}{0.7}
\tableofcontents
\end{spacing}

\clearpage

%
\section{Introduction}
\label{sec:intro}
%

Sampling techniques play an important role in various aspects of data
analysis, whether it is to generate sample points at which to collect
data, or identify samples from a larger data set as a representative
subset for analysis, or add new samples to increase the sampling
density in a sub-region of the data. While many diverse fields have
independently devised sampling algorithms that meet various criteria
specific to their fields, it is often not straightforward to apply
these algorithms to new tasks or to existing tasks when new
constraints are imposed.  For example, an algorithm that generates an
initial set of samples that are well separated in a rectangular
domain, may not be suitable for generating well-separated samples in a
non-rectangular domain. Or, an algorithm to extract a
uniformly-distributed subset of samples in an existing data set may
not scale to massive data sets where we are allowed only one pass
through the data.

A typical solution to this mis-match between a sampling algorithm and
any newly-imposed constraints on the analysis task is to modify the
existing algorithm, even if the results are less than satisfactory.
An alternative would be to consider or combine ideas developed for
sampling in other disciplines, or for other tasks, as such algorithms
may be easier to modify for our needs without any associated
drawbacks. We refer to this concept as ``{\it intelligent sampling}''
to indicate that we select sampling algorithms that best meet our
needs with the goal of making better use of our computational
resources and reducing the time required for generating and analyzing
the data.

To motivate our work, we start by describing how a popular algorithm
used to generate samples for surrogate modeling does not meet the
current requirements of this task (Section~\ref{sec:background}).
Next, we consider various data analysis tasks, ranging from surrogate
modeling and hyperparameter optimization to data exploration, that
require sampling and identify the characteristics that a good sampling
algorithm should possess (Section~\ref{sec:roles}).  We then briefly
describe several commonly-used sampling algorithms in various
disciplines (Section~\ref{sec:algorithms}) and show how we can easily
modify some of these algorithms to generate samples for tasks of
interest (Section~\ref{sec:adapting}).  This allows us to
qualitatively compare the algorithms and identify the ones that meet
most of our requirements. We focus on this subset of algorithms in
Section~\ref{sec:comparison} and quantitatively compare the quality of
sampling and the time required for generating samples as we vary the
number of dimensions and number of samples. We summarize our
observations in Section~\ref{sec:conclusions}.

%
\section{Background and motivation}
\label{sec:background}
%

Sampling is an important task in many aspects of data processing,
including the generation, exploration, visualization, and analysis of
the data.  In a broad sense, we can think of sampling in two ways:
\begin{itemize}
\item {\bf Generating $N$ samples:} Here, we are interested in
  generating $N$ samples at points or locations in a $d$-dimensional
  space at which we want to collect or generate data. The $i$-th
  sample is defined as ${\bf x}_i = ( {x_{i1}, \ldots, x_{id}} )$. We
  assume that the $d$-dimensional input space is defined over
  $[0,1]^d$. If the range of any dimension is different, it can be
  scaled to $[0,1]^d$, the samples generated, and the coordinates then
  mapped back to the original range.

\item {\bf Selecting a subset of $N$ samples:} Here, we already have a large
  data set with $M$ samples and we want to select a subset of $ N <
  M $ samples for use.  Each sample is a $d$-dimensional vector ${\bf x}_i = (
  {x_{i1}, \ldots, x_{id}} )$, with $i = 1, \ldots, M$. 

\end{itemize}
As we show later, in either of these two options, there are often
constraints placed on the generation or the selection of the samples
depending on the task for which the samples will be used.

Sampling is a very active area of research and, as we describe in
Section~\ref{sec:algorithms}, sampling techniques have been developed
independently in vastly different fields ranging from surrogate-based
design and analysis of computational experiments, graphics,
visualization, image processing, applied mathematics, data analysis,
combinatorial optimization, as well as the many application domains
where sampling is used. Despite this wealth of ideas, there is little
cross fertilization among these fields. When there is a need for
sampling in a new context, an existing technique that is used within a
discipline is often adapted to this new context, even if it results in
a less-than-optimal solution, but potentially better-suited algorithms
developed in a different and unrelated field are not considered.

Consider, for example, the generation of samples for the Design and
Analysis of Computer Experiments, also referred to as
DACE~\cite{sacks1989:dace,garud2017:doereview,santner2018:book,bhosekar2018:surrogate}.
This counterpart to the traditional design of
experiments~\cite{montgomery12:book} focuses on simulations, also
called computer experiments, that are increasingly being used as a
cost-effective alternative to, or in conjunction with, real-life
experiments~\cite{crombecq2011:spacefilling,joseph2016:spacefilling}.
The basic idea is to generate a number of sample points in the space
spanned by the input parameters of the simulation, run the simulation
at these points to generate output variables, and analyze the
corresponding inputs-outputs to understand the behavior of the
phenomenon being simulated. When the input space is high dimensional,
the number of samples used to cover the space adequately is large,
making it prohibitively expensive to run many simulations, especially
if each one takes hours or days on a high-performance computing
system. An often-used solution is to build a ``code surrogate'' or
``meta-model'', which is a regression model built by
using the inputs and their corresponding outputs as a training set.
This regression model, which could be either a polynomial model or a
machine-learning model, can then be used as a fast interpolator.

\begin{figure}[htb]
\centering
\begin{tabular}{cccc}
\includegraphics[trim = 2.9cm 0cm 2.9cm 0cm, clip = true,width=0.18\textwidth]{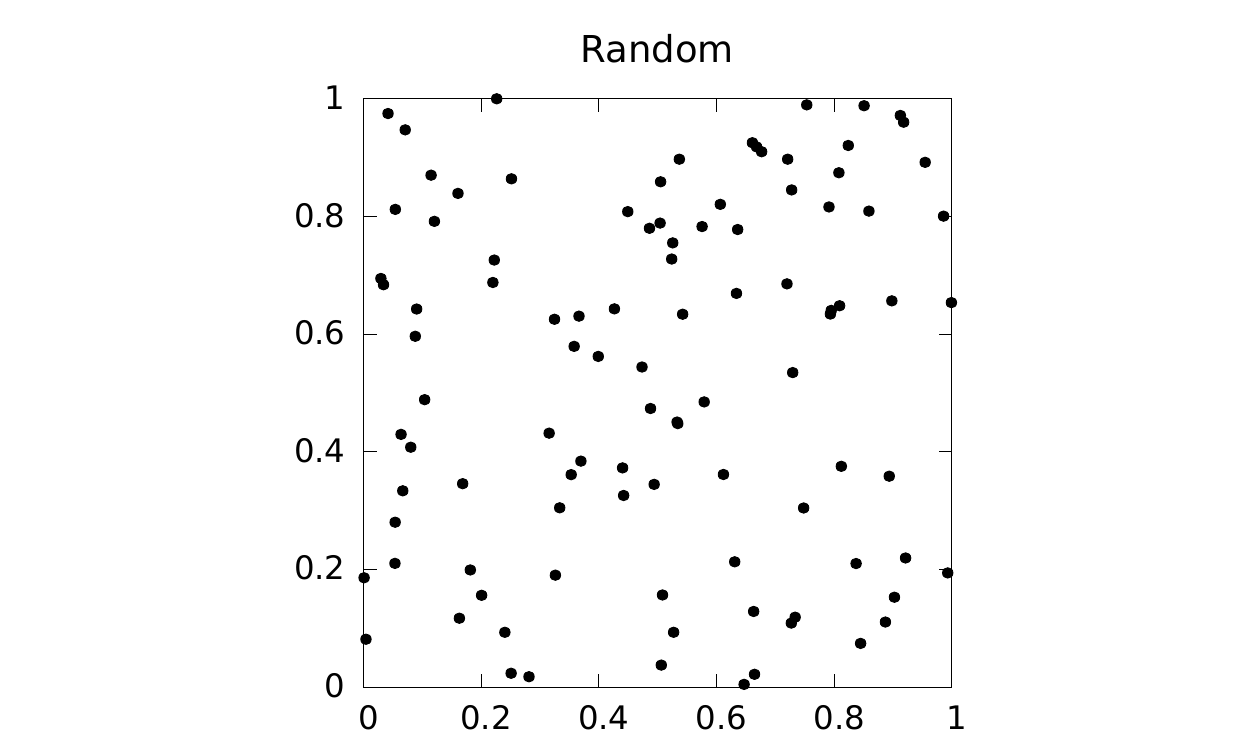} &
\includegraphics[trim = 2.9cm 0cm 2.9cm 0cm, clip = true,width=0.18\textwidth]{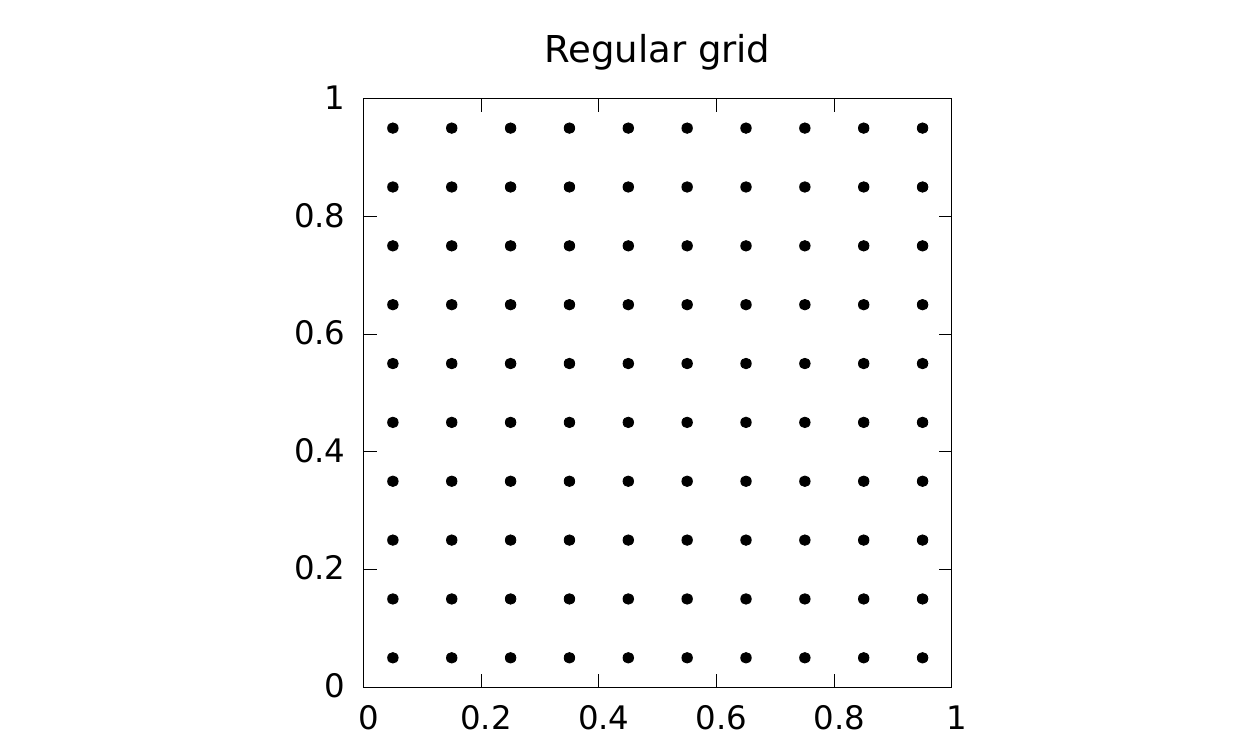} &
\includegraphics[trim = 2.9cm 0cm 2.9cm 0cm, clip = true,width=0.18\textwidth]{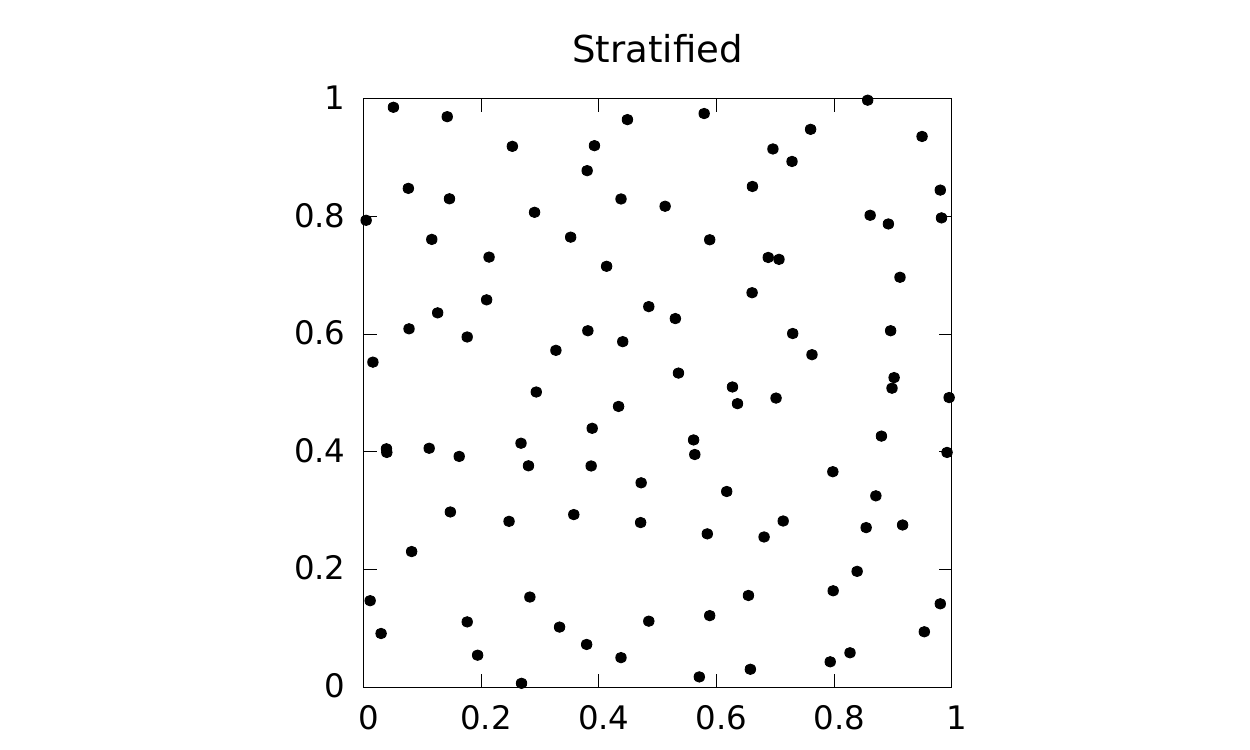} &
\includegraphics[trim = 2.9cm 0cm 2.9cm 0cm, clip = true,width=0.18\textwidth]{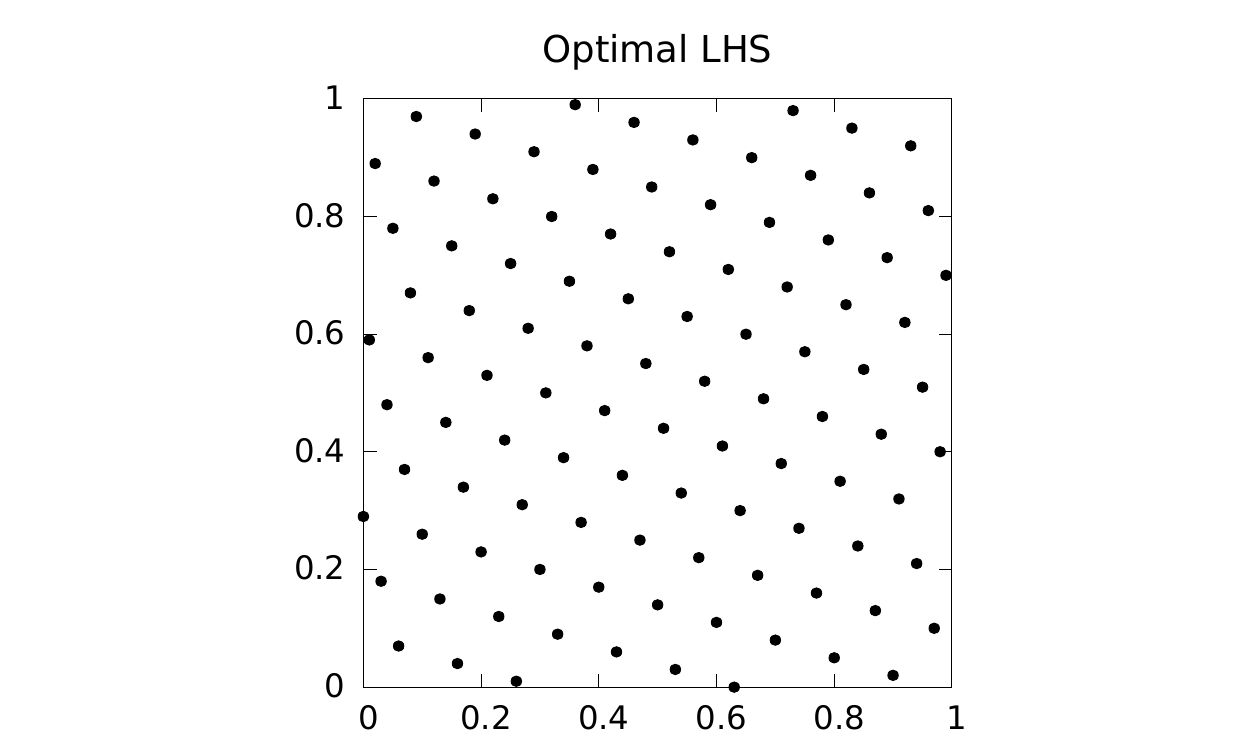} \\
(a) & (b) & (c) & (d) \\
\end{tabular}
\vspace{-0.2cm}
\caption{Sampling in two dimensions: 100 points using (a) random
  sampling; (b) grid sampling and (c) random stratified sampling, both
  with ten bins along each dimension; and (d) optimal Latin hypercube
  sampling from https://spacefillingdesigns.nl/.}
\label{fig:sampling_example}
\end{figure}

The accuracy of these code surrogates depends not only on degree of
the polynomial model or the machine learning model used, but also on
the samples at which the simulation is run. In the absence of any
prior information to guide the placement of a limited number of
samples, an intuitive approach is to start by placing the samples
uniformly and randomly in the input space.
Figure~\ref{fig:sampling_example} shows possible solutions to placing
100 samples in a two-dimensional domain using four different schemes -
uniform random sampling, grid sampling, stratified random sampling,
and an optimal Latin hypercube sampling, which is a highly popular
method used in surrogate modeling. We observe that the randomly-placed
samples have over- or under-sampled regions. In contrast, grid
sampling covers the space well, but the number of samples is the
product of the number of bins along each dimension, which limits our
ability to generate an arbitrary number of samples, making the method
inflexible especially in high-dimensional spaces. Grid sampling also
has the drawback that only a few coordinate values are sampled in each
dimension. This issue is somewhat mitigated by the stratified random
sampling, which places a sample randomly in each grid cell instead of
at the corners, but suffers from the same over- and under-sampling
issue as uniform random sampling.

The Latin hypercube sampling (LHS), one of the most popular sampling
designs used in code surrogates~\cite{santner2018:book}, can generate
an arbitrary number of samples that are well spread out in the input
space and uniformly cover the coordinate values along each dimension.
Despite this desirable distribution of sample points, as the use of
code surrogates gains wider acceptance, certain drawbacks of LHS 
become apparent. As we discuss in Section~\ref{sec:algo_lhs},
generating an optimal LHS can be prohibitively expensive, especially
as the number of dimensions and samples increases. Further, as code
surrogates are adapted for use in a sequential setting, where the
samples are added incrementally, we find that LHS does not lend itself
to the addition of new samples while maintaining the desirable
distribution of sample points and the use of the previously-generated
samples.

Our experiences trying to unsuccessfully adapt the LHS algorithm to
meet our needs in surrogate modeling prompted us to consider sampling
techniques developed in other fields. We were interested in algorithms
that either directly met our needs or could be easily modified to do
so. We also wanted to create solutions in situations with limited time
and computational resources.  We refer to this approach, where we
devise algorithms specifically tailored to the requirements of a task,
as {\it intelligent sampling}.  The name refers not to a specific
algorithm, such as the smart sampling algorithms described
in~\cite{loyola2016:smartsampling,garud2018:smartsampling}, but an
approach to sampling.  We focus not just on building surrogate models,
but also other tasks in data analysis that we describe in more detail
in the next section, along with their sampling requirements.

Our contributions in this paper are as follows: Our major contribution
is to show how ideas from different disciplines can be combined and
modified to create algorithms that meet the sampling needs of tasks in
surrogate modeling, hyperparameter optimization, and data analysis. We
develop new algorithms for i) fast generation of space-filling Latin
hypercube samples using efficient interchanges; ii) space-filling
sampling that supports both progressive and incremental generation of
samples; iii) generating samples distributed in accordance with a
probability density function; and iv) generating random but
uniformly-distributed samples in strangely shaped regions that are
space filling relative to any previously generated samples in the
region.  Our minor contribution in this paper is to make the data
analysis community aware of the sampling algorithms developed in other
fields so they can utilize these algorithms to meet their sampling
requirements. Our intent is to show that some of these ideas are worth
considering and that simple algorithms can be competitive, perhaps
even better than more sophisticated solutions.

In this paper, we will use the term ``regularly-spaced'' or ``regular
sampling'' to refer to sampling where the sample points are on a grid,
and the term ``random uniformly-spaced'' or ``random uniformly distributed''
to refer to sampling where the sample points are placed randomly, but
such that the distance of each sample to the nearest neighbor is
roughly the same. In other words, the former is structured, but the
latter is not.

%
\section{The role of sampling in data analysis}
\label{sec:roles}
%

We next describe the ways in which sampling techniques are used in
various data analysis tasks, along with the requirements for these
techniques. In the tasks considered, we exclude the transmission and
display of data; these topics, often considered in graphics and
visualization~\cite{brandt2019:gpufp}, are restricted to two or three
dimensions and require the real-time generation of a far larger number
of samples, in hundreds of thousands or greater, than the tasks of
interest in this paper. We, however, do borrow extensively from the
ideas developed in these fields.  We conclude this section with a list
of desirable attributes that a sampling algorithm should have to solve
problems in surrogate modeling, hyperparameter optimization, and data
analysis.

%
\subsection{Surrogate modeling}
\label{sec:task_surrogate}
%

We start by discussing the task of building surrogate models as it has
some of the more diverse requirements for sampling algorithms.  As
explained in Section~\ref{sec:background}, a code surrogate is a
regression model that is built by first running an expensive computer
simulation at a few carefully selected sample points in the domain
spanned by the input parameters of the simulation, and then using the
inputs and outputs of the simulations as a training data set to build
the surrogate model. The number of input parameters to a simulation
can be small, equal to two or three, but it can also be quite large,
numbering in the tens.  In the context of DACE (design and analysis of
computer experiments) the input parameter space is referred to as the
{\it design space} and the set of sample points is the {\it
  experimental design}, the {\it sampling}, or the {\it sample set}.
As the dimension of the input space increases, the number of samples
required to cover the space adequately to build an accurate surrogate
increases exponentially.

The regression model built in DACE can be used in many different
ways~\cite{kamath2018:small,kamath2021:inverse}: as an interpolator to
predict the simulation outputs at new sample points in the domain, to
identify samples at which the outputs have certain values (that is,
solving an inverse problem), to identify regions where additional
samples are necessary to improve the accuracy of the surrogate, and to
identify regions of interest to explore further with improved
simulations or experiments.

Traditionally, a {\it one-shot} approach has been used to build a
surrogate model where all the samples for a computer experiment are
identified at the start, the simulation run at these sample points,
and an appropriate model built using the simulation inputs and
outputs. But as the ideas behind DACE are adopted in a range of
science and engineering problems, the one-shot approach is no longer
the best option to address the practical requirements of these
problems.  For example, we may be able to run only a limited number of
computationally expensive simulations due to constraints on computer
resources available for use, or due to constraints on time, when a
fast response to a question is needed. If, in addition, the phenomenon
being modeled is poorly understood, it is more efficient to use an
incremental approach that would allow us to learn about the phenomenon
as we go along and place samples appropriately. Or, if a
scientist does not know the exact range of each input parameter, a
small number of simulations run using an initial guess at these ranges
could indicate whether the ranges need to be shrunk or expanded. Or,
we may find that the ranges are correct, but the initial simulations
indicate that the solution to the problem being addressed lies in a
small sub-region of the initial domain and a denser sampling in just
this sub-region would give a more accurate surrogate. This sub-region
may not necessarily be rectangular. We may also find that our initial
sampling is too coarse and we need to add samples over the entire
domain to build a more accurate model.

In light of these requirements, a sequential design that would allow a
trade-off between {\it exploration} and {\it exploitation} of the
design space, is
preferred~\cite{crombecq2011:spacefilling,eason2014:adaptive}. In
exploration, we would use the surrogate to explore the entire design
space further, with the goal of finding regions of interest, such as
discontinuities or steep slopes in the output. In exploitation, we
would focus on regions already determined to be interesting, perhaps
in a previous exploration step, and try to improve our understanding
of the problem. For example, we may want to investigate the variation
in the region around a minimum, or to determine if a region with a
steep slope is a discontinuity.

Unfortunately, popular sampling algorithms do not lend themselves to
easy modification to meet all these new requirements and we need to
consider new algorithms that would enable us to make better use of our
computational resources and create more accurate surrogate models in
less time than otherwise.

%
\subsection{Hyperparameter optimization}
\label{sec:task_hyperparam}
%

The next task we consider is hyperparameter optimization, where we
want to find the optimal parameters for a machine learning algorithm.
These parameters are sometimes divided into two types: model
hyperparameters that define the model that is built using the data,
such as the topology and the size of a neural network, or algorithm
hyperparameters that influence the speed and quality of the learning
process, such as the learning rate or mini-batch size. 
In hyperparameter optimization, we optimize a loss function, which may be a
measure of the error in prediction, such as the cross-validation error
for a regression model built using a specific set of parameters.
Typically, a tuple of hyperparameters is considered, a model is built
with this tuple, and the resulting loss function is evaluated. A
search through the space of hyperparameter tuples will yield the
optimal value of the tuple.  It has been
observed~\cite{bergstra2012:hyperparam} that for most data sets, only
a few of the hyper-parameters really matter, but that different
hyper-parameters are important on different data sets, making random
search a better option than grid search.

We observe that building an accurate surrogate for an expensive
computer simulation, as described in Section~\ref{sec:task_surrogate},
and finding the optimal hyperparameters for an algorithm are very
similar tasks, despite having very different motivations. In
Table~\ref{tab:surrogate_hyperparam}, we compare the steps in the task
of building a Gaussian process (GP) surrogate for a simulation with
multiple inputs and a single output and the corresponding steps in the
task of finding optimal hyperparameters for a neural network (or any
other algorithm) using Bayesian
optimization~\cite{frazier2018:bopttutorial}, which also uses a
Gaussian process.

\renewcommand{\arraystretch}{1.2}
\begin{table}[htb]
  \begin{center}
    \begin{tabular}{|p{7cm} | p{7cm} |} \hline
      Surrogate modeling & Hyperparameter optimization \\
      \hline
      Task: Build a GP surrogate model to accurately predict simulation output in a given range of input values & Task: Find the optimal hyperparameters for a neural network using Bayesian optimization \\
      \hline
      Generate samples in the high-dimensional input space of the simulation & Generate samples in the high-dimensional input space of hyperparameters \\
      \hline
      Run the simulation at the sample points and generate the simulation output & Train a neural network for each tuple of hyperparameters and calculate loss function for each \\
      \hline
      Build the GP surrogate using the training set formed by the inputs and the corresponding  output of the simulations & Build a GP model using the training set formed by the hyperparameter tuples and the corresponding loss function \\
      \hline
      Use the leave-one-out error in prediction and the uncertainty in prediction to identify locations of new samples to be used as inputs to the simulation & Use the GP model at a set of samples to evaluate an acquisition function that identifies the next hyperparameter sample to be evaluated \\
      \hline
Continue adding samples in simulation input space until desired accuracy of the surrogate is obtained & Continue adding samples in hyperparameter tuple space until the loss function is minimized \\
      \hline
    \end{tabular}
  \end{center}
  \vspace{-0.2cm}
  \caption{Similarities between the steps used for building a Gaussian process (GP) 
    surrogate for a simulation and for finding optimal hyperparameters for a neural 
    network using Bayesian optimization.}
  \label{tab:surrogate_hyperparam}
\end{table}

Based on the similarity between the two tasks, the characteristics of
sampling algorithms for hyperparameter optimization are the same as
those for surrogate modeling described in
Section~\ref{sec:task_surrogate}. In both cases we need to sample a
high-dimensional input space such that we can optimize an
appropriately defined quantity by evaluating the value of an expensive
function at these samples. Needless to say, we want to perform this
optimization using as few samples as possible.

%
\subsection{Other data analysis tasks}
\label{sec:task_other}
%

In addition to surrogate modeling and hyperparameter optimization,
there are several other data analysis tasks where sampling is required
and that would benefit from improved algorithms. These tasks include:

\begin{itemize}

\item {\bf Data compression:} Sampling can also be used in compressing
  data sets using machine learning.  For example, given a data set
  with each data point described by several inputs and one or more
  outputs, we can take a random sample of the data points to create a
  training set and use it to build a machine learning model and
  predict the values of the outputs at the remaining data points. If
  the error in prediction at points not in the training set is high,
  we can enhance the training set using suitably selected new points
  from the data set.  As only the training data needs to be saved,
  this process effectively compresses the data set.

  This idea for compression has been applied both to image
  data~\cite{cheng07:compress} and to time varying output from
  simulations~\cite{kamath17:compression,kamath18:compression}. In
  both cases, it is desirable that the points in the initial training
  set be randomly selected and as far apart as possible. Selecting new
  data points with high reconstruction error to add to the training
  set must be done carefully. While there may be many such points in a
  region, including just one (or a few) of these points may be
  sufficient to improve the error at all points in the region.

\item {\bf Data exploration: } Another area where sampling techniques
  play an important role is in data exploration, especially for
  massive data sets. Typically in data analysis, we start by exploring
  the data set to determine its characteristics so we can select
  suitable analysis algorithms and their parameters.  For large
  data sets, it may not be possible to visualize or plot the full
  data set, or even load it into the memory of a computer. An
  alternative is to explore the data set by focusing on a subset of the
  instances. A random sampling of instances is a computationally fast
  way to generate this subset, but suffers from under- and
  over-sampling issues (see Figure~\ref{fig:sampling_example}(a)). It
  is preferable to select subset instances that are farther apart from
  each other, but many algorithms to generate such a sampling require
  the entire data set in memory, which is infeasible for large
  data sets.  In previous work~\cite{kamath19:bigdata}, we showed that
  we can modify an algorithm from computer graphics to select samples
  far from each other in one pass through the data set.  When these
  samples are used to probe the data set, the results are more
  insightful than using random sampling.

\item {\bf Selecting initial centroids in clustering:} Some clustering
  algorithms, such as $k$-means
  clustering~\cite{morissette2013:kmeans}, start with an initial
  selection of $k$ centroids and iteratively refine the locations of
  these centroids. Selecting the initial centroids randomly from the
  instances that form the data set, or at random points in the domain
  of the data set, could lead to slower convergence due to the under-
  and over-sampling issues. A better initial guess may be to select
  the centroids to be far from each
  other~\cite{gonzalez1985:clustering}.

\item {\bf Selection of next sample in active learning or sensor
    placement:} In some tasks, such as active learning for
  regression~\cite{settles2009:active} or placement of
  sensors~\cite{zack2010:esa}, the currently available data are
  analyzed to identify the next sample to label or the next location
  at which to place a sensor. The approach usually involves generating
  an appropriate metric, such as the uncertainty in prediction at
  possible candidate samples or locations, and selecting as the next
  sample or location, the one with the highest uncertainty. As
  discussed earlier in sampling for data compression, when these
  high-uncertainty samples are close together, it may be more useful
  to select one sample from the highest-uncertainty region and select
  the next sample farther away from the first to make better use of
  resources.

\end{itemize}

%
\subsection{Characteristics of good sampling algorithms}
\label{sec:task_goodalgo}
%

Based on the discussion of the role sampling plays in surrogate
modeling, hyperparameter optimization, and data analysis, we 
identify the following desirable characteristics of a sampling algorithm:

\begin{itemize}

\item {\bf User-specified number of samples:} We should be able to
  specify any number of samples without constraints such as the number
  of samples being a power of two, or a product of a set of numbers.
  Further, for ease of use, the number of samples should be a direct
  input to the sampling algorithm rather than being specified
  indirectly, for example, as the minimum distance between samples.

\item {\bf Random, with no structure in the sampling:} Randomness in
  the locations of the samples generated is an important attribute as
  in some tasks, such as hyperparameter optimization, a random
  sampling finds as good or better models than grid search in a small
  fraction of the computation time~\cite{bergstra2012:hyperparam}.
  This is because only a few dimensions are usually important for a
  data set, and a grid-based sampling, by replicating values in
  lower-dimensional projections, makes poor use of the
  samples~\cite{joseph2016:spacefilling}.  In surrogate modeling and
  graphics, regularly placed samples can lead to aliasing problems and
  limit the complexity of the functions being modeled, per the
  Shannon-Nyquist theorem~\cite{mitchell1987:antialiased}.

\item {\bf Space-filling sampling:} The samples should be {\it
    space-filling}~\cite{joseph2016:spacefilling}, covering the input
  domain uniformly.  By uniformly, we mean that the sample density is
  approximately constant, with the sample set providing an equal
  amount of information about every part of the
  domain~\cite{eldar1997:farthest}. This is especially true for the
  initial set of samples in surrogate modeling because, in the absence
  of any knowledge about how the simulation outputs are related to the
  inputs, a sample design that is spread out over the input domain
  would give a better model than one that over- or under-sampled
  parts of the region.
 
\item {\bf Progressive sampling: } The space filling property should
  be progressive --- as the samples are generated, they should
  initially cover the entire input domain coarsely, with the coverage
  becoming finer as more samples are added. By processing the samples
  in order, this would allow us to quickly identify incorrect input
  ranges before we expend too many resources. This constraint would
  prefer algorithms that generate a sequence of samples, where a
  sample is added one at a time, instead of algorithms that generate a
  set of samples, where all the samples are generated at the
  same time.

\item {\bf Incremental sampling: } This characteristic is closely
  related to progressive sampling, but instead of focusing on the
  coverage of the domain, we consider the number of samples. If we
  initially start with a set of random, uniformly distributed samples,
  we should be able to add an arbitrary number of additional samples
  to generate a denser sampling of the region.  The region over which
  the samples are added may or may not be rectangular.

\item {\bf Non-collapsing sampling:} A sampling with this
  characteristic has sample points that have unique values along any
  of the input dimensions~\cite{crombecq2011:spacefilling}. This also
  means that when the samples in a $d$-dimensional space are projected
  onto a $(d-1)$-dimensional space, no two points are projected onto
  each other. This is useful in data sets where it is not known if one
  of the input dimensions has little to no effect on the output; when
  this is the case, two samples that differ only in this one
  dimension, will be considered the same point, resulting in a waste
  of resources.

\item {\bf Sampling based on a probability density function (PDF):} We
  should be able to specify a PDF  to sample some
  regions of the domain more densely or sparsely than others while
  generating samples that are random and uniformly distributed. This
  capability is useful when part of the domain is of greater interest
  and would benefit from denser sampling than other regions of less
  interest.

\item {\bf Sampling in non-rectangular regions:} While most sampling
  algorithms assume that we are working in a rectangular domain, there
  are two situations where the domain may not be rectangular.  First,
  in generating an initial set of samples, we may know a priori that
  some parts of the domain are out-of-bounds or nonviable and we
  should not place any samples there. In this case, we will assume that
  we have access to a function that indicates if a sample in a certain
  location in a domain is allowed or not.  Second, in problems where
  we have generated an initial set of samples we may find that we need
  additional samples only in a small region of the original domain.
  This would be the case when we want to explore this region further
  or create more accurate surrogates in this region by increasing the
  sampling density.  These regions can often be quite thin and narrow,
  occupying a small percentage of a rectangular domain, but spanning
  the full extent of each dimension (see for example, Figure 6
  in~\cite{kamath2018:small}).

\item {\bf Sampling when domain extents are modified:} This capability
  is useful when an initial sampling indicates that the domain extents
  selected were incorrect and we need to either shrink the domain by
  removing points or extend the domain and add new points. The
  resulting sampling should still satisfy the random, uniform coverage
  of the space.

\item {\bf Sampling in high dimensional spaces:} We should be able to
  generate samples in high dimensional spaces, not just two or three
  dimensions. The number of samples required to cover a
  high-dimensional space is large, and careful placement of samples
  would be critical for efficient data analysis.

\item {\bf Sampling a subset from a previously generated set of
    samples: } This situation arises when we want to explore a large
  data set by extracting a subset of samples and analyzing them to
  understand the larger data set.

\item {\bf Efficient generation of large number of samples:} In some
  problems, instead of a small number of carefully selected samples,
  we may want to generate efficiently a large number of samples that
  randomly cover the domain uniformly, for example, when we want to
  use a surrogate model to predict the output at points in a domain.

\end{itemize}

In addition, as with any algorithm, we want our sampling algorithms to
be computationally efficient, easy to implement, depend on only a few
parameters, and be relatively robust to the choice of these
parameters.  Understandably, this is a long list of requirements to be
met by a sampling algorithm and it may not be possible for a single
algorithm to meet all these requirements or be amenable to
modifications so they can meet the ones that are important to our
application. However, as we show in Section~\ref{sec:adapting}, we can
make simple modifications to some algorithms to enable many of these
requirements to be met.

%
\section{An overview of common sampling algorithms}
\label{sec:algorithms}
%

We next describe some of the sampling algorithms commonly in use in
various disciplines including design of computer experiments, image
processing, combinatorial optimization, and computer science. In each
case, we describe the basic algorithm assuming that we are generating
$N$ samples at points or locations in a $d$-dimensional space, that
is, the $i$-th sample is ${\bf x}_i = ( {x_{i1}, \ldots, x_{id}} )$.
We further assume that the $d$-dimensional input space is defined over
$[0,1]^d$.  As mentioned earlier, if the range of any dimension is
different, it can be scaled to [0,1], the samples generated, and the
coordinates mapped back to the original range.  We also show the
sampling results for the generation of 100 samples in two-dimensions.
This allows us to visually evaluate the quality of the samples. Later,
in Section~\ref{sec:comparison}, we quantitatively compare the
performance of the more promising algorithms for generating a larger
number of samples in higher dimensions.

In our work, we do not include the quasi-Monte Carlo algorithms, which
include the Sobol, Halton, and Hammarsley sequences.  These techniques
generate sequences that cover the sampling space well, and perform
better than standard Monte Carlo methods using random or pseudo-random
sequences, but they have problems in high-dimensions resulting in poor
two-dimensional projections that have regular structure or large
regions with no
samples~\cite{morokoff1994:quasirandom,chi2005:halton}.

%
\subsection{Uniform random sampling}
\label{sec:algo_random}
%

Uniform random sampling is the simplest form of sampling where the
coordinates of each sample point in a $d$-dimensional space are
generated randomly.  However, as shown in
Figure~\ref{fig:sampling_example}(a) for 100 samples in a
two-dimensional space, the sample set does not cover the space
uniformly, resulting in  over- or under-sampled regions.
However, the method is simple, very fast, and can be easily
parallelized, making it suitable for generating a large number of
initial samples for use in more complex sampling algorithms.

%
\subsection{Grid sampling}
\label{sec:algo_grid}
%

In contrast to uniform random sampling, where we focus on the
``randomness'' of a sample set, in grid sampling, we focus on samples
that are far from each other and cover the space uniformly. As shown
in Figure~\ref{fig:sampling_example}(b) for 100 samples in a
two-dimensional space, we sacrifice the randomness of the data points,
sampling only a few coordinates along each dimension. Randomness can
be reintroduced using stratified random sampling, with a sample point
selected randomly in each cell of the regular grid, as shown in
Figure~\ref{fig:sampling_example}(c). However, this still has the
over- and under-sampling problem of uniform random sampling. For both
grid and stratified random sampling, the total number of samples is
the product of the number of bins along each dimension, resulting in
less flexibility in selecting the number of samples. Further,
having generated a sample set with $N$ samples, adding $M$ more
samples while maintaining the grid structure is not possible except
for specific values of $M$.

%
\subsection{Latin hypercube sampling and variants}
\label{sec:algo_lhs}
%

\begin{figure}[htb]
\centering
\begin{tabular}{ccc}
\includegraphics[trim = 2.8cm 0cm 2.7cm 0cm, clip = true,width=0.23\textwidth]{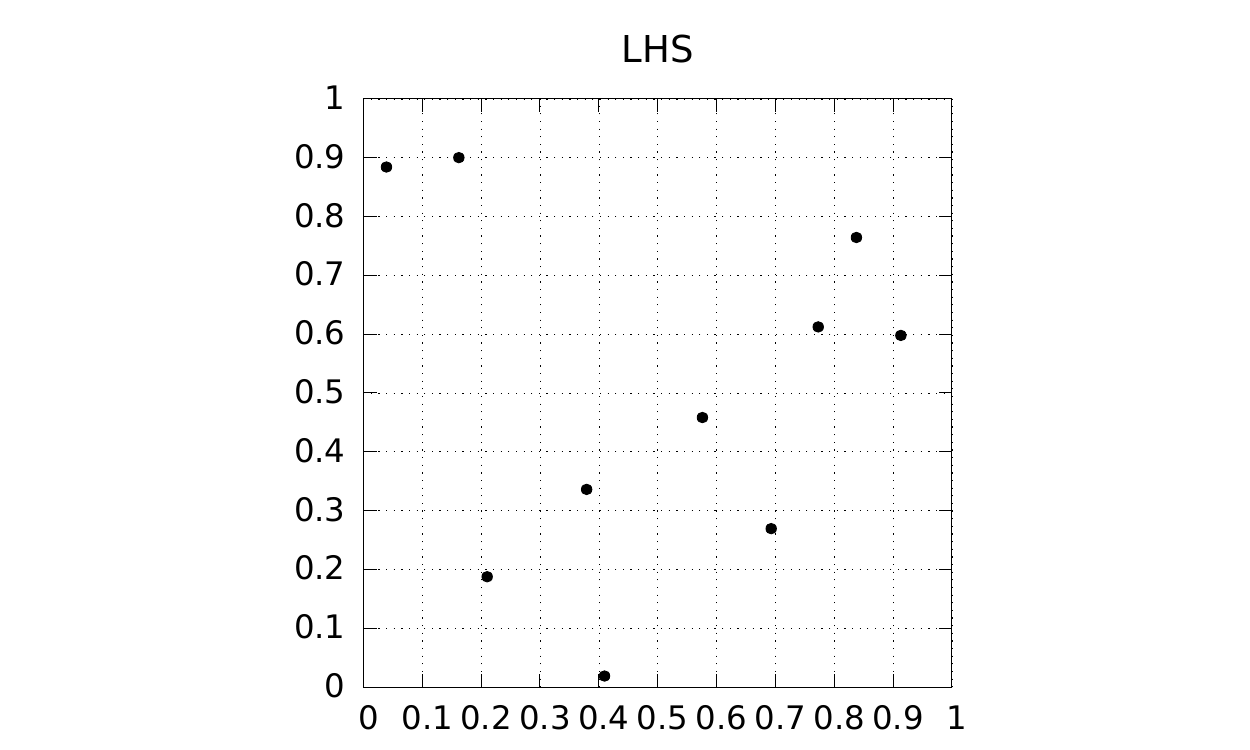} &
\includegraphics[trim = 2.8cm 0cm 2.7cm 0cm, clip = true,width=0.23\textwidth]{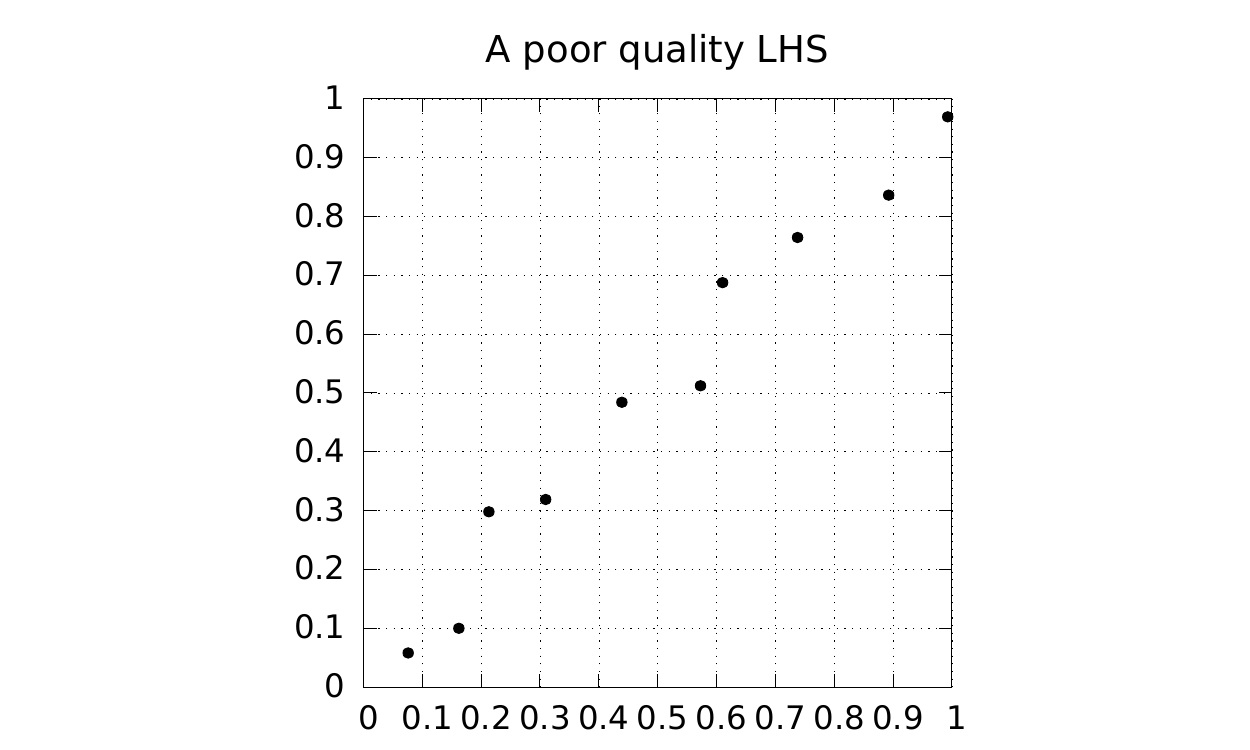} & 
\includegraphics[trim = 2.8cm 0cm 2.7cm 0cm, clip = true,width=0.23\textwidth]{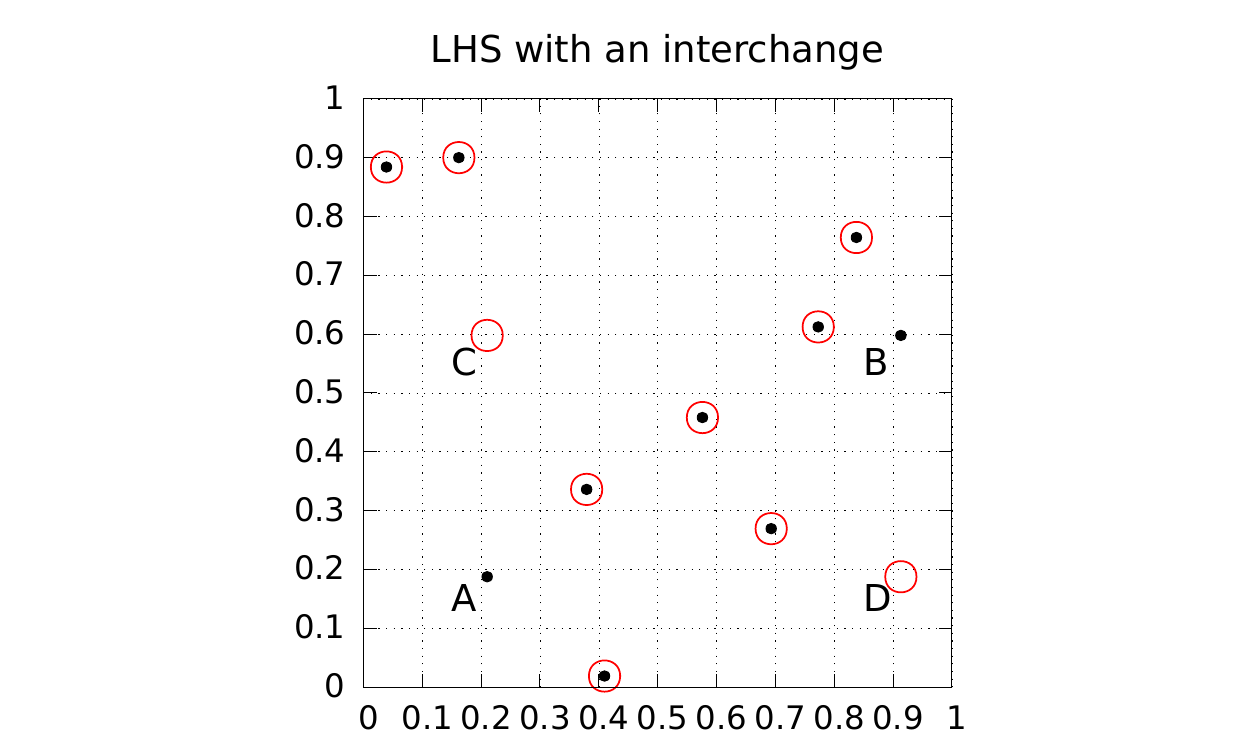} \\
(a) & (b) & (c) \\
\end{tabular}
\vspace{-0.2cm}
\caption{LHS using 10 points in two dimensions. (a) A sample LHS -
  each bin along the two dimensions has only one value. (b) A valid
  but poor-quality LHS that leaves large regions of the domain with no
  samples. (c) Effect of interchanging the second coordinates of
  samples A and B, resulting in samples C and D, respectively. The
  original sample set (black dots) becomes the new sample set (open
  red circles) after the interchange.}
\label{fig:simple_lhs}
\end{figure}

The Latin hypercube sampling (LHS) is one of the most popular sampling
algorithms for generating the initial set of samples for building code
surrogates~\cite{viana2016:lhstutorial,santner2018:book}.  Originating
in the statistics community in the early work of W.J. Conover in 1975
at Los Alamos National Laboratory~\cite{iman2006:lhs}, it first
appeared in a journal article in 1979~\cite{mckay1979:lhs}.
The basic idea in LHS (see Algorithm~\ref{algo:lhs_basic}) is quite
simple. To generate $N$ samples, we first divide the range of each of
the $d$ dimensions into $N$ equally-spaced intervals, and select a
coordinate value in each interval along a dimension, either randomly,
or at the center of the interval. This gives us $N$ coordinate values
in $d$ dimensions. Then, a sample point, composed of $d$ coordinate
values, is formed by randomly selecting one coordinate value (without
replacement) from each dimension.  As a result, for a two-dimensional
region, each row and column has a single sample, as shown in
Figure~\ref{fig:simple_lhs}(a). The name, Latin hypercube, is a
generalization of the term 'Latin square', in which a square grid of
sample positions in two dimensions has one and only one sample in each
row and each column. Such a sampling is said to have the Latin
property. By construction, LHS generates a set, not a sequence of
samples, that is, all samples are generated at the same time.

\begin{algorithm}
\caption{Basic Latin hypercube sampling }
\label{algo:lhs_basic}
\begin{algorithmic}[1]

  \STATE Goal: generate $N$ samples in a $d$-dimensional space
  
  \STATE Let $x_{ij}$ denote the $j$-th coordinate of the $i$-th sample ${\bf x}_i$.
  
  \FOR{$j = 1$ to $d$}
  
  \STATE Generate $P_j$, a random permutation of the set $ \{ 1, \ldots, N \}$, with the i-th element denoted by $P_j(i)$.
  
  \ENDFOR
  
  \FOR{$j = 1$ to $d$}
  
  \FOR{$i = 1$ to $N$}
  
  \STATE Generate  $x_{ij} = \frac{P_j(i) - U_j(i)}{N}$, where $U_j(i)$ is a uniform random number in $[0,1]$.
  
  \ENDFOR
  
  \ENDFOR

\end{algorithmic}
\end{algorithm}

The basic LHS algorithm has several advantages: it is very simple and
computationally inexpensive; unlike grid sampling, we can generate any
arbitrary number of samples; and, by virtue of its construction, the
samples have good projective property as the projection of the samples
on any dimension uniformly covers the range of values along that
dimension.  However, there is no guarantee that a valid LHS will
uniformly cover the design space as shown in
Figure~\ref{fig:simple_lhs}(b); in fact, generating a good-quality LHS
is non-trivial, especially in high-dimensional spaces or for a large
number of samples.

As a result, much of the research in the design of LHS has focused on
ways to generate a sampling that not only has good coverage along each
dimension individually, but also has good coverage of the overall
design space. This is typically accomplished by modifying an initial
LHS using interchanges to optimize a suitably-defined quality metric.
The simple idea of interchanges first identifies a dimension in the
data randomly and then interchanges the coordinate values of two
randomly-selected rows in that dimension, as shown in
Figure~\ref{fig:simple_lhs}(c). This preserves the Latin property of
the sampling. An interchange is accepted if it improves the quality of
the sampling. A common metric is the minimum distance between any two
samples, with the optimal sampling being one that maximizes the
minimum distance, that is, the {\it maximin}
criterion~\cite{johnson1990:maximin}.

A particular challenge in this optimization process is that the space
of these interchanges is quite large. If we start with an initial LHS
of $N$ samples in a $d$ dimensional space, then for each column, the
first row can be interchanged with any of the remaining $(N-1)$ rows,
the second row can be interchanged with any of the remaining $(N-2)$
rows, and so on. Thus, there are $N!$ distinct ways in which the
values in a column can be arranged. With $d$ columns, this results in
$(N!)^d$ possible interchanges. For even a small sampling of 10
samples in 4 dimensions, this is 1.734e+26 interchanges, making the
determination of an optimal LHS computationally
expensive~\cite{husslage2006:phdthesis}.
There are two approaches to reduce this cost: i) improve the
optimization by using algorithms such as simulated
annealing~\cite{morris95:design} and evolutionary
algorithms~\cite{jin2005:optimallhs} or limit the number of
interchanges to obtain an approximation to the optimal design; and ii)
define alternate objective functions that allow for faster
evaluation~\cite{jin2005:optimallhs} or give better designs by
Latinizing across multiple dimensions~\cite{joseph2015:maxpro}.
Further, to make it easier for users, pre-generated sample sets have
also been created for specific number of points and dimensions and are
available in the public domain~\cite{husslage2011:lhs}.
Understandably, this is restricted to a small number of samples in low
dimensions and is of little help to users requiring a moderate to
large number of samples in high dimensions. Some authors strongly
discourage optimizing an LHS on the fly for surrogate modeling as it
can be quite expensive~\cite{crombecq2011:spacefilling}; instead they
recommend an algorithm that is similar to the farthest-point algorithm
discussed later in this section as it is better and faster than LHS.

In Algorithm~\ref{algo:lhs_opt}, we propose a simple way to generate
an LHS that satisfies the maximin criterion.  We first show by
contradiction that to improve the value of this objective function,
any interchange must involve one of the two samples between which the
distance is minimum.  Suppose that,
in a randomly selected dimension, we interchange the coordinate values
of two samples, neither of which belongs to the minimum distance pair.
This interchange modifies only the distances between the two new
samples and the remaining samples. So, the minimum distance in the new
sampling is either the old minimum distance (as the minimum distance
pair was left unchanged), or a value even smaller than the old minimum
distance. In either case, we would reject the interchange as it does
not increase the minimum distance. Therefore, to increase the
possibility of an acceptable interchange, we only consider
interchanges where one sample is from the minimum distance pair and
the other is selected randomly. We refer to this algorithm, outlined
in Algorithm~\ref{algo:lhs_opt}, as Approximate Maximin LHS, to
distinguish it from other maximin algorithms.

\begin{algorithm}
\caption{Approximate Maximin Latin Hypercube Sampling }
\label{algo:lhs_opt}
\begin{algorithmic}[1]

  \STATE Goal: generate and LHS with $N$ samples in a $d$-dimensional
  space, where the minimum distance between samples is maximized.

  \STATE Set $\it nTries$, the number of tries and $\it
  nInterchanges$, the number of interchanges per try.

  \FOR{$j = 1$ to $\it nTries$}

  \STATE Generate $N$ samples using the basic Latin hypercube sampling
  (Algorithm~\ref{algo:lhs_basic}).

  \STATE Let $index_1$ and $index_2$ denote the indices  of the samples
  with the minimum distance.

  \FOR{$i = 1$ to $\it nInterchanges$}

  \STATE Randomly select $index$ to be either $index_1$ or $index_2$.

  \STATE For a random integer, $r_{col}$ in $[1, \ldots, d]$, and a
  random row, $r_{row}$ in $[1, \ldots, N]$, interchange the values in
  column $r_{col}$, rows $index$ and $r_{row}$.

  \STATE If the resulting sampling increases the minimum distance
  between samples, accept the interchange, and update the values
  of $index_1$ and $index_2$.

  \ENDFOR

  \STATE If the sampling on this try has a larger minimum distance
  between samples than the previous try, select the samples in this
  try as the current best sampling.

\ENDFOR

\end{algorithmic}
\end{algorithm}

\begin{figure}[!htb]
\centering
\begin{tabular}{cccc}
\includegraphics[trim = 2.8cm 0cm 2.8cm 0cm, clip = true,width=0.16\textwidth]{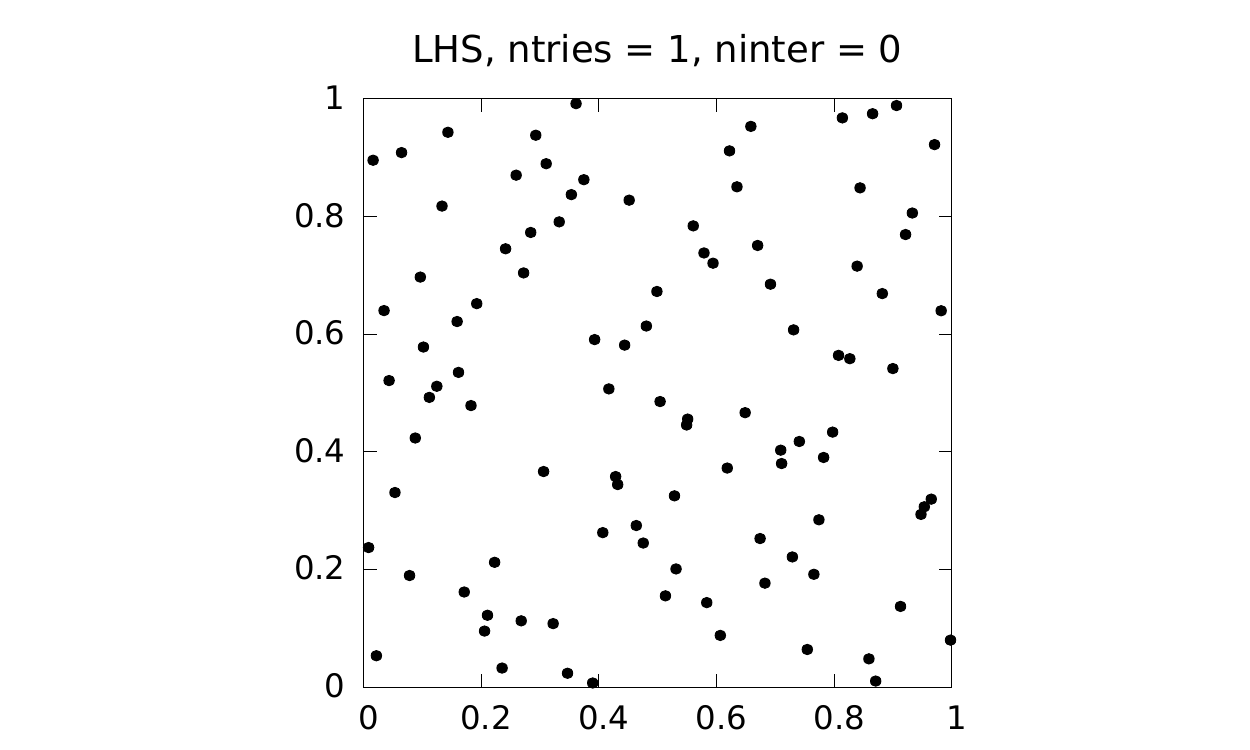} &
\includegraphics[trim = 2.8cm 0cm 2.8cm 0cm, clip = true,width=0.16\textwidth]{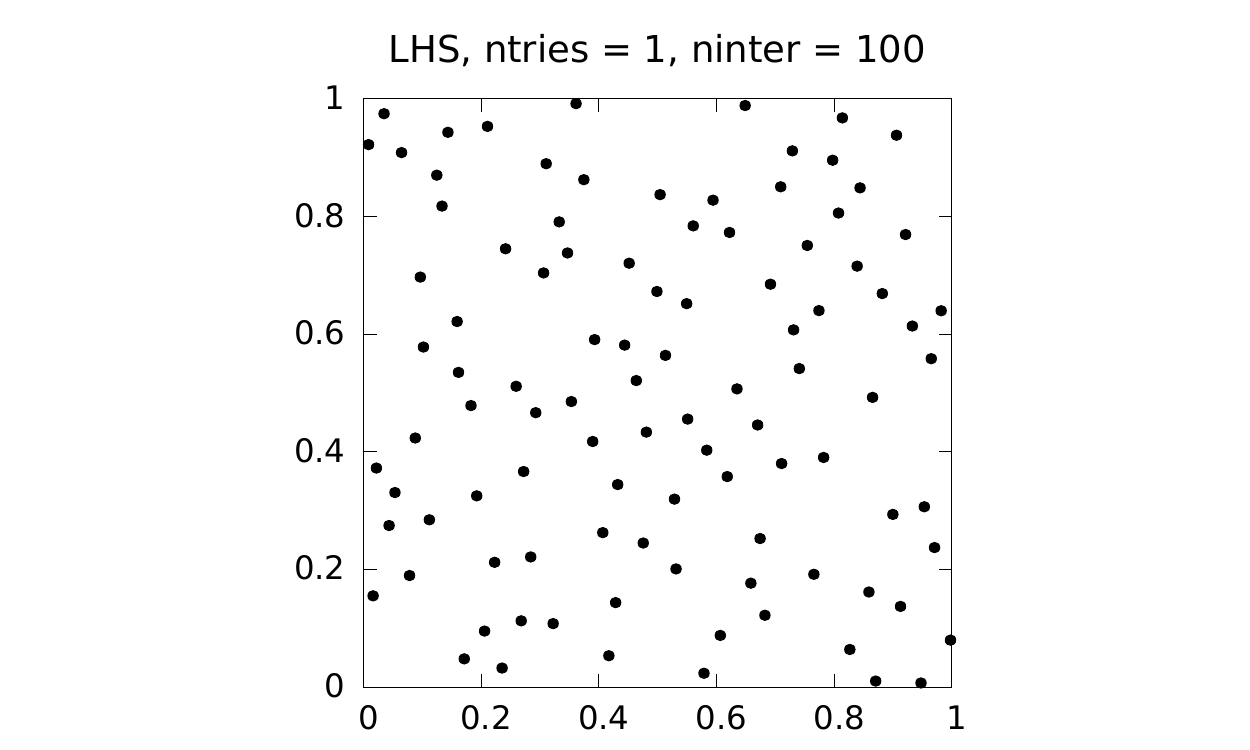} & 
\includegraphics[trim = 2.8cm 0cm 2.8cm 0cm, clip = true,width=0.16\textwidth]{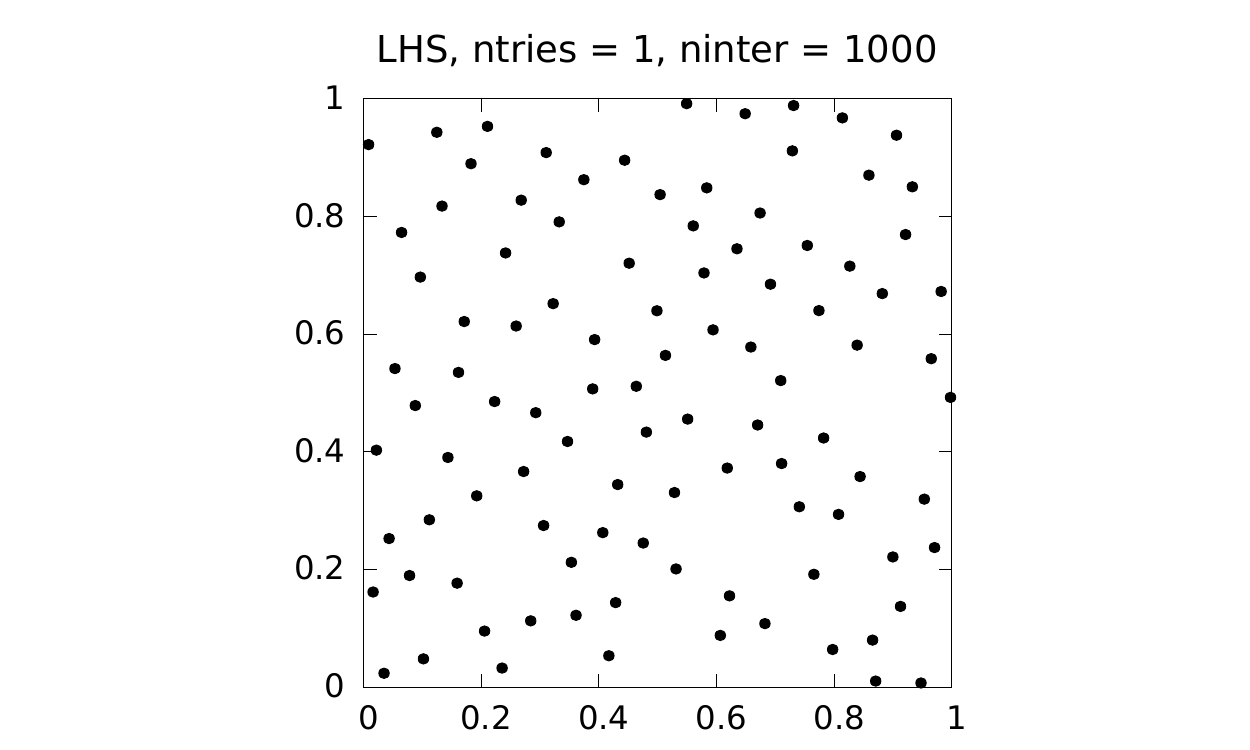} & 
\includegraphics[trim = 2.8cm 0cm 2.8cm 0cm, clip = true,width=0.16\textwidth]{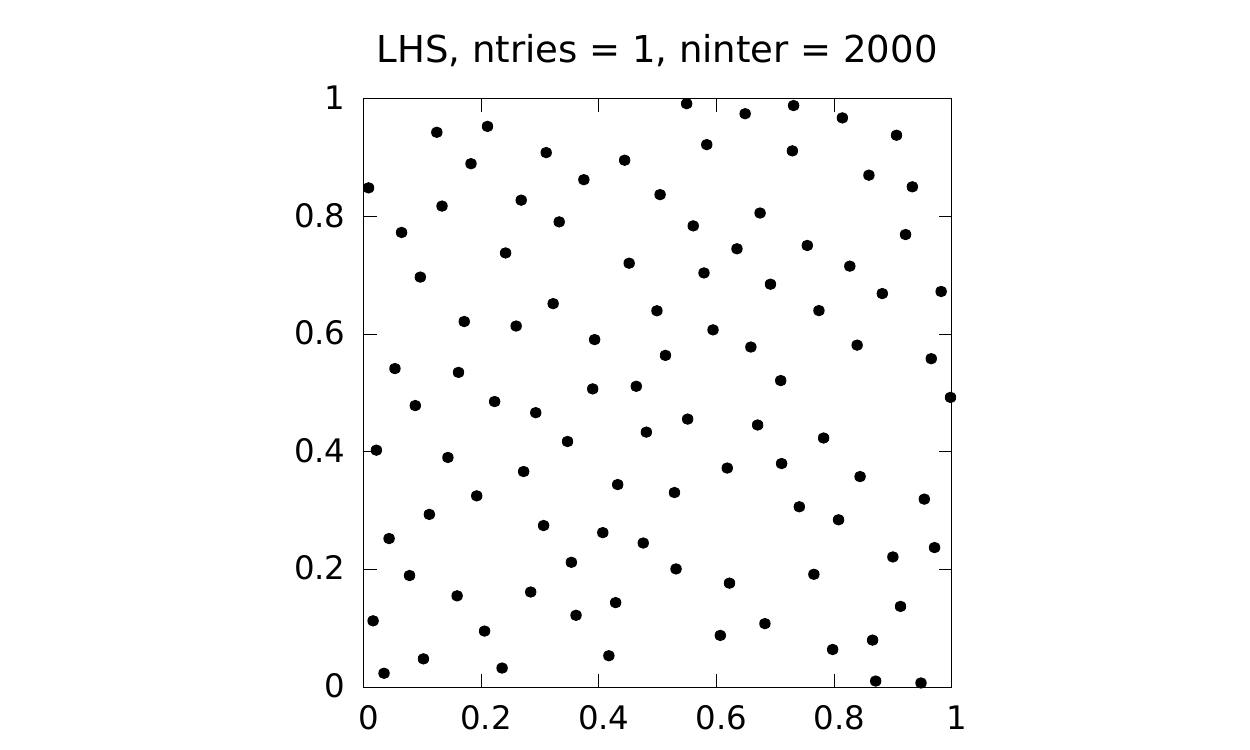}  \\
(a) & (b) & (c) & (d) \\
\multicolumn{4}{c}{\includegraphics[trim = 0.0cm 0cm 0.0cm 0cm, clip = true,width=0.40\textwidth]{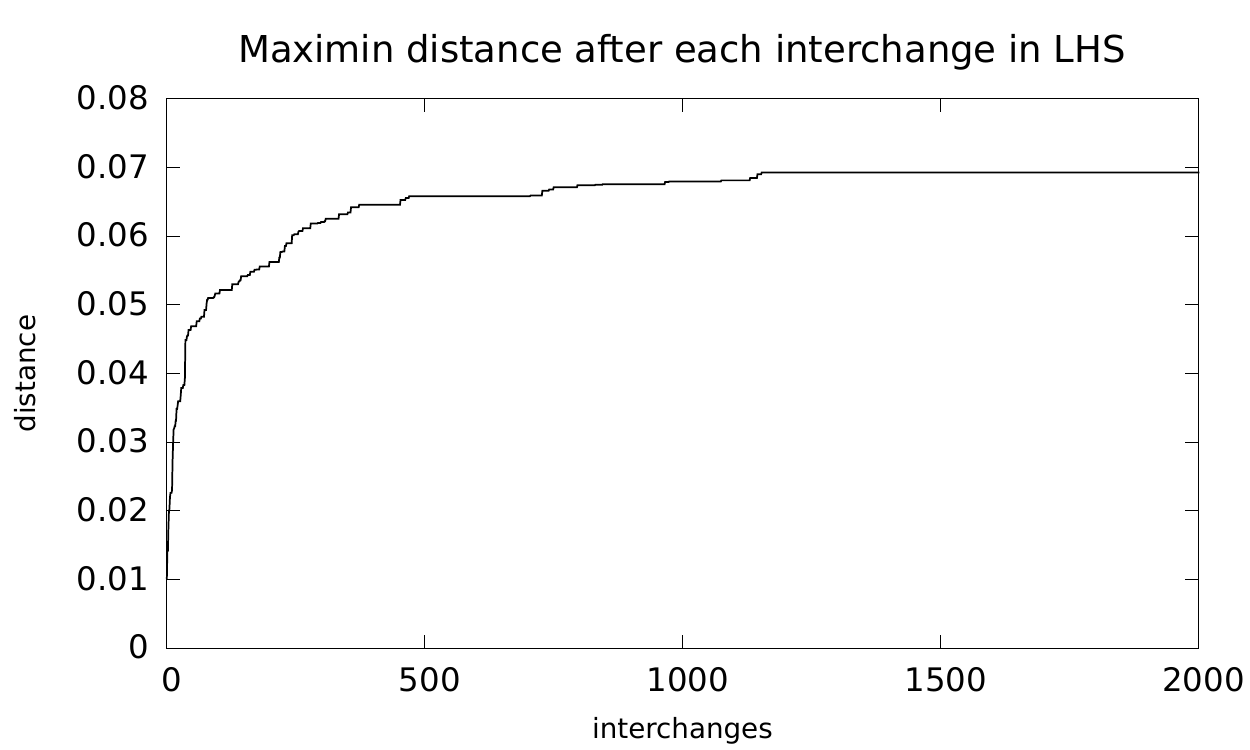}} \\
\multicolumn{4}{c}{(e)}\\
\end{tabular}
\vspace{-0.2cm}
\caption{Effect of interchanges in the Approximate Maximin LHS
  algorithm (Algorithm~\ref{algo:lhs_opt}): (a) 100 initial samples
  before interchanges and after (b) 100, (c) 1000, and (d) 2000
  interchanges. (e) The maximum value of the minimum distances between
  samples as a function of the number of interchanges indicates that
  the early interchanges are the most useful. The distance plateaus at
  0.07, which is less than optimal; as shown in
  Section~\ref{sec:algo_poisson} on Poisson disk sampling, we get
  roughly 100 samples in 2D when we set the disk radius to 0.08.}
\label{fig:samples_lhs1}
\end{figure}

\begin{figure}[!htb]
\centering
\begin{tabular}{ccc}
\includegraphics[trim = 2.5cm 0cm 2.5cm 0cm, clip = true,width=0.16\textwidth]{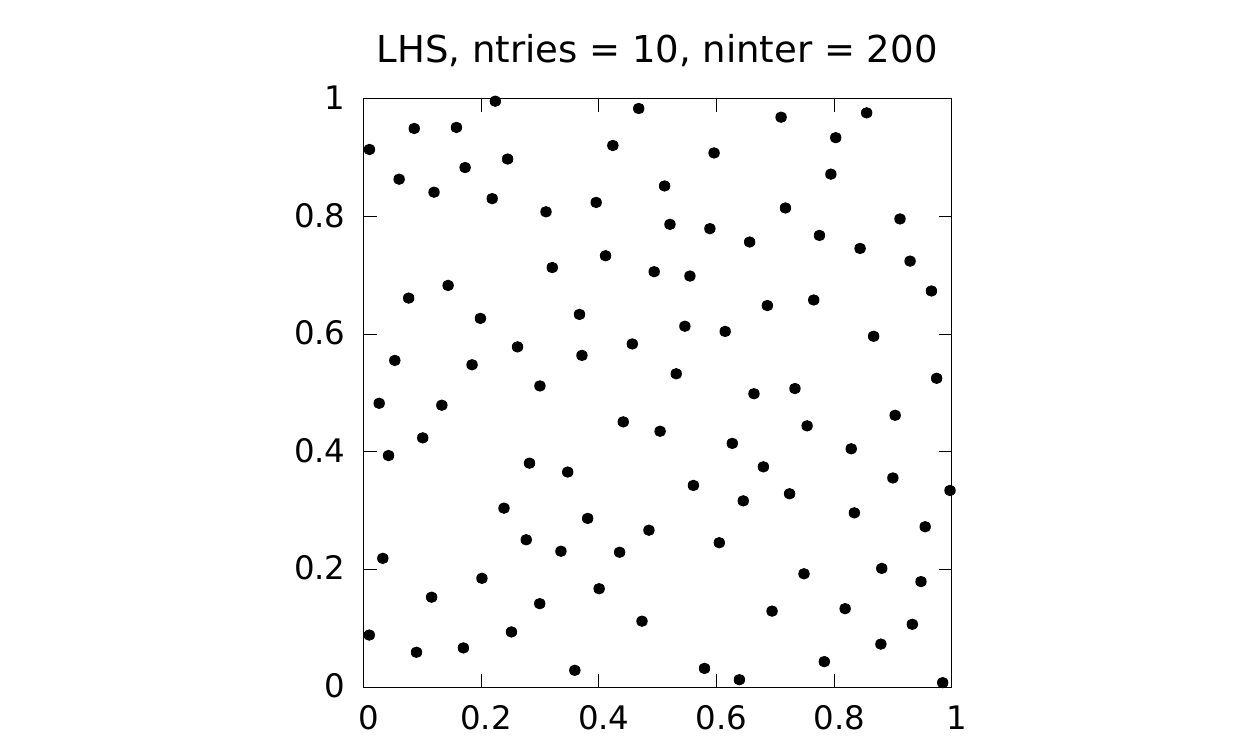} &
\includegraphics[trim = 2.5cm 0cm 2.5cm 0cm, clip = true,width=0.16\textwidth]{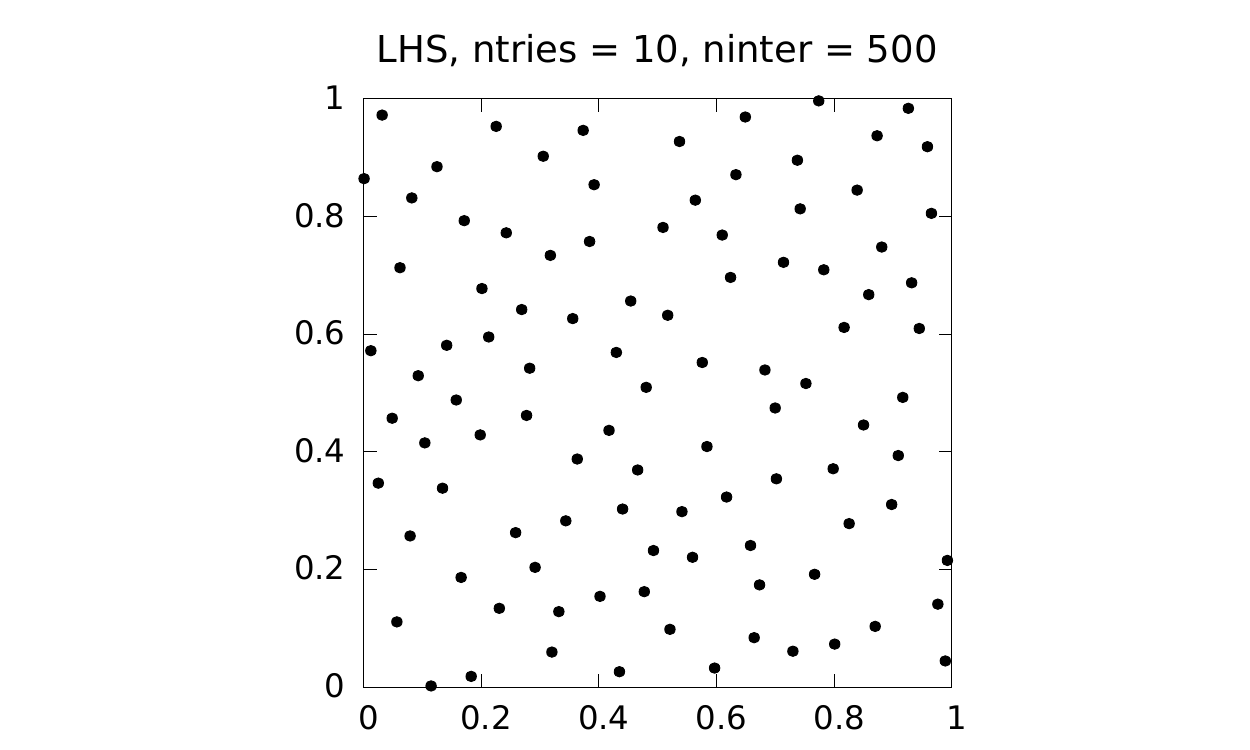} & 
\includegraphics[trim = 2.5cm 0cm 2.5cm 0cm, clip = true,width=0.16\textwidth]{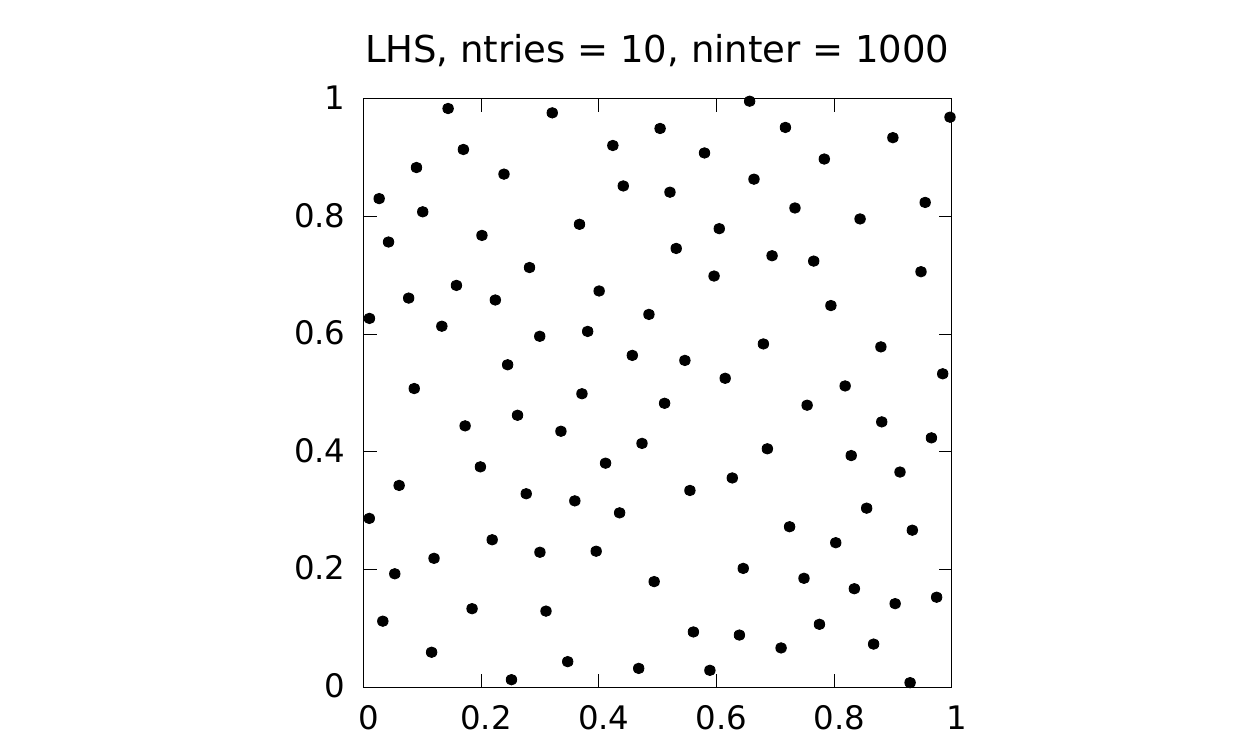} \\
\includegraphics[trim = 2.5cm 0cm 2.5cm 0cm, clip = true,width=0.16\textwidth]{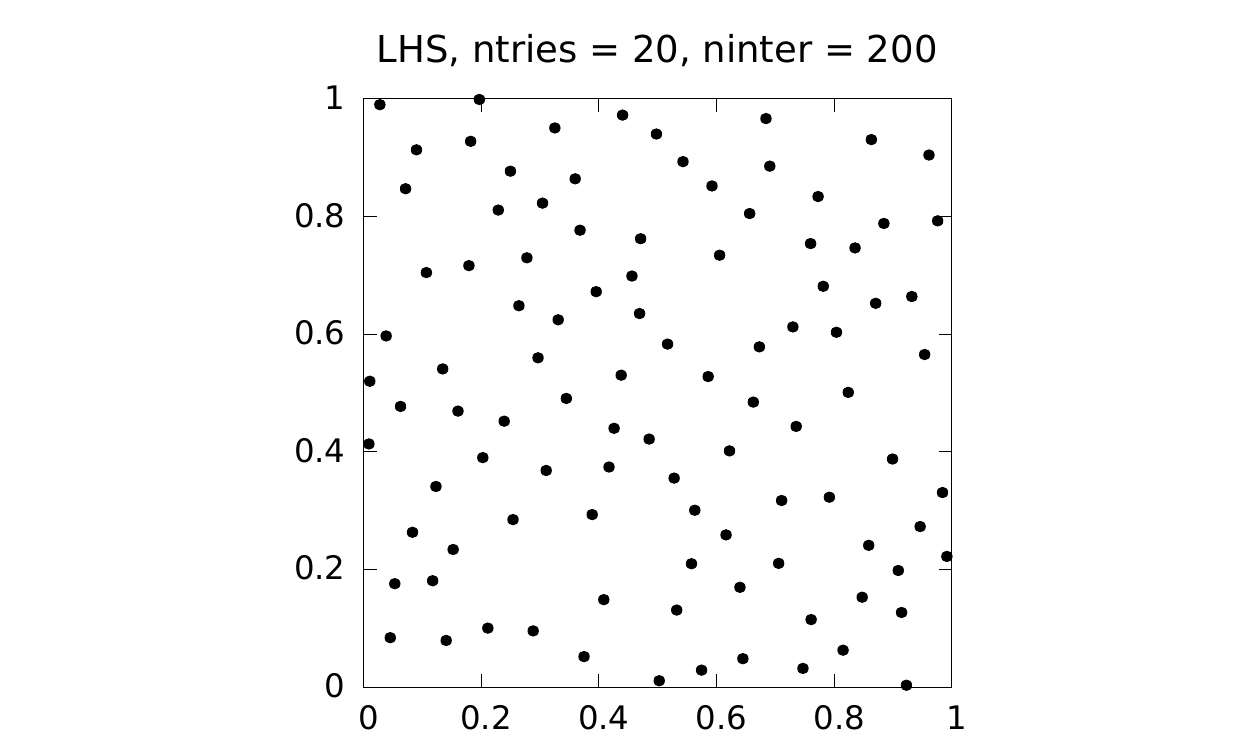} &
\includegraphics[trim = 2.5cm 0cm 2.5cm 0cm, clip = true,width=0.16\textwidth]{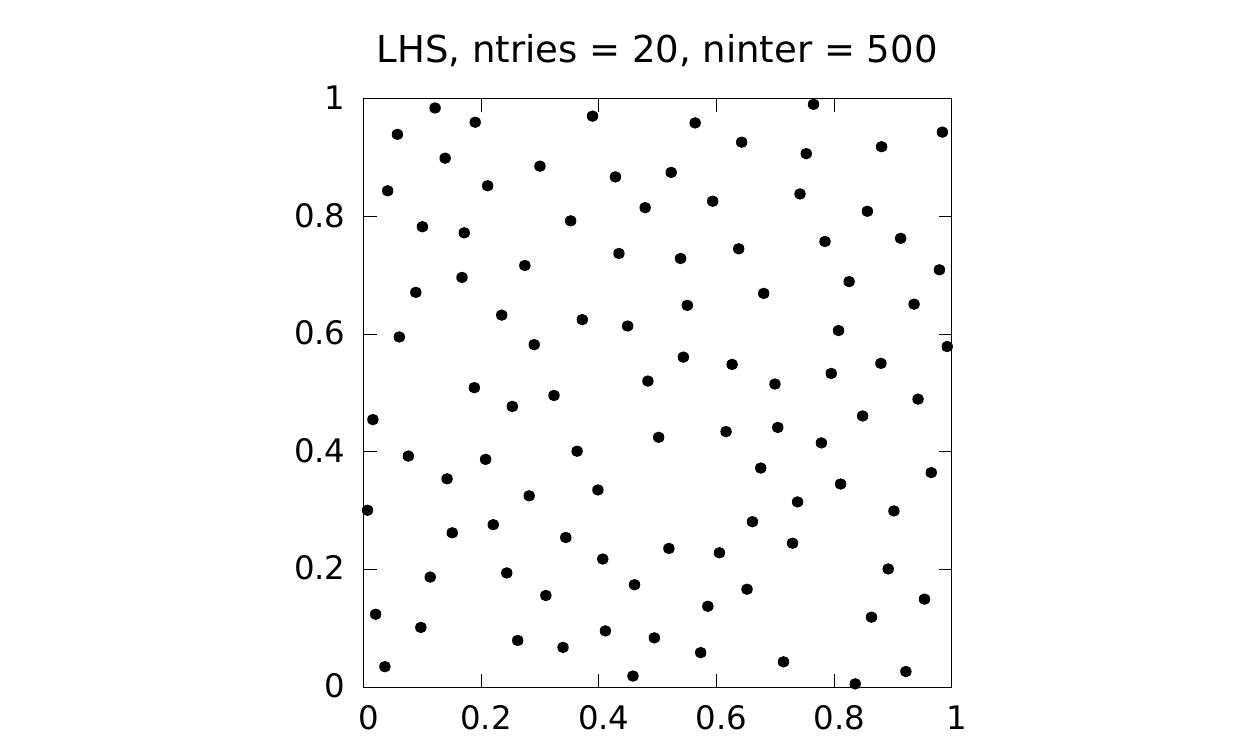} & 
\includegraphics[trim = 2.5cm 0cm 2.5cm 0cm, clip = true,width=0.16\textwidth]{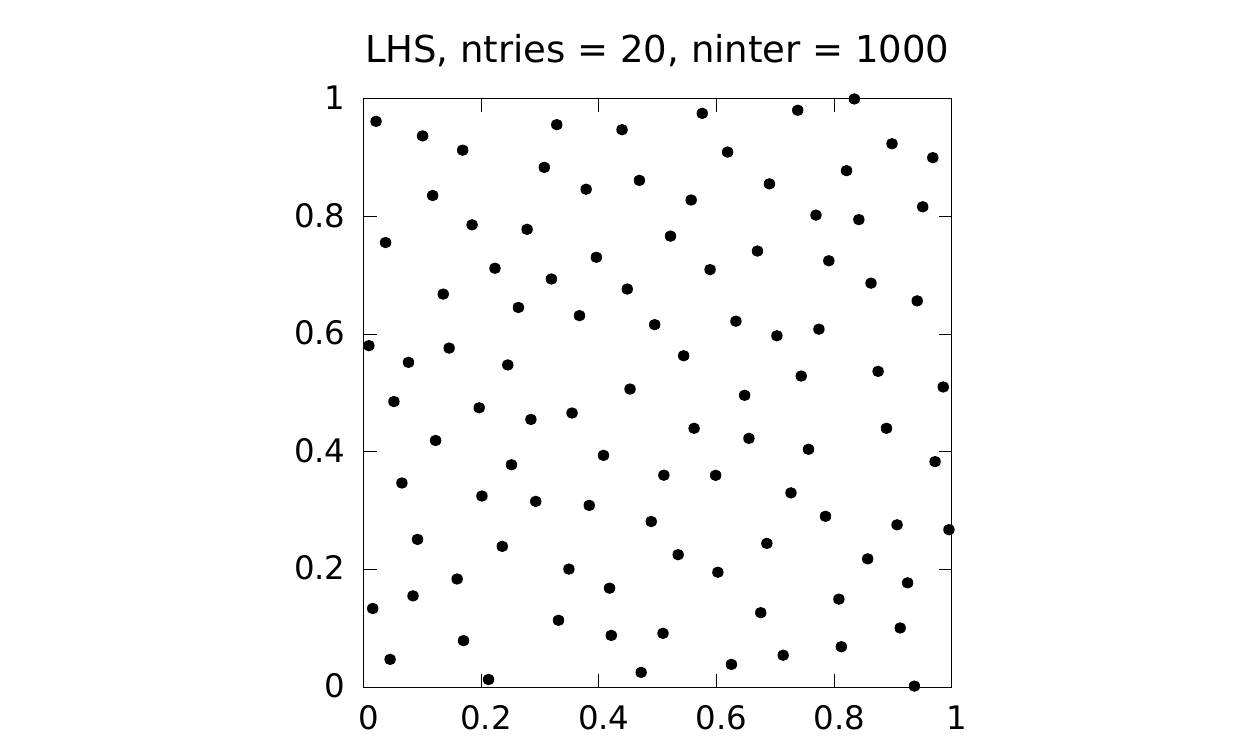} \\
(a) & (b) & (c) \\
\end{tabular}
\vspace{-0.2cm}
\caption{Effect of parameters in the Approximate Maximin LHS algorithm
  (Algorithm~\ref{algo:lhs_opt}): Top row - best of 10 tries; bottom
  row - best of 20 tries. 100 samples after (a) 200, (b) 500, and (c)
  1000 interchanges. For a given number of interchanges, increasing
  the number of tries has minimal effect on the quality of the
  sampling.}
\label{fig:samples_lhs2}
\end{figure}

Figure~\ref{fig:samples_lhs1} shows an initial LHS of 100 samples, as
well as the sampling after 100, 1000, and 2000 interchanges. We found
that the initial interchanges increase the minimum distance, with
diminishing returns as the number of interchanges increases (see
Figure~\ref{fig:samples_lhs1}(e)). To avoid generating a poor-quality
LHS resulting from a poor initial sample set that is difficult to
improve with a limited number of interchanges, we repeat the algorithm
for a number of tries, each time starting with a new LHS. As shown in
Figure~\ref{fig:samples_lhs2}, for a given number of interchanges,
increasing the number of tries has little effect on the quality of the
sampling.

%
\subsubsection{Latinizing a sample set}
\label{sec:algo_latinizing}
%

A typical approach to generating a good quality LHS is to start with a
sampling that has the Latin property and then use interchanges, which
maintains this property, to improve the quality of the sampling.
However, the idea of {\it Latinization of point sets}, introduced in
conjunction with the centroidal Voronoi
tessellation~\cite{saka2007:latinized}, allows us to first generate a
sampling with desired properties and then ``Latinize'' the set of
points, instead of the converse which involves an expensive
optimization.

\begin{algorithm}
\caption{Latinizing a sampling, adapted from~\cite{saka2007:latinized} }
\label{algo:latinizing}
\begin{algorithmic}[1]

  \STATE Goal: given $N$ samples in a $d$-dimensional space, transform
  the sampling into one that satisfies the Latin hypercube property.
  
  \FOR{$j = 1$ to $d$}

  \STATE Reorder the samples in increasing order of the values in the
  $j$-th dimension, with ties arbitrarily broken.

  \STATE Divide the range of values in the $j$-th dimension into $N$
  equispaced bins.

  \FOR{$i = 1$ to $N$}

  \STATE If the value in row $i$, column $j$ of the reordered samples
  is not in bin $i$, then shift it to a random position in bin $i$.

  \ENDFOR

\ENDFOR

\end{algorithmic}
\end{algorithm}

Algorithm~\ref{algo:latinizing} describes the simple procedure of
Latinization, where the sample values in each dimension are first
re-ordered and then shifted so that there is a value in each bin in
that dimension. Note that this implementation reorders the samples,
but if we associate a key with each sample point, the original order
of samples can be restored at the end, if necessary.  Typically, when
a random sample set is space filling, the shifts required to Latinize
the set of points are relatively small, as shown in
Figure~\ref{fig:samples_latinize} in
Section~\ref{sec:adapt_noncollapsing}.
Algorithm~\ref{algo:latinizing} is thus a simple way to introduce the
Latin property into sample sets as required.

%
\subsection{Centroidal Voronoi tessellation (CVT) sampling}
\label{sec:algo_cvt}
%

Generating a random, space-filling sampling using a centroidal Voronoi
tessellation (CVT)~\cite{du1999:cvt} is an idea common to many fields,
including applied mathematics, engineering, and computer science.  The
technique is closely connected to the Delaunay triangulation used for
mesh generation in scientific
computing~\cite{shewchuk2012:delaunaynotes} and the k-means algorithm
used for clustering in data mining~\cite{morissette2013:kmeans}. Given
a set of points, called the generators, and a distance function, a
Voronoi tessellation is the subdivision of another set of points into
subsets such that the points in a subset are closer to one of the
generators than any other generators~\cite{ju2002:parallelcvt}.  When
the generators coincide with the center of mass of each subset, the
tessellation is referred to as a centroidal Voronoi tessellation as
the generators are the centroids of each Voronoi cell, which is the
region defined by the points in the subset assigned to that generator.

The very definition of a CVT suggests that to generate $N$ samples
with the space-filling property, we can first generate a large number
of randomly distributed points, enough to cover the domain densely,
and then identify the $N$ generators of these points as the CVT
sampling. The initial locations of the generators are selected
randomly and iteratively refined until they become the true generators
for the CVT of the domain.  The iterations are stopped either when the
locations of the generators stabilize or a maximum number of
iterations is reached.  By construction, CVT generates a set, not a
sequence of samples, that is, all samples are generated at the same
time. We also observe that selecting the initial locations of the
generators is, in itself, a sampling problem, a common solution to
which is discussed in Section~\ref{sec:algo_fps}.

There are many algorithms to generate a CVT. We implemented a
probabilistic algorithm (Algorithm 3 from~\cite{ju2002:parallelcvt})
that also allows the $N$ samples to be generated using a non-uniform
density of points, as discussed n Section~\ref{sec:adapt_pdf}. For the
algorithm summarized in Algorithm~\ref{algo:cvt}, we use $\alpha_1 =
\beta_1 = 0$ and $\alpha_2 = \beta_2 = 1$, which, in step 9,
essentially replaces each centroid with the mean of the random points
assigned to it. The randomness of the points generated in Step 6 makes
the mean a probabilistic approximation to the centroid; a larger
number of points will result in a better approximation.

\begin{algorithm}
  \caption{Centroidal Voronoi tessellation, adapted from
    ~\cite{ju2002:parallelcvt}(Algorithm 3). }
\label{algo:cvt}
\begin{algorithmic}[1]

  \STATE Goal: generate $N$ samples in a $d$-dimensional space that
  are the centroids of a centroidal Voronoi tessellation over the
  space, with a distribution defined by the probability density
  function, $ \rho({\bf x})$, at any point $\bf x$ in the domain.
  
  \STATE Set $ niter$, the number of iterations and $PPI$, the number
  of random points to generate per iteration.

  \STATE Set the parameters $\alpha_1, \alpha_2, \beta_1$, and
  $\beta_2$ such that $\alpha_2 > 0$, $\beta_2 > 0$, $\alpha_1 +
  \alpha_2 = 1$ and $\beta_1 + \beta_2 = 1$. These determine how the
  centroids are updated in each iteration.

  \STATE Choose an initial set of $N$ samples, ${\bf x}_i, i = 1,
  \ldots, N$ at random in the domain as the generators or
  the centroids.  Set $m_i = 1, i = 1, \ldots, N$, which indicates the
  number of times each centroid is updated.

  \FOR{$iter = 1$ to $niter$}

  \STATE Generate $PPI$ number of samples, ${\bf y}_k, k = 1, \ldots,
  PPI$, randomly in the $d$-dimensional domain  according to the
  probability density function $ \rho({\bf x})$.

  \FOR{$i = 1$ to $N$}

  \STATE Collect the set $W_i$ of all points ${\bf y}_k$ that are
  closest to ${\bf x}_i$ and compute their mean ${\bf u}_i$.

  \STATE If the set $W_i$ is empty, do nothing, else update the
  centroid ${\bf x}_i$ with a weighted average of the old value and the mean of
  the random points assigned to the centroid:
  \begin{equation}
    {\bf x}_i = \frac{(\alpha_1 m_i + \beta_1){\bf x}_i + (\alpha_2 m_i + \beta_2){\bf u}_i }{m_i+1}  \qquad , \qquad m_i = m_i + 1  \notag
  \end{equation}

  \ENDFOR

  \STATE If a convergence criterion, such as the maximum change in the
  centroids from a previous $iter$ is less than a threshold, is
  satisfied, terminate.

\ENDFOR

\end{algorithmic}
\end{algorithm}

\begin{figure}[htb]
\centering
\begin{tabular}{cccc}
\includegraphics[trim = 2.8cm 0cm 2.8cm 0cm, clip = true,width=0.16\textwidth]{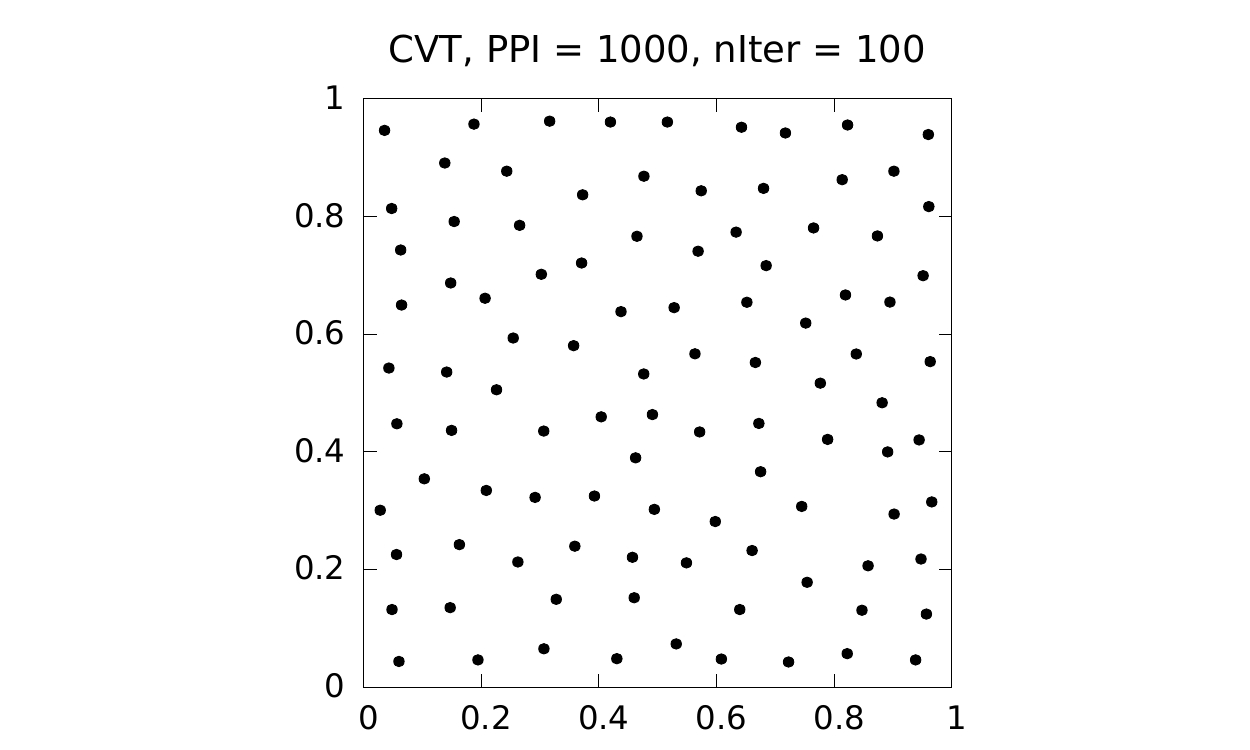} &
\includegraphics[trim = 2.8cm 0cm 2.8cm 0cm, clip = true,width=0.16\textwidth]{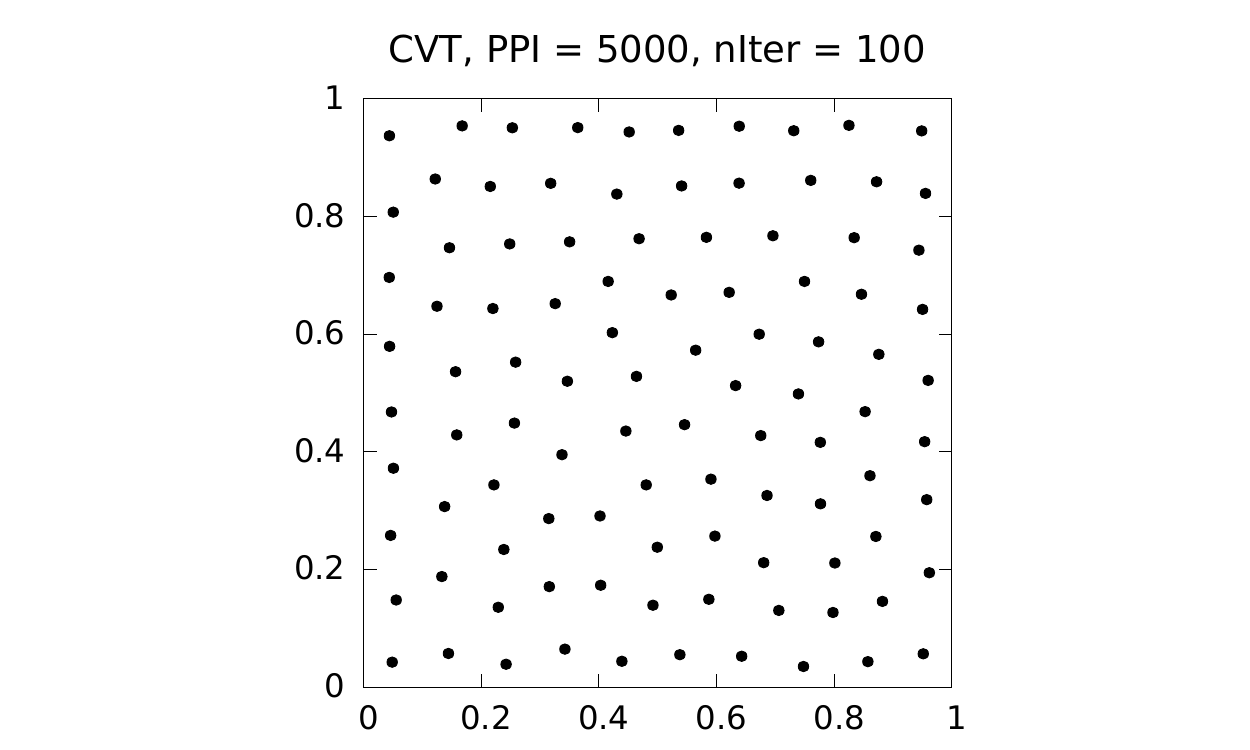} & 
\includegraphics[trim = 2.8cm 0cm 2.7cm 0cm, clip = true,width=0.16\textwidth]{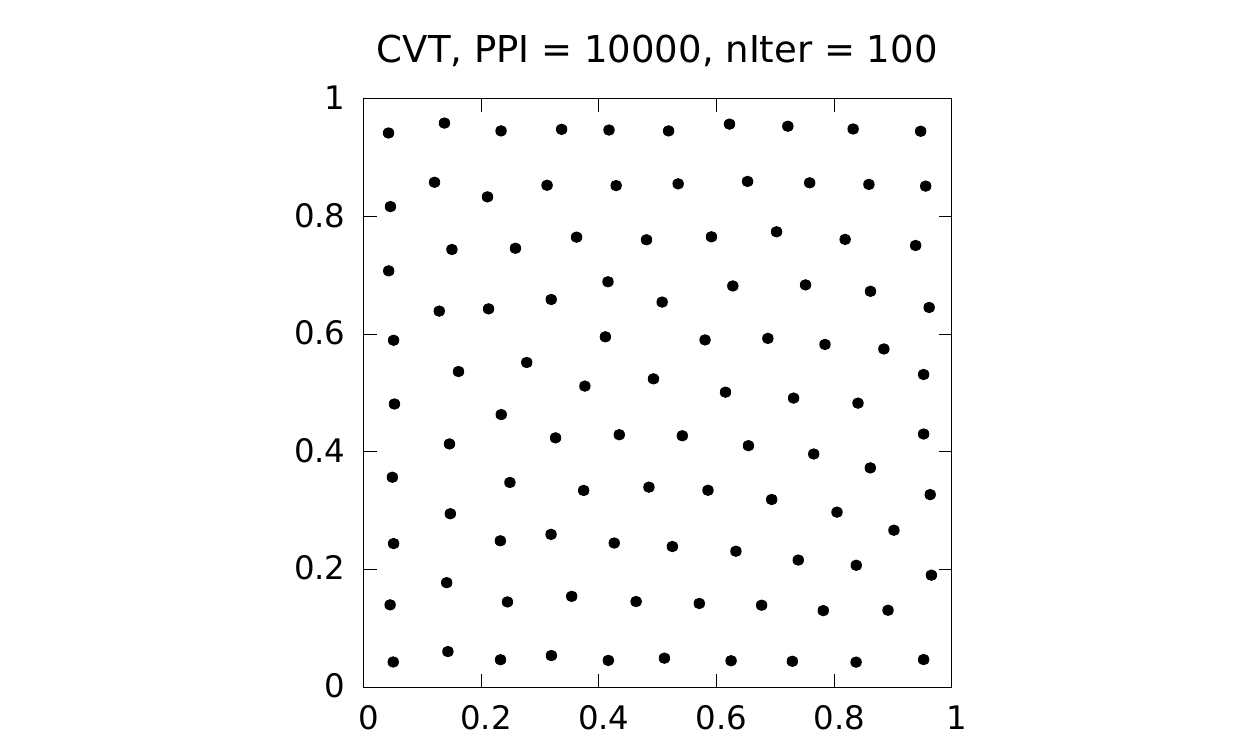} & 
\includegraphics[trim = 2.8cm 0cm 2.7cm 0cm, clip = true,width=0.16\textwidth]{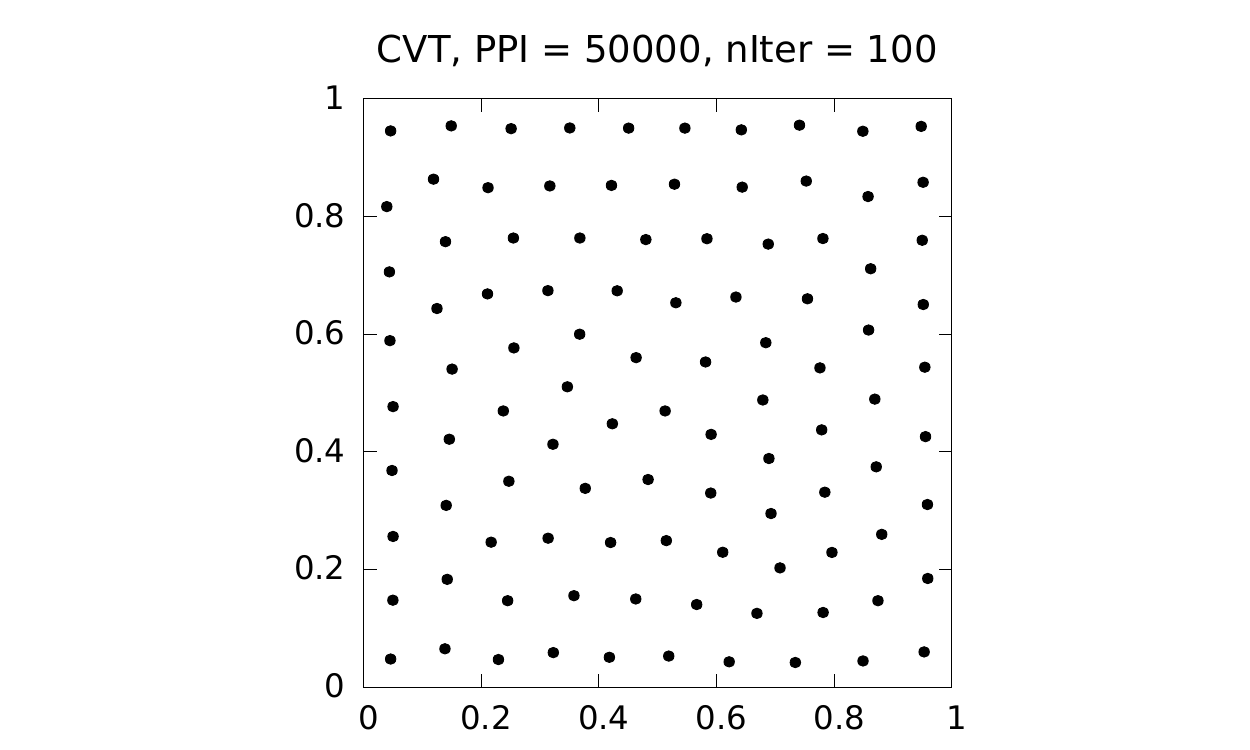} \\
\includegraphics[trim = 2.8cm 0cm 2.8cm 0cm, clip = true,width=0.16\textwidth]{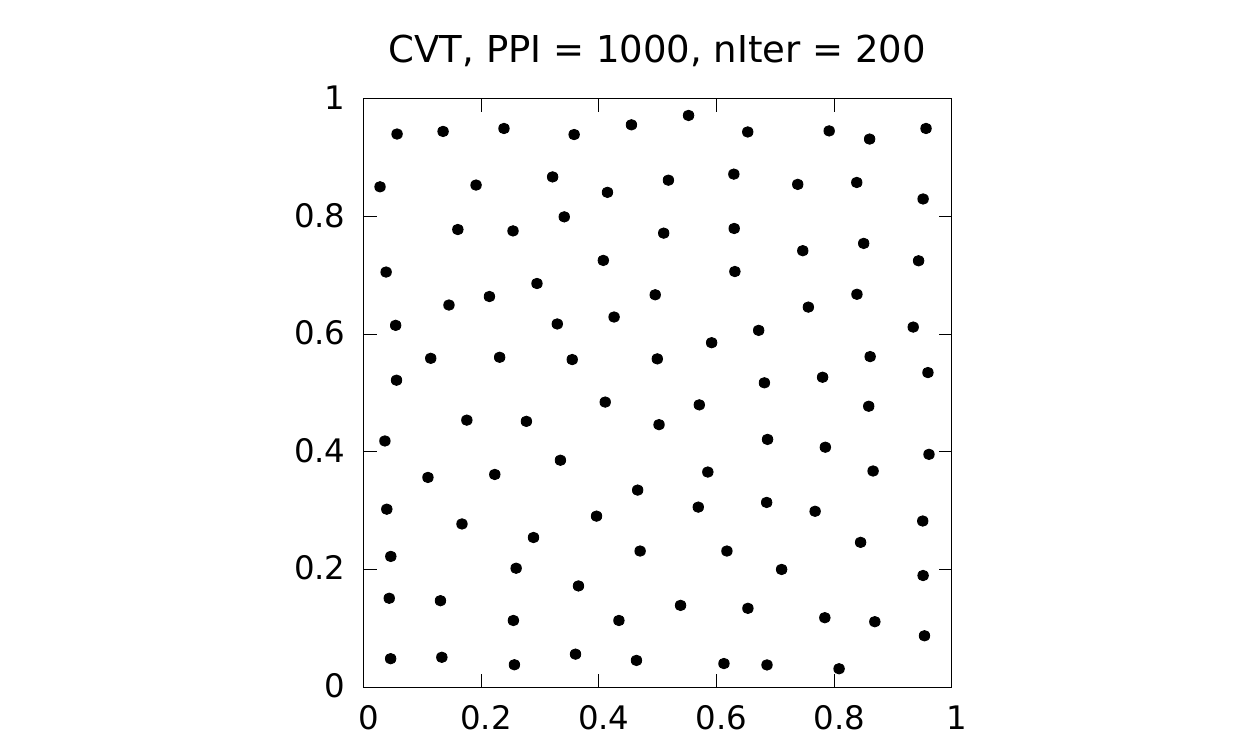} &
\includegraphics[trim = 2.8cm 0cm 2.8cm 0cm, clip = true,width=0.16\textwidth]{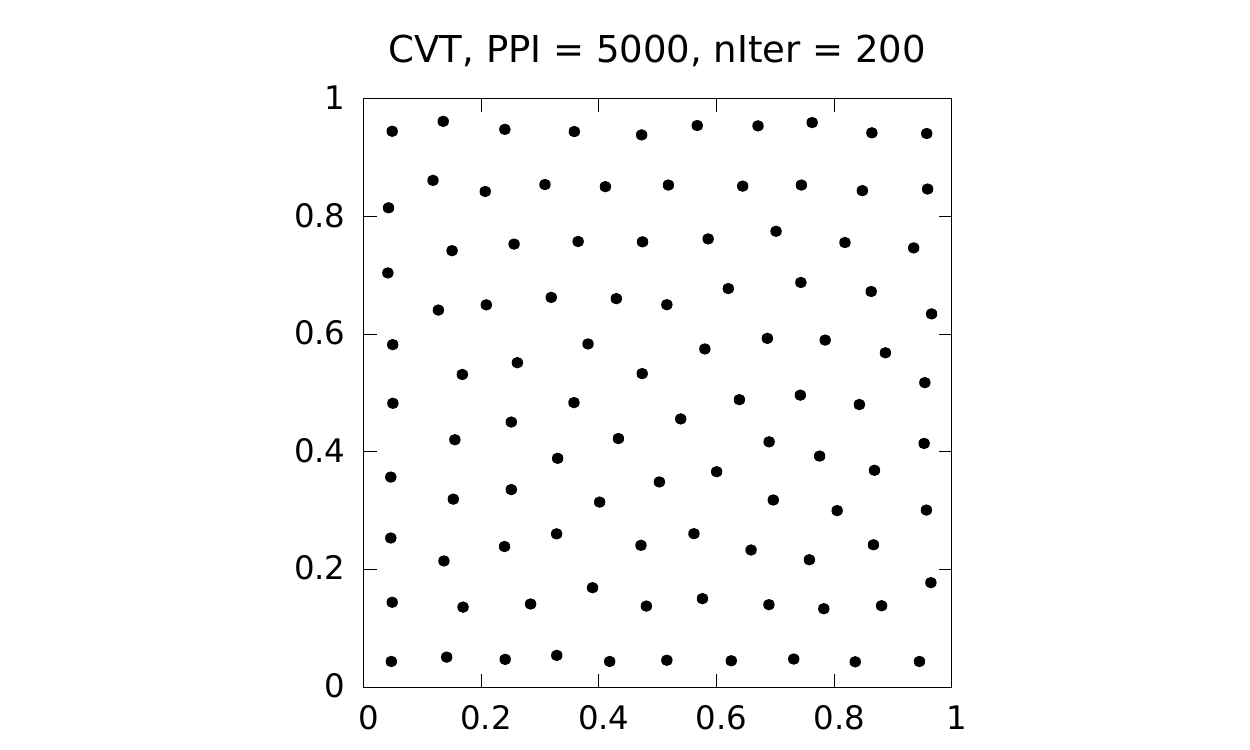} & 
\includegraphics[trim = 2.8cm 0cm 2.7cm 0cm, clip = true,width=0.16\textwidth]{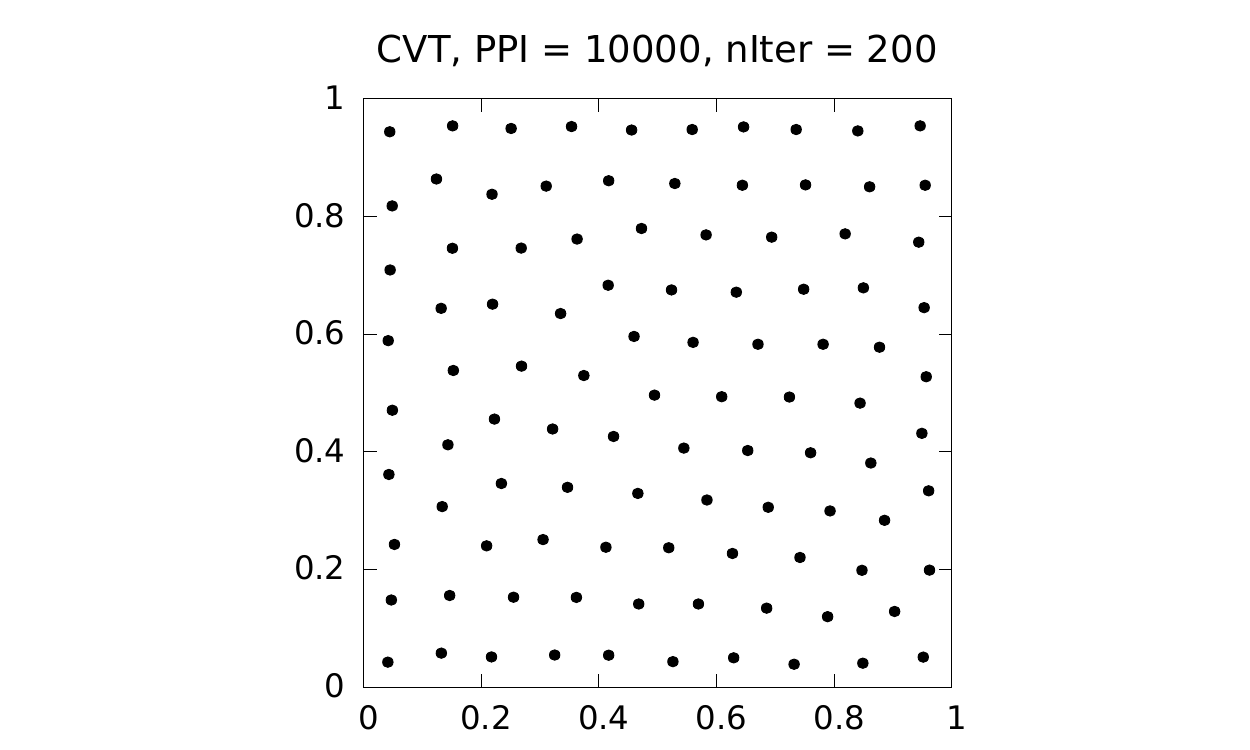} & 
\includegraphics[trim = 2.8cm 0cm 2.7cm 0cm, clip = true,width=0.16\textwidth]{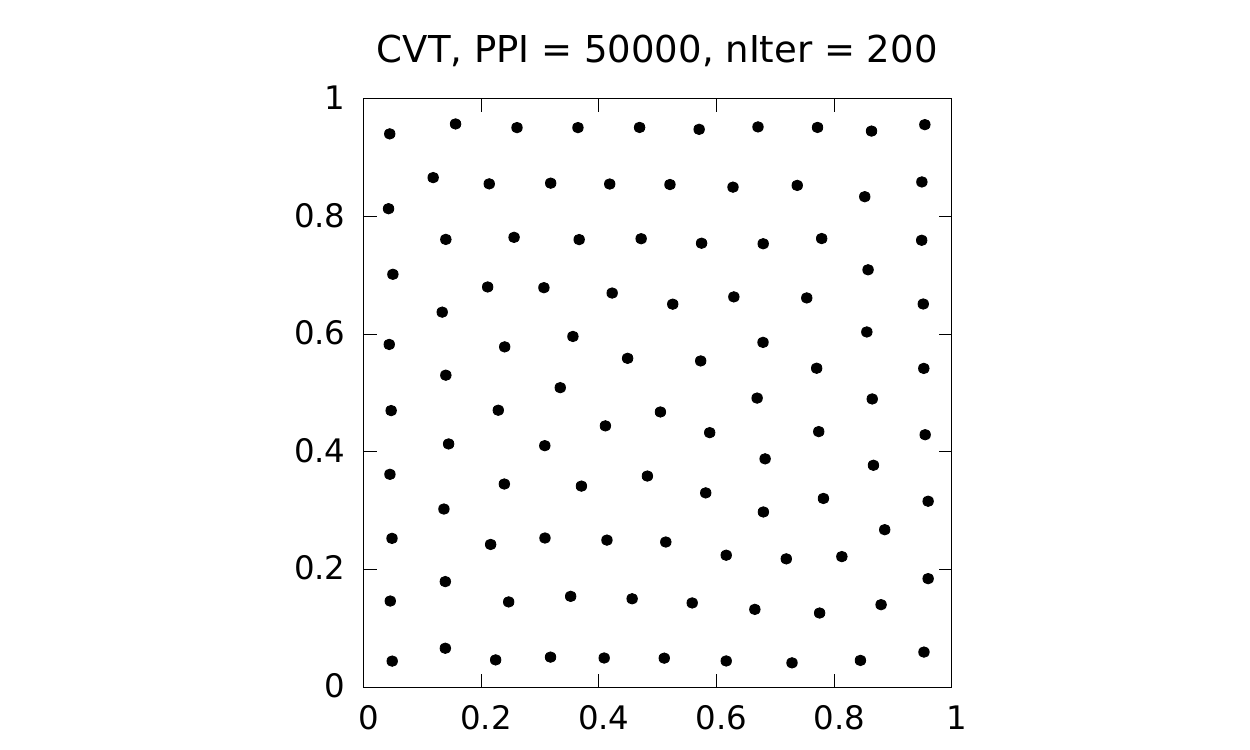} \\
(a) & (b) & (c) & (d)\\
\end{tabular}
\vspace{-0.2cm}
\caption{Centroidal Voronoi tessellation sampling: 100 samples
  generated using (a) 1000 (b) 5000 (c) 10000; and (d) 50000 points
  per iteration ($PPI$). Top row - 100 iterations; bottom row - 200
  iterations. As $PPI$ increases, the samples
  become more ordered, with alignment along the boundaries seen as
  early as 5000 points per iteration, but increasing the number of
  iterations has less of an affect on the space filling property.}
\label{fig:samples_cvt}
\end{figure}

Figure~\ref{fig:samples_cvt} shows the effect of the parameters
$niter$ (the number of iterations) and $PPI$ (the number of
points per iteration) on the quality of 100 samples generated in 
two dimensions. When $PPI$ is small, the points are more
random, with non-uniform distances between them. Increasing $PPI$
leads to a more ordered set, as can be seen by the points that are
aligned along the boundaries of the domain.  This regularity of the
CVT sample set can be addressed by Latinizing the samples
generated~\cite{saka2007:latinized} using
Algorithm~\ref{algo:latinizing}.  In addition, we observe that when
$PPI$, the points per iteration, is large, a larger number of
iterations does not improve the space-filling property.

%
\subsection{Poisson disk sampling}
\label{sec:algo_poisson}
%

Poisson disk sampling is a technique that originated in the fields of
image processing and graphics, where it was proposed as a solution to
the problem of aliasing. As described in the landmark paper by
Cook~\cite{cook1986:stochastic}, aliasing arises because the size of
the discrete image pixels, which are regularly spaced, determines an
upper limit, called the Nyquist limit, to the frequencies that can be
displayed.  As a result, when we attempt to display frequencies
greater than this limit, we see artifacts such as Moir\'e patterns or
jagged edges of objects. An anti-aliasing approach is to sample the
data at appropriate non-uniformly spaced locations rather than at
regularly spaced locations like a
grid~\cite{mitchell1987:antialiased}. As described earlier, such
non-regularly spaced locations are also preferred in design of
computational experiments~\cite{joseph2016:spacefilling} and
hyperparameter optimization~\cite{bergstra2012:hyperparam}, driven not
by the need for anti-aliasing, but by the desire to make better use of
samples when one (or more) dimensions in a high-dimensional domain is
less important than others.

The Poisson disk algorithm was motivated by the distribution of
photoreceptors in the human eye~\cite{cook1986:stochastic}.  Despite
the limited number of photoreceptors, our eyes are not prone to
aliasing, suggesting that a similar distribution of samples could
alleviate aliasing effects.  The photoreceptors follow a Poisson disk
distribution, wherein randomly placed samples are no closer than a
certain distance, which is the radius of the disk.  The Fourier
spectrum of this distribution is in the ``blue'' spectrum, with a
spike at the origin for the DC component, and noise beyond the Nyquist
limit; the high frequency components appear as noise, and as our eyes
are less sensitive to noise than to aliasing, the appearance of
artifacts in the image is minimized. Such a sampling, also referred
to as ``blue noise'' sampling, is desirable in many applications.

There are many exact and approximate solutions to generating a Poisson
disk distribution. Cook~\cite{cook1986:stochastic} proposed a simple
dart throwing algorithm, where random points are generated
sequentially and if they are too close to existing samples, they are
discarded; if not, they are accepted as a new sample.  We consider the
version of Poisson disk sampling by Bridson~\cite{bridson07:sampling}
described in Algorithm~\ref{algo:poisson_disk}. Starting with a
randomly selected sample in the domain, we maintain a list of active
samples. A sample is moved from this list to the final-samples list if
it is not possible to insert a candidate sample around it that is more
than $r$, the disk radius, away from all the samples currently in the
final-samples list.  Otherwise, the candidate sample is appended to
the active-samples list as it indicates a region to grow around. For
each sample in the active-samples list, the number of candidates
evaluated is limited to a user-defined number $ncand$.

\begin{algorithm}
  \caption{Poisson disk sampling algorithm, adapted from~\cite{bridson07:sampling}}
\label{algo:poisson_disk}
\begin{algorithmic}[1]

  \STATE Goal: generate samples in a $d$-dimensional space that
  are at least a distance $r$  from each other.

  \STATE Set the values of $r$ - the minimum distance between samples
  and $ncand$ - the maximum number of candidates to consider around
  each active sample before moving this sample to the final samples list.

  \STATE Select the initial sample randomly in the domain and insert
  it into the active list of samples.

  \WHILE{{\it samples are present in active list}}

  \STATE Select a sample randomly from the active list 

  \FOR{$i = 1$ to $ncand$}

  \STATE Generate a candidate sample within a torus of radius between $r$ and
  $2r$ around the selected sample.

  \STATE If candidate sample is less than a distance $r$
  from the samples in the final list of samples, continue.

  \STATE If candidate sample is farther than distance $r$ from the
  samples in the final list of samples, add candidate sample to
  active list and exit {\bf for} loop.

  \ENDFOR

  \STATE If all $ncand$ candidate samples around the selected sample
  were within distance $r$ from the samples in the final list, move
  the selected sample from the active list to the final samples list.

  \ENDWHILE

\end{algorithmic}
\end{algorithm}

Algorithm~\ref{algo:poisson_disk} generates a sequence of samples
incrementally; however, as these samples are generated near existing
samples, we create an advancing front of samples that slowly grows
larger and larger to eventually cover the entire domain (see
Figure~\ref{fig:samples_inc}). To progressively add samples and fill
the domain with an increasing density of samples, we can vary the
radius $r$, starting with a larger value, and when no new samples can
be added, reducing it by a small factor, and continuing the
process~\cite{mccool1992:poisson}. Or, we could use the best-candidate
sampling~\cite{mitchell91:sampling}, which is discussed separately in
Section~\ref{sec:algo_bc} as it does not strictly enforce the
criterion that no two samples can be closer than a certain distance.

A major drawback of the Poisson disk algorithm for use in tasks such
as surrogate modeling or hyperparameter optimization is that the user
provides a minimum distance between samples, not a number of samples;
the relationship between the two is difficult to obtain in general for
a high-dimensional domain. Further, efficient techniques for
identifying candidates within a distance $r$ and $2r$ from a sample
are available only for two and three dimensions and cannot be easily
extended to higher dimensions~\cite{roberts2021:fastpoissondisk}.
As a result, Poisson disk sampling is used mainly in graphics or image
processing applications where the number of samples is not of primary
concern.

\begin{figure}[htb]
\centering
\begin{tabular}{ccc}
\includegraphics[trim = 2.8cm 0cm 2.8cm 0cm, clip = true,width=0.16\textwidth]{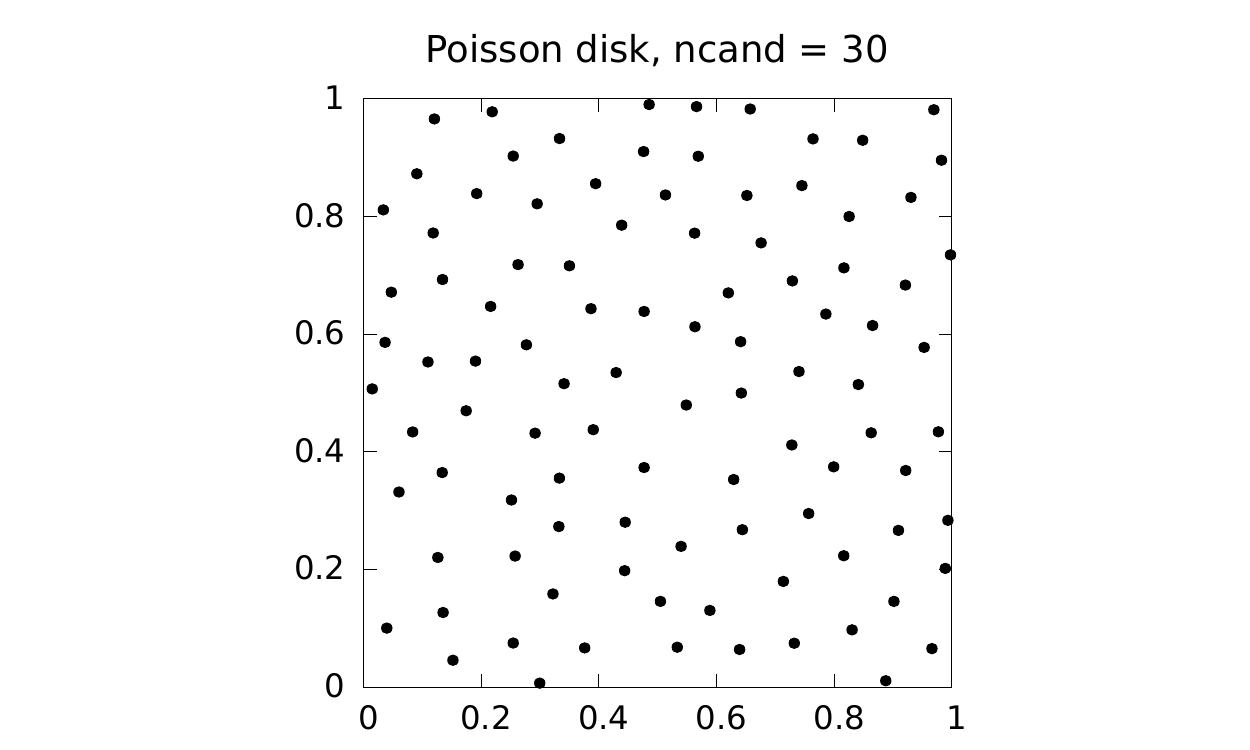} &
\includegraphics[trim = 2.8cm 0cm 2.8cm 0cm, clip = true,width=0.16\textwidth]{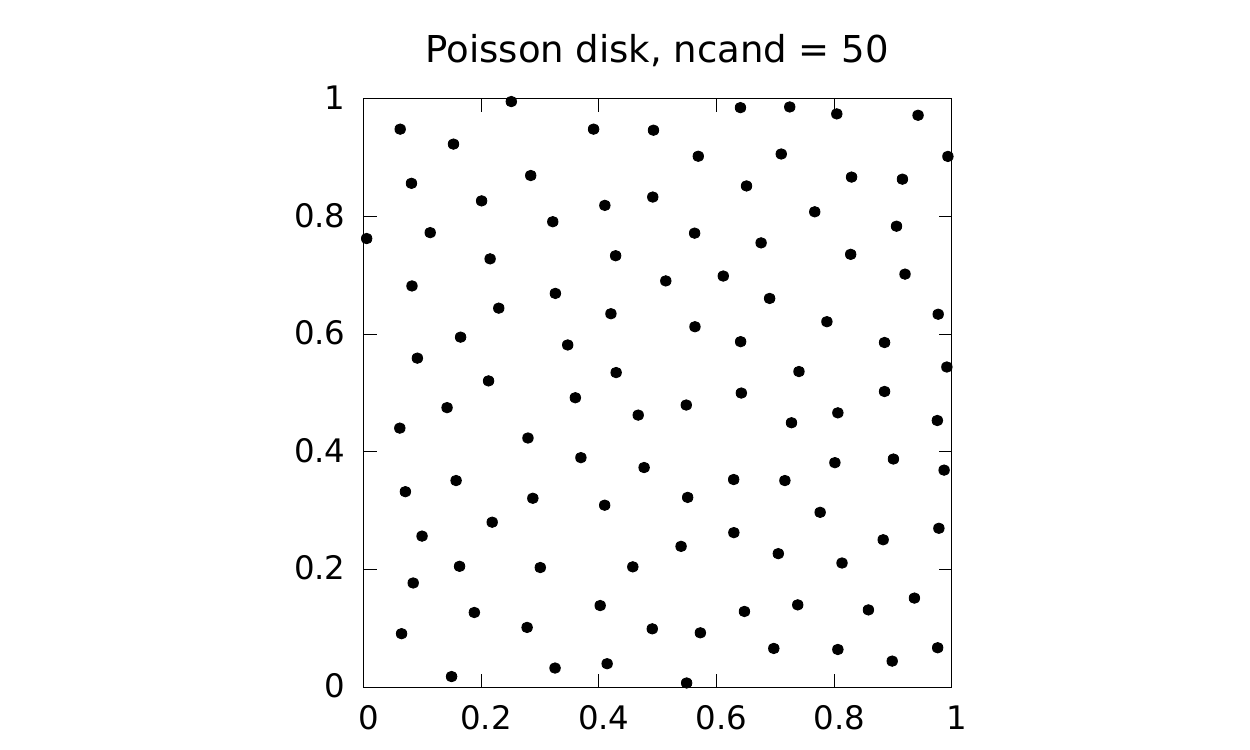} & 
\includegraphics[trim = 2.8cm 0cm 2.8cm 0cm, clip = true,width=0.16\textwidth]{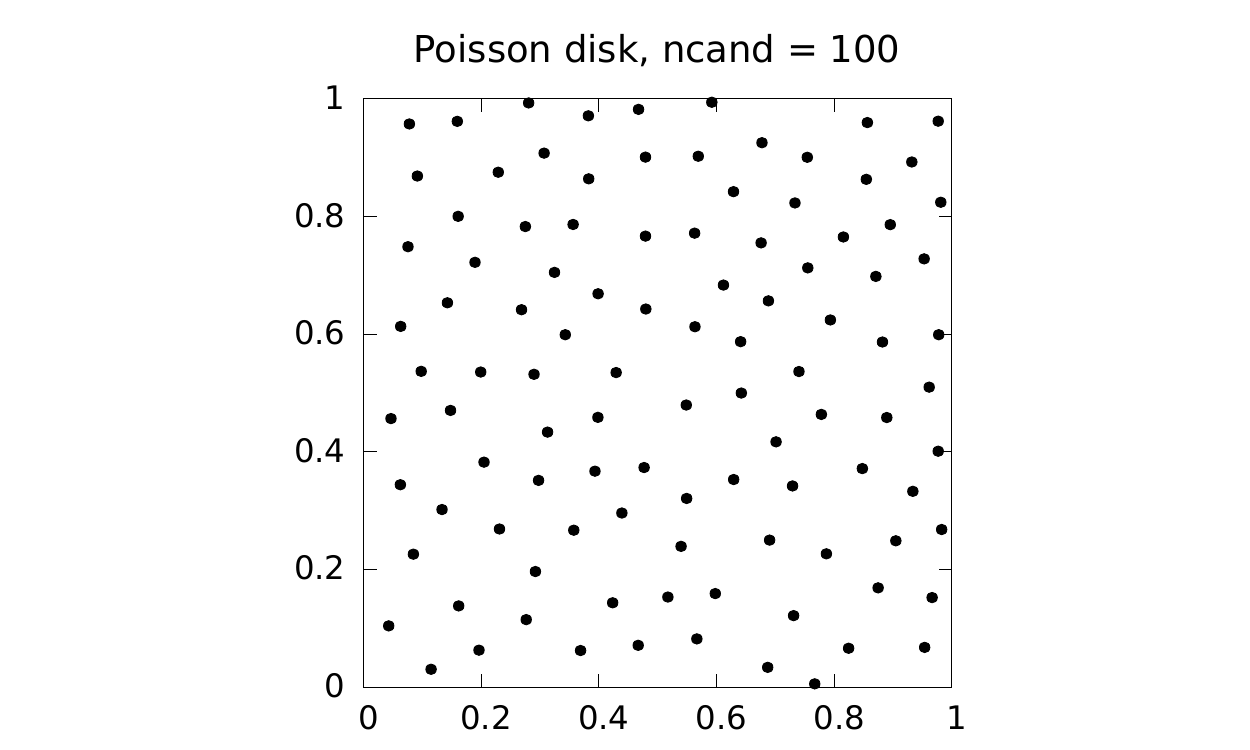} \\
(a) & (b) & (c) \\
\end{tabular}
\vspace{-0.2cm}
\caption{Poisson disk sampling: 100 samples generated using (a) 30;
  (b) 50; and (c) 100 candidates for each active point. The radius of
  the disk was set to 0.08, generating 99, 104, and 103 samples,
  respectively. For this example, the sampling quality appears
  independent of the number of candidates considered, provided it is
  not too small. }
\label{fig:samples_poisson}
\end{figure}

Figure~\ref{fig:samples_poisson} shows the locations of approximately
100 samples in a two-dimensional space as the number of candidates
$ncand$ is varied. The radius of the disk was determined by trial and
error such that it would result in close to 100 samples.  We observe
that a small value of $ncand = 30$ suffices for a good quality
sampling, as observed by Bridson~\cite{bridson07:sampling}.

%
\subsection{Farthest point sampling and the GreedyFP algorithm}
\label{sec:algo_fps}
%

The Farthest Point (FP) sampling algorithm, a contribution from the
image processing community, was proposed in 1997 by Eldar et
al.~\cite{eldar1997:farthest}. The original goal was a progressive
sub-sampling algorithm to generate a uniform approximation to an image
for situations where it took too long to transmit all the
data. A regular sub-sampling of the image on rectangular grids of
increasing resolution would have also generated a uniform
approximation progressively, but would have resulted in
aliasing effects discussed in Section~\ref{sec:algo_poisson}.

The basic idea in the FP sampling algorithm is to add a sample point
at a time to an existing set of samples and place it in the
least-known image area, that is, farthest from the existing set.
Eldar et al. proposed an efficient implementation using an incremental
version of the Voronoi diagram~\cite{aurenhammer1991:voronoi}.  They
proved that if we start with samples at the four corners of the image
and a randomly selected point, the next sample to be added must be at
one of the vertices of the bounded Voronoi diagram of the samples
selected thus far. By selecting the vertex that is farthest from the
current samples, we have a progressive algorithm that samples the
space with increasing density as more samples are generated. As the
number of samples does not have to be predetermined, this
space-filling algorithm is ideal for building a code surrogate
sequentially.

The FP sampling algorithm was proposed for images and two-dimensional
data, but given the challenges of constructing a Voronoi diagram, is
difficult to extend to higher dimensional spaces. However, the
construction of the algorithm suggests that we can create an
approximate FP sampling algorithm by starting with a large number of
random points, or candidates, in the domain, selecting the first one
randomly, and then selecting as the next sample the point that is
farthest from the samples already selected. This
greedy version of a farthest point algorithm is described in
Algorithm~\ref{algo:greedyfp}. It is similar to the idea proposed by
Gonzalez~\cite{gonzalez1985:clustering} to select the initial cluster
centers for a clustering algorithm that minimizes the maximum
inter-cluster distance. It is also related to the method proposed in
the k-means++ algorithm for selecting initial cluster
centers~\cite{arthur2007:kmeanspp},  as well as the farthest-first
traversal algorithm used in tasks such as the traveling salesman
problem.

\begin{algorithm}
  \caption{Greedy Farthest Point (GreedyFP) algorithm, adapted from~\cite{gonzalez1985:clustering}}
\label{algo:greedyfp}
\begin{algorithmic}[1]

  \STATE Goal: generate $N$ samples in a $d$-dimensional space that
  are far from each other.

  \STATE Set $ scale$, a multiplying factor that determines the number
  of random samples, or candidates.

  \STATE Generate $N * scale$ candidate samples in the $d$-dimensional
  domain randomly.

  \STATE Select the first sample randomly from the $N * scale$ candidate samples.

  \FOR{$i = 2$ to $N$}

  \STATE Select the $i$-th sample from the $N * scale$ candidate samples such
  that it is farthest from the $(i-1)$ samples selected thus far.

  \ENDFOR

\end{algorithmic}
\end{algorithm}

\begin{figure}[htb]
\centering
\begin{tabular}{ccccc}
\includegraphics[trim = 3.0cm 0cm 2.9cm 0cm, clip = true,width=0.16\textwidth]{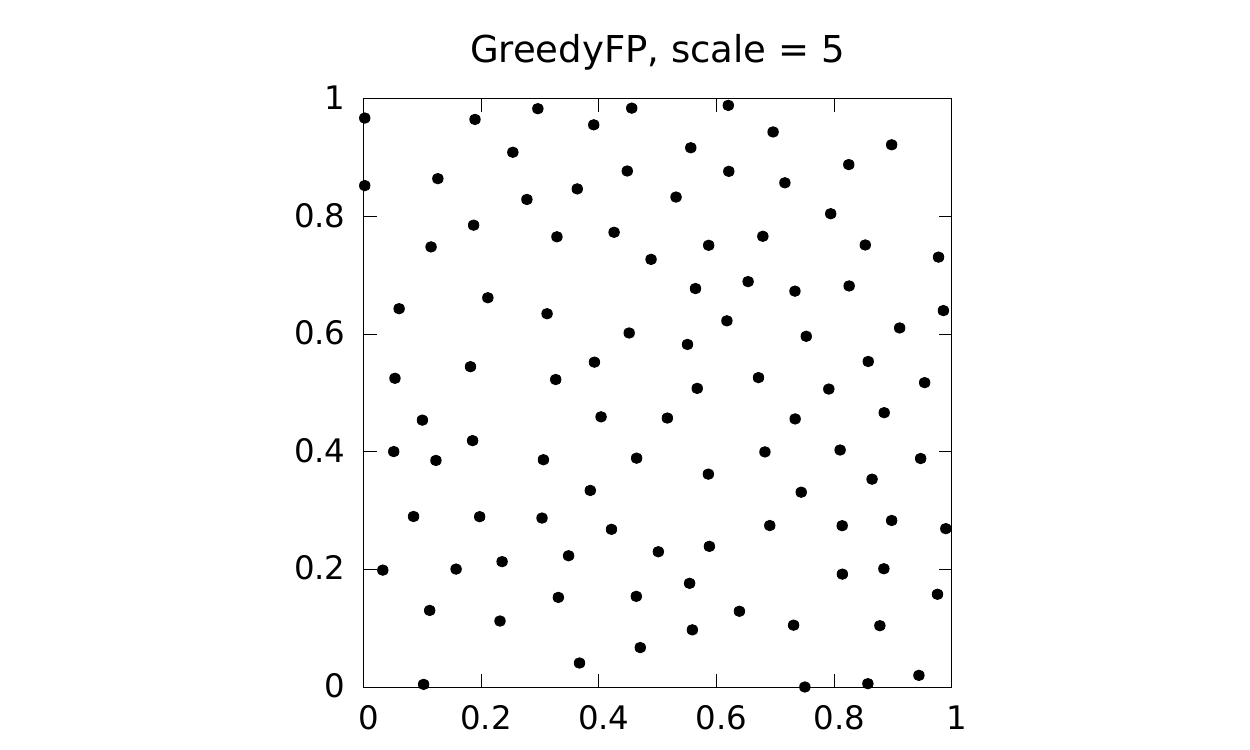} &
\includegraphics[trim = 3.0cm 0cm 2.9cm 0cm, clip = true,width=0.16\textwidth]{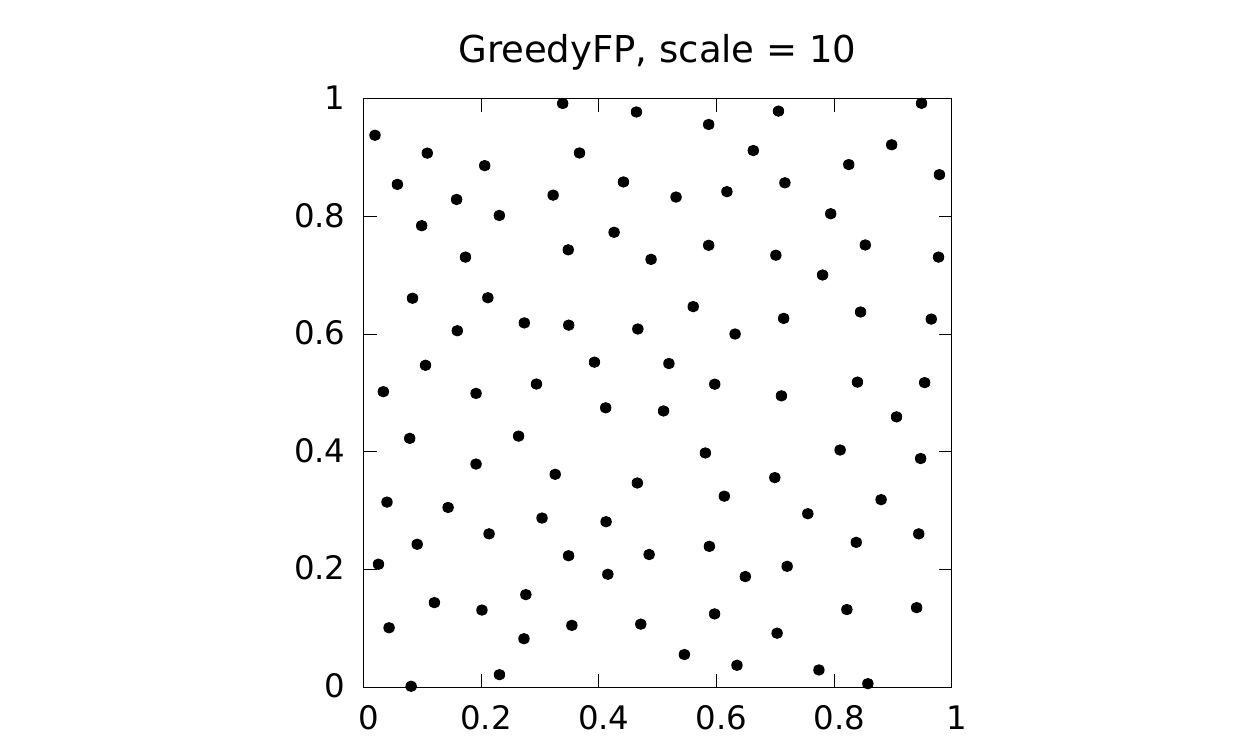} & 
\includegraphics[trim = 3.0cm 0cm 2.9cm 0cm, clip = true,width=0.16\textwidth]{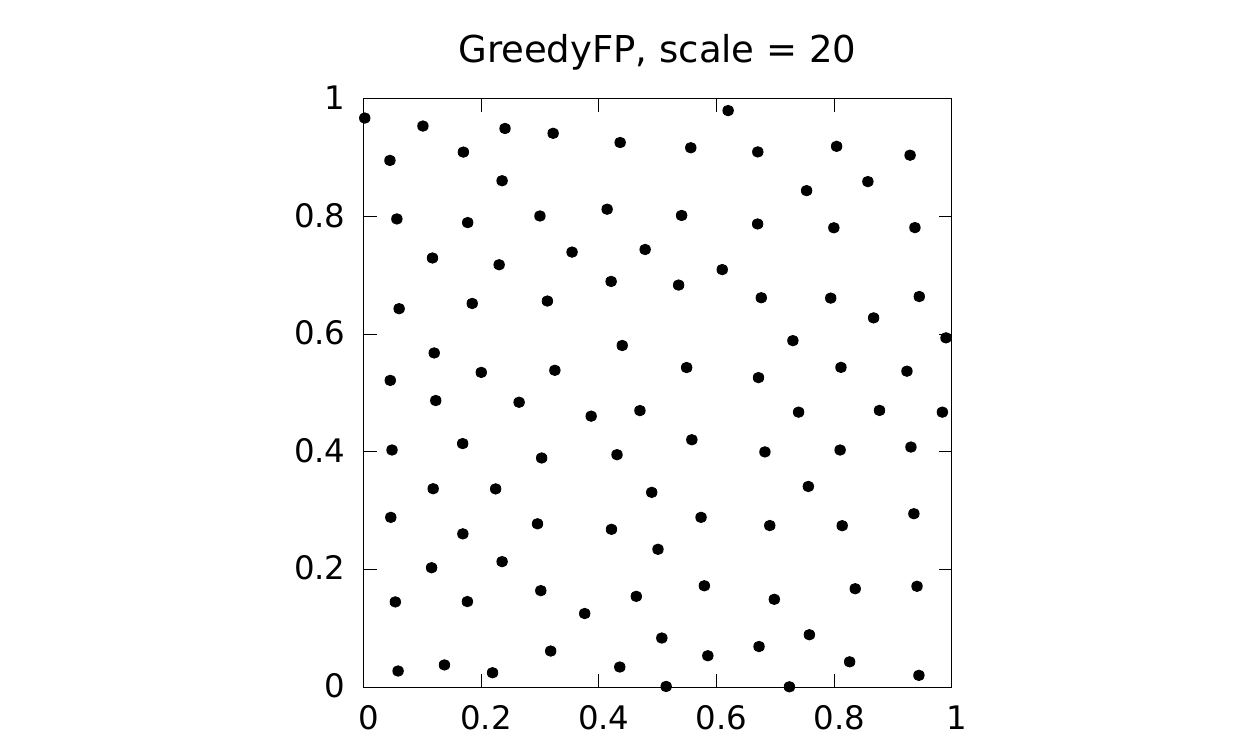} &
\includegraphics[trim = 3.0cm 0cm 2.9cm 0cm, clip = true,width=0.16\textwidth]{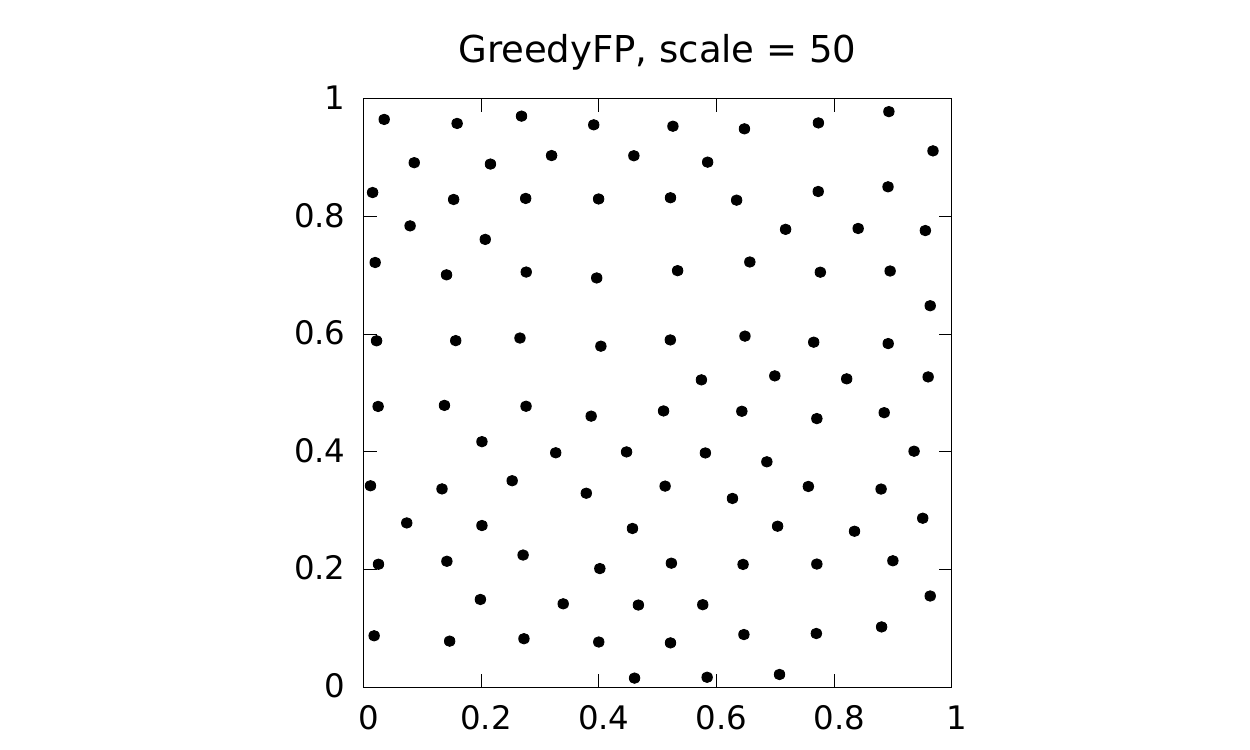} &
\includegraphics[trim = 3.0cm 0cm 2.9cm 0cm, clip = true,width=0.16\textwidth]{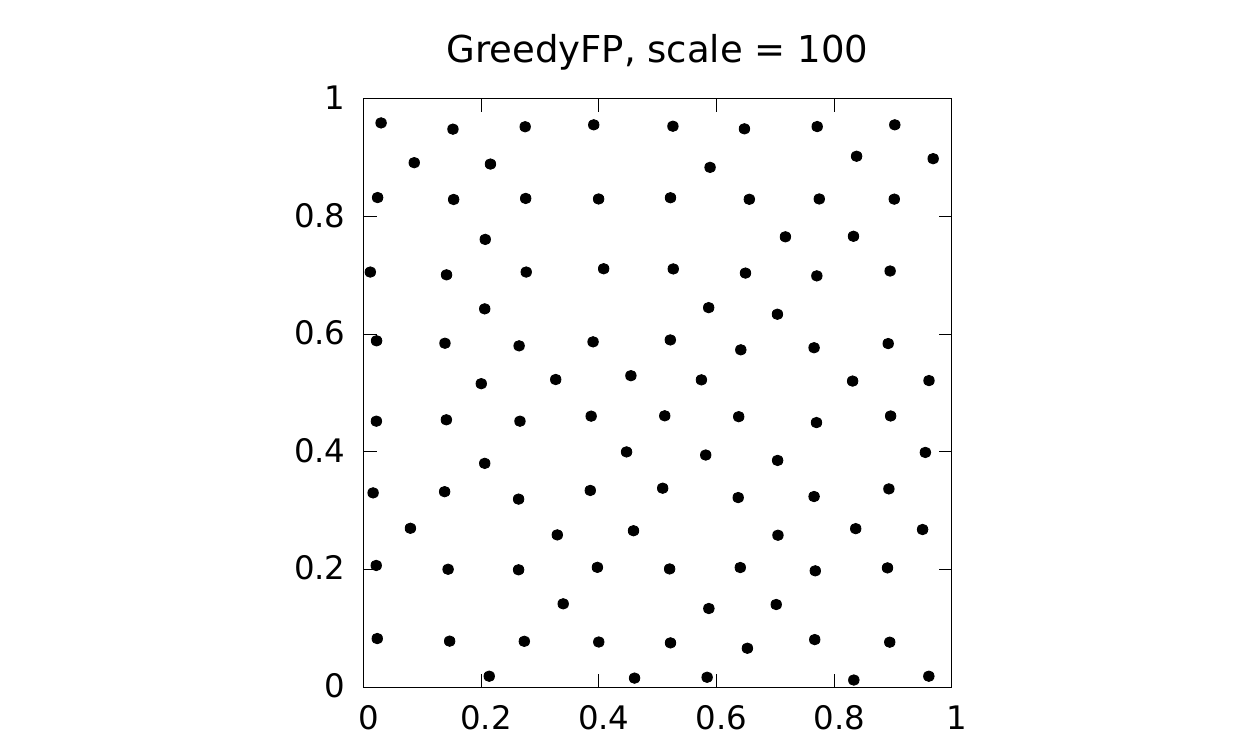} \\
(a) & (b) & (c) & (d) & (e) \\
\end{tabular}
\vspace{-0.2cm}
\caption{GreedyFP sampling: 100 samples generated using scale factor
  (a) 5; (b) 10; (c) 20; (d) 50; and (e) 100. Increasing the
  scale factor results in a more ordered, less random, sampling.}
\label{fig:samples_greedyfp}
\end{figure}

Figure~\ref{fig:samples_greedyfp} shows the 100 samples generated
using the GreedyFP algorithm in a two-dimensional space as the scale
factor is varied.  When $scale$ is small (5-10), the samples are
random in their location. However, as the candidates are fixed at the
start of the algorithm, if there is no candidate in a small sub-region,
there will not be a sample in the sub-region, possibly
resulting in under-sampling. 
As the $scale$ increases (50 and
higher), the samples are increasingly aligned. This is to be expected
as a very large number of candidates fills the domain completely, so 
candidates selected to maximize the distance between samples tend to
be aligned into a grid pattern.  There are also regions where the
spacing appears larger than elsewhere; these regions will eventually
contain samples as more samples are added. These observations suggest
that we should select a moderate value for $scale$ for a random, yet
space-filling sample set.

%
\subsection{Best candidate (BC) sampling}
\label{sec:algo_bc}
%

The best candidate (BC) algorithm, by
Mitchell~\cite{mitchell91:sampling}, was proposed in 1991 as an
approximate version of the Poisson disk sampling for use in graphics
and visualization. The algorithm generates a sequence of samples by
proposing a number of candidate samples and selecting the one farthest
away from the current set of samples. Thus, it can be also be
considered as an approximation to the farthest point
algorithm~\cite{eldar1997:farthest}.  However, unlike the
independently developed GreedyFP algorithm
(Section~\ref{sec:algo_fps}), where all candidate samples are
generated at the start, BC sampling generates a new set of candidates
for each sample; this results in slightly different statistical
properties of the samples generated (see
Section~\ref{sec:comparison}). BC sampling is not strictly Poisson
disk as there is a small chance that two samples could be very close
to each other.

A key issue in BC sampling is the number of candidates to generate for
each sample. To generate the $i$-th sample, Mitchell suggested using
$i * scale$ candidates, where $scale$ is a constant parameter, as this
maintains a constant ratio of candidates to samples already generated,
resulting in statistics that remain similar as the sampling density
increases.  But, this is computationally expensive for a large
number of samples and also results in the regular pattern of samples
we see in the GreedyFP sampling for a large number of candidates.
Therefore, the number of candidates is often capped to a maximum,
$maxCand$, as outlined in Algorithm~\ref{algo:bc}.

\begin{figure}[htb]
\centering
\begin{tabular}{cccc}
\includegraphics[trim = 2.9cm 0cm 2.7cm 0cm, clip = true,width=0.16\textwidth]{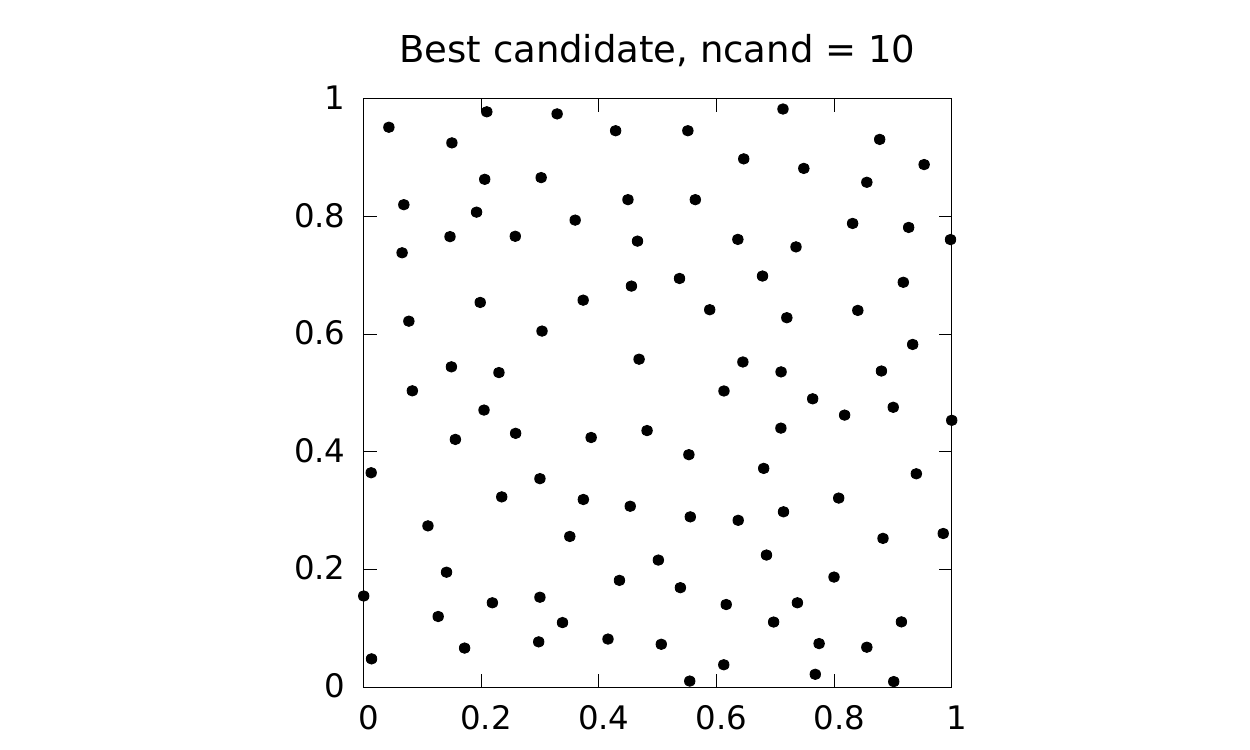} &
\includegraphics[trim = 2.9cm 0cm 2.7cm 0cm, clip = true,width=0.16\textwidth]{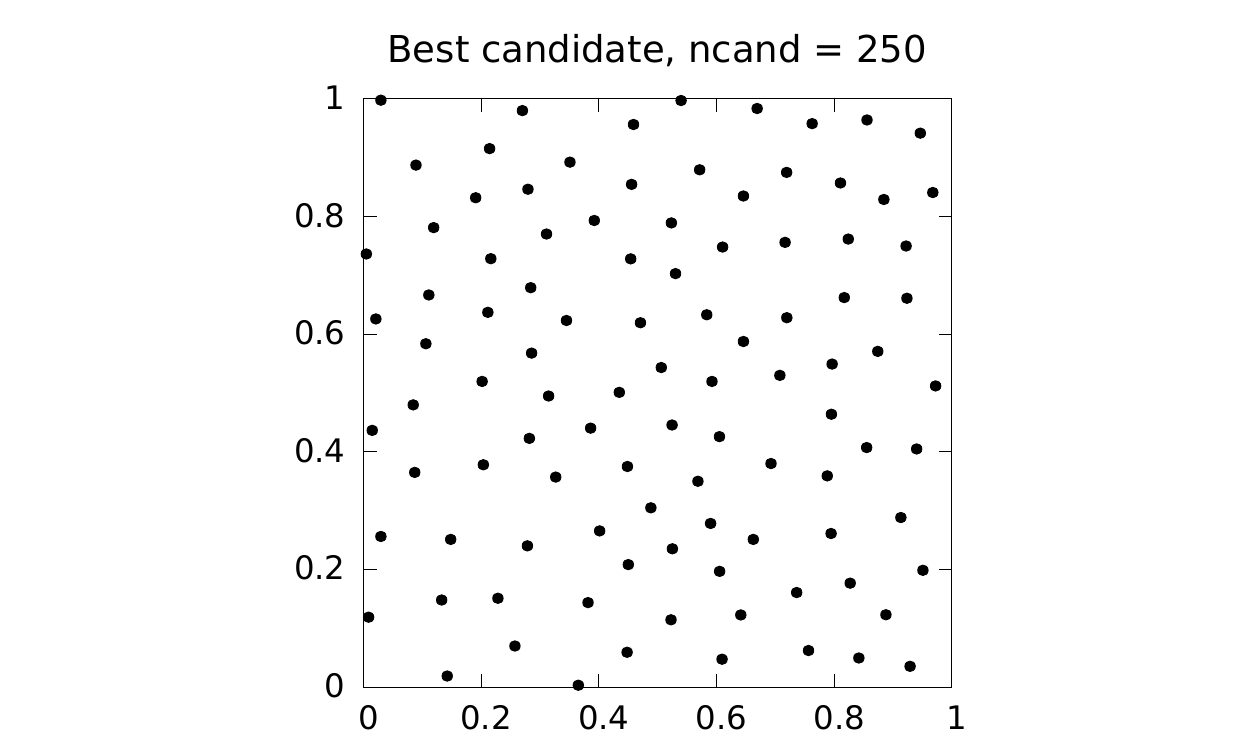} &
\includegraphics[trim = 2.9cm 0cm 2.7cm 0cm, clip = true,width=0.16\textwidth]{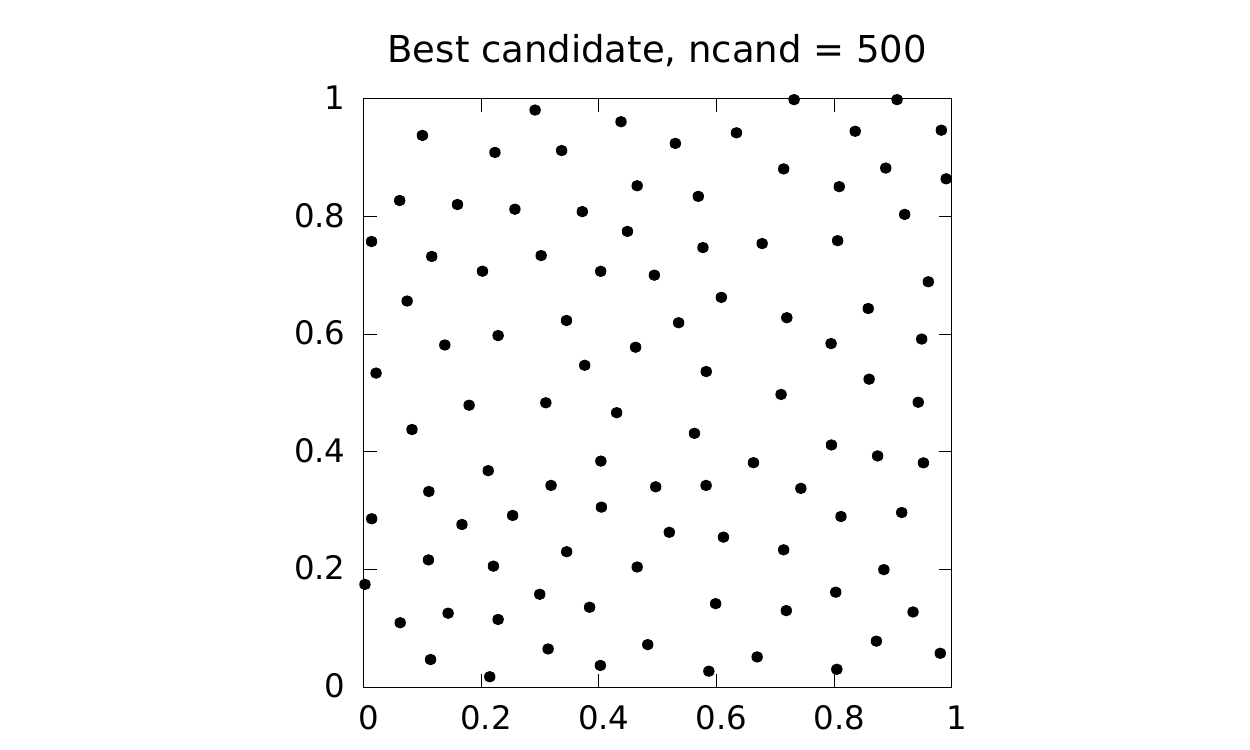} & 
\includegraphics[trim = 2.9cm 0cm 2.7cm 0cm, clip = true,width=0.16\textwidth]{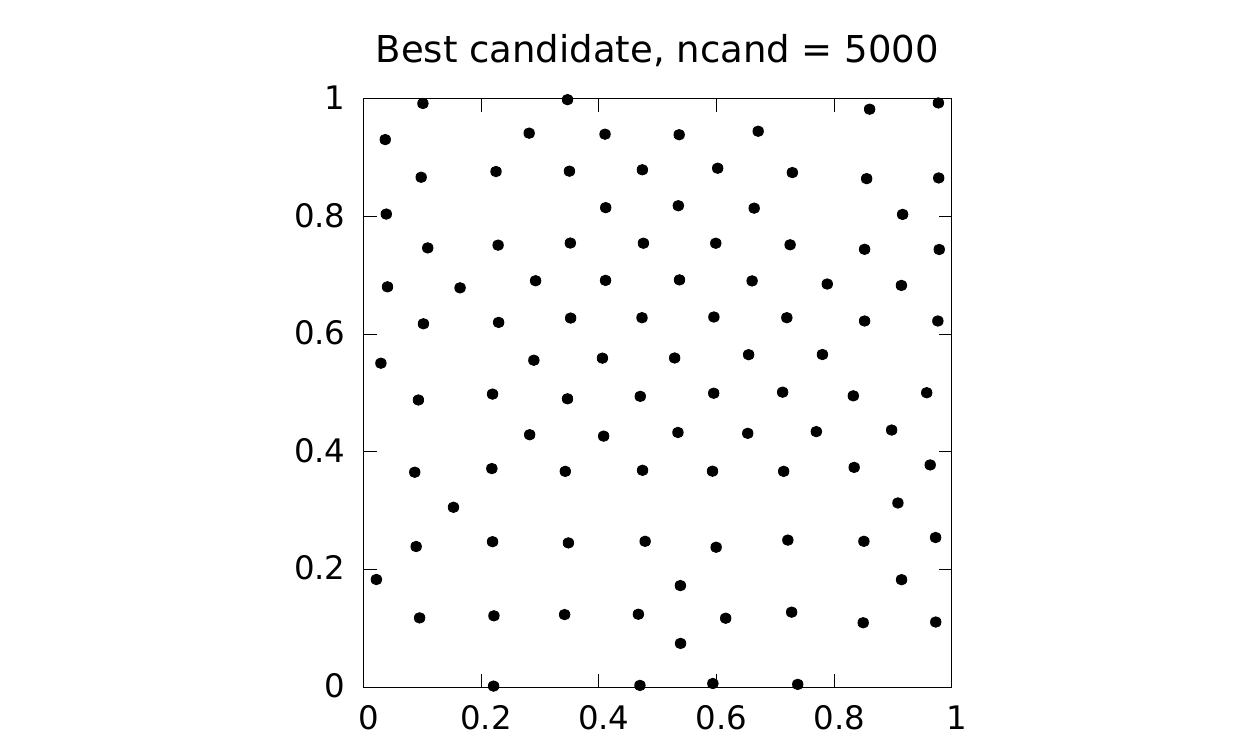} \\
(a) & (b) & (c) & (d)\\
\end{tabular}
\vspace{-0.2cm}
\caption{Best candidate sampling: 100 samples
  generated using (a) 10; (b) 250; (c) 500; and (d) 5000 candidates
  for each sample. With too few candidates, there is a large
  variation in the distance of a sample to the nearest neighbor, and
  with a large number of candidates, the sample locations lose their
  randomness and become more regular.}
\label{fig:samples_bc}
\end{figure}

In this paper, we present results using a modified best-candidate
algorithm, where we used a fixed number of candidates, {\it nCand},
for each new sample. Figure~\ref{fig:samples_bc} shows examples of 100
samples generated with different values of {\it nCand}.  As expected,
for a large number of candidates, the sampling becomes more regular.
Using a very small number of candidates results in a faster algorithm,
but as there are few candidates to choose from, the distances to the
nearest neighbor of a sample can vary. A moderate value of {\it
  nCand} is preferred for an efficient algorithm without any regular
patterns seen with a large number of candidates.

\begin{algorithm}
  \caption{Best candidate algorithm~\cite{mitchell91:sampling}}
\label{algo:bc}
\begin{algorithmic}[1]

  \STATE Goal: generate $N$ samples in a $d$-dimensional space that
  are far from each other.

  \STATE Set $maxCand$, the maximum number of random candidate samples,
  from which each new sample is selected and $scale$, a scale factor.

  \STATE Select the first sample randomly in the domain.

  \FOR{$i = 2$ to $N$}

  \STATE Set $nCand = \min \{ scale * i, maxCand \}$.

  \STATE Generate $nCand$ candidate samples in the $d$-dimensional
  domain randomly.

  \STATE Select the $i$-th sample from the $nCand$ candidate samples such
  that it is farthest from the $(i-1)$ samples selected thus far.

  \ENDFOR

\end{algorithmic}
\end{algorithm}

A sampling algorithm similar to the best-candidate and the GreedyFP
methods was independently developed by Crombecq et al. in 2011 for use
in surrogate
modeling~\cite{crombecq2011:spacefilling,crombecq2011:hybrid,crombecq2011:phdthesis}.
Their approach is based on Monte Carlo methods, where repeated random
sampling is used to compute a result. To adapt this idea to sequential
design, they propose generating a large number of candidate points in
the design space, computing a criterion for these points, and
selecting the optimum one as the next sample. The criteria include the
distance of a candidate to the existing samples and a combination that
excludes candidates that are within a certain projected distance of
existing samples.  Interestingly, like Mitchell's best candidate
algorithm, they scale the number of candidates with the number of
samples selected thus far in the sequential process. This approach is
similar to identifying the next instance to label in active
learning~\cite{settles2009:active}.

%
\subsection{Hybrid BC-GreedyFP sampling}
\label{sec:algo_bc_greedyfp}
%

Both the GreedyFP and the BC algorithms generate candidate samples and
select the next sample as the one that is farthest from the current
set of selected samples. The BC sampling is slower because a new set
of candidates is generated for each new sample and the minimum
distance of the candidates to the already-selected samples is
calculated.  The GreedyFP algorithm generates all candidates at the
start using random sampling, but due to under-sampling, there may be
small sub-regions with no candidates, and consequently, no sample. To
ensure equal probability of selecting any sample point in the domain,
we must choose a large enough scale factor, which results in a more
regularly-spaced set of samples.

These observations suggest that a hybrid BC-GreedyFP algorithm could
be faster than the BC algorithm and have better randomness properties
than the GreedyFP algorithm.  We could either speed up the BC
algorithm by selecting more than one sample from the set of candidates
that is generated, or make the GreedyFP algorithm more random by
periodically regenerating the candidate samples. We therefore
introduce a parameter, refreshCount, to indicate the number of samples
after which we generate a new set of candidates. This parameter acts
as a dial between the BC and GreedyFP algorithms; a small value makes
the hybrid closer to the BC algorithm, while a large value makes it
closer to the GreedyFP algorithm. We do not explicitly outline this
hybrid algorithm as it is a minor variation on
Algorithm~\ref{algo:greedyfp} or~\ref{algo:bc}.

%
\section{Adapting sampling algorithms to meet requirements}
\label{sec:adapting}
%

We next consider the sampling algorithms described in the previous
section and discuss their pros and cons in the context of the desired
criteria identified in Section~\ref{sec:task_goodalgo} for algorithms
used in surrogate modeling, hyperparameter optimization, and data
analysis. We show how some algorithms can be easily modified to meet
these criteria, while others, by virtue of their construction, cannot
support some of our requirements.

%
\subsection{Generating user-specified number of samples}
\label{sec:adapt_number}
%
 
Almost all of the sampling algorithms described in
Section~\ref{sec:algorithms} allow a user to specify any arbitrary
number of samples. The grid and stratified sampling algorithms require
the number of samples to be a product of the number of bins along each
dimension, which can result in a large number for high-dimensional
problems. The Poisson disk sampling, which uses the disk radius as
input, is more suitable for applications such as graphics, where the
number of samples is not the main consideration, but the distances
between them is.

%
\subsection{Generating randomly-placed samples}
\label{sec:adapt_random}
%

All the sampling algorithms considered generate randomly-placed
samples, with the exception of grid sampling, though stratified
sampling addresses this drawback. The CVT sampling can also result in
regularly-spaced samples aligned along the domain boundaries, but this
issue can be addressed by the use of the Latinizing algorithm
(Algorithm~\ref{algo:latinizing} in Section~\ref{sec:algo_cvt}). The
GreedyFP and BC sampling techniques could result in regularly-spaced
samples when the number of candidate samples is set too large; again
this issue can be addressed by using fewer candidates or through
Latinization of the sampling.  Recall that we do not consider
Quasi-Monte Carlo algorithms, such as the Sobol, Halton, and
Hammarsley sequences, precisely because they can have regular structure
when projected in some dimensions.

%
\subsection{Space-filling sampling}
\label{sec:adapt_spacefilling}
%

Many of the sampling techniques considered are space filling, which
was one of the main reasons for their consideration in this paper,
with the exception of the random sampling and the stratified random
sampling that can result in under- or over-sampled regions. A similar
problem can occur in an initial LHS, requiring the use of optimization
to increase the spacing between samples as discussed in
Section~\ref{sec:algo_lhs}.

%
\subsection{Progressive sampling}
\label{sec:adapt_progressive}
%

\begin{figure}[htb]
\centering
\begin{tabular}{ccc}
\includegraphics[trim = 2.9cm 0cm 2.8cm 0cm, clip = true,width=0.18\textwidth]{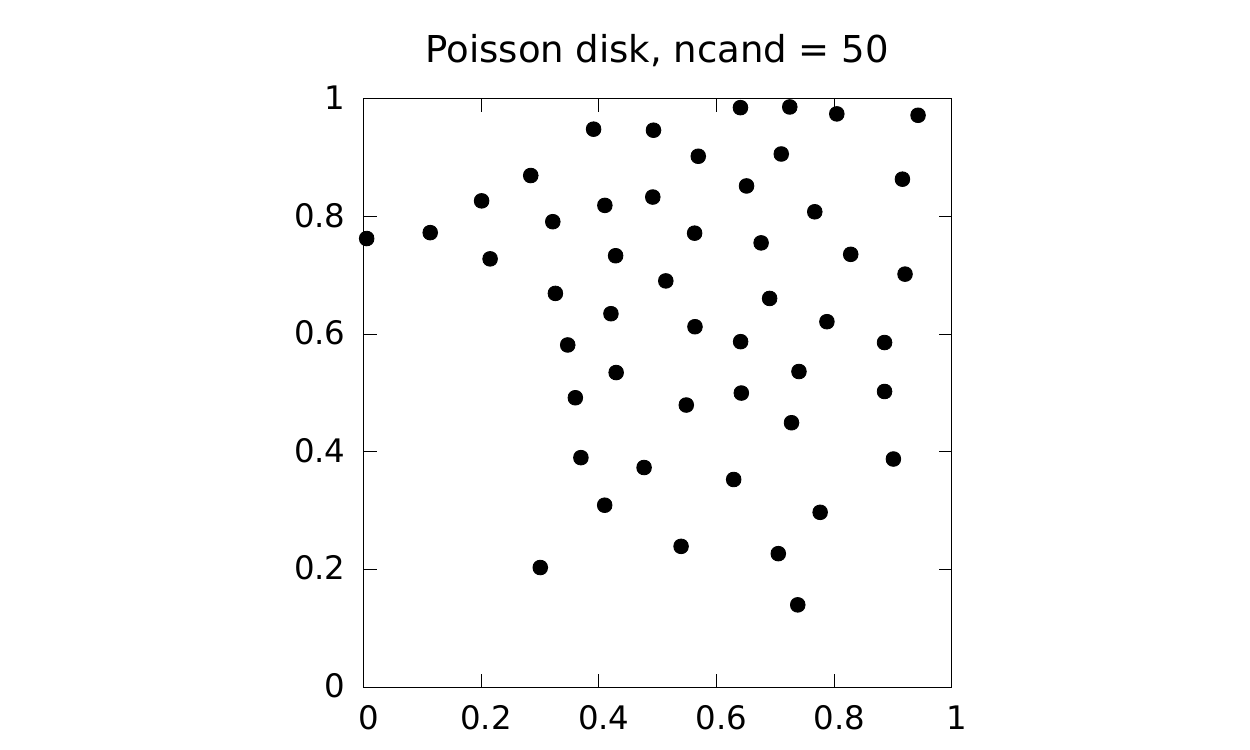} &
\includegraphics[trim = 2.9cm 0cm 2.8cm 0cm, clip = true,width=0.18\textwidth]{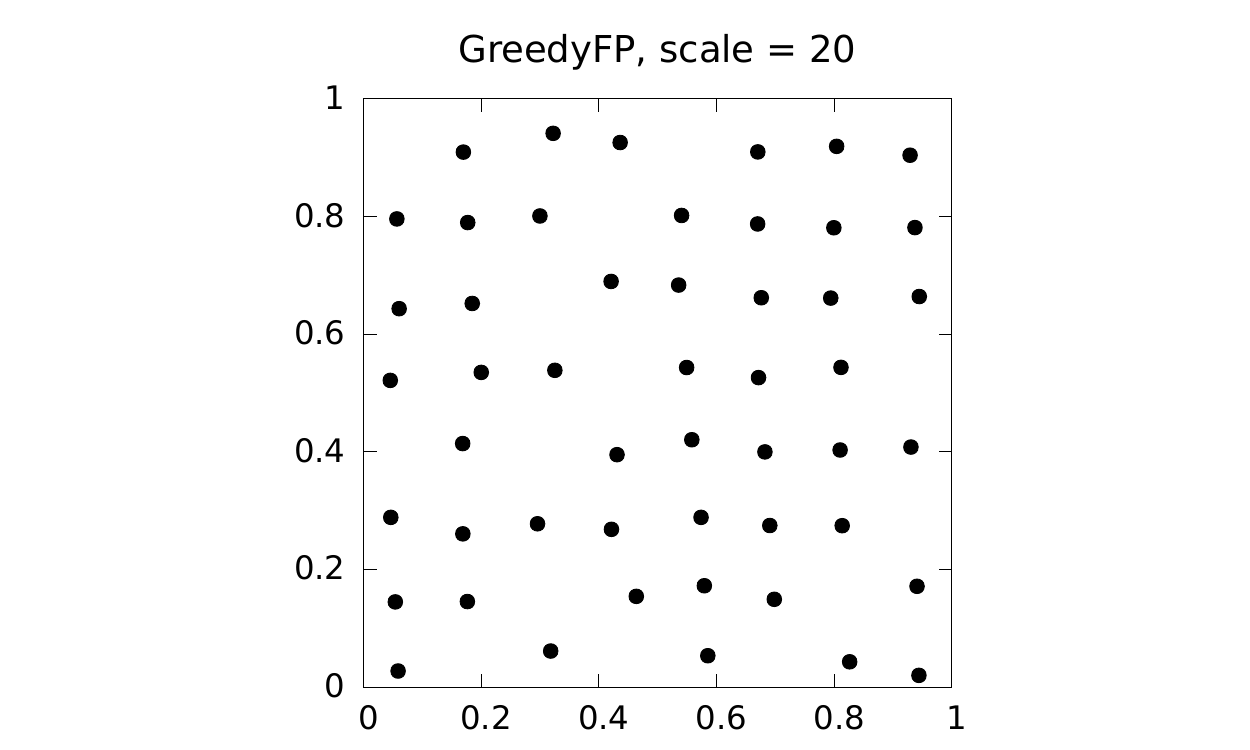} & 
\includegraphics[trim = 2.9cm 0cm 2.8cm 0cm, clip = true,width=0.18\textwidth]{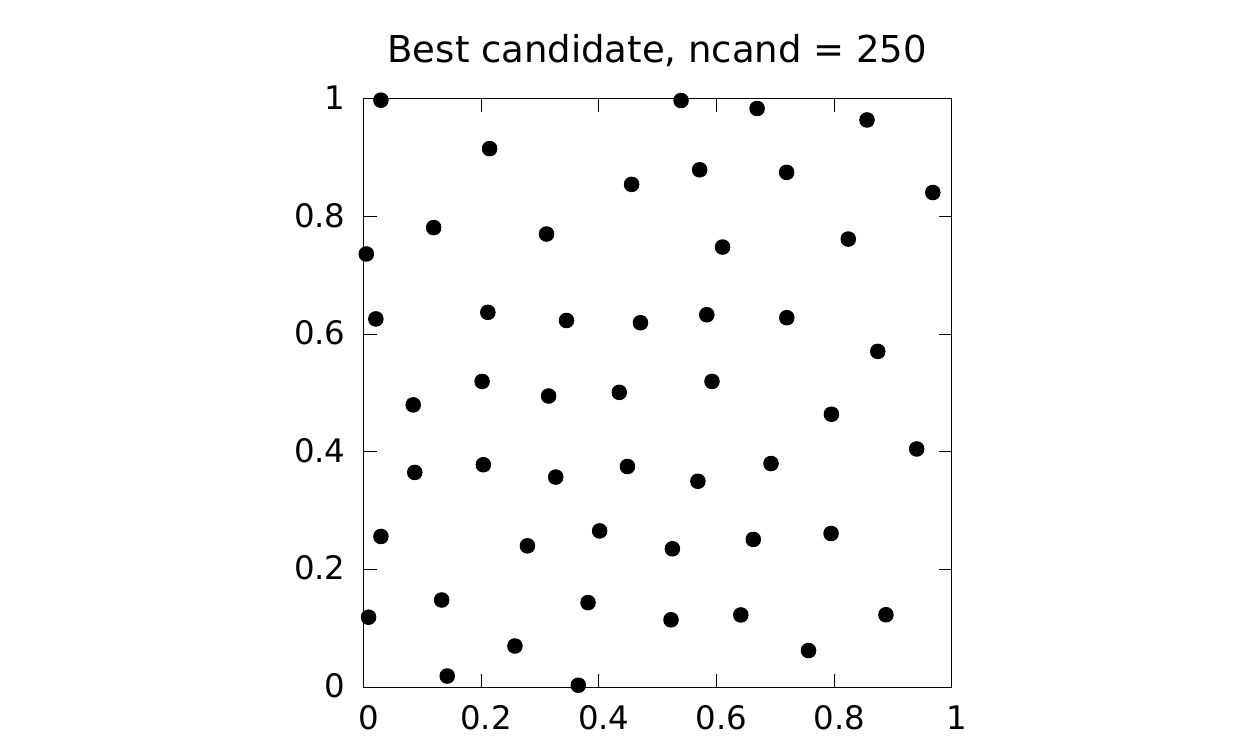} \\
\includegraphics[trim = 2.9cm 0cm 2.8cm 0cm, clip = true,width=0.18\textwidth]{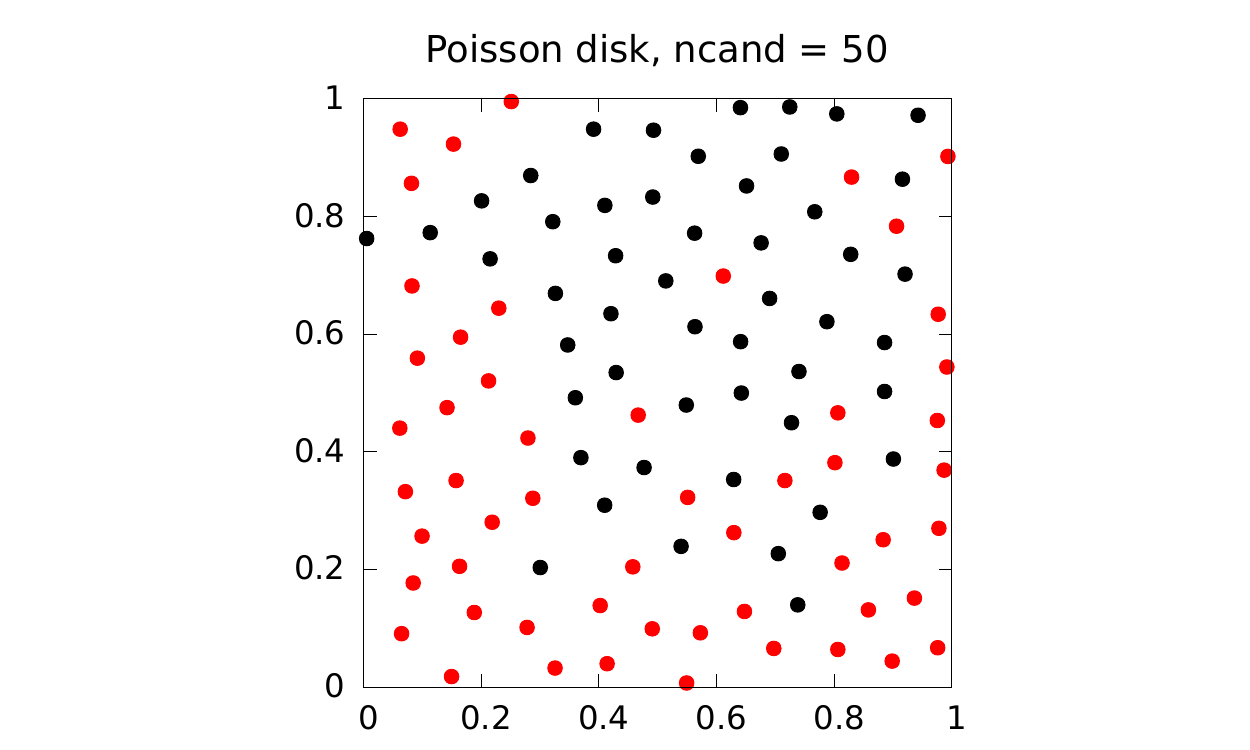} &
\includegraphics[trim = 2.9cm 0cm 2.8cm 0cm, clip = true,width=0.18\textwidth]{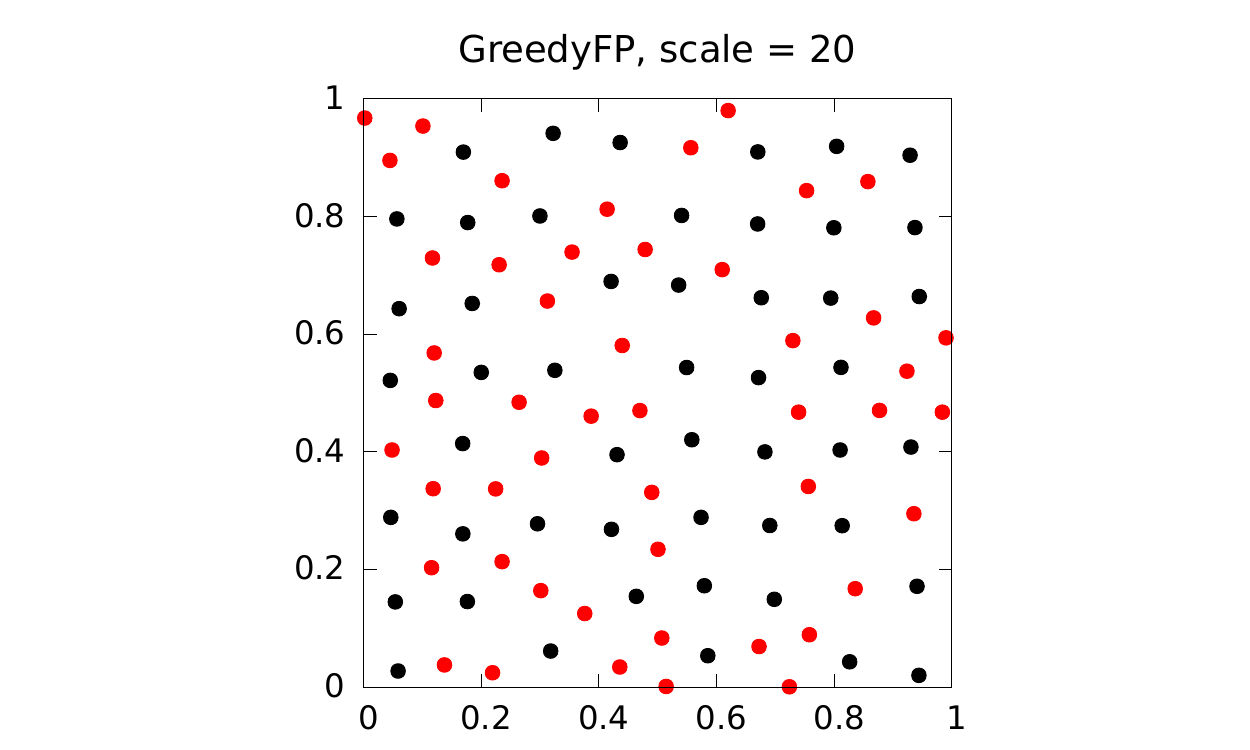} & 
\includegraphics[trim = 2.9cm 0cm 2.8cm 0cm, clip = true,width=0.18\textwidth]{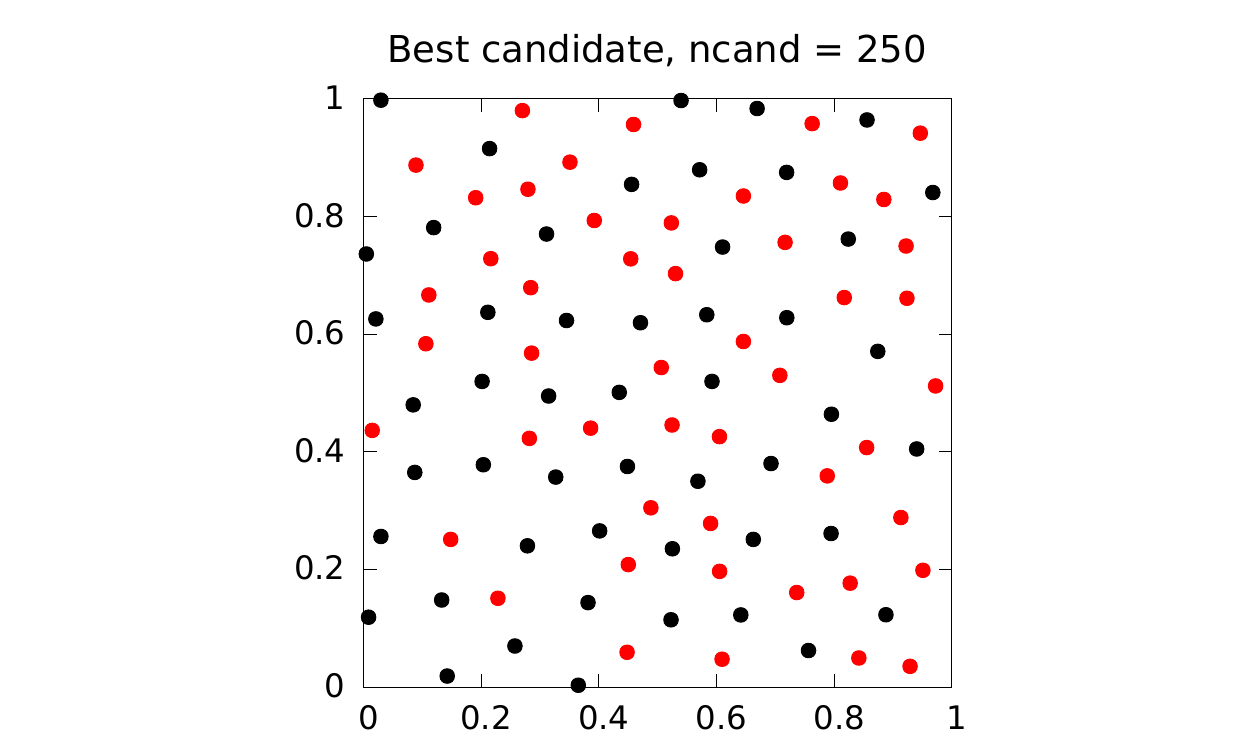} \\
(a) & (b) & (c) \\
\end{tabular}
\vspace{-0.2cm}
\caption{Progressive sampling: 100 Samples of (a) Poisson disk, (b)
  GreedyFP, and (c) BC, earlier shown in
  Figures~\ref{fig:samples_poisson}(b), \ref{fig:samples_greedyfp}(c),
  and \ref{fig:samples_bc}(b), respectively, with the first 50 samples
  (top, in black) and the next 50 samples (bottom, in red). The BC and
  GreedyFP algorithms cover the space uniformly at 50 samples, with a
  denser sampling at 100 samples, while Poisson disk adds the later
  samples mainly around the advancing front of initial samples.  The
  first 50 samples of GreedyFP are structured with a scale factor of
  just 20, while BC samples ({\it nCand} =250) remain random.}
\label{fig:samples_inc}
\end{figure}

Progressive sampling is a highly desirable property of a sampling
algorithm when used for tasks such as surrogate modeling and
hyperparameter optimization, where expensive computation is to be
performed at each sample point. In progressive sampling, as the
samples are generated, they start by covering the entire input domain
coarsely, with the coverage becoming finer with additional samples.
Obviously, techniques that generate a set of samples, such as grid,
stratified, LHS, and CVT sampling, do not have the progressive
property. However, even techniques that generate a sequence of samples
may not possess the progressive property, as shown in
Figure~\ref{fig:samples_inc} for the Poisson disk sampling which
samples along an advancing front created by the current set of
samples. In contrast, both GreedyFP and BC sampling generate samples
that cover the domain with increasingly finer resolution as the
sequence of samples is generated.

%
\subsection{Incremental sampling}
\label{sec:adapt_incremental}
%

This requirement is closely related to progressive sampling, but
instead of focusing on the coverage of the domain, we consider the
number of samples.  If we initially start with a set of random,
uniformly-placed samples, and need to add an unrestricted number of
samples to generate a denser sampling of the region, we should be able
to do so while retaining the space-filling property. Note that we do
not want to discard the original set of samples, but use them in
addition to the new samples.

Several sampling schemes do not support this capability. Both the grid
and the stratified sampling schemes can only double the number of bins
along each dimension. In CVT sampling, where the locations of the
generators (that is, the samples) are iteratively modified to satisfy
the space-filling requirement, we cannot add an unrestricted number of
new generators and change only their locations, while keeping the
locations of the original generators fixed.  In Poisson disk sampling,
where the input parameter is a radius, not a number of samples, we
will need to use a smaller radius to add new samples and fill-in
around the existing samples. As with the original samples, we have no
control over how many new samples this will create.

LHS sampling does not easily lend itself to the addition of an
unrestricted number of new samples. If the number of new samples is an
integer multiple of the number of original samples, then the new
samples can easily be added by splitting each interval along a
dimension into multiple bins and adding new samples into these bins,
such that the combined set of original and new samples satisfy the
Latin property. However, to ensure that the new sampling is still
space filling, the interchanges (Section~\ref{sec:algo_lhs}) have to
be applied to just the new samples as the original samples cannot
change their positions. This may result in a less than satisfactory
sampling in terms of its space-filling property.

Adding an arbitrary number of new samples to an already existing
LHS is a lot more challenging. None-the-less, as LHS is a very popular
sampling technique in surrogate modeling, there have been attempts to
address this requirement.  One solution is to divide each dimension
into the new number of intervals, and if there are multiple samples in
any cell, keep only one, and increment the number of new samples to be
added accordingly~\cite{nuchit2013:samplesize}. This means that the
processing already performed at the sample locations that were removed
would be wasted.  Alternately, one could leave the duplicates in, and
generate the additional new samples to meet the LHS criterion of one
sample in each interval in each dimension\cite{yan2011:inclhs}. While
no processing at previously-generated samples is wasted, this approach
results in some intervals having zero or multiple samples, so the
resulting design does not satisfy the Latin property.  Any optimization to
make the resulting sampling more space filling is now hampered by
these empty cells and the minimum distance between samples will be
determined by the cells with multiple samples.

In contrast to the other algorithms, the random, GreedyFP, and BC
techniques, by their construction, all allow additional samples to be
added very easily to an existing sample set without any restrictions
to the number of samples, though in the GreedyFP algorithm, the
candidate samples will need to be regenerated.

%
\subsection{Non-collapsing sampling}
\label{sec:adapt_noncollapsing}
%

\begin{figure}[!htb]
\centering
\begin{tabular}{ccc}
\includegraphics[trim = 2.5cm 0cm 2.0cm 0cm, clip = true,width=0.30\textwidth]{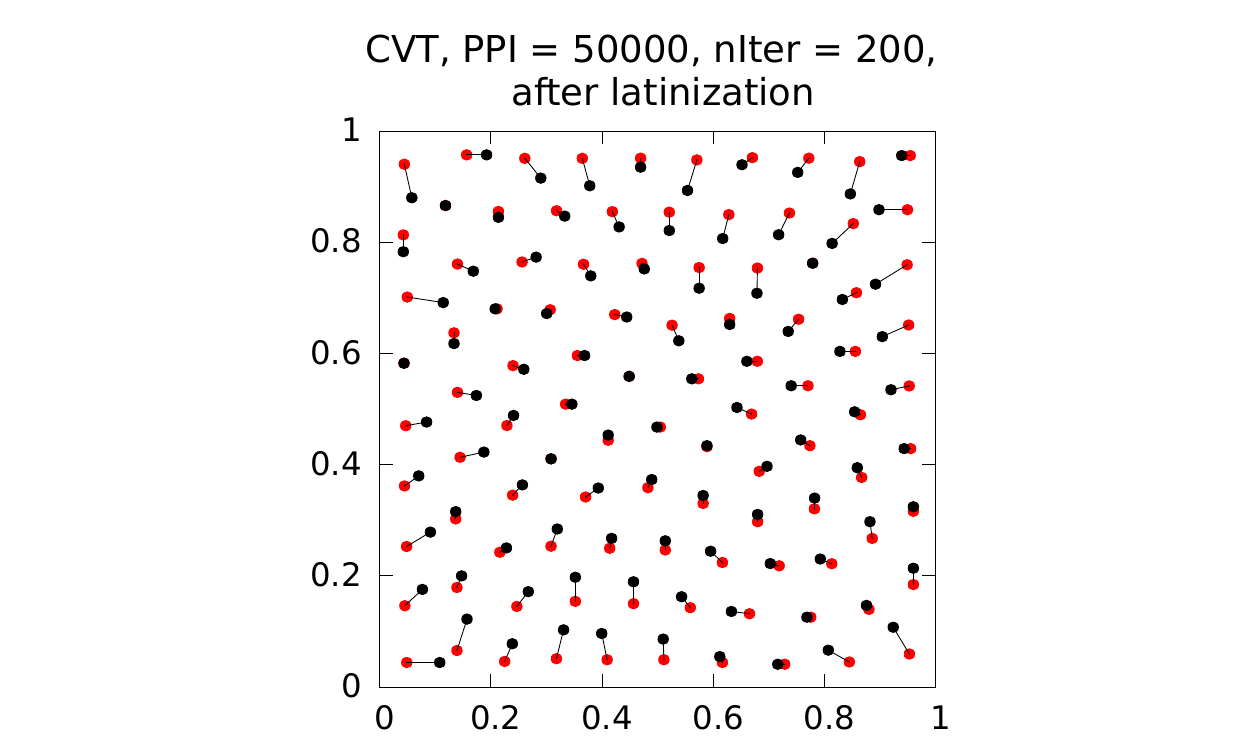} &
\includegraphics[trim = 2.5cm 0cm 2.5cm 0cm, clip = true,width=0.28\textwidth]{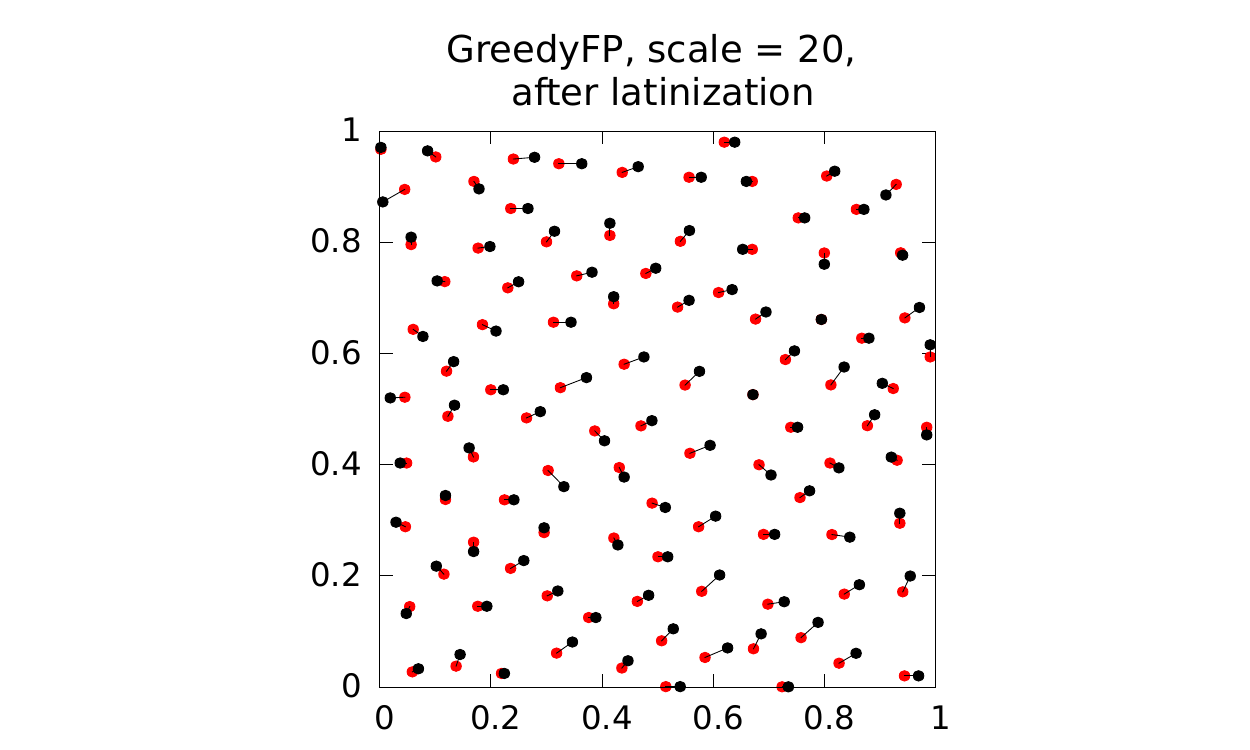} & 
\includegraphics[trim = 2.5cm 0cm 2.0cm 0cm, clip = true,width=0.30\textwidth]{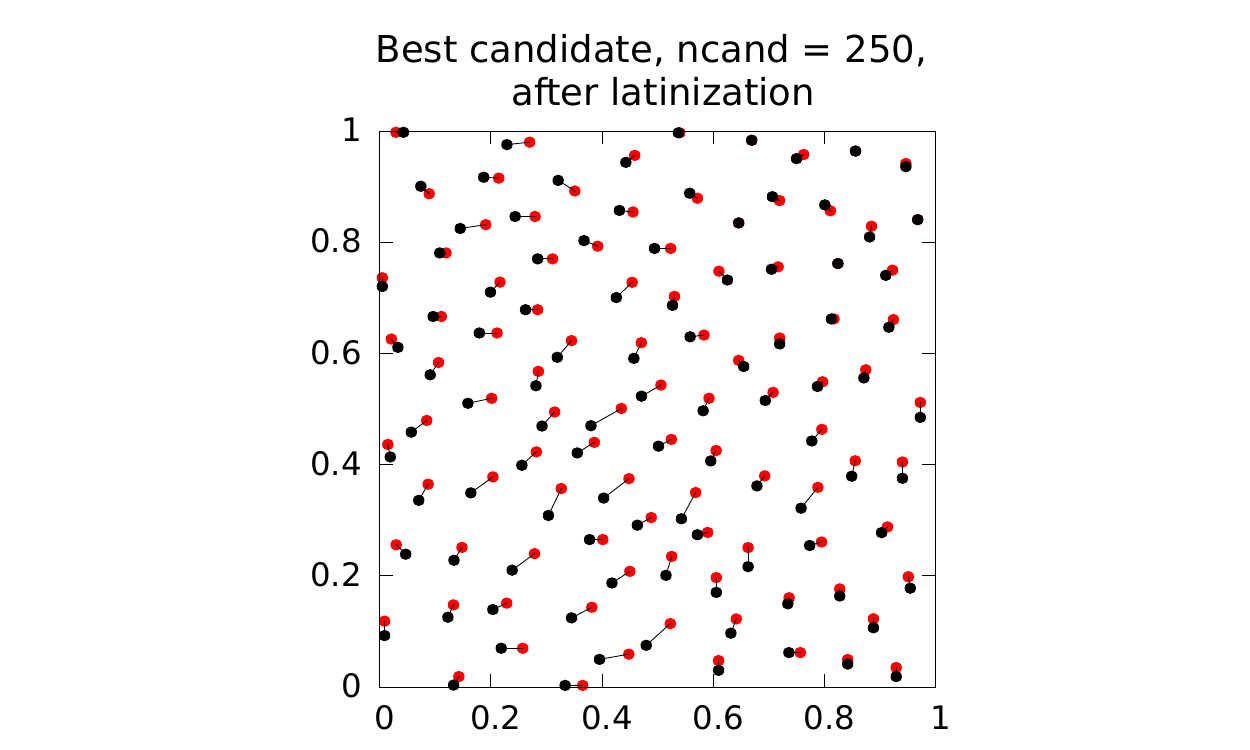} \\
\includegraphics[trim = 0.0cm 0cm 0.5cm 0cm, clip = true,width=0.31\textwidth]{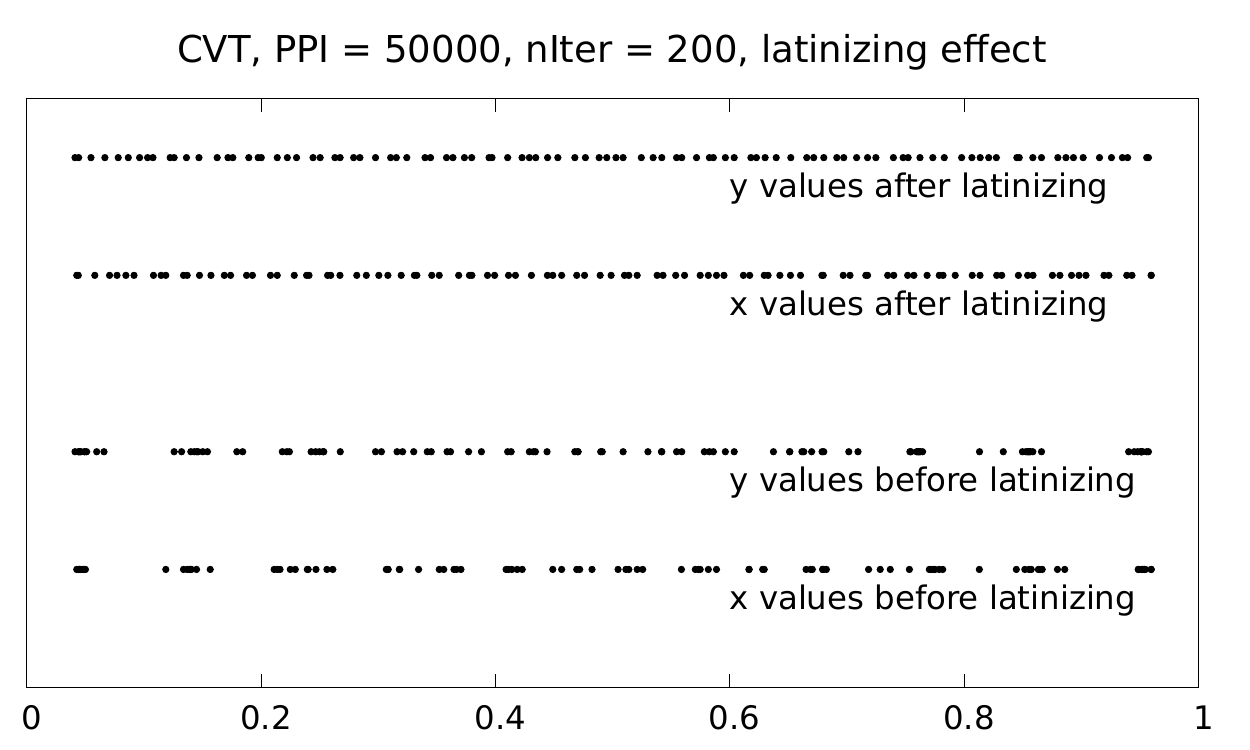} &
\includegraphics[trim = 0.0cm 0cm 0.5cm 0cm, clip = true,width=0.31\textwidth]{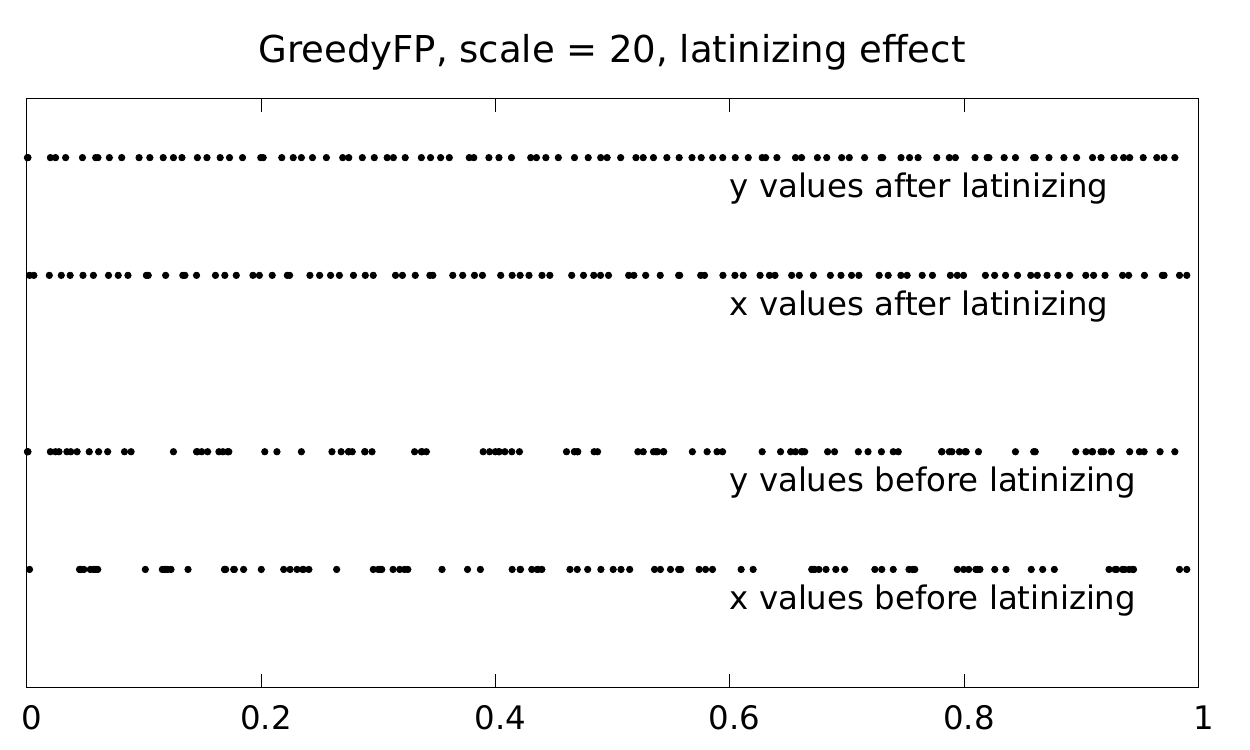} &
\includegraphics[trim = 0.0cm 0cm 0.5cm 0cm, clip = true,width=0.31\textwidth]{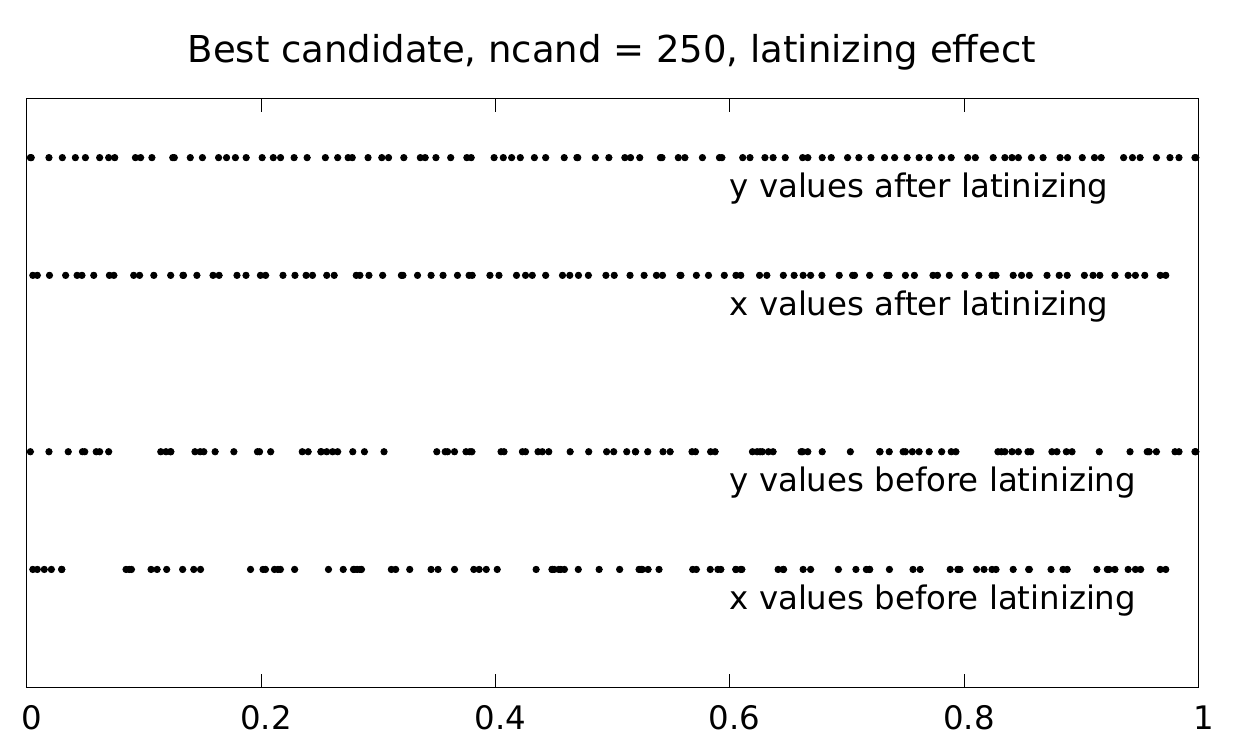} \\
(a) & (b) & (c) \\
\end{tabular}
\vspace{-0.2cm}
\caption{Effect of Latinizing 100 samples obtained using (a) CVT with
  {\it PPI} = 50000 and {\it nIter} = 200; (b) GreedyFP with {\it
    scale} = 20; and (c) BC with {\it ncand} = 250. The original and
  final locations of the samples are shown in red and black,
  respectively; the line indicates the distance moved.  Latinizing can
  move samples closer to each other as seen in BC sampling for some
  points near the center. The plots in the bottom row show the
  coordinates of the samples in each dimension before Latinizing
  (bottom two rows) becoming more uniform after Latinizing (top two
  rows). }
\label{fig:samples_latinize}
\end{figure}

A non-collapsing sampling is one where the sample points have unique
values along any input dimension, so that when the samples are
projected into the dimension, no two overlap. Some algorithms, such as
LHS, by definition, satisfy this requirement. In other
algorithms, such as grid sampling, the requirement in clearly
violated, while in CVT the regular structure of the sample points may
result in sample values close to each other in any dimension. One
solution to this is to add a post-processing step that jitters the
samples, using an approach from graphics for randomizing
samples~\cite{cook1986:stochastic}. Alternately, one can use the
Latinizing algorithm (Algorithm~\ref{algo:latinizing}) from
Section~\ref{sec:algo_lhs}.  This is a fast algorithm and can be used
to transform a sampling that meets other requirements into one that
also satisfies the non-collapsing property.
Figure~\ref{fig:samples_latinize} shows the effect of Latinizing
samples generated using CVT, GreedyFP, and BC sampling techniques, all
of which generate space-filling designs. Note that the samples do not
change their position by a large amount, though Latinizing may reduce
the distance between samples.

We observe that while Latinizing could create a non-collaping
sampling when a sample set is first created, it may be difficult to
continue to satisfy this property when new samples are added to an
existing set. An option may be to jitter only those new samples that
do not have unique coordinate values along a dimension.

%
\subsection{Sampling based on a probability density function}
\label{sec:adapt_pdf}
%

Sometimes, we need to generate samples in accordance with a
probability density function (PDF) such that some regions of the
domain are sampled more densely or sparsely than others, while at the
same time, generating samples that are random and uniformly
distributed.  This capability is useful when part of the domain
is of greater interest and would benefit from denser sampling than the
regions of lesser interest.

At first glance, many of the sampling algorithms considered in this
paper lend themselves to generating samples based on a PDF. However, a
closer look indicates that there are drawbacks. For example, both grid
and stratified sampling can accommodate a PDF by changing the spacing
of the grid appropriately, but as the grid spans the entire range of
each dimension, this may create regions with unnecessarily fine
sampling, unless the spacing is changed only locally as required,
which can be non-trivial to implement in high dimensions. For LHS, we
can suitably modify the grid spacing along each dimension to account
for the PDF, but the follow-on task of optimizing the distances using
interchanges will be more complex as not all interchanges will be
allowed due to the constraint of the PDF. For Poisson disk sampling, a
variable density of sampling points can be generated using the ideas
proposed by McCool and Fiume~\cite{mccool1992:poisson}.

\begin{figure}[!htb]
\centering
\begin{tabular}{ccc}

\includegraphics[trim = 2.9cm 0cm 2.7cm 0cm, clip = true,width=0.22\textwidth]{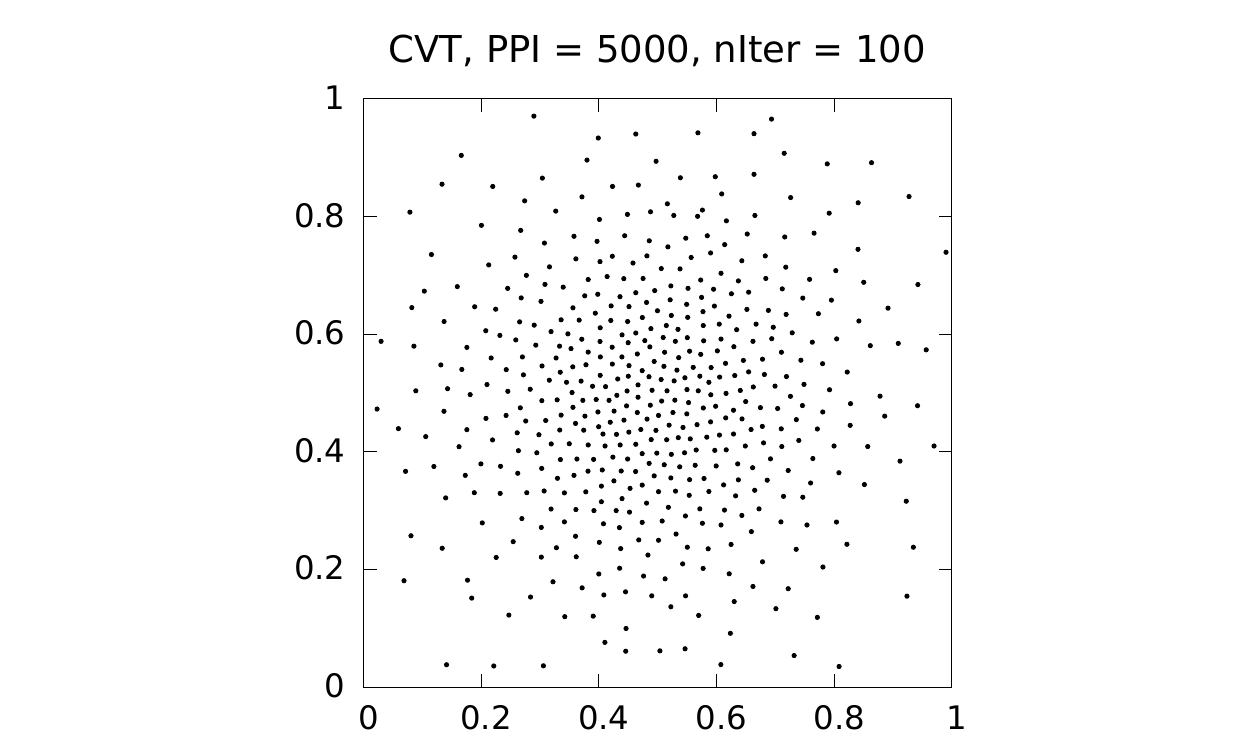} & 
\includegraphics[trim = 2.9cm 0cm 2.7cm 0cm, clip = true,width=0.22\textwidth]{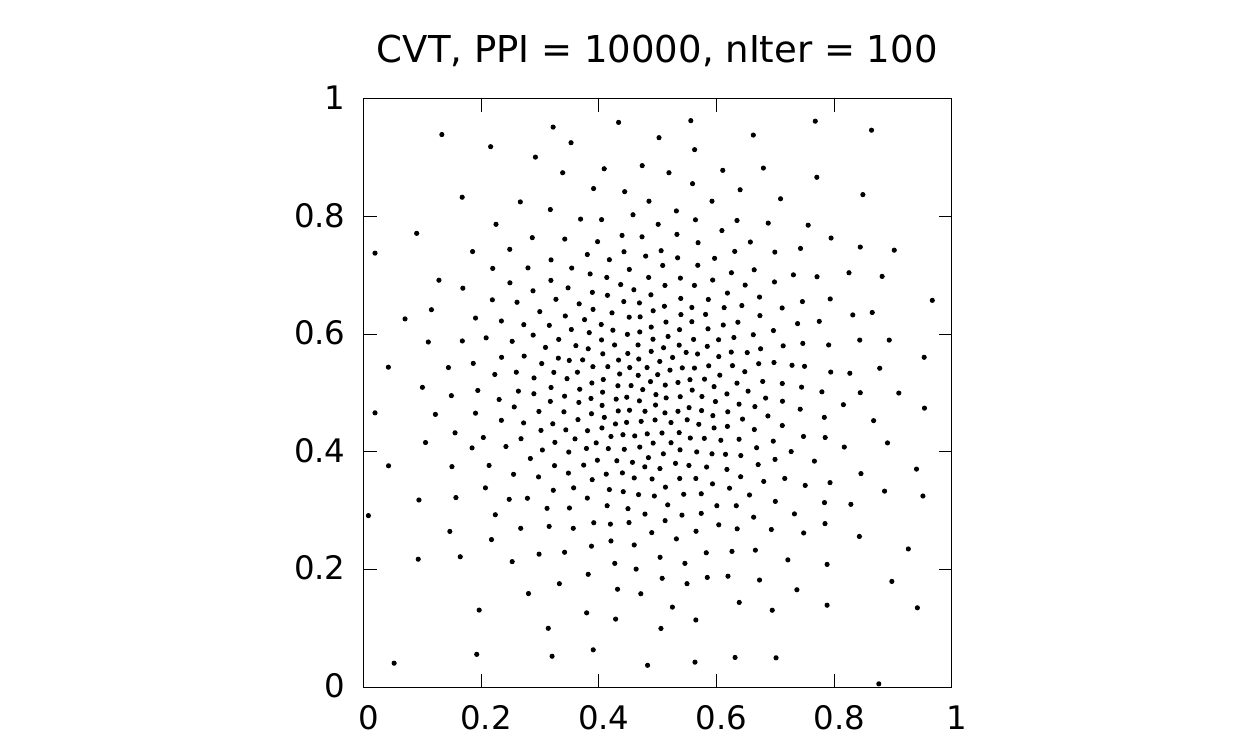} & 
\includegraphics[trim = 2.9cm 0cm 2.7cm 0cm, clip = true,width=0.22\textwidth]{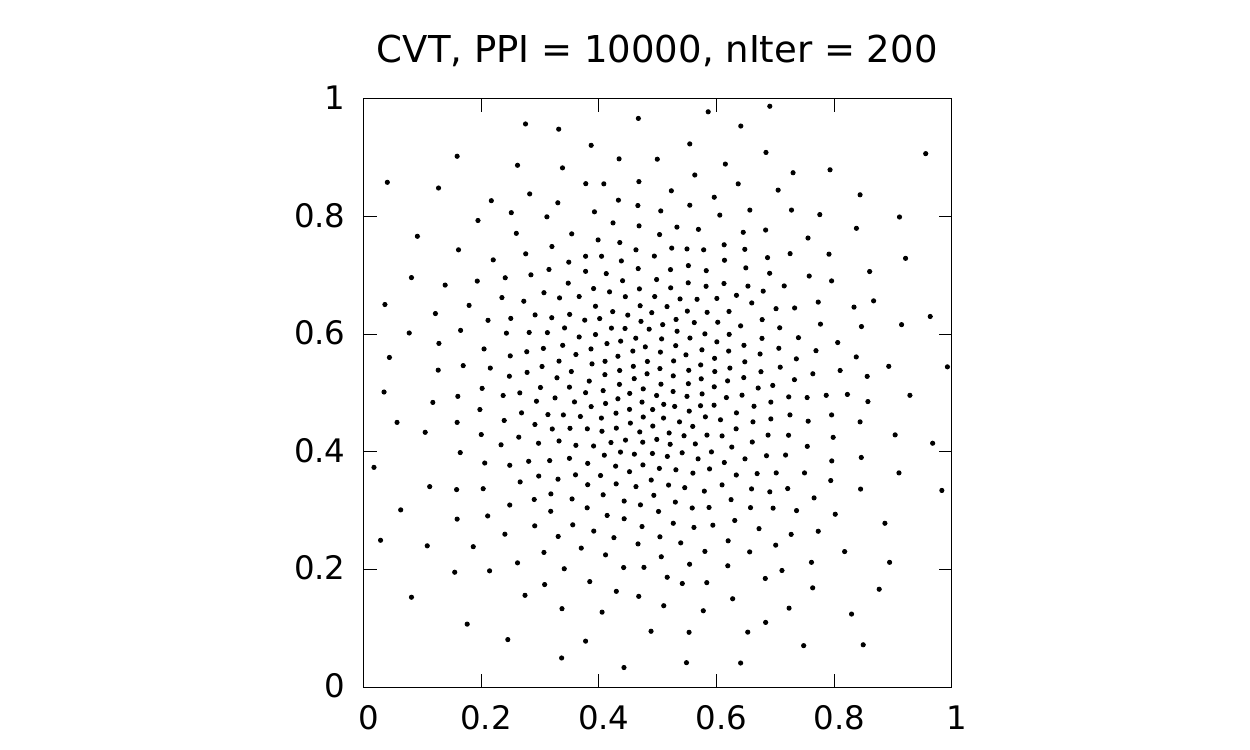} \\
(a) & (b) & (c) \\
\includegraphics[trim = 2.9cm 0cm 2.7cm 0cm, clip = true,width=0.22\textwidth]{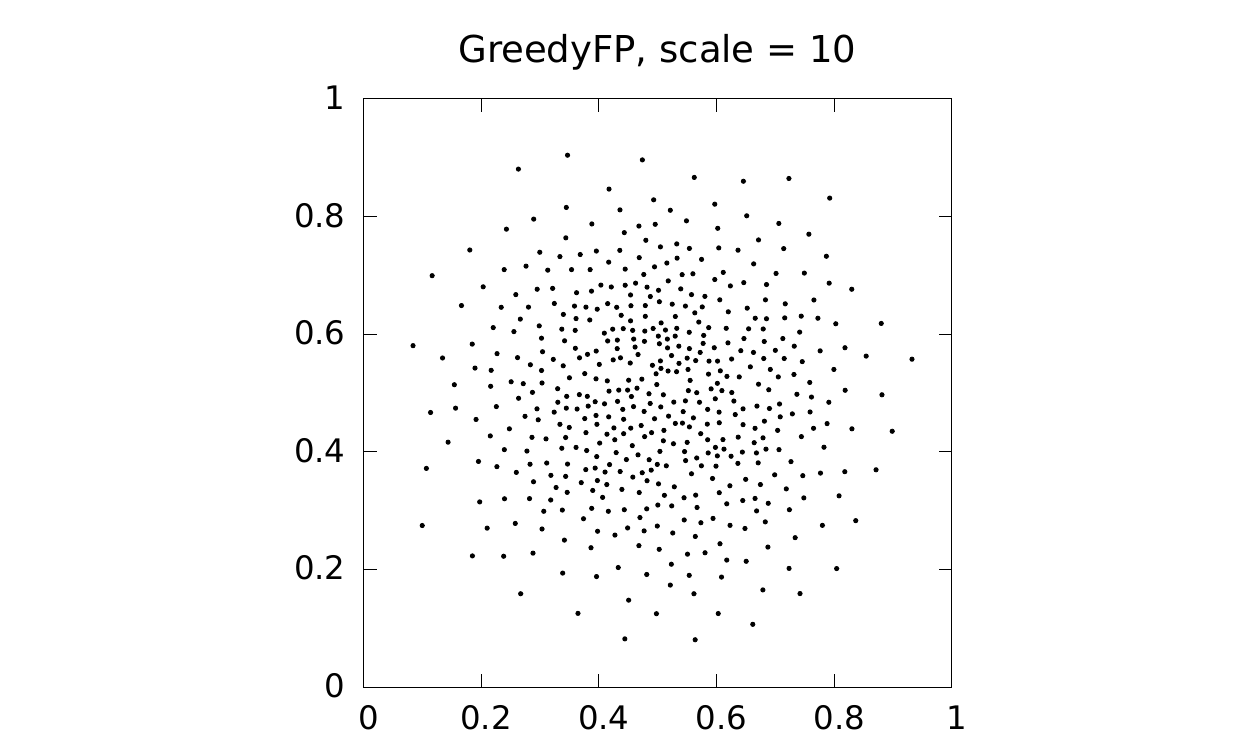} & 
\includegraphics[trim = 2.9cm 0cm 2.7cm 0cm, clip = true,width=0.22\textwidth]{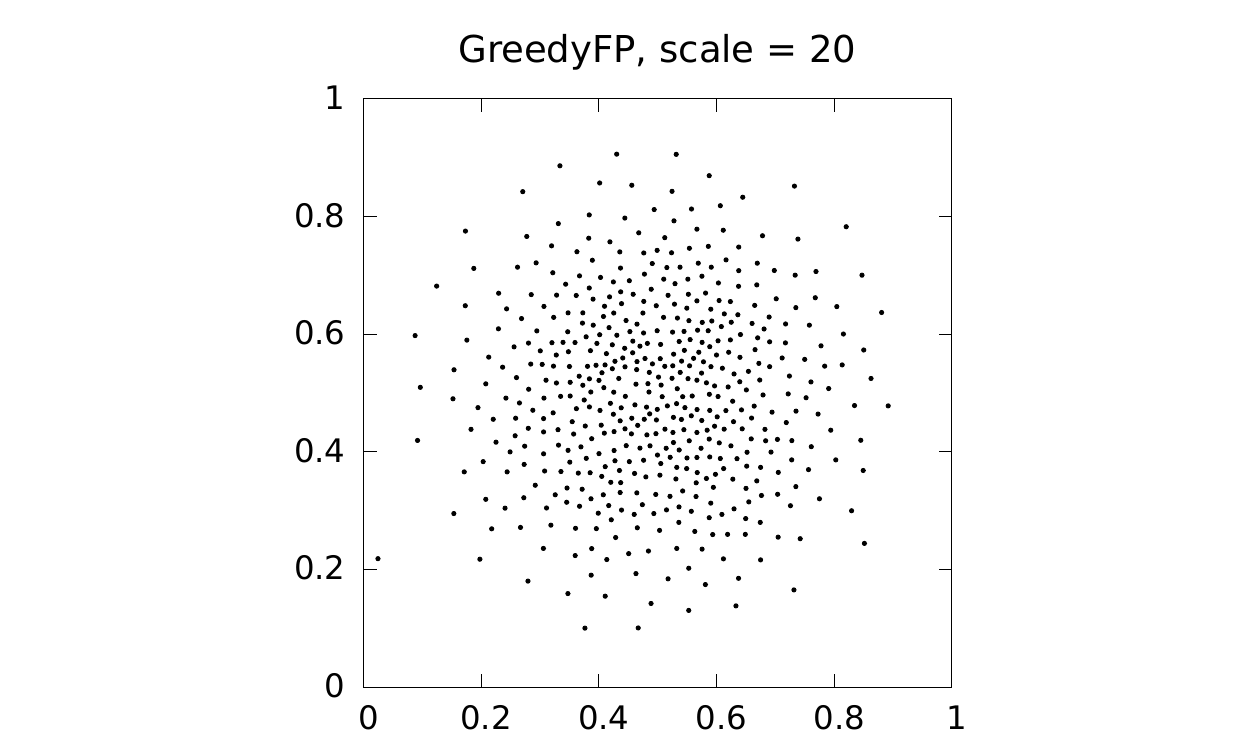} & 
\includegraphics[trim = 2.9cm 0cm 2.7cm 0cm, clip = true,width=0.22\textwidth]{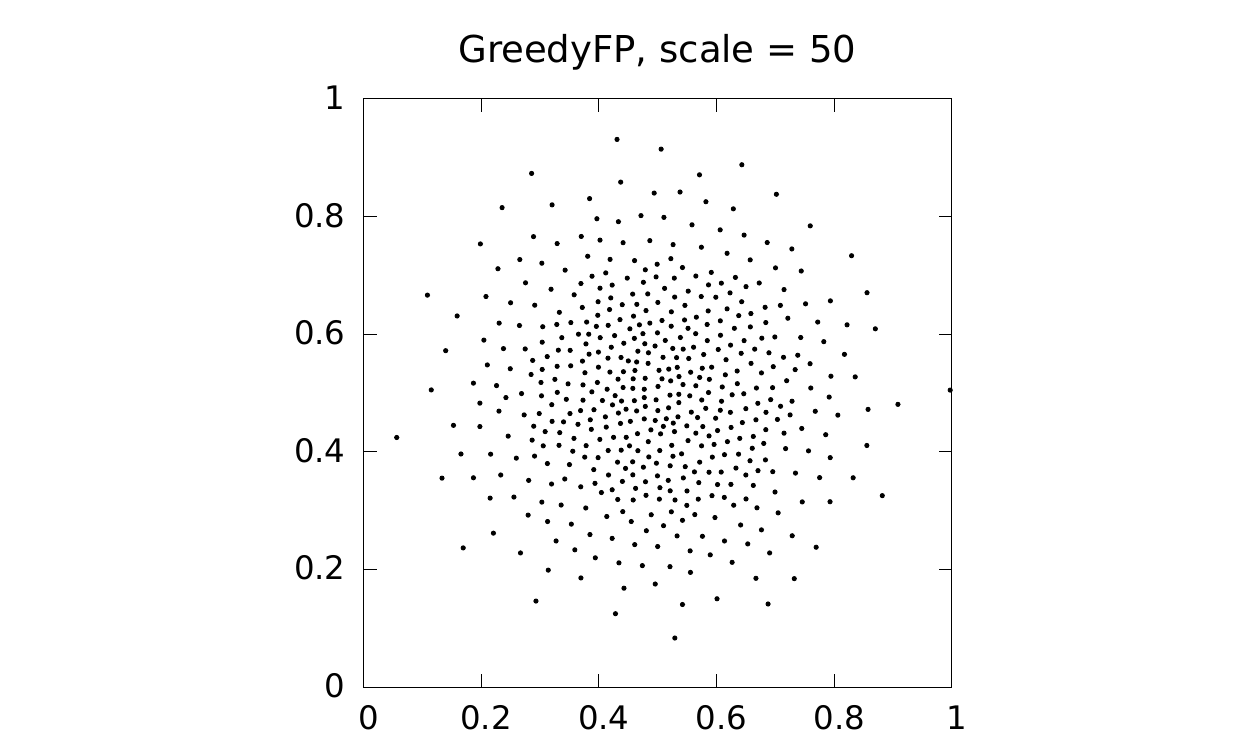} \\
(d) & (e) & (f) \\
\includegraphics[trim = 2.9cm 0cm 2.7cm 0cm, clip = true,width=0.22\textwidth]{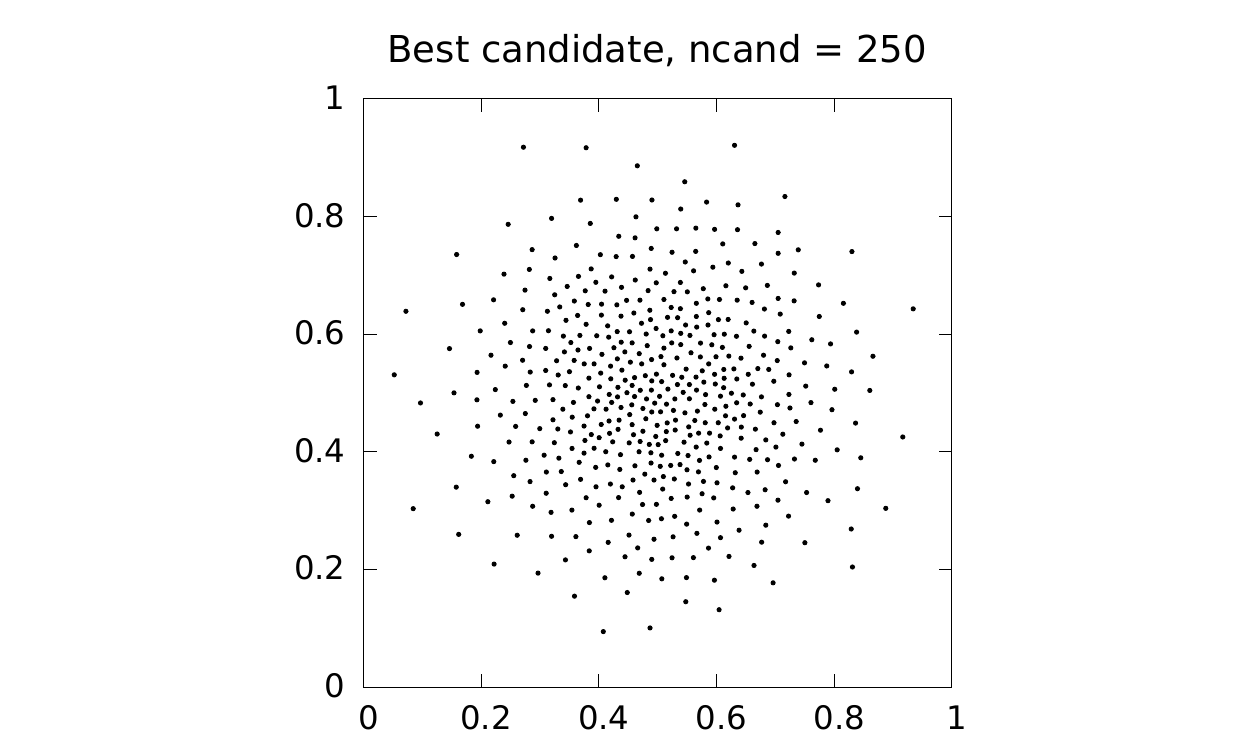} & 
\includegraphics[trim = 2.9cm 0cm 2.7cm 0cm, clip = true,width=0.22\textwidth]{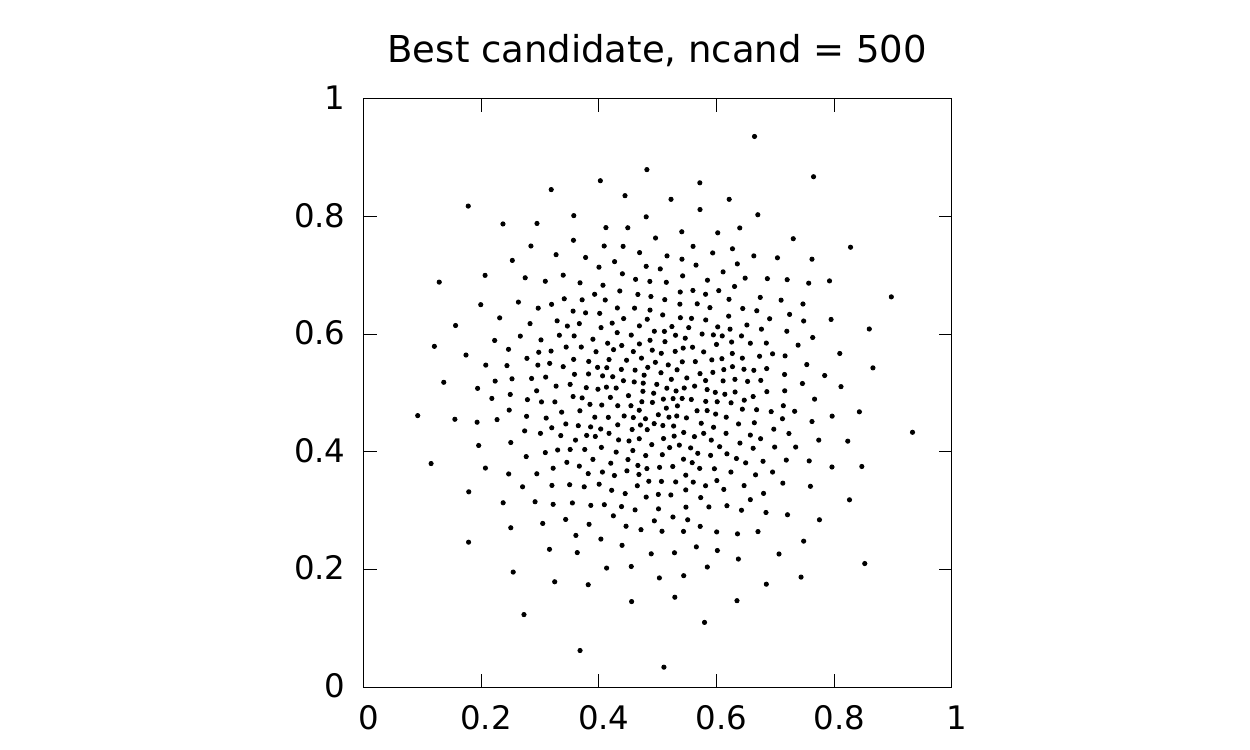} & 
\includegraphics[trim = 2.9cm 0cm 2.7cm 0cm, clip = true,width=0.22\textwidth]{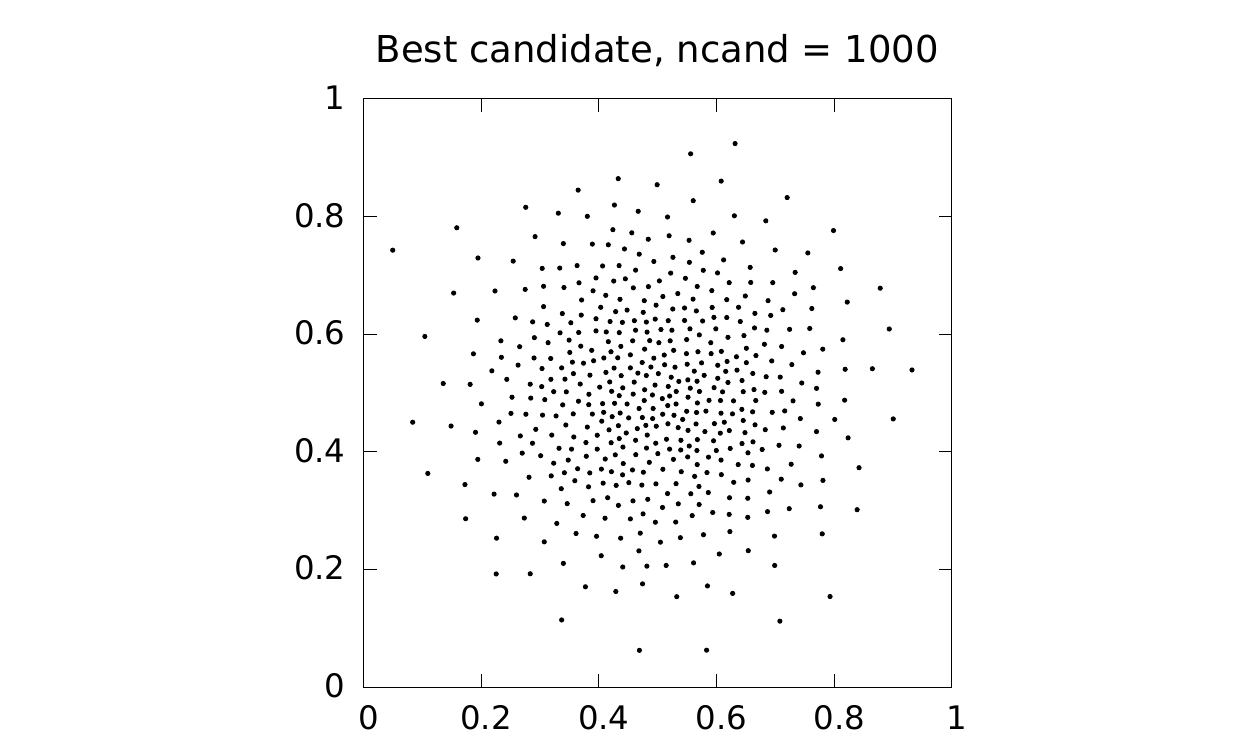} \\
(g) & (h) & (i) \\
\end{tabular}
\vspace{-0.2cm}
\caption{ 500 samples distributed according to a PDF: $pdf(x_1,x_2) =
  -20.0 * \exp\{(x_1-0.5)^2 + (x_2-0.5)^2\}$, with high probability
  regions having a greater density of samples.  (a) - (c) CVT with {\it PPI}
  = 5000, {\it nIter} = 100; {\it PPI} = 10000, {\it nIter} = 100; and {\it PPI} = 10000,
  {\it nIter} = 200, respectively.  (d) - (f) GreedyFP with {\it scale} = 10,
  20, and 50, respectively.  (g) - (i) best candidate with {\it ncand} =
  250, 500, and 1000, respectively.}
\label{fig:samples_pdf}
\end{figure}

It is relatively simple to modify CVT to generate samples in
accordance with a PDF as shown in Algorithm~\ref{algo:cvt}, step 6.
We can use the rejection method (\cite{larson81:book} Section 7.1.3)
to generate the {\it PPI} samples from the user-defined PDF.  In our
work, we made a minor modification to Algorithm~\ref{algo:cvt} to also
generate the initial $N$ generators (Algorithm~\ref{algo:cvt}, step 4)
using the same PDF to aid in the convergence of the algorithm.  The
results for 500 samples with a simple PDF are shown in
Figure~\ref{fig:samples_pdf}(a)-(c).

We can modify the GreedyFP and BC algorithms to generate samples in
accordance with a PDF by selecting a candidate in a denser region of the
PDF  over a candidate in a sparse region. One way to achieve this is to 
multiply the distance between a candidate and an already-selected
sample by the density at the location of the candidate, and 
select the candidate with the largest minimum distance as the next
sample. The results for 500 samples using
a simple PDF are shown in Figure~\ref{fig:samples_pdf}(d)-(i).

These results show that for the CVT,
GreedyFP, and BC sampling algorithms, different values for the
parameters seems to have relatively little effect on the quality of
the sampling, though a larger value of {\it PPI} in CVT, a larger {\it
  scale} factor in GreedyFP, and a larger {\it ncand} in BC sampling,
result in slightly better distributed samples. CVT tends to create
samples that are better separated in the dense regions, and cover the
four corners of the region better; this is the result of the generators
being moved during each iteration, unlike the GreedyFP and BC
sampling, where a sample, once selected, does not change its location.

The well-separated samples produced by CVT makes it useful in {\it
  stippling}, which is an artistic technique to place dots on paper
such that their density gives the impression of tone, where points
placed closer together form dark regions, while those placed further
apart form lighter regions~\cite{secord2002:stippling}.  We can
interpret the lighter and darker regions of an image as a PDF and the
dots as samples as being generated from this PDF.

%
\subsection{Sampling in non-rectangular regions}
\label{sec:nonrectangular}
%
 
Many, if not most, sampling algorithms assume that the samples are to
be generated in a rectangular domain, which may not always be the
case.  For example, in generating an initial set of samples for a
surrogate model, we may know a priori that some parts of the domain are
out-of-bounds or nonviable and we should not place any samples there.
Or, having generated an initial set of samples, we may need additional
samples only in a small region of the original domain, either to
explore this region further or to create more accurate surrogates in
this region by increasing the sampling density. These regions can
often be quite narrow, spanning the range of each dimension, but
occupying only a small part of the rectangular domain (see, for example,
Figure 6 in~\cite{kamath2018:small}).  We consider these two cases
separately and show how some of the sampling algorithms considered in
this paper can be used to address these requirements.

\begin{figure}[!htb]
\centering
\begin{tabular}{ccc}
\includegraphics[trim = 2.9cm 0cm 2.8cm 0cm, clip = true,width=0.18\textwidth]{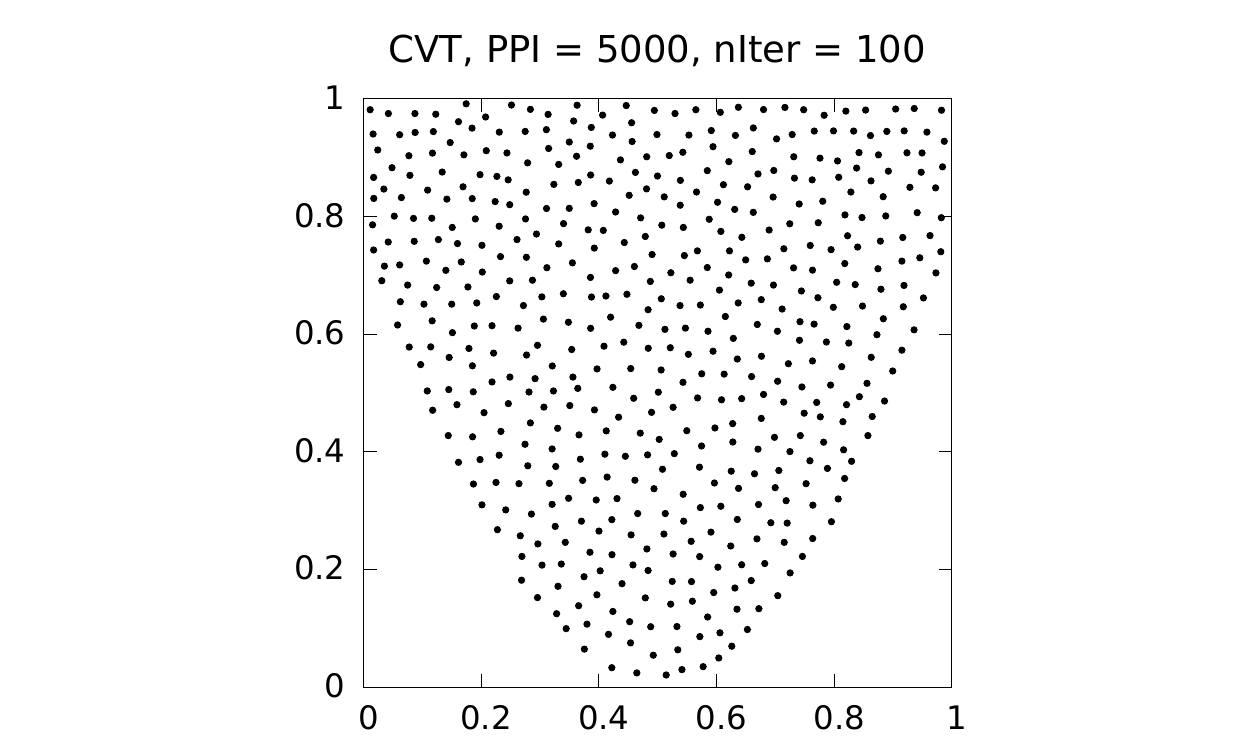} & 
\includegraphics[trim = 2.9cm 0cm 2.8cm 0cm, clip = true,width=0.18\textwidth]{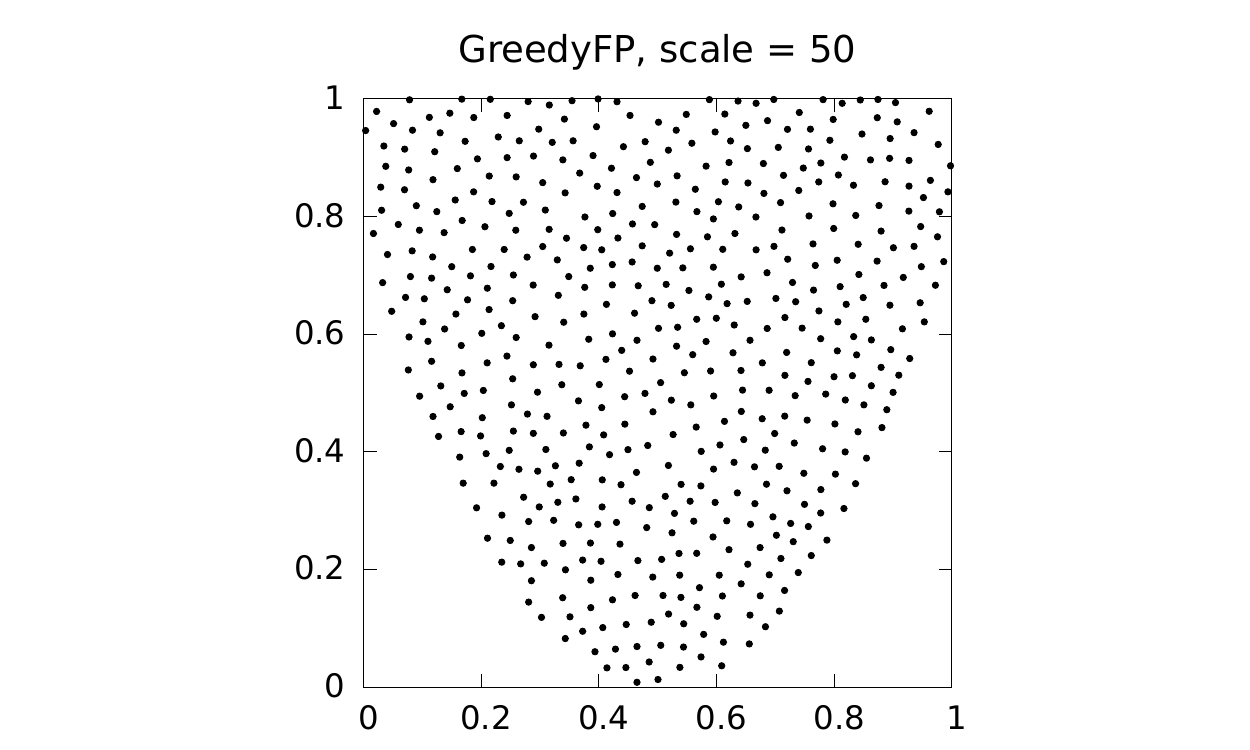} & 
\includegraphics[trim = 2.9cm 0cm 2.8cm 0cm, clip = true,width=0.18\textwidth]{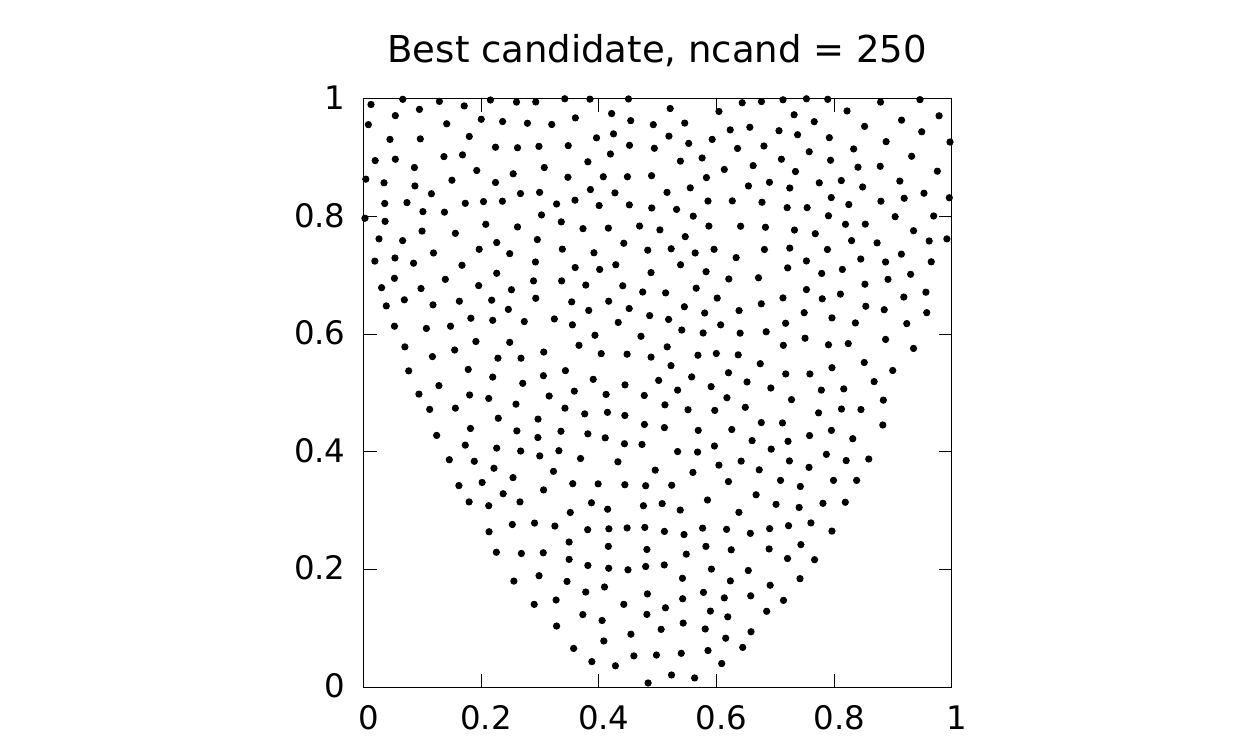} \\
\includegraphics[trim = 2.9cm 0cm 2.8cm 0cm, clip = true,width=0.18\textwidth]{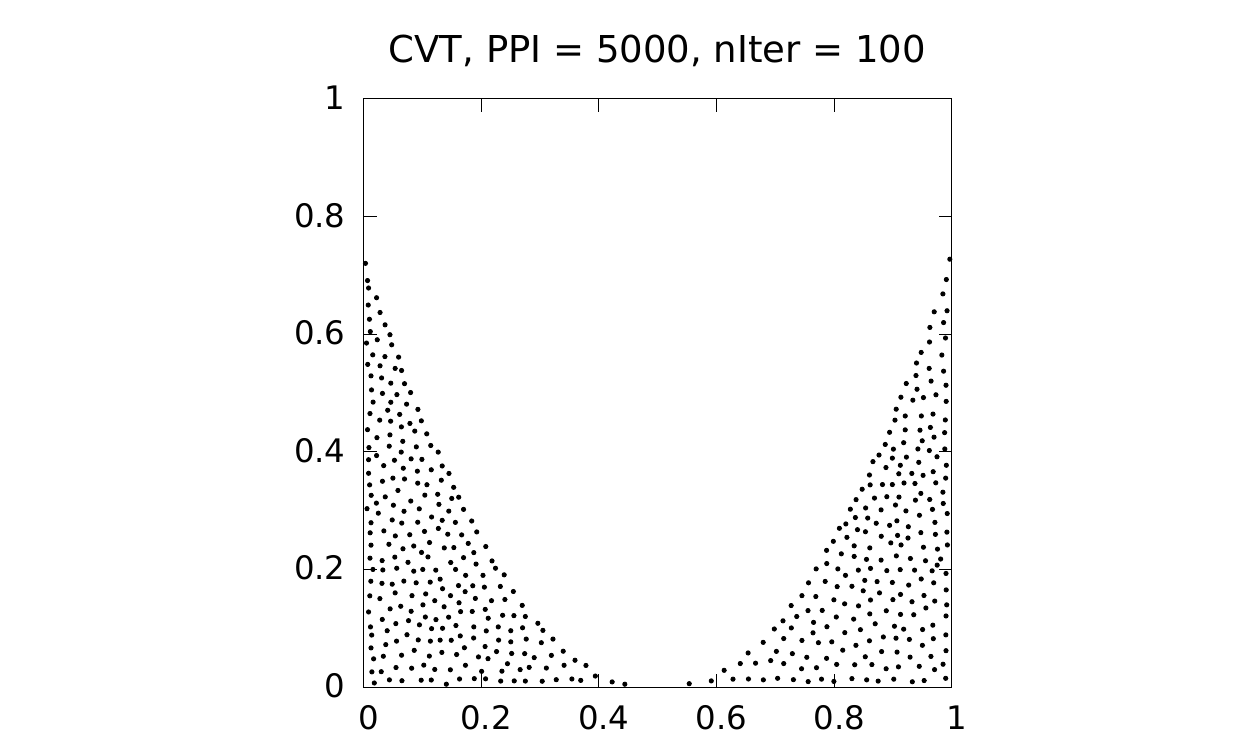} & 
\includegraphics[trim = 2.9cm 0cm 2.8cm 0cm, clip = true,width=0.18\textwidth]{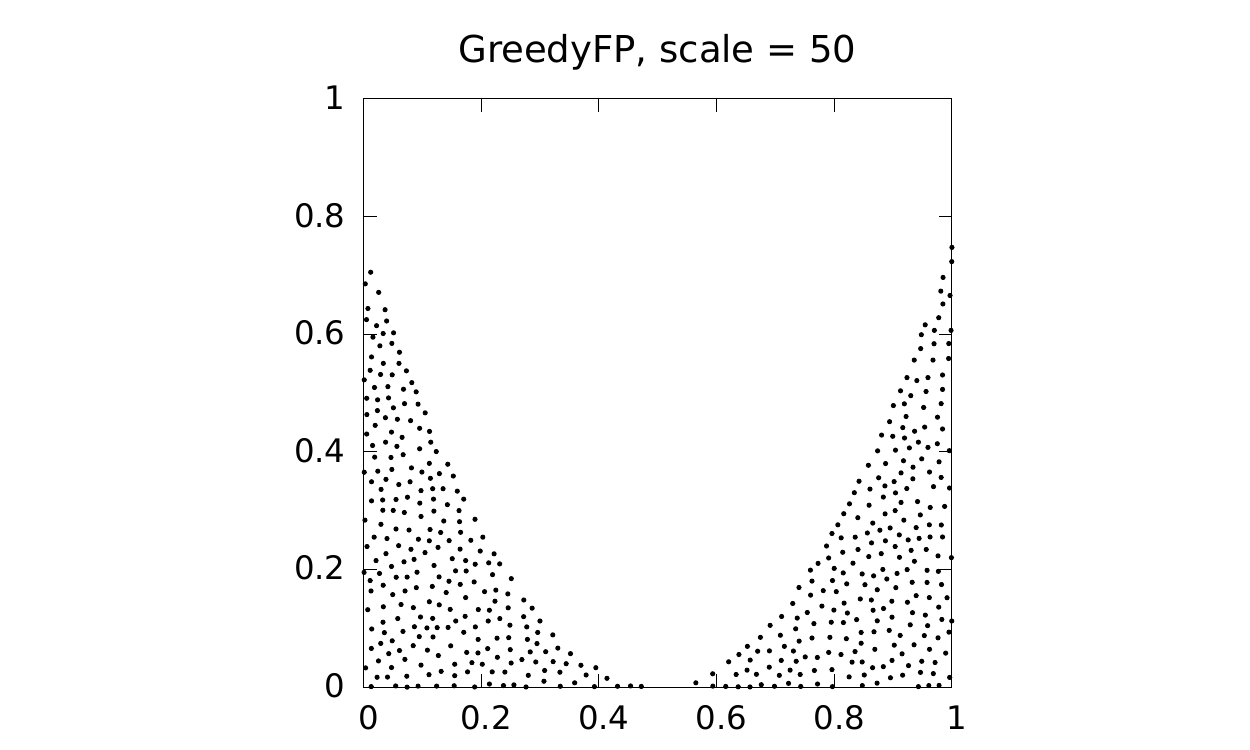} & 
\includegraphics[trim = 2.9cm 0cm 2.8cm 0cm, clip = true,width=0.18\textwidth]{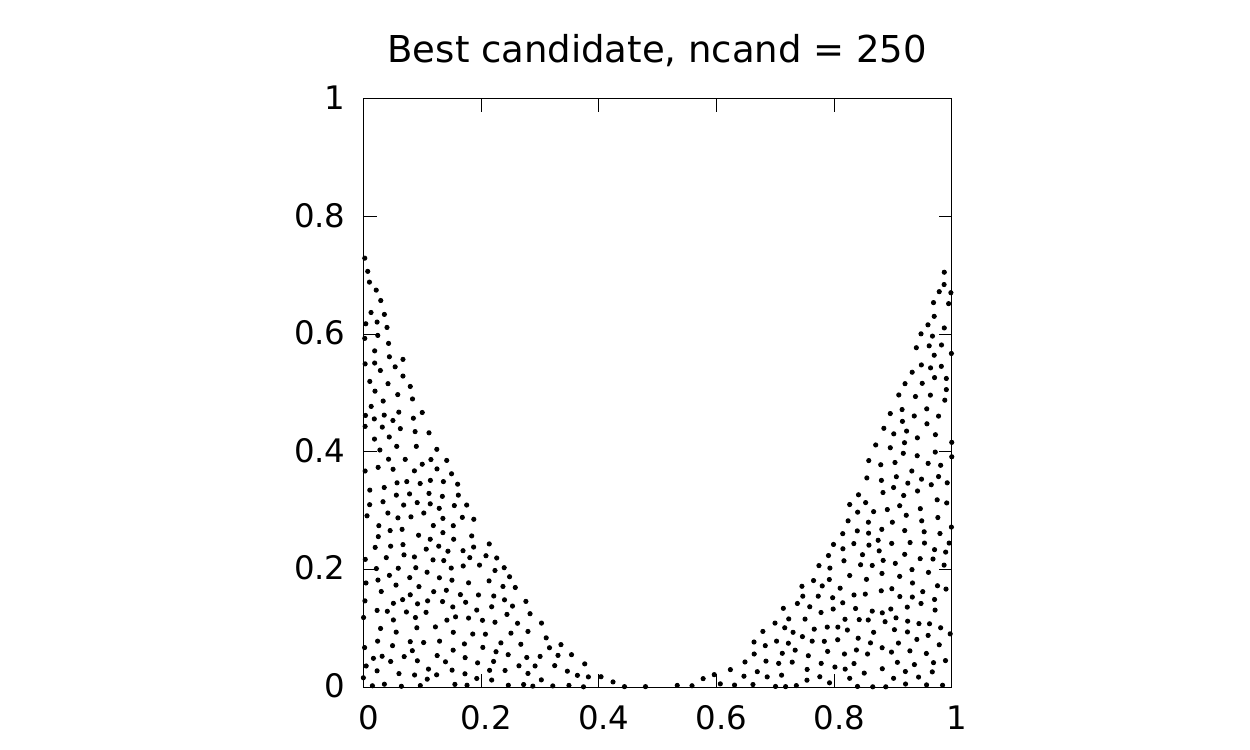} \\
(a) & (b) & (c) \\
\end{tabular}
\vspace{-0.2cm}
\caption{Uniformly distributed samples in a non-rectangular region:
  500 samples generated using (a) CVT with {\it PPI} = 5000, {\it
    nIter} = 100, (b) GreedyFP with {\it scale} = 50, and (c) BC with
  {\it ncand} = 250, respectively. Convex (top row) and concave
  (bottom row) regions defined by $x2 = 3*(x1-0.5)^2$, where $x1$ and
  $x2$ are the two dimensions.}
\label{fig:samples_non_rect}
\end{figure}

In the first case, where we want to generate or add to samples in a
viable region that is non-rectangular, assume that we have a function
that indicates whether a sample at a specific location in a domain is
allowed or not.  Figure~\ref{fig:samples_non_rect} shows an example of
a sampling with samples above or below the parabola $x2 =
3.0*(x1-0.5)^2$, where $0 \le x1 \le 1$ and $x2$ are the two
dimensions. It is straightforward to use random sampling to create such
a sample set with a specified number of samples. A grid sampling
allows us to place samples in the allowable region, but calculating
how many bins to use along each direction to generate the specified
number of samples is non-trivial. For LHS, the generation of the
initial samples, before interchanges, is a problem because all samples
are generated at the same time, and removing the non-viable ones later
on would reduce the number of samples and violate the Latin property.
Therefore, it is preferable to use an algorithm that starts by
putting samples only in the viable
region~\cite{mitry2014:surrogatethesis}.

This can easily be achieved for the CVT, GreedyFP, and BC algorithms,
as shown in Figure~\ref{fig:samples_non_rect} for 500 samples.  The
GreedyFP and BC algorithms generate candidate samples only in the
viable region, while in the CVT algorithm, we can restrict the
generation of the {\it PPI} samples to the viable region so the
generators would lie within the region as well.  The Poisson disk
algorithm, which is also based on candidate samples, can  be
similarly modified; however, it is now even more difficult to estimate
a disk radius for a user-defined number of samples in an odd-shaped
region.  If we want to add to already-existing samples in a
non-rectangular region, the GreedyFP and BC algorithms allow the
incremental addition, while CVT and LHS do not support this capability
(as explained in Section~\ref{sec:adapt_incremental}).

In the second case, we want to add new samples to a non-rectangular
region defined not by a viability function, but by existing samples.
This can be considered a special case of incremental sampling, but
instead of a rectangular region as discussed in
Section~\ref{sec:adapt_incremental}, the region could have an odd
shape. The task is especially challenging when this region occupies a small
fraction of the rectangular domain spanned by  the
coordinates of the samples that define the region.  For example, in
our previous work (Figure 6 in~\cite{kamath2018:small}), we found an
approximate solution to an inverse problem in the form of a curve,
defined by a small number of sample points, in a two dimensional
space.  We wanted to add new sample points, uniformly but randomly,
close to the curve so we could build an accurate surrogate in this
region and use it to find the exact solution to the inverse problem,
which we expected to be in the vicinity of the approximate solution.

\begin{figure}[!htb]
\centering
\begin{tabular}{ccc}
\includegraphics[trim = 1.5cm 0cm 1.0cm 0cm, clip = true,width=0.3\textwidth]{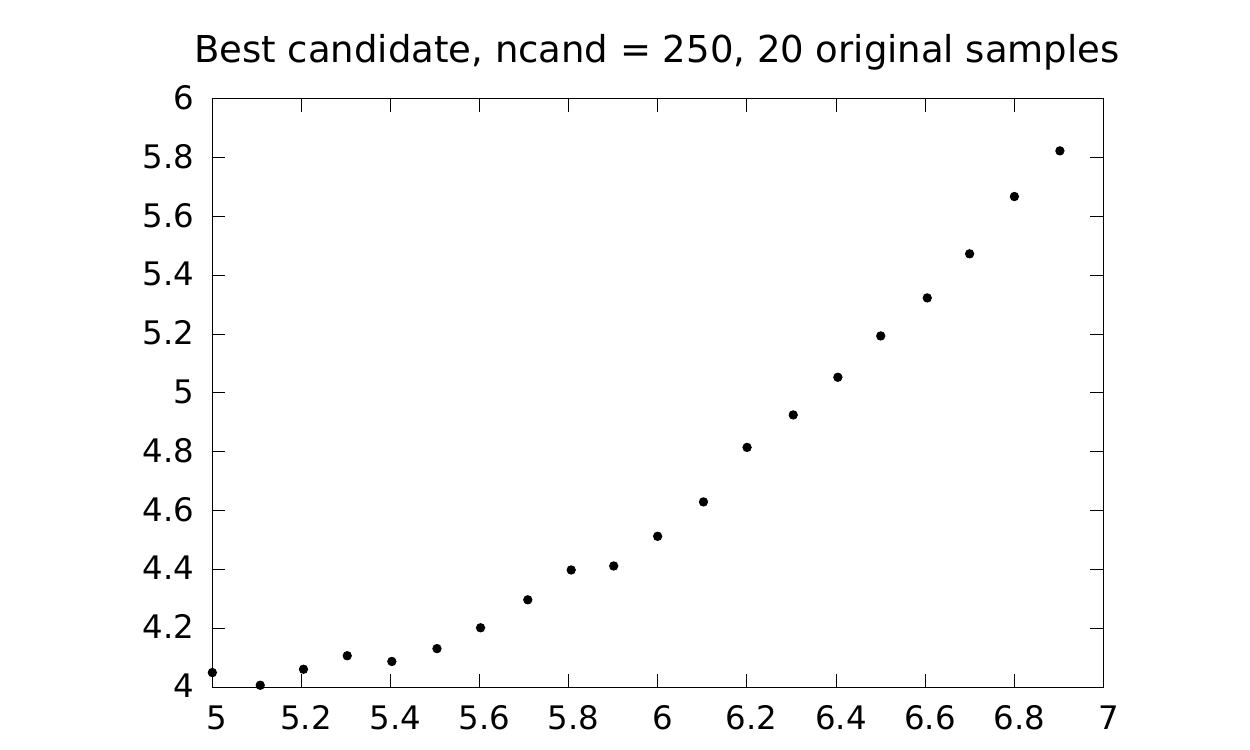} &
\includegraphics[trim = 1.5cm 0cm 1.0cm 0cm, clip = true,width=0.3\textwidth]{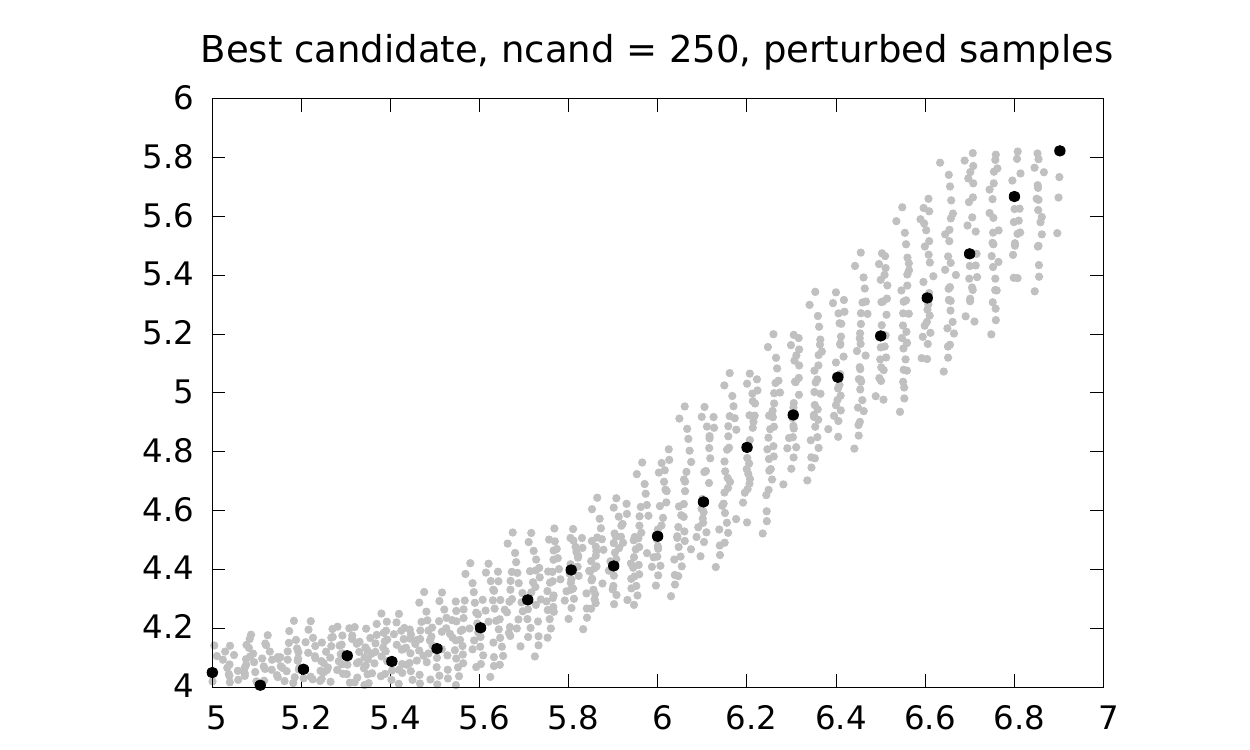} &
\includegraphics[trim = 1.5cm 0cm 1.0cm 0cm, clip = true,width=0.3\textwidth]{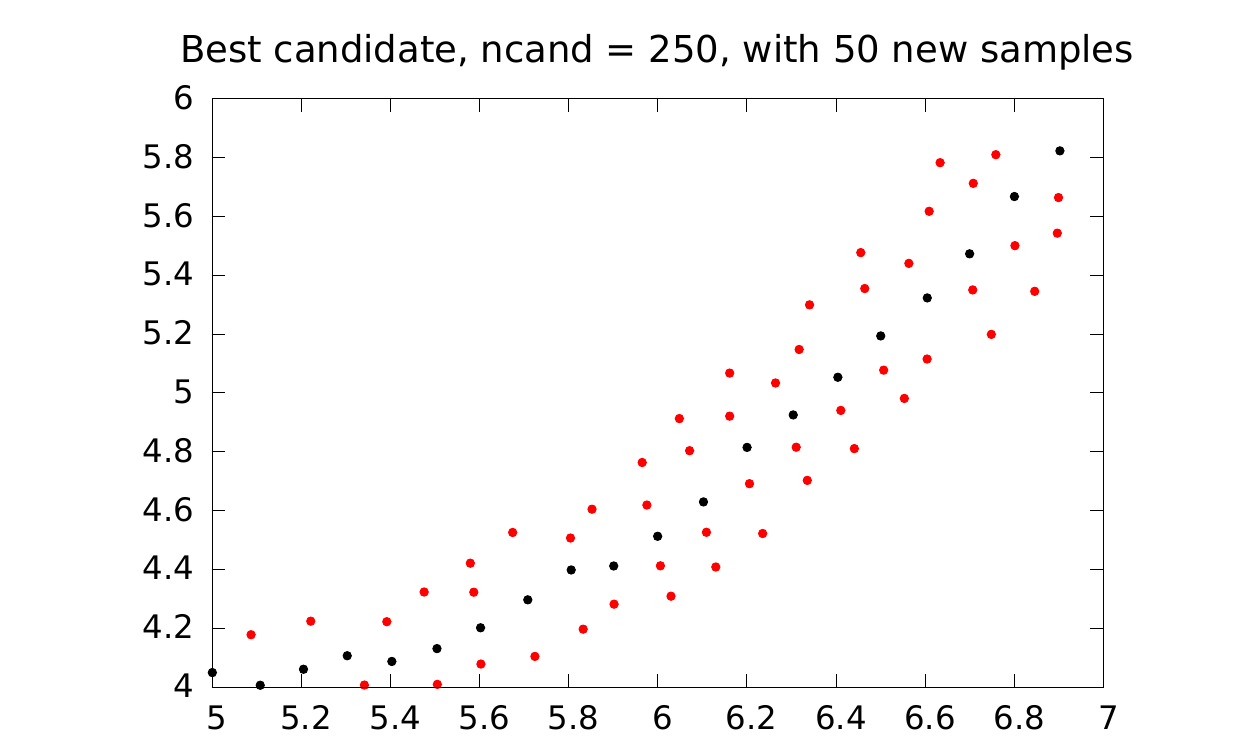} \\
(a) & (b) & (c)\\
\end{tabular}
\vspace{-0.2cm}
\caption{(a) 20 samples, in two dimensions, indicating the region
  around which we want to add new samples. (b) Additional 1000
  candidate samples (in gray) generated in a small region around each
  curve sample.  (c) 50 new space-filling samples (in red) selected
  from the 1000 candidate samples. These 70 samples together spatially
  fill the curved region and can be used to build a good-quality
  surrogate in the region around the original 20 samples.}
\label{fig:samples_append_non_rect}
\end{figure}

We illustrate the challenges in this task using the curve described by
20 samples shown in Figure~\ref{fig:samples_append_non_rect}(a).  As
the curve occupies a small part of the entire domain, identifying
possible samples by generating a large number of samples in the
domain, and keeping only those close to the curve, is wasteful. We
also want the new samples to be space filling in the region around the
curve, while allowing for the option to include or exclude the curve
samples themselves. 

One solution is to first generate many candidate samples randomly in a
small region around each curve sample
(Figure~\ref{fig:samples_append_non_rect}(b)); this can be done using
random, GreedyFP, or BC sampling. The size of the region around each
curve sample can be chosen to reflect the maximum distance, as defined
by the user, of the new samples from the existing ones.  Then, from
the candidates generated, we can select the requisite number of samples that
are farther away from each other and optionally, the curve samples
(Figure~\ref{fig:samples_append_non_rect}(c)).
In Figure~\ref{fig:samples_append_non_rect}, we use the BC algorithm
to generate 50 candidate samples in a rectangular region around each
of the 20 curve samples. The half-width of the rectangular region in
each dimension is defined to be equal to 3\% of the coordinate value
of the curve sample. We then select 50 samples from the resulting 1000
candidates.  This approach also works in higher dimensions, though it
is harder to visualize the results. We also observe that it is not
straightforward to support this capability with the LHS, CVT or
Poisson disk algorithms.

%
\subsection{Sampling when domain extents are modified}
\label{sec:adapt_modifydomains}
%

In both surrogate modeling and hyperparameter optimization, we start
with a domain where we expect to find the solution and create an
initial set of samples in this domain. However, this initial set might
indicate that the domain extents selected were incorrect and we need
to either shrink or expand the domain by removing or adding new
samples, while retaining any original samples that lie in the new
domain. Ideally, the resulting sampling should still satisfy the
random, uniform coverage of the space.

Some sampling algorithms support this functionality better than
others. For example, for all algorithms, we can shrink the domain by
removing points near a boundary, though in the case of LHS, the
remaining samples would no longer satisfy the Latin property.  Adding
samples along one or more dimensions, while retaining the already processed 
samples, is more challenging. In random sampling, there is no
guarantee that the new samples will be in the newly expanded region
unless we include a constraint to ignore samples not in this region.
In grid sampling, we can expand the grid, though we cannot expand by
an arbitrary number of samples. For LHS, expanding the domain along
one or more dimensions is problematic as the algorithm generates a set
of samples by dividing the extent of each dimension into the same
number of bins. So, in a two dimensional problem with $N$ samples, if
we expand the extent along one of the dimensions to add $M$ new
samples, there is no guarantee that when the other dimension is
divided into $(M+N)$ bins, we can insert the $M$ samples so that the
combined $(M+N)$ samples satisfy the Latin property.

\begin{figure}[!htb]
\centering
\begin{tabular}{ccc}
\includegraphics[trim = 1.0cm 0cm 1.0cm 0cm, clip = true,width=0.3\textwidth]{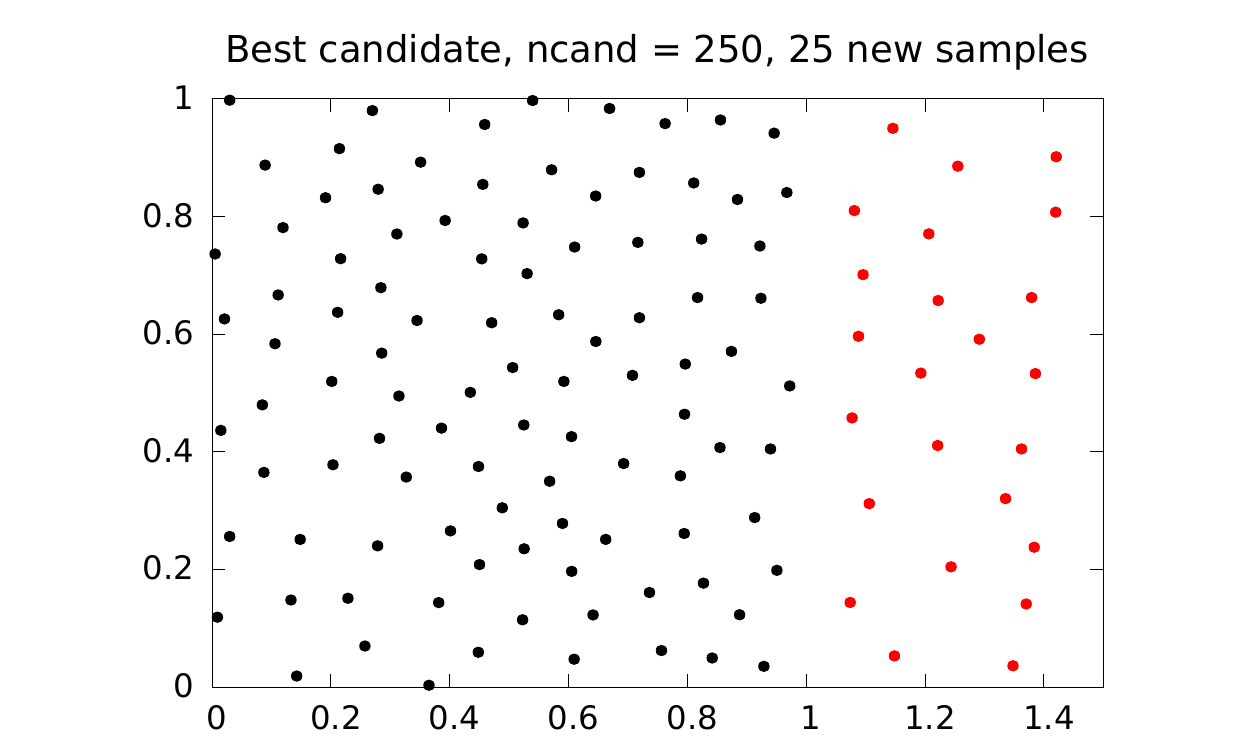} &
\includegraphics[trim = 1.0cm 0cm 1.0cm 0cm, clip = true,width=0.3\textwidth]{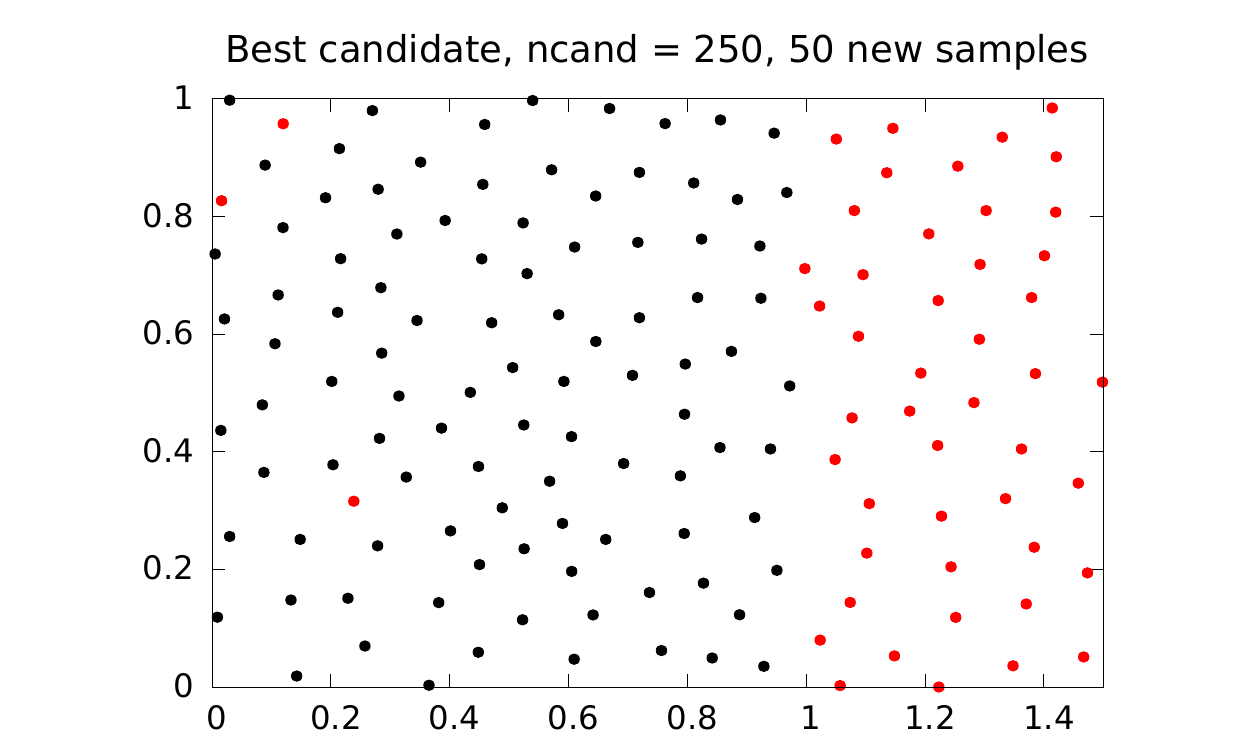} & 
\includegraphics[trim = 1.0cm 0cm 1.0cm 0cm, clip = true,width=0.3\textwidth]{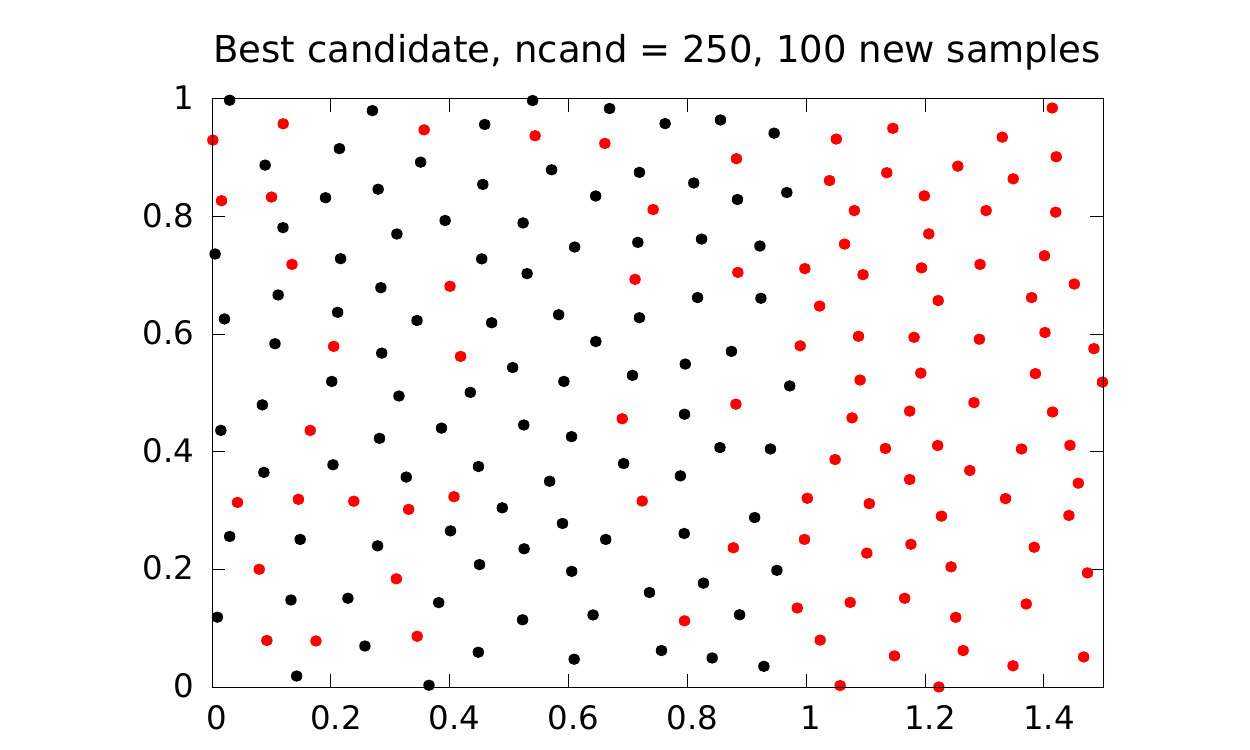} \\
(a) & (b) & (c) \\
\end{tabular}
\vspace{-0.2cm}
\caption{Expanding the domain from $[0,1] \times [0,1]$ to $[0,1.5]
  \times [0,1]$ to add new samples using the BC algorithm: 100 samples
  (in black) in the original domain. (a) The first 25 new samples (in
  red) are all in the new region. (b) With 50 new samples, the sample
  density in the right third of the domain approaches that of the old
  region. (c) With 100 new samples, the sampling density in the
  combined region increases compared to the original.}
\label{fig:samples_expand_domain}
\end{figure}

Expanding the domain to add new samples using the CVT algorithm has
the same issues as incrementally adding new samples to the original
domain (Section~\ref{sec:adapt_incremental}). The remaining three
algorithms --- Poisson disk, GreedyFP, and BC --- are however more
amenable to a change in the domain extents. The Poisson disk algorithm
can start with all existing samples as active samples and then expand
preferentially around the points at the boundary being expanded, which
will form the new ``front''. In both the GreedyFP and BC algorithms,
we can generate candidate samples only in the new region, and use
all existing samples to identify the farthest one.  However, depending
on the number of new samples added, this could result in different
sampling densities in the new expanded region.  In
Figure~\ref{fig:samples_expand_domain}, using BC sampling as an
example, we show how we can maintain the sampling density in both
regions. We start with 100 samples in $[0,1] \times [0,1]$ and expand
the region to $[0,1.5] \times [0,1]$. As we incrementally add samples
to this new domain, the initial new samples are all in the new region.
But once the sampling density in the expanded region approaches that
of the original region, the new samples appear in the combined region.

%
\subsection{Sampling in high dimensional spaces}
\label{sec:adapt_highdim}
%

The tasks of surrogate modeling and hyperparameter optimization
usually involve high-dimensional spaces, where we want to sample the
input parameter space of the simulation for which the surrogate model
is being developed, or the parameter space of an algorithm for which
we want to determine the optimum values that minimize the error in the
algorithm for a given data set. The number of parameters could number
in the tens or more.

All the sampling techniques considered can be easily extended to
higher dimensions, at least from an algorithm viewpoint.  Some
algorithms, such as the Poisson disk or the exact farthest point
algorithm based on the Voronoi diagram (Section~\ref{sec:algo_fps}),
were motivated by two-dimensional images or three-dimensional objects
and can be difficult to extend to high dimensions easily.  Despite
this, some algorithms, such as the spoke-dart method that shoots rays
from existing samples and selects points along these
rays~\cite{mitchell2018:spokedart}, have been proposed to extend
Poisson disk to high dimensions. The drawback of having to convert the
number of samples desired into a distance in high dimensions for the
disk radius, however, remains, as does the challenge of extending
efficient implementations from two and three dimensions to higher
dimensions (Section~\ref{sec:algo_poisson}).

%
\subsection{Sampling a subset from an existing set of samples}
\label{sec:adapt_subset}
%

One of the first steps in many data analysis efforts is the
exploration of data.  When the data sets are too large to fit into
memory or to visualize in a meaningful way, a common solution is to
work with a subset of the data and explore this subset to understand
the characteristics of the data. To avoid any bias, this subset must
be randomly sampled from the existing data points. For large data
sets, the sampling algorithm must process the data set without loading
all of it into memory and ideally, must be able to extract the samples
in one pass through the data set.

Given these constraints, some of the sampling algorithms considered in
this paper do not lend themselves to easy modification. While CVT,
based on the k-means algorithm, would need to use the k-medioids
algorithm~\cite{schubert2021:kmediods} to select actual data
instances, the algorithm is iterative, not a one pass algorithm. Grid
sampling is also not applicable as there may not be any points at
the grid locations or within a cell.  For Poisson disk sampling, it
would be a challenge to identify the instances in a torus, without
reading in the whole data set, or using multiple passes through the
data. LHS is also not an option as there may be empty bins along a
dimension, making the initial generation of an LHS sampling, and
subsequent optimization through interchanges, challenging.

Both the GreedyFP and the BC algorithms appear, at first glance, to allow
the selection of a subset of instances as we can consider all of the
instances to be potential candidate samples. However, in GreedyFP, for
each new sample selected, we need to evaluate the distance of all
candidate samples to the samples already selected, which would require
multiple passes through the data set. In contrast, as the BC
algorithm considers only a small number of candidate samples at a
time, we can read in a small segment of the data that fits into
memory, consider all instances in this segment as possible candidates,
and select one as the next sample in the subset (see Algorithm 3
in~\cite{kamath19:bigdata}). Random sampling provides an even faster
solution as we can just generate a set of random indices and extract
the corresponding instances in one pass through the data set.
However, unlike the subset extracted using the BC algorithm, the
subset samples in random sampling will not be far apart from each
other.

%
\subsection{Efficient real-time generation of large number of samples}
\label{sec:adapt_largenumber}
%

In some problems, instead of a small number of carefully selected
samples, we may want to generate a large number of samples that
randomly cover the domain uniformly, for example, when we want to use
a surrogate model to predict values of an output at sample points in a
domain. Or, we may need fast generation of samples to satisfy time
constraints. To meet both these requirements, a sampling algorithm
should ideally be computationally efficient. While this is true of
some sampling algorithms, such as random sampling, others may need
modification.

There are two ways to efficiently generate a large number of samples
with desired properties. The first is through parallelization, which
may be helpful in the more time consuming sampling algorithms, such as
CVT~\cite{ju2002:parallelcvt} and Poisson disk
sampling~\cite{wei2008:parallelpd,ying2013:parallelpd,brandt2019:gpufp}.
Another approach is to create small sub-domains with samples that
satisfy the desired constraint, and then use the sub-domains as tiles
to cover the entire domain.  If we have a single tile, repeating it to
cover the domain would result in a periodic structure in the samples.
Instead, multiple tiles are generated and placed next to each other
using the concept of Wang
tiles~\cite{kopf2006:recursivewang,lagae2008:comparisonpd} that allows
for the generation of infinite non-periodic tilings.  Alternately, the
tiles can be placed with some distance separating them, and the region
in between the tiles filled with sample points in a follow-on
step~\cite{patel2018:fastpoissondisk}. The tiling approach has been
used mainly for sampling in two dimensions; its extension to higher
dimensions may require ingenuity.

%
\subsection{Summary of adaptability of different algorithms}
\label{sec:adapt_summary}
%

We next summarize our observations regarding the
adaptability of various algorithms to meet the sampling needs of 
surrogate modeling, hyperparameter optimization, and data analysis.
Table~\ref{tab:adapt_summary} lists the different algorithms
considered and identifies which of the desired properties of sampling
each can support. We do not explicitly list the hybrid BC-GreedyFP
algorithm as its capabilities are similar to those of the BC and
GreedyFP algorithms.

\renewcommand{\arraystretch}{1.5}
\begin{table}[htb]
\centering
\begin{tabular}{|p{4.9cm}|p{.6cm}|p{.6cm}|p{.6cm}|p{0.6cm}|p{0.6cm}|p{0.6cm}|p{0.6cm}|}
\hline
\backslashbox{\strut Adaptability \\ to}{\strut Sampling \\ method} & 
      \rotatebox[origin=c]{90}{Random} & 
      \rotatebox[origin=c]{90}{\parbox[c]{1.8cm}{Grid  (stratified)}} & 
      \rotatebox[origin=c]{90}{LHS} & 
      \rotatebox[origin=c]{90}{CVT} & 
      \rotatebox[origin=c]{90}{\parbox[c]{2cm}{\centering Poisson disk}} & 
      \rotatebox[origin=c]{90}{GreedyFP} & 
      \rotatebox[origin=c]{90}{BC} \\
\hline
Arbitrary number of samples & \checkm & \xmark & \checkm & \checkm & \xmark & \checkm & \checkm \\
\hline
Randomness & \checkm & \checkm & \checkm & \checkm & \checkm & \checkm & \checkm \\
\hline
Space filling property & \xmark & \xmark & \checkm&  \checkm & \checkm & \checkm & \checkm \\
\hline
Progressive sampling & \checkm & \xmark & \xmark & \xmark & \xmark &  \checkm & \checkm \\
\hline
Incremental sampling & \checkm & \xmark & \xmark & \xmark & \xmark &  \checkm  & \checkm \\
\hline
Non-collapsing property & \checkm & \checkm & \checkm & \checkm & \checkm &  \checkm  & \checkm \\
\hline
Probability density function & \checkm & \xmark & \checkm & \checkm & \xmark &  \checkm  & \checkm \\
\hline
Non-rectangular regions & \checkm & \xmark & \xmark  & \checkm & \checkm &  \checkm  & \checkm \\
\hline
Modified domain extents & \checkm & \xmark & \xmark & \xmark & \checkm & \checkm & \checkm   \\
\hline
High dimensional sampling & \checkm & \checkm & \checkm & \checkm & \checkm &  \checkm  &  \checkm \\
\hline
Subset sampling in one pass & \checkm & \xmark & \xmark & \xmark & \xmark & \xmark & \checkm \\
\hline
Large number of samples & \checkm & \checkm & \checkm & \checkm & \checkm & \checkm & \checkm \\
\hline
\end{tabular}
\caption{Summary of the adaptability of various sampling algorithms to
  the requirements of surrogate modeling, hyperparameter optimization, and
  data analysis. The hybrid BC-GreedyFP algorithm is not listed
  separately as it combines the capabilities of both methods.}
\label{tab:adapt_summary}
\end{table}

Based on this analysis, we conclude that in terms of their ability to
meet our needs, some of the more popular algorithms, such as LHS and
CVT, do not possess some key capabilities, such as progressive and
incremental sampling. In contrast, some of the lesser-known methods,
such as GreedyFP and BC sampling, that are relatively unknown in the
surrogate modeling and hyperparameter optimization communities, fare
much better. Surprisingly, the very simple random sampling, despite its poor
space-filling capability, supports all other requirements. All the
algorithms considered were relatively simple to implement, have few
parameters, and did not require extensive fine tuning of these
parameters. We next describe a more quantitative comparison of the
better adapted algorithms, focusing on their performance on larger
number of samples in higher dimensions, including the time to
generate the samples.

%
\section{Quantitative comparison of sampling algorithms}
\label{sec:comparison}
%

In this section, we complement the qualitative comparison of sampling
algorithms from Section~\ref{sec:adapt_summary} with a more
quantitative comparison. We focus mainly on the better performing
algorithms from Table~\ref{tab:adapt_summary} --- Greedy-FP, BC, and
hybrid BC-GreedyFP (abbreviated as the hybrid algorithm) --- as well
as random sampling, due to its simplicity, and LHS, due to its
popularity in surrogate modeling. We excluded CVT, despite its
excellent space-filling capabilities, mainly due to its inability to
support progressive and incremental sampling. We also excluded Poisson
disk for the same reason and for requiring the specification of the
minimum distance between samples as an input parameter instead of
number of samples.

Prior studies comparing different sampling methods have differed in
terms of the methods considered, the task for which the
samples are generated, and the metric used for comparison. For
example, Cho et al.~\cite{cho2017:samplingcomparison} compared optimal
LHS, CVT, and quasi-random sequences. For metrics, they used
space-filling and projective properties of the samples generated, as
well as the root-mean-squared error (RMSE) of a Gaussian process
surrogate created for several test functions with dimensions ranging
from two to ten. Their study indicated that CVT, which focuses on
space filling, performs better than optimal LHS which focuses first on
the projective property and then uses an optimality criterion to make
the sampling more space filling. The RSME results for the Gaussian
process surrogate using a one-shot sampling approach indicated that
the best sampling algorithm depended on the properties of the
function, as would be expected.

In another paper, Romero et al.\cite{romero2006:latinized} also
conducted a similar comparison with the same set of sampling
algorithms, and the Latinized variant of CVT. For their test problem
and choice of metrics, they found Hammersley sampling to be the
consistently best performer, with Latinized CVT, CVT, and LHS as the
next best in that order. However, they observe that Hammersley does
not allow incremental addition of sample points and might suffer from
the spurious correlation effects that have been observed in
quasi-Monte-Carlo methods~\cite{morokoff1994:quasirandom}. They also
acknowledge that the non-incremental nature of CVT can reduce its
potential advantage over simple random sampling and some of the
quasi-random sequences.

In their work on image sampling with quasicrystals, Grundland et
al.~\cite{grundland2009:quasicrystal} introduced a new sampling
algorithm that produces space-filling, non-periodic sets of points
that are uniformly discrete and relatively dense. They compared their
approach with a range of sampling methods, mainly from graphics,
including farthest point, jittered sampling, quasi-random sampling,
and random sampling. Given their focus on sampling for use in
visualization, they compared the results visually, using qualitative
factors such as accurate reconstruction, progressive refinement,
uniform coverage, isotropic distribution, and blue noise spectrum.
They recommend a blue noise sampling strategy, such as the farthest
point sampling (Section~\ref{sec:algo_fps}) for general use in image
representation as it performs well on their comparison criteria. Their
proposed method based on quasicrystals was a close second, and was
significantly simpler to implement and calculate than the farthest
point algorithm that required complex data structures for the Voronoi
diagram.

Lagae and Dutr\'e~\cite{lagae2008:comparisonpd} also compared a range
of algorithms specifically for generating Poisson disk samples using
radial statistics and spectral properties, which are important
criteria for such distributions. Though they considered sampling in
graphics and visualization, they noted that different applications
have different requirements, such as real-time performance or the
generation of a very large number (100,000 or more) of samples, and
recommend that for a given application, certain methods should not be
discarded a priori.

Finally, Loyola et. al~\cite{loyola2016:smartsampling} and Garud et.
al~\cite{garud2018:smartsampling} have both used a variety of test
functions to compare ways in which new samples can be incrementally
added to an existing set, in a concept they call ``{\it smart
 sampling}''.

%
\subsection{Metrics used for comparison}
\label{sec:metrics}
%

There are several metrics we can use to compare our five sampling
algorithms. Our selection of metrics was guided by the comparison
paper of Romero et al.\cite{romero2006:latinized}, who observed that
their results are somewhat tied to the specific metric used and
cautioned that empirical studies are only point glimpses of the
relative accuracy tendencies of one method over another under a very
specific set of conditions. We therefore chose not to compare our
sampling algorithms based on the accuracy of surrogates on test
functions because such comparisons are very dependent on the test
function, number of samples, the choice of one-shot or sequential
sampling, and the randomness in the samples generated. We also chose
not to focus on the projective property in one dimension as it can not
only be addressed by Latinizing an initial set of samples, but is
difficult to enforce when we want to add new samples to existing ones
or sample in strangely shaped domains.

Instead, we considered a two step approach to compare the suitability
of different sampling algorithms to meet our needs in surrogate
modeling, hyperparameter optimization, and data analysis. First, as
described in the previous section, we used a qualitative comparison to
determine whether we could modify an algorithm to meet each specific
need. This comparison indicated that a key desirable property is that
the samples be placed randomly, far apart from each other, in a
space-filling manner. Now, in the second step, we evaluate the quality
of samples generated using three metrics that capture this property in
different ways:

\begin{itemize}

\item{\bf Nearest neighbor distance:} A simple way to intuitively
  determine whether the samples generated by an algorithm are spread
  out or not, is to obtain the distance of each sample to its nearest
  neighbor. In our work, we consider the Euclidean distance, and
  calculate the average, minimum, and maximum values of the nearest
  neighbor distances across the samples in the set.
  A good algorithm will generate samples that are all roughly the same
  distance from each other, that is, the range of distances to the
  nearest neighbor of each sample in the set should be small. Also,
  larger minimum and average distances to the nearest neighbor are
  preferred.

\item{\bf $\phi_p$ criterion:} This metric~\cite{morris95:design}
  extends the maximin distance criterion, which maximizes the minimum
  distance between samples in a sample set, to consider all distances
  between samples in a set.  It defines the quantity $\phi_p$

  \begin{equation}
    \phi_p = \left[ \sum_{1 \leq i < j \leq N} (1/d_{ij})^p \right]^{1/p}
    \label{eqn:phi50}
  \end{equation}

  where $d_{ij}$ is the Euclidean distance between samples $i$ and $j$
  and $p$ is a positive integer, which we set to
  50~\cite{jin2005:optimallhs}.  For very large $p$, the $\phi_p$
  criterion is equivalent to the maximin distance criterion. A
  sampling is called $\phi_p$-optimal if it minimizes $\phi_p$; thus,
  we are interested in smaller values of $\phi_p$.

\item{\bf Centered $L_2$ discrepancy criterion (CL2):} The $L_2$
  discrepancy is a measure of the non-uniformity of a sampling.
  Following~\cite{jin2005:optimallhs}, we use the centered $L_2$
  discrepancy defined by Hickernell~\cite{hickernell1998:cl2} as follows:
  \begin{equation}
    \begin{split}
      CL2 = \Bigg\{ \Big( \frac{13}{12} \Big)^d - 
      & \frac{2}{N} \sum_{i=1}^{N} \prod_{k=1}^d 
      \bigg[ 1 + 0.5 \Big( | x_{ik} -0.5 | - | x_{ik} - 0.5 |^2 \Big) \bigg]  + \\
      & \frac{1}{N^2} \sum_{i,j=1}^{N} \prod_{k=1}^d  
      \bigg[ 1 + 0.5 \Big( | x_{ik} -0.5 | + | x_{jk} - 0.5 | - |x_{ik} - x_{jk} | \Big) \bigg]  
      \Bigg\}^{1/2}
    \end{split}
    \label{eqn:cl2}
  \end{equation}

  We are interested in a sampling with a lower value of discrepancy.

\end{itemize}

These three metrics were selected to evaluate the space-filling
property of the sample set generated by an algorithm. They may not
necessarily capture all desirable properties, or help in identifying
any drawbacks of a sample set, such as a structured distribution of
samples or large empty spaces that are found in some two-dimensional
projections of quasi Monte-Carlo sample
sets~\cite{morokoff1994:quasirandom,chi2005:halton}.

%
\subsection{Experimental results and discussion}
\label{sec:results}
%

We next use the metrics identified in Section~\ref{sec:metrics} to
evaluate the five sampling algorithms: random, LHS, GreedyFP, BC, and
the hybrid BC-GreedyFP, using the parameters listed in
Table~\ref{tab:params} for each algorithm.  In this section, we
refer to the GreedyFP, BC, and the hybrid BC-GreedyFP algorithms
collectively as the {\it farthest point algorithms} as they are all
based on the identification of the farthest point, making their
behavior very similar to each other, but different from the other two
algorithms considered.

\renewcommand{\arraystretch}{1.2}
\begin{table}[htb]
  \begin{center}
    \begin{tabular}{|l|l|} 
      \hline
      Sampling method & Parameters used \\
      \hline
      LHS & {\it nTries} = 10; {\it nInterchanges} = 100 \\
      \hline
      GreedyFP & {\it scale} = 10 \\
      \hline
      Best candidate & {\it nCand} = 250 \\
      \hline
      Hybrid BC-GreedyFP & {\it scale} = 10; {\it refreshCount} = 100 \\
      \hline
    \end{tabular}
  \end{center}
  \vspace{-0.2cm}
  \caption{Parameters used for the sampling algorithms considered in the quantitative comparison.}
  \label{tab:params}
\end{table}

\begin{figure}[!htb]
\centering
\begin{tabular}{ccc}
\includegraphics[trim = 0.0cm 0cm 0.0cm 0cm, clip = true,width=0.31\textwidth]{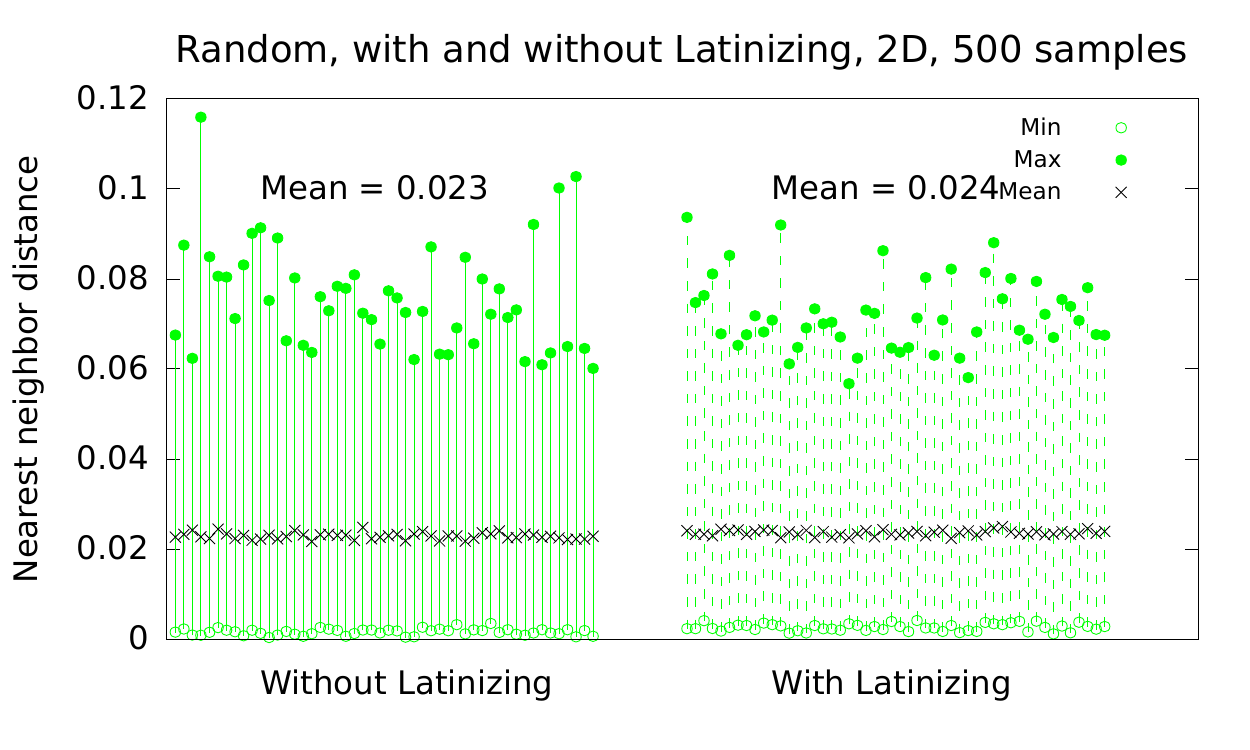} &
\includegraphics[trim = 0.0cm 0cm 0.0cm 0cm, clip = true,width=0.31\textwidth]{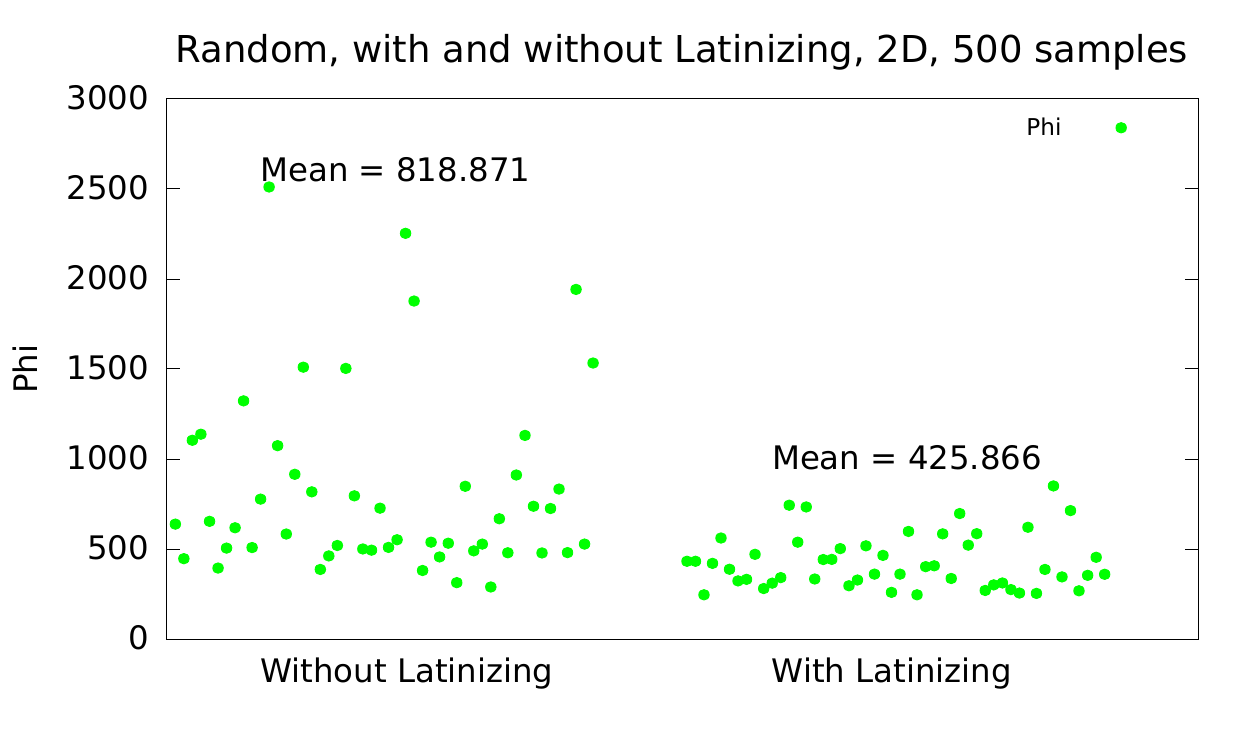} &
\includegraphics[trim = 0.0cm 0cm 0.0cm 0cm, clip = true,width=0.31\textwidth]{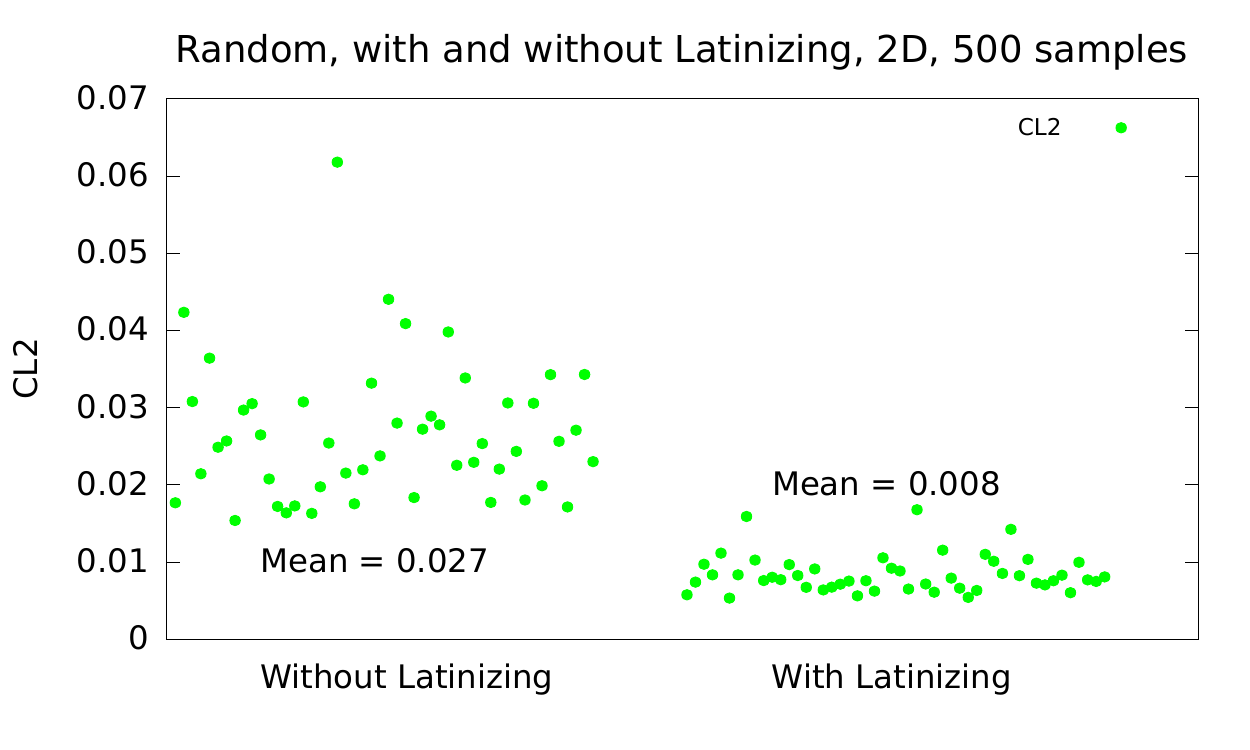} \\
\includegraphics[trim = 0.0cm 0cm 0.0cm 0cm, clip = true,width=0.31\textwidth]{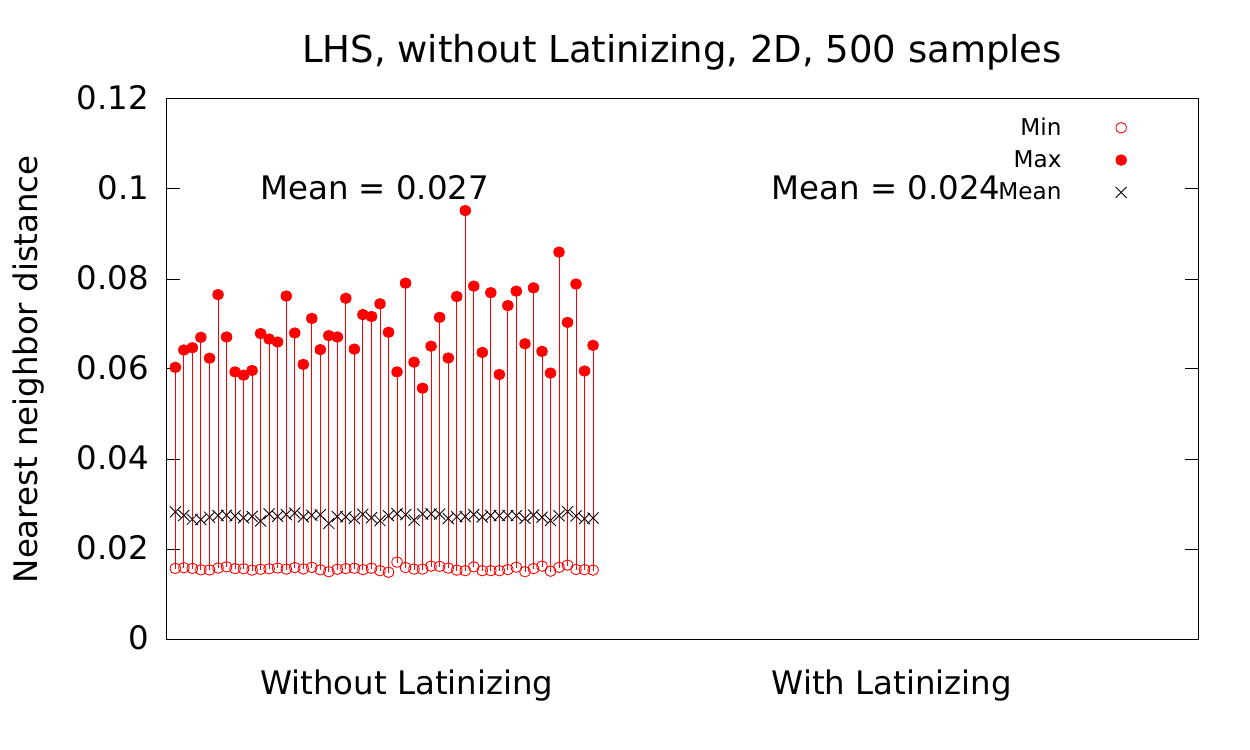} &
\includegraphics[trim = 0.0cm 0cm 0.0cm 0cm, clip = true,width=0.31\textwidth]{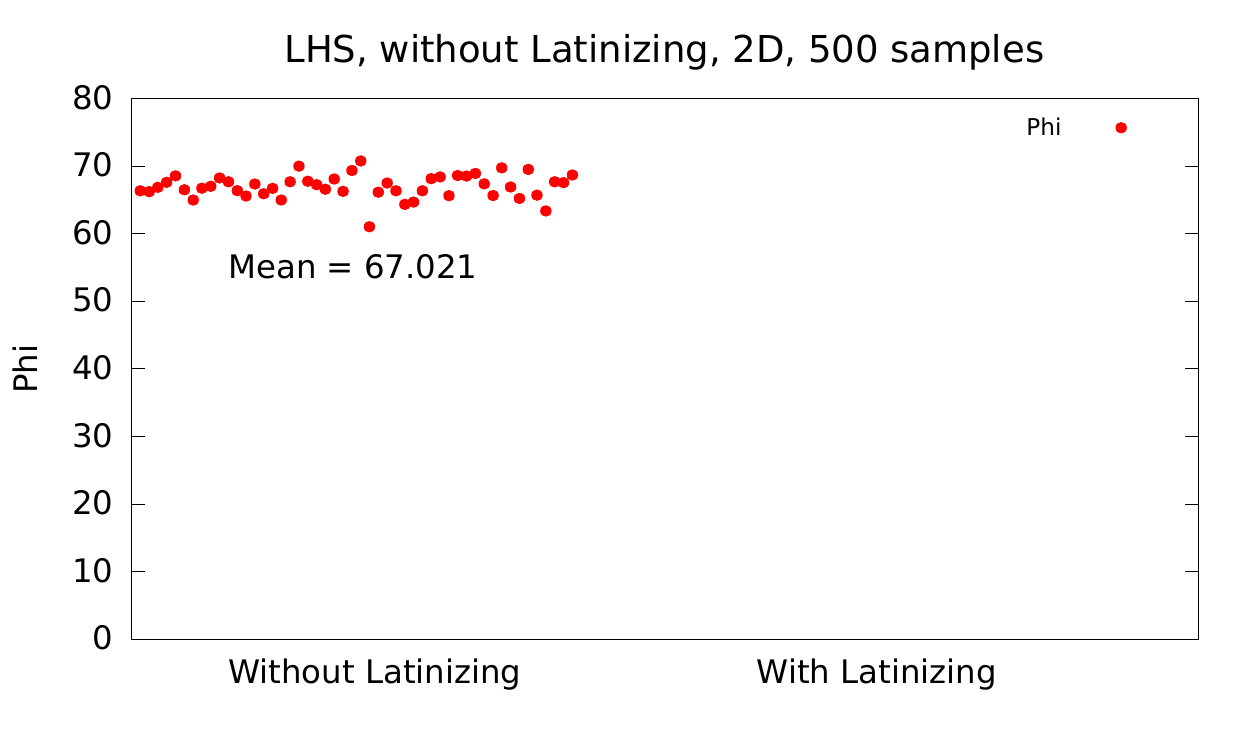} &
\includegraphics[trim = 0.0cm 0cm 0.0cm 0cm, clip = true,width=0.31\textwidth]{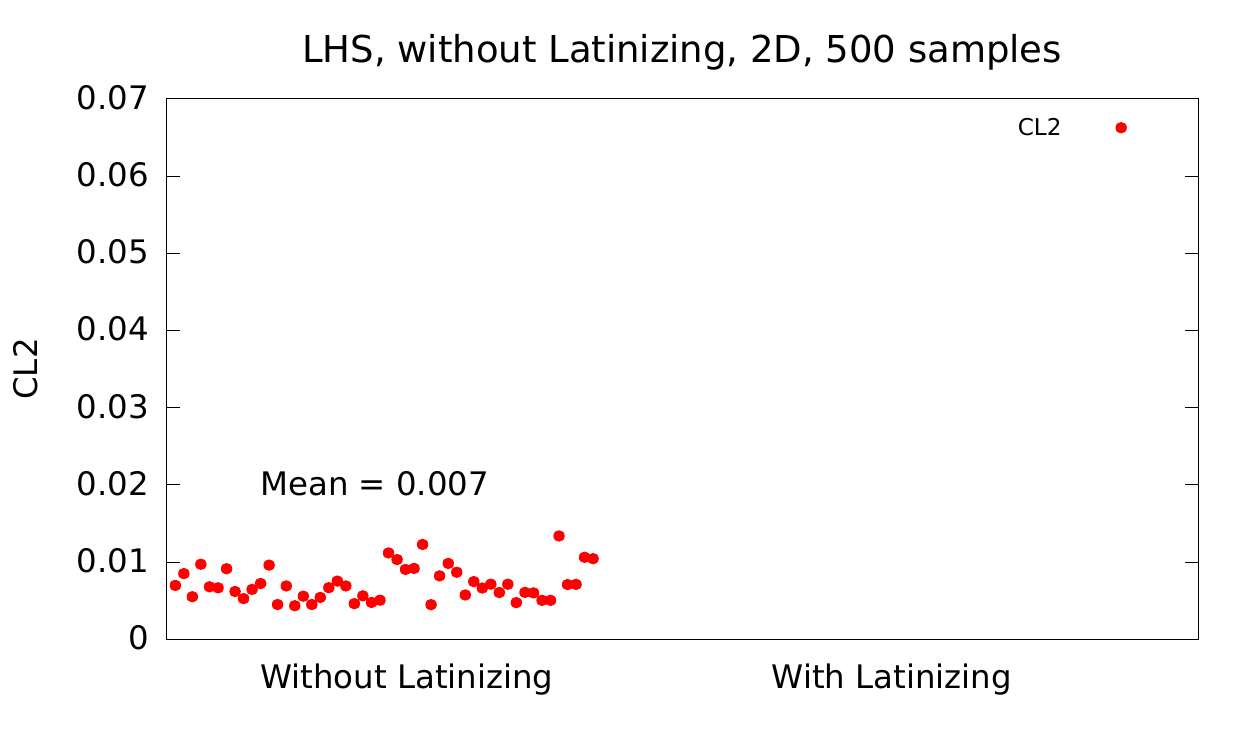} \\
\includegraphics[trim = 0.0cm 0cm 0.0cm 0cm, clip = true,width=0.31\textwidth]{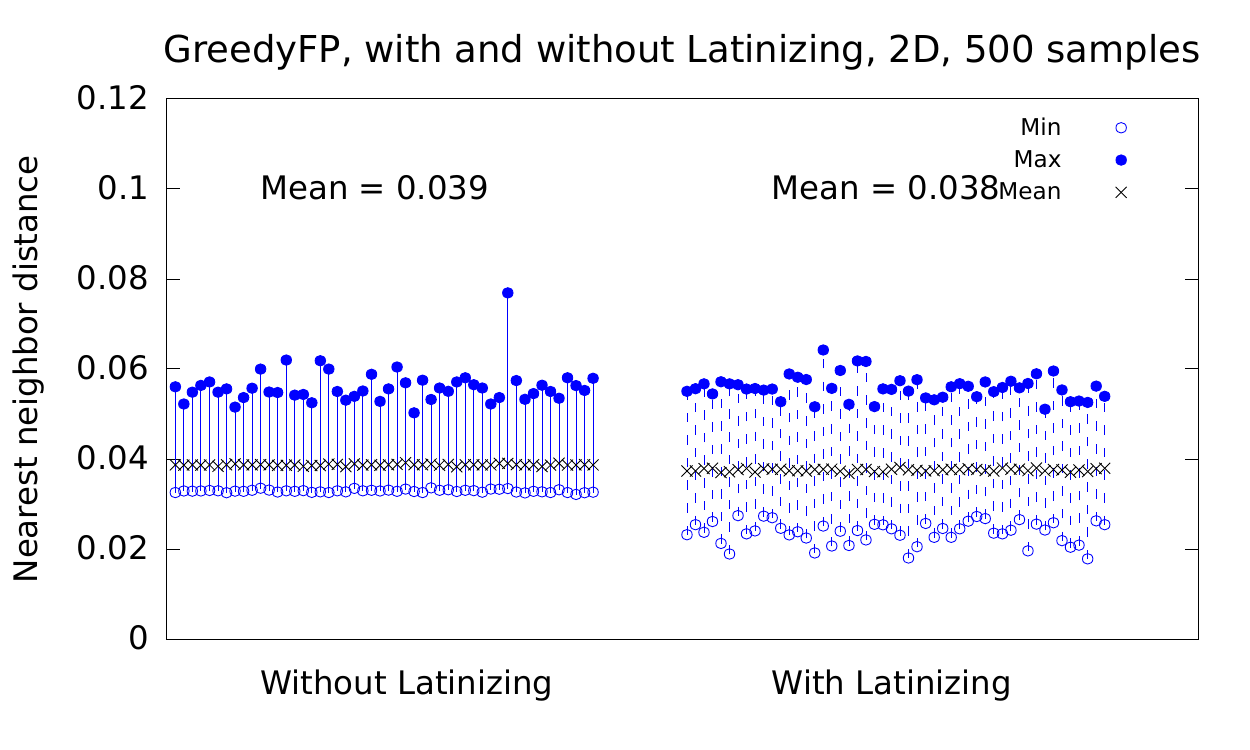} &
\includegraphics[trim = 0.0cm 0cm 0.0cm 0cm, clip = true,width=0.31\textwidth]{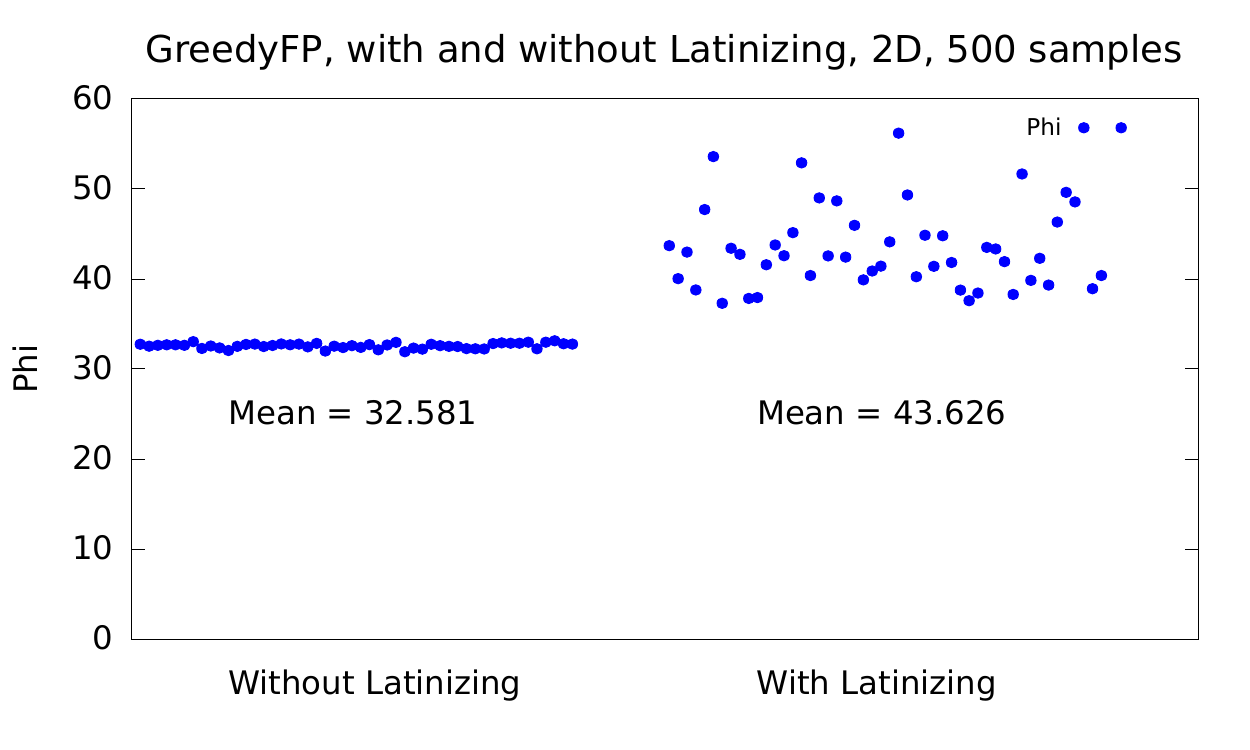} &
\includegraphics[trim = 0.0cm 0cm 0.0cm 0cm, clip = true,width=0.31\textwidth]{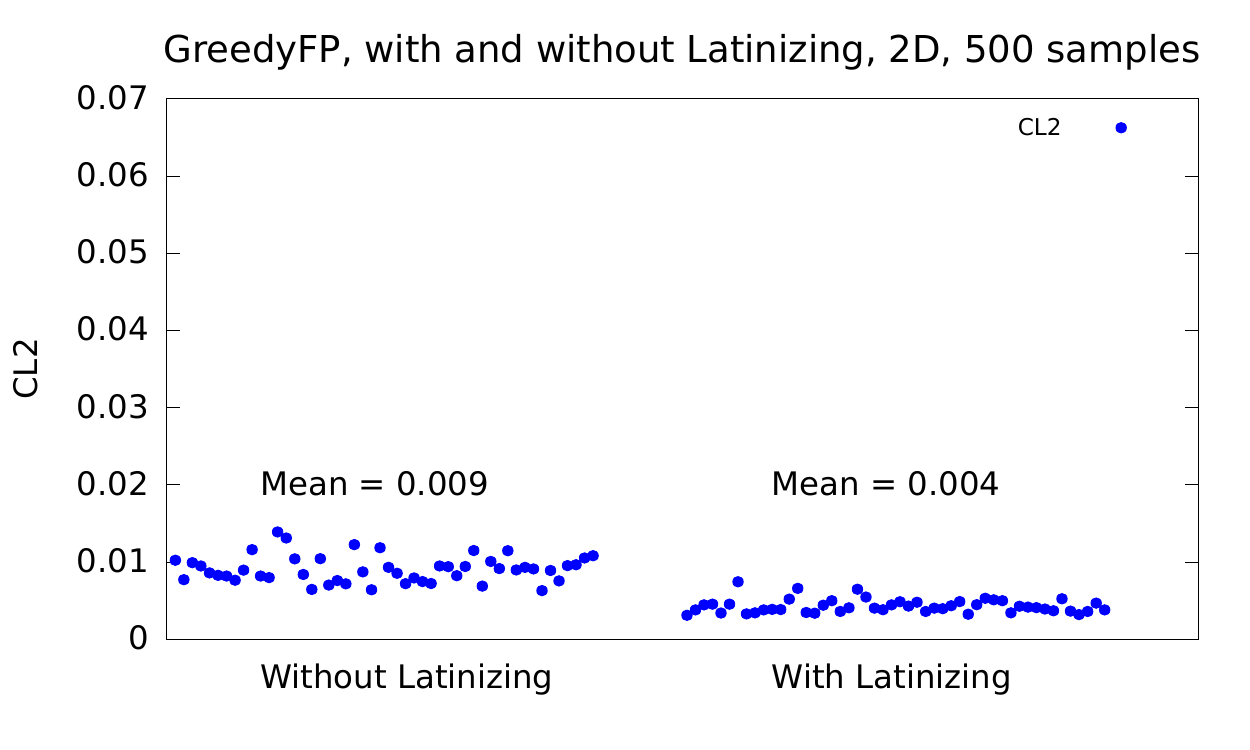} \\
\includegraphics[trim = 0.0cm 0cm 0.0cm 0cm, clip = true,width=0.31\textwidth]{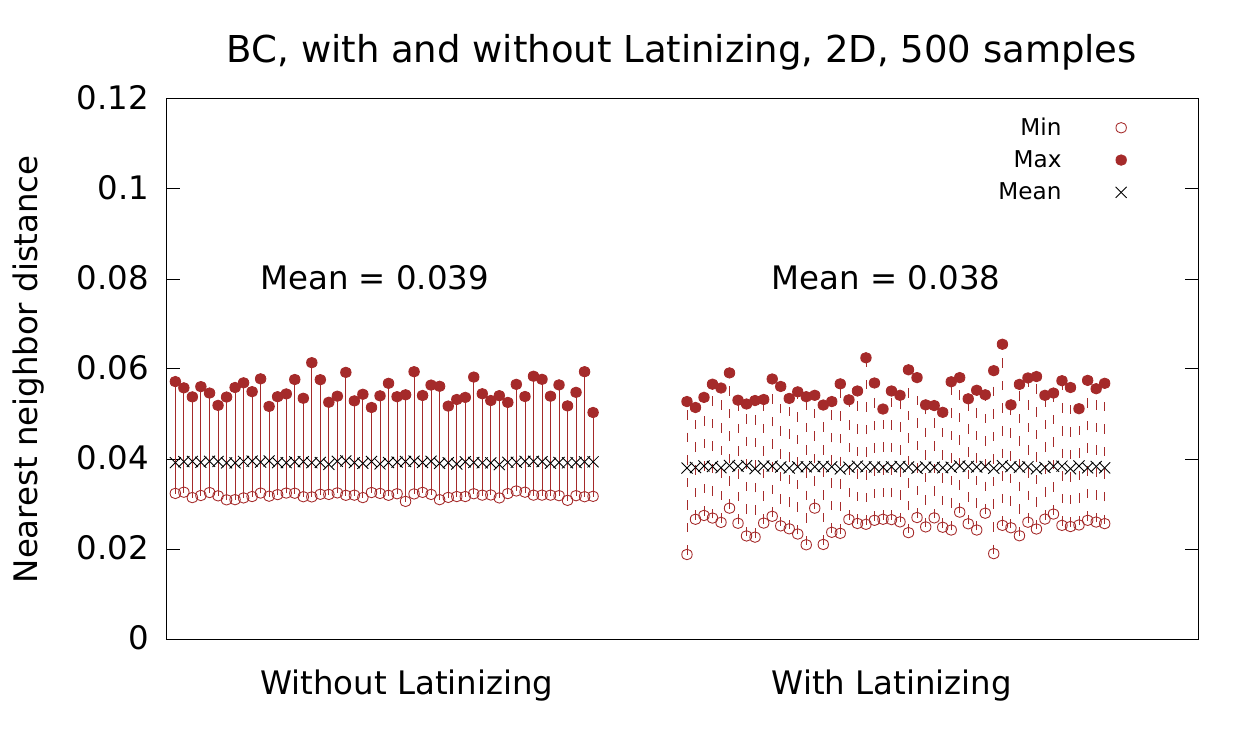} &
\includegraphics[trim = 0.0cm 0cm 0.0cm 0cm, clip = true,width=0.31\textwidth]{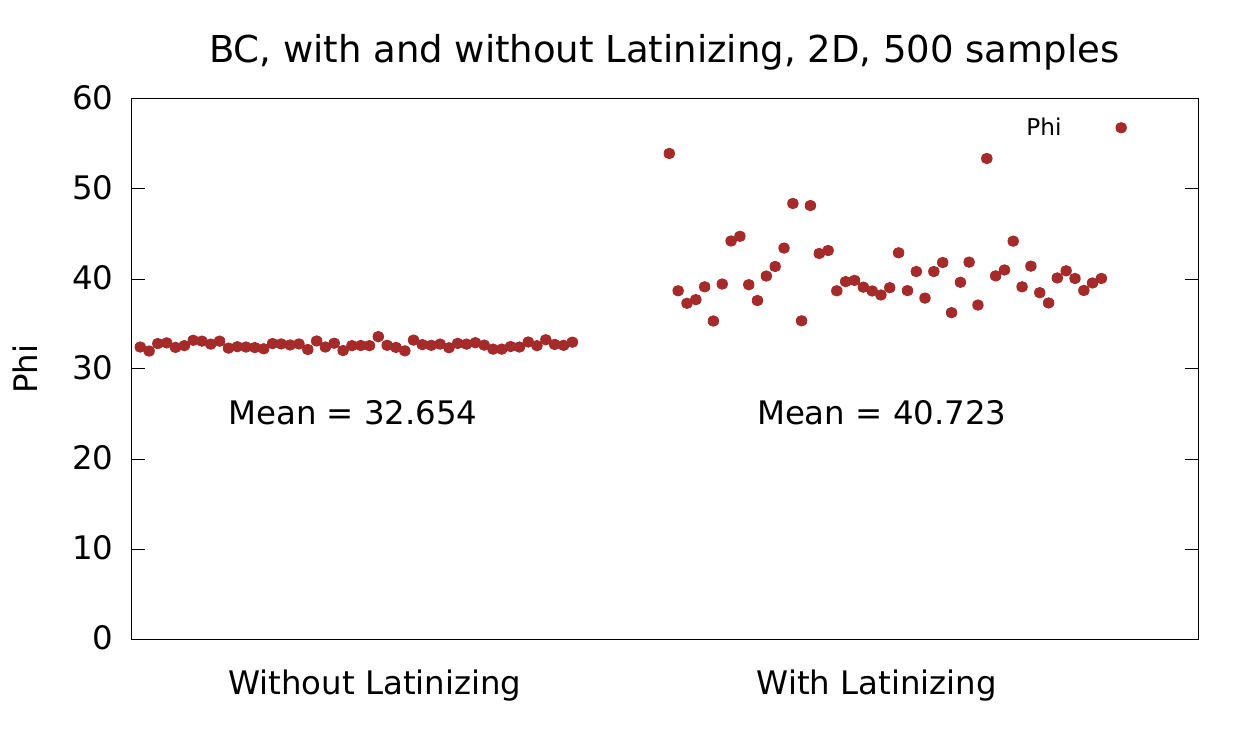} &
\includegraphics[trim = 0.0cm 0cm 0.0cm 0cm, clip = true,width=0.31\textwidth]{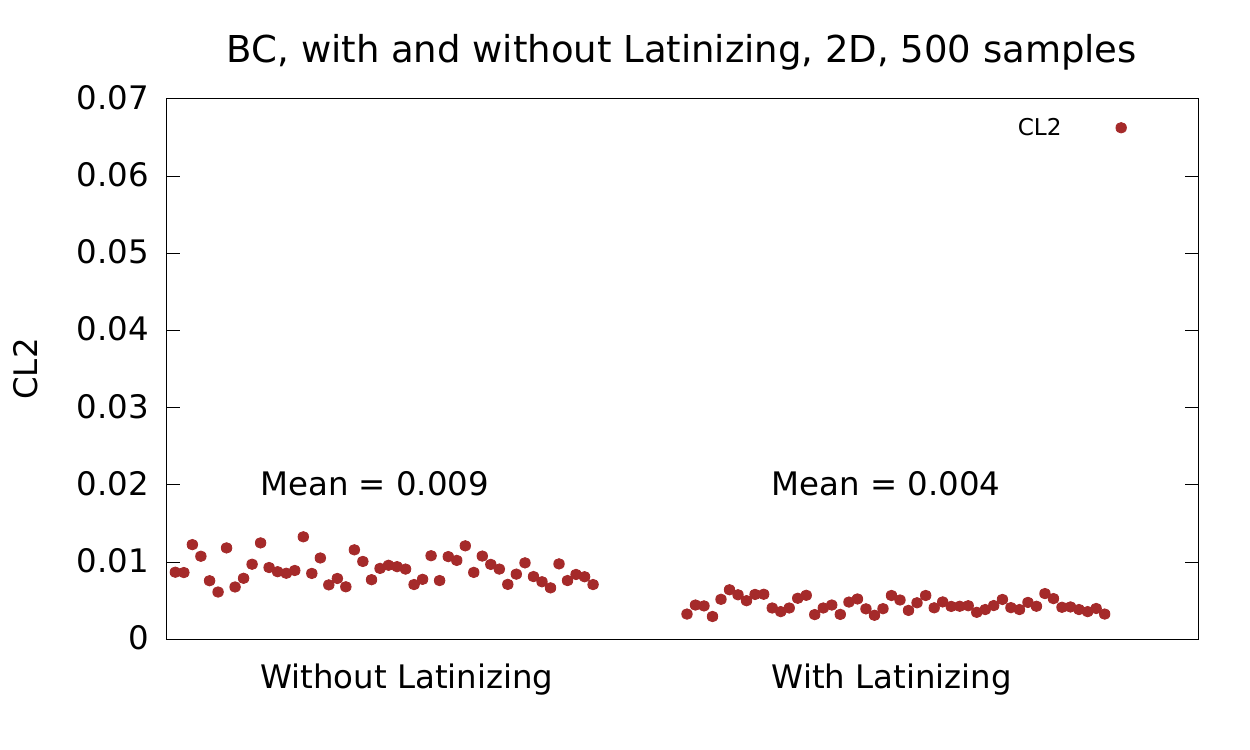} \\
\includegraphics[trim = 0.0cm 0cm 0.0cm 0cm, clip = true,width=0.31\textwidth]{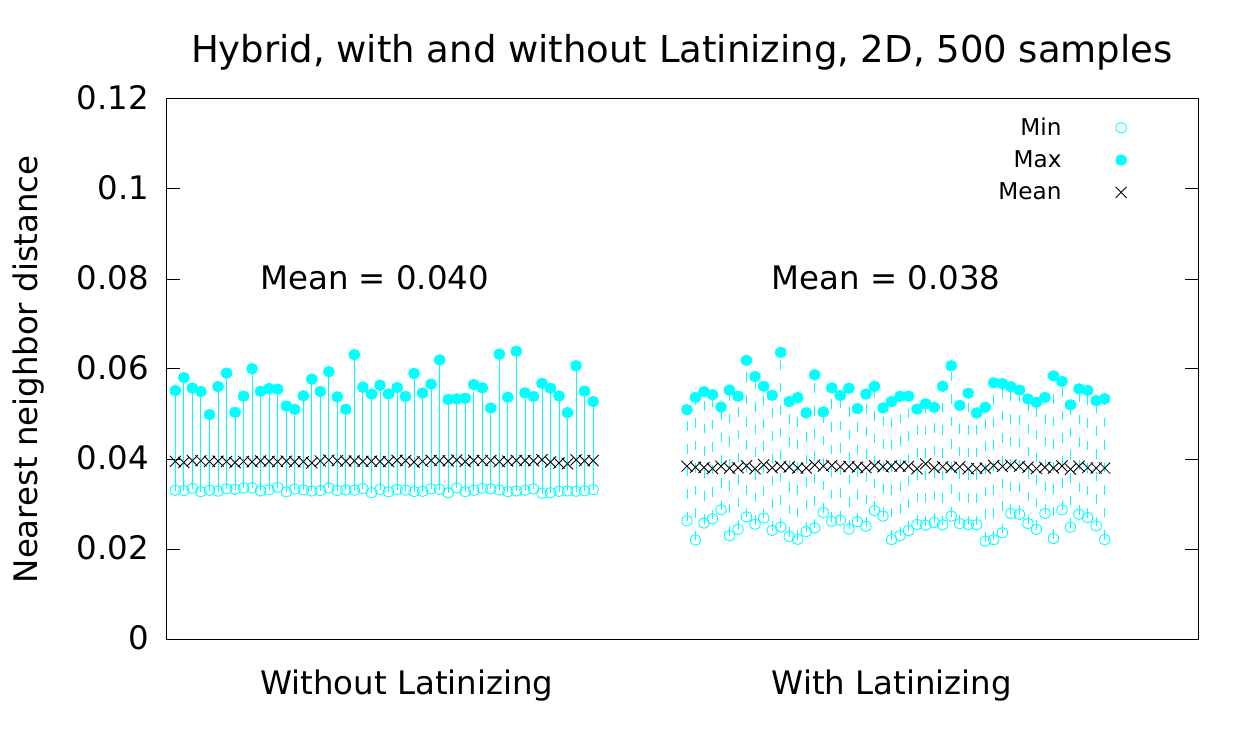} &
\includegraphics[trim = 0.0cm 0cm 0.0cm 0cm, clip = true,width=0.31\textwidth]{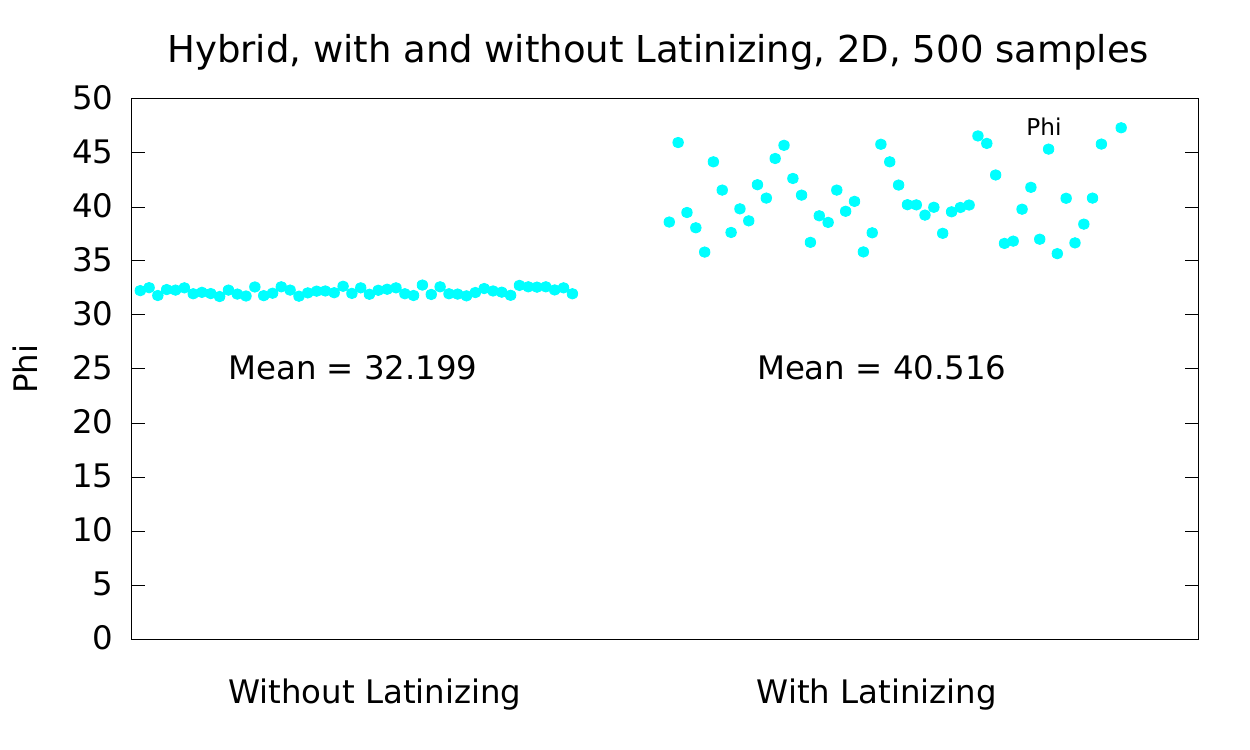} &
\includegraphics[trim = 0.0cm 0cm 0.0cm 0cm, clip = true,width=0.31\textwidth]{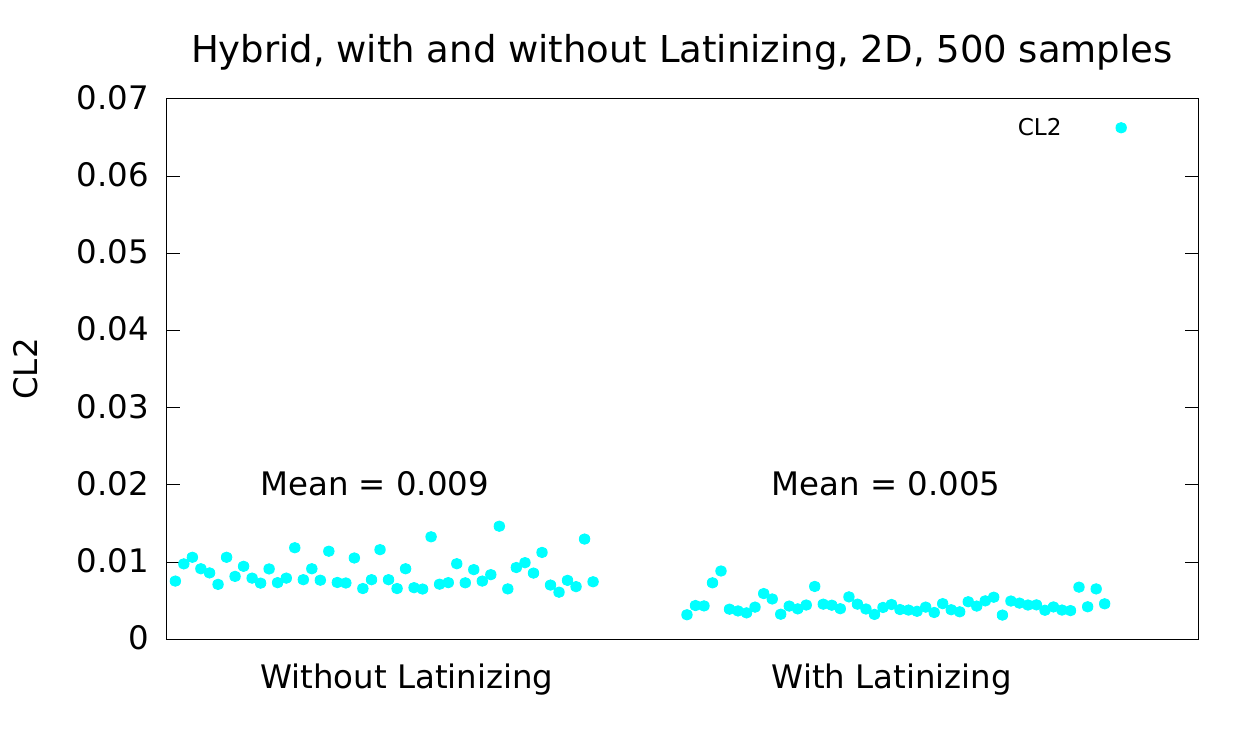} \\
(a) & (b) & (c) \\
\end{tabular}
\vspace{-0.2cm}
\caption{2D-500 experiment: Statistics on 50 repetitions of 500 samples in two dimensions
  generated using (top to bottom) random, LHS, GreedyFP, BC, and the
  hybrid methods. (a) distance to the nearest neighbor; (b)
  $\phi_{50}$ criterion; and (c) the CL2 discrepancy. Each plot shows
  the results without Latinizing (left) and with Latinizing (right);
  there are no Latinizing results for LHS.  The mean value of the metrics
  --- the average of the nearest neighbor distance in each sample set,
  the $\phi_{50}$ criterion and the CL2 discrepancy --- over the 50
  repetitions is also included. The range of values in the plots for
  $\phi_{50}$ are different for each algorithm. }
\label{fig:metrics_2d_500}
\end{figure}

\begin{figure}[!htb]
\centering
\begin{tabular}{ccc}
\includegraphics[trim = 0.0cm 0cm 0.0cm 0cm, clip = true,width=0.31\textwidth]{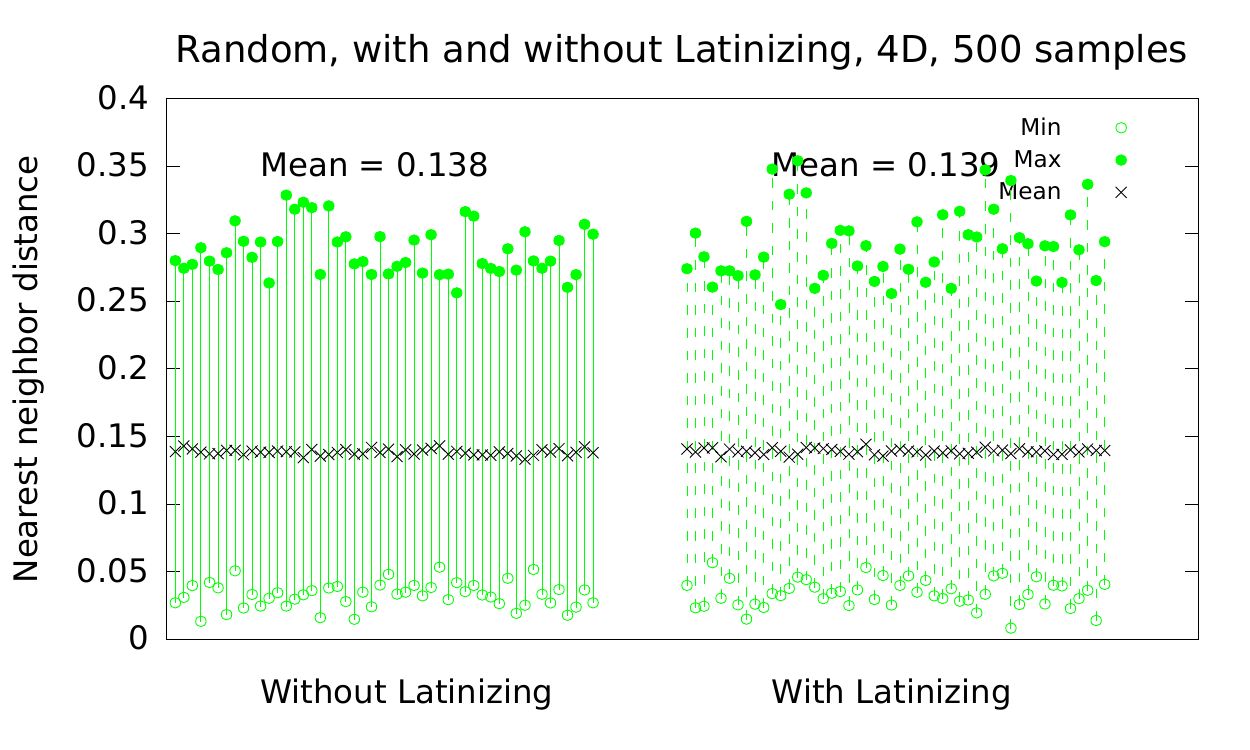} &
\includegraphics[trim = 0.0cm 0cm 0.0cm 0cm, clip = true,width=0.31\textwidth]{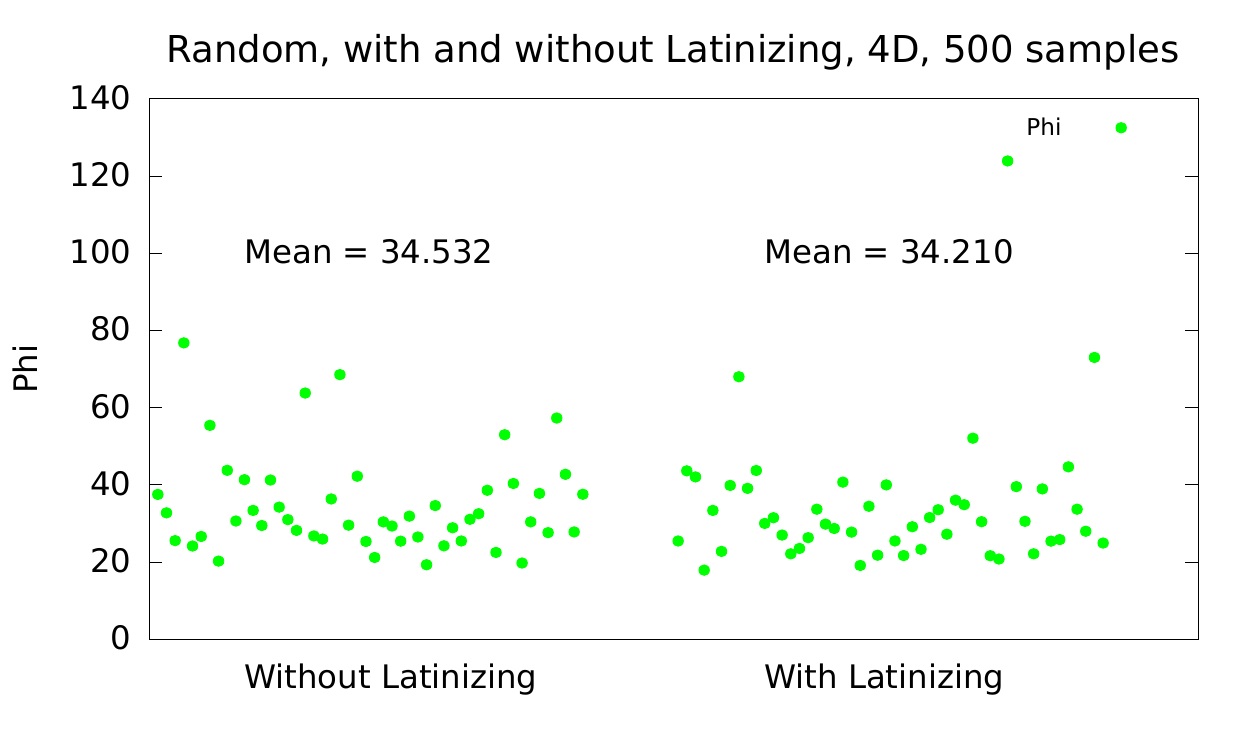} &
\includegraphics[trim = 0.0cm 0cm 0.0cm 0cm, clip = true,width=0.31\textwidth]{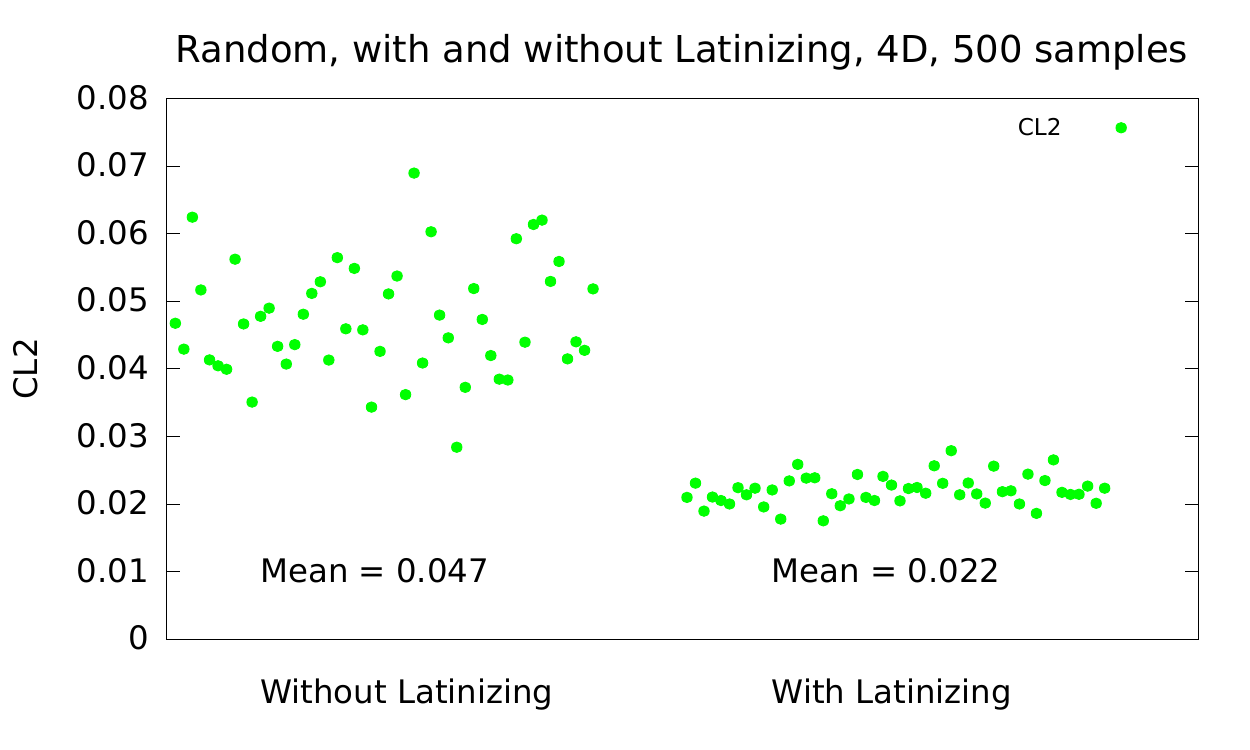} \\
\includegraphics[trim = 0.0cm 0cm 0.0cm 0cm, clip = true,width=0.31\textwidth]{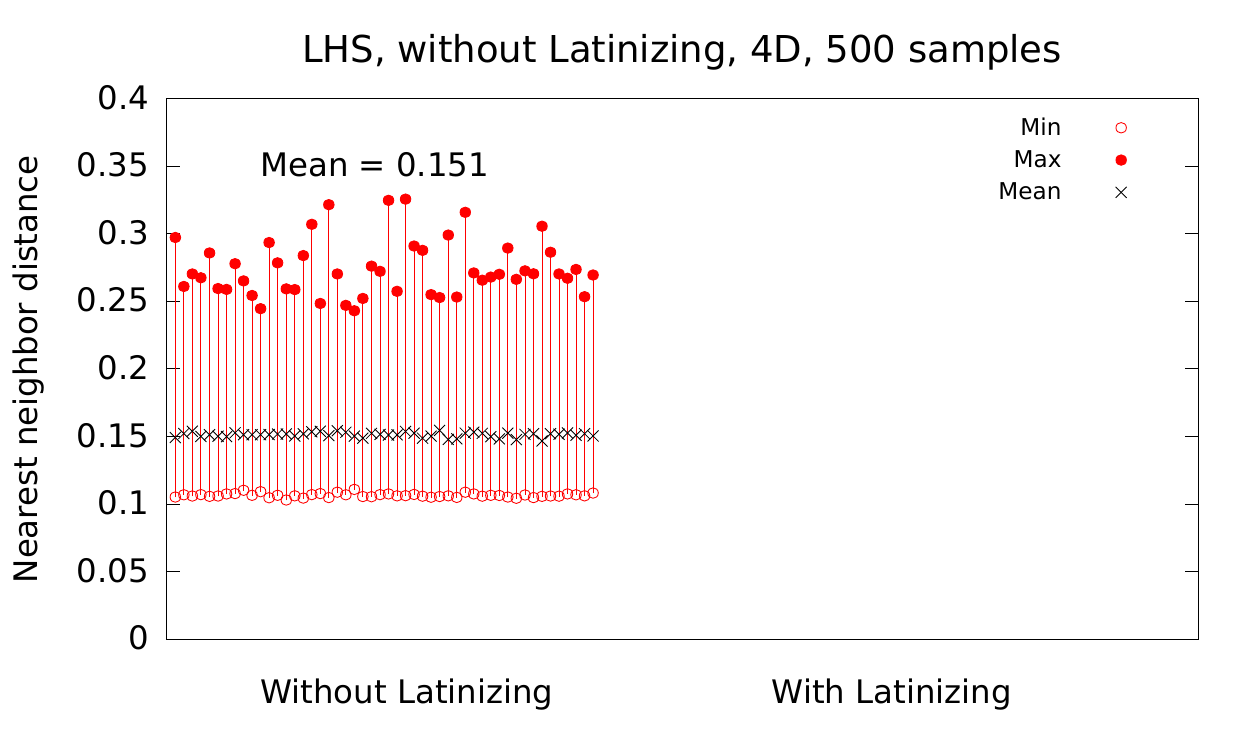} &
\includegraphics[trim = 0.0cm 0cm 0.0cm 0cm, clip = true,width=0.31\textwidth]{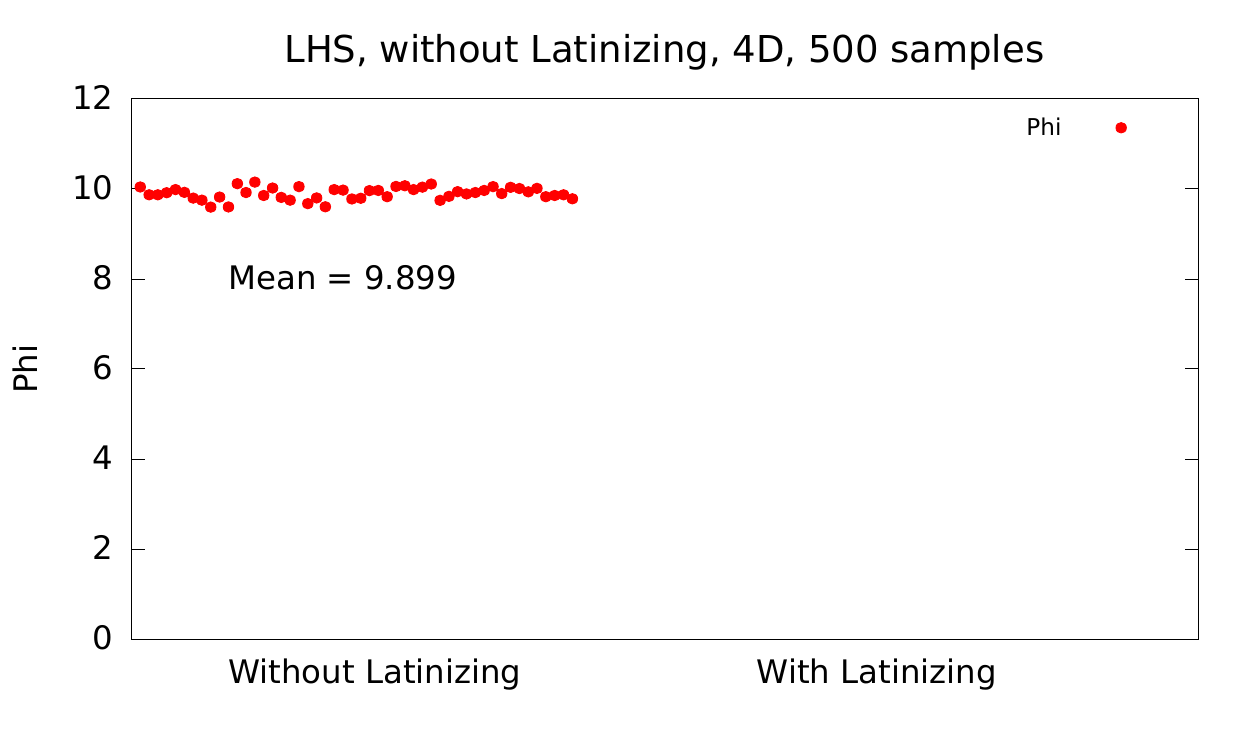} &
\includegraphics[trim = 0.0cm 0cm 0.0cm 0cm, clip = true,width=0.31\textwidth]{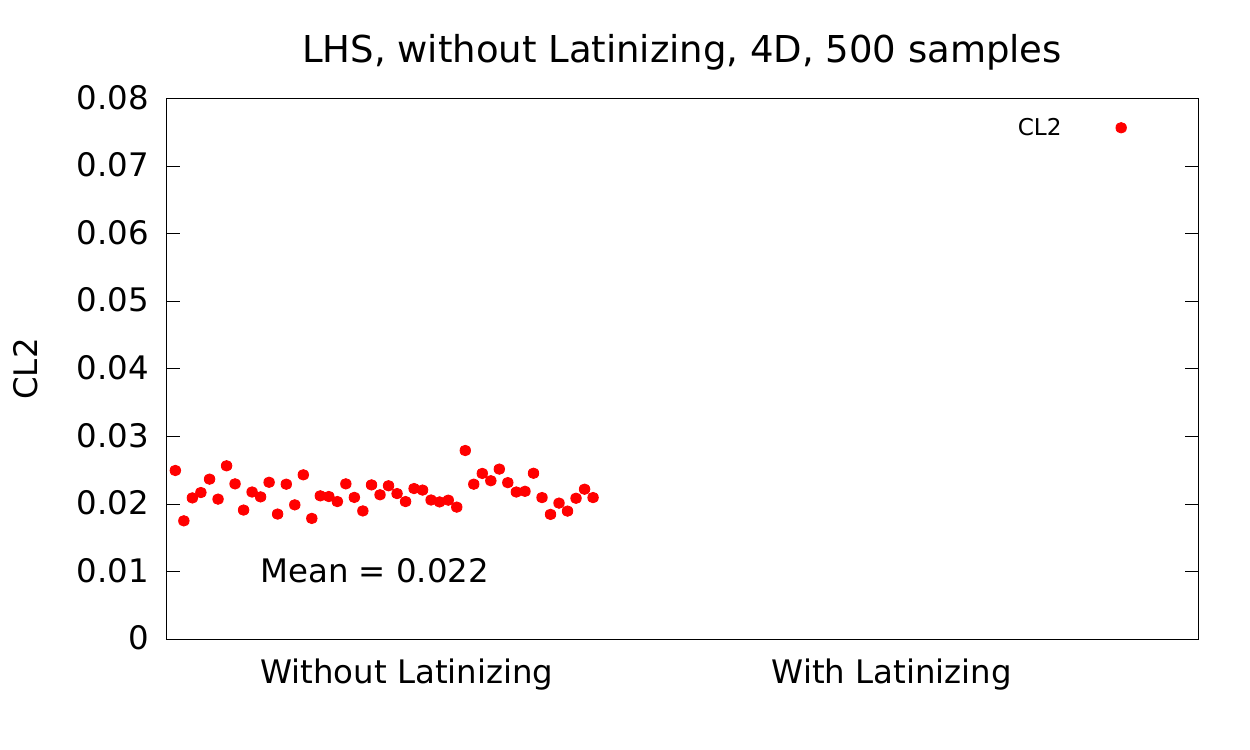} \\
\includegraphics[trim = 0.0cm 0cm 0.0cm 0cm, clip = true,width=0.31\textwidth]{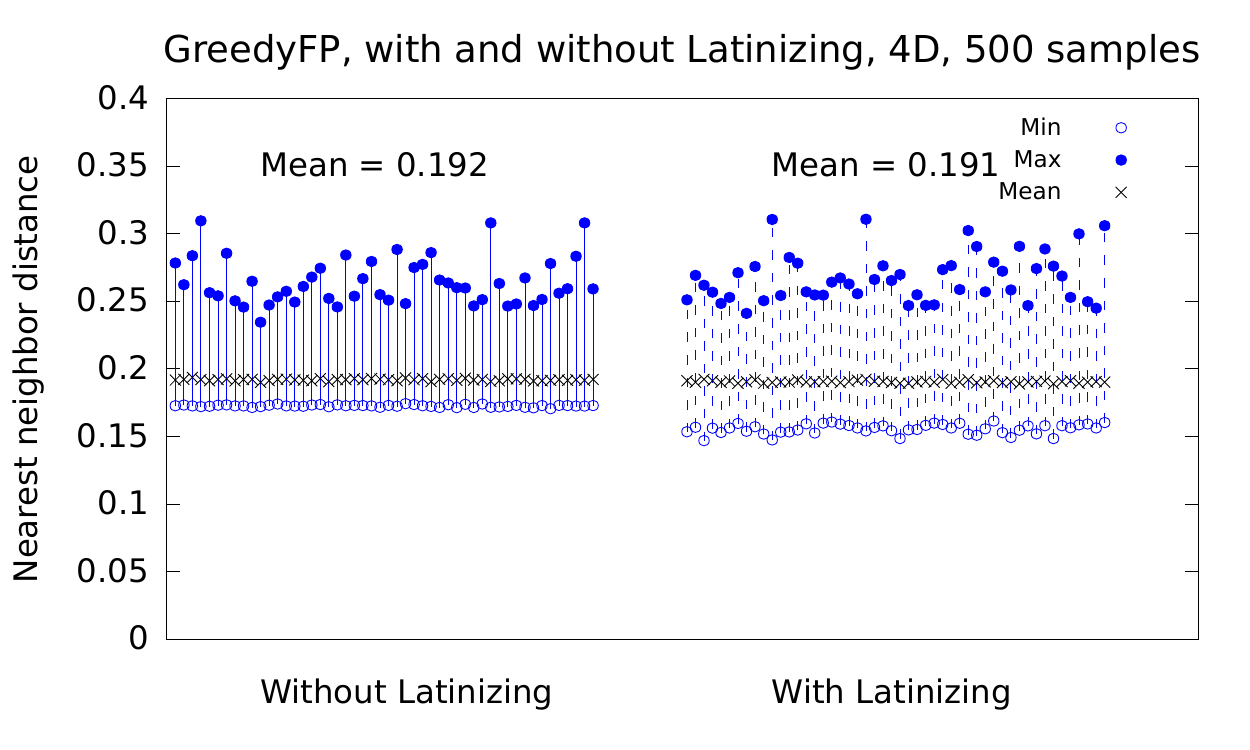} &
\includegraphics[trim = 0.0cm 0cm 0.0cm 0cm, clip = true,width=0.31\textwidth]{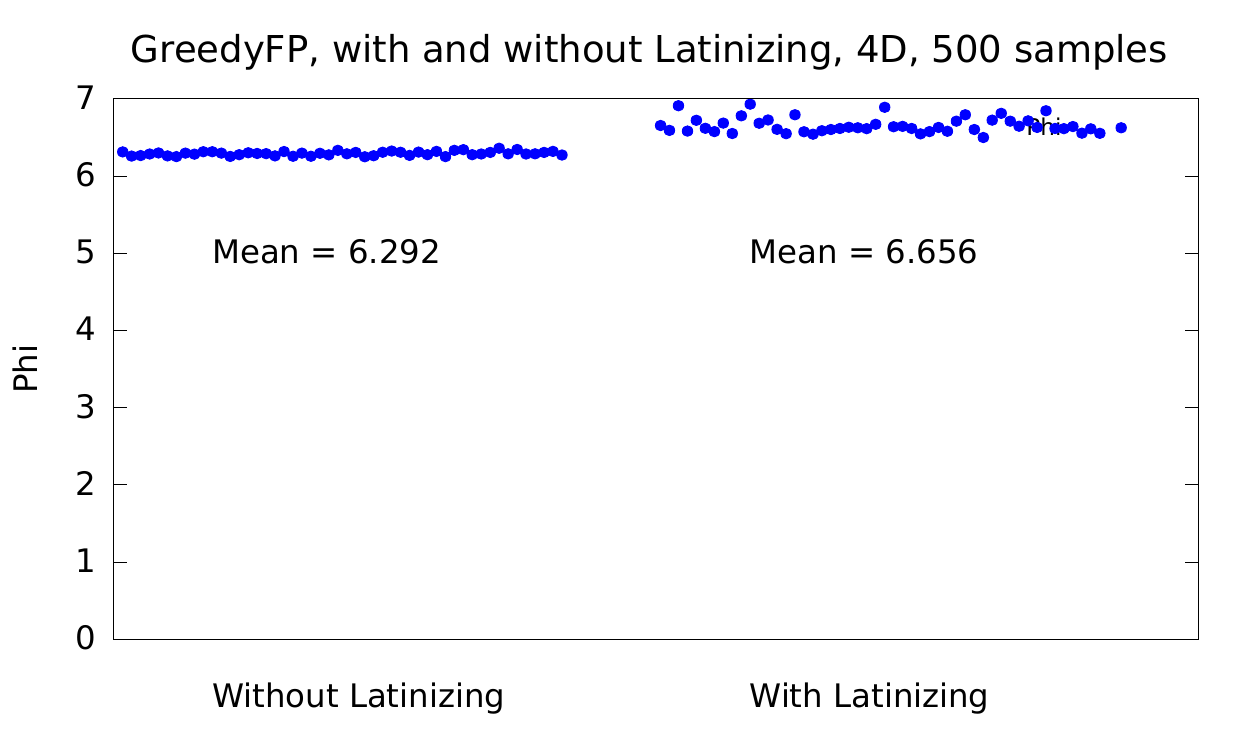} &
\includegraphics[trim = 0.0cm 0cm 0.0cm 0cm, clip = true,width=0.31\textwidth]{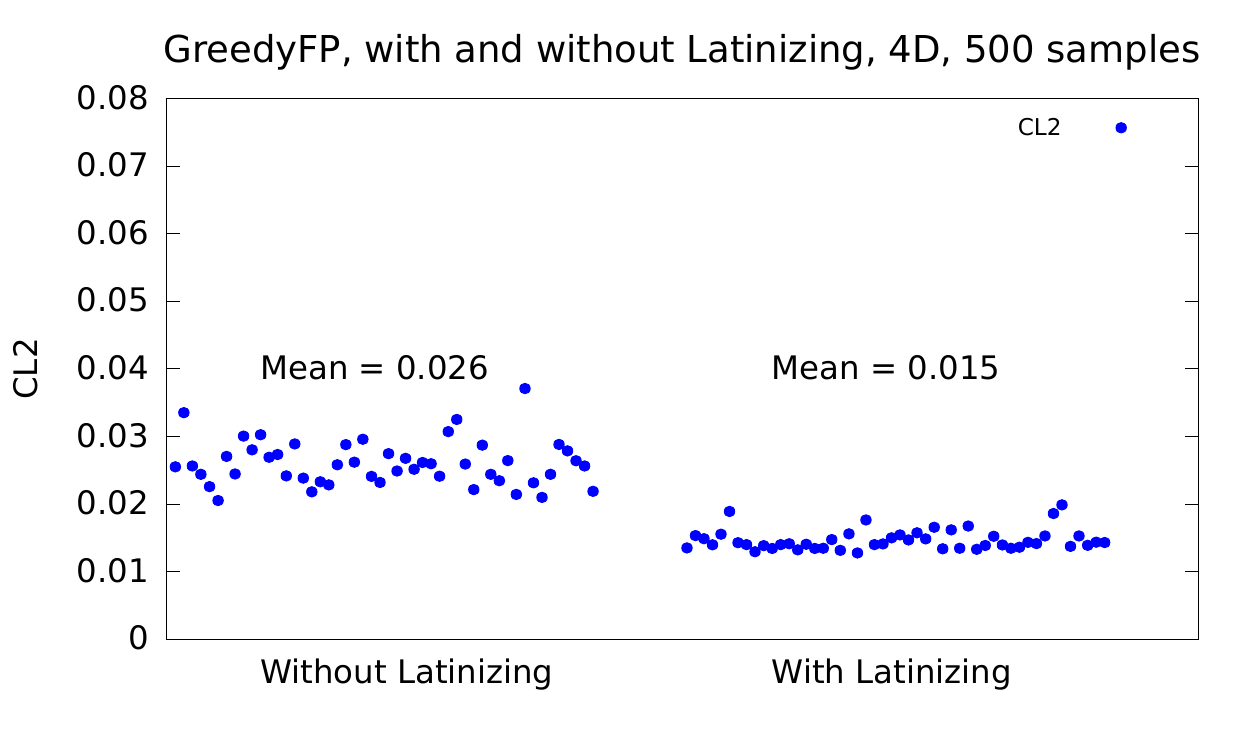} \\
\includegraphics[trim = 0.0cm 0cm 0.0cm 0cm, clip = true,width=0.31\textwidth]{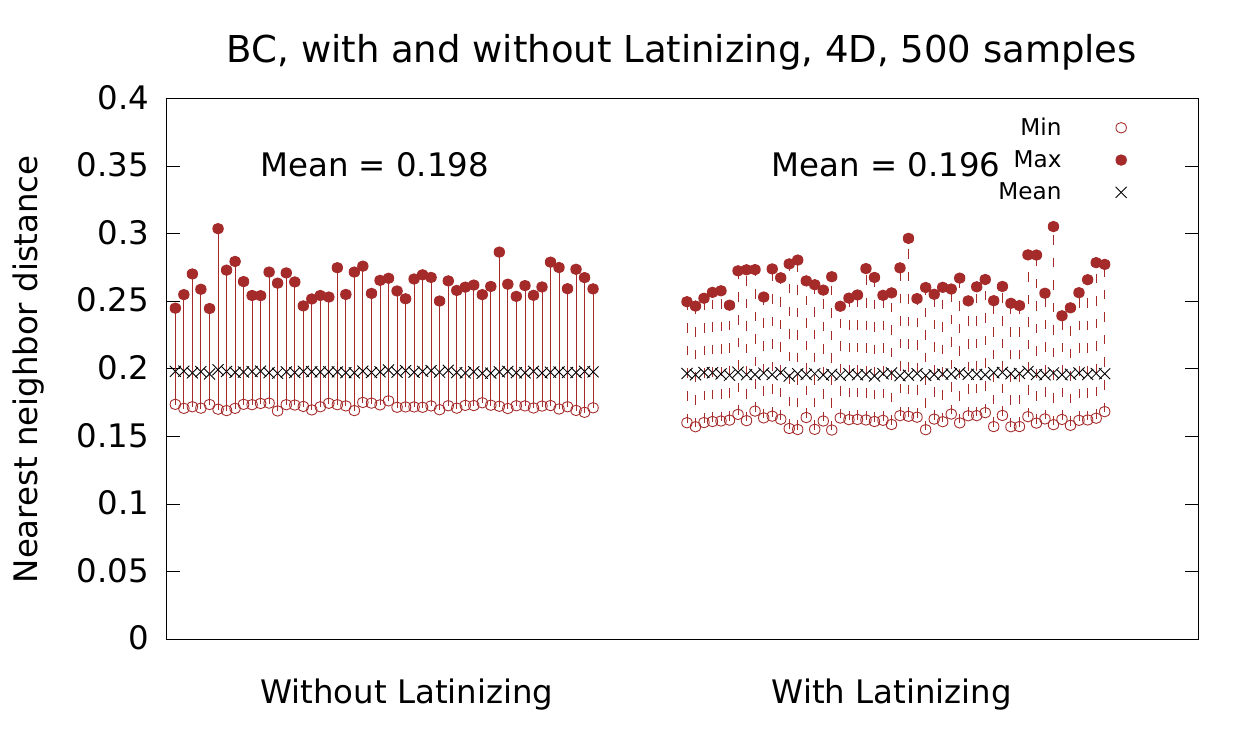} &
\includegraphics[trim = 0.0cm 0cm 0.0cm 0cm, clip = true,width=0.31\textwidth]{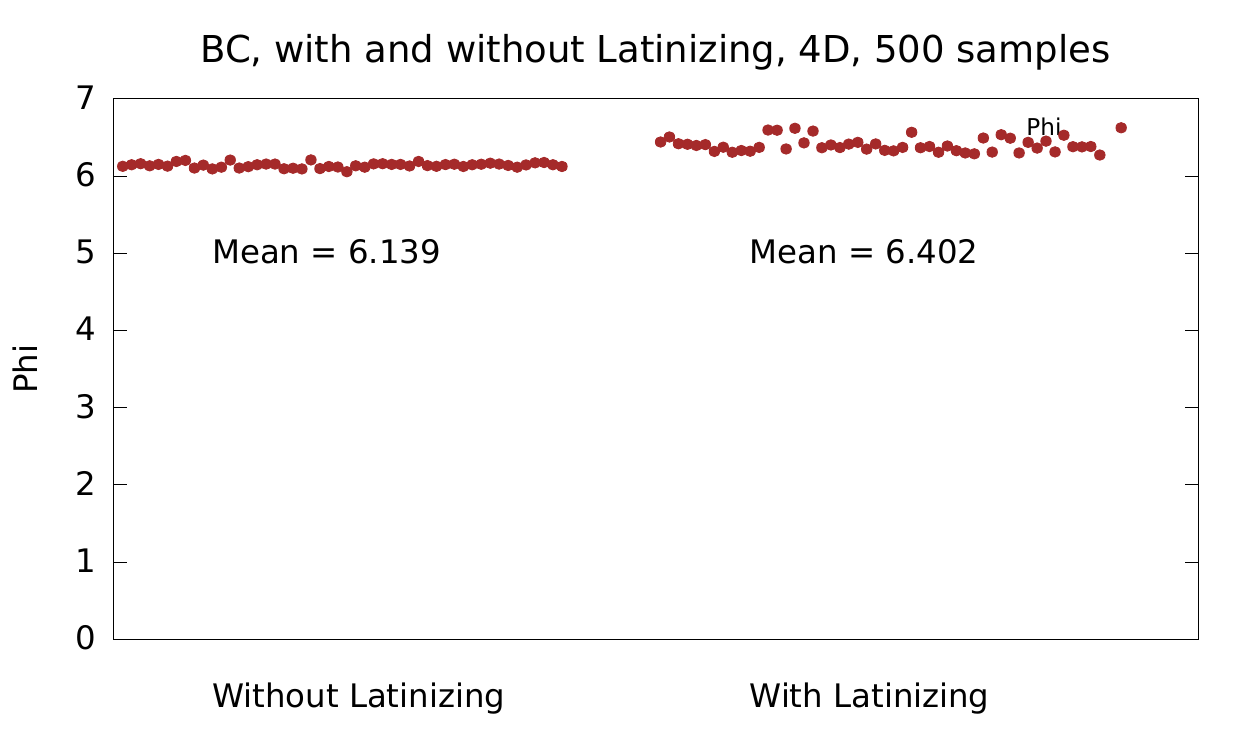} &
\includegraphics[trim = 0.0cm 0cm 0.0cm 0cm, clip = true,width=0.31\textwidth]{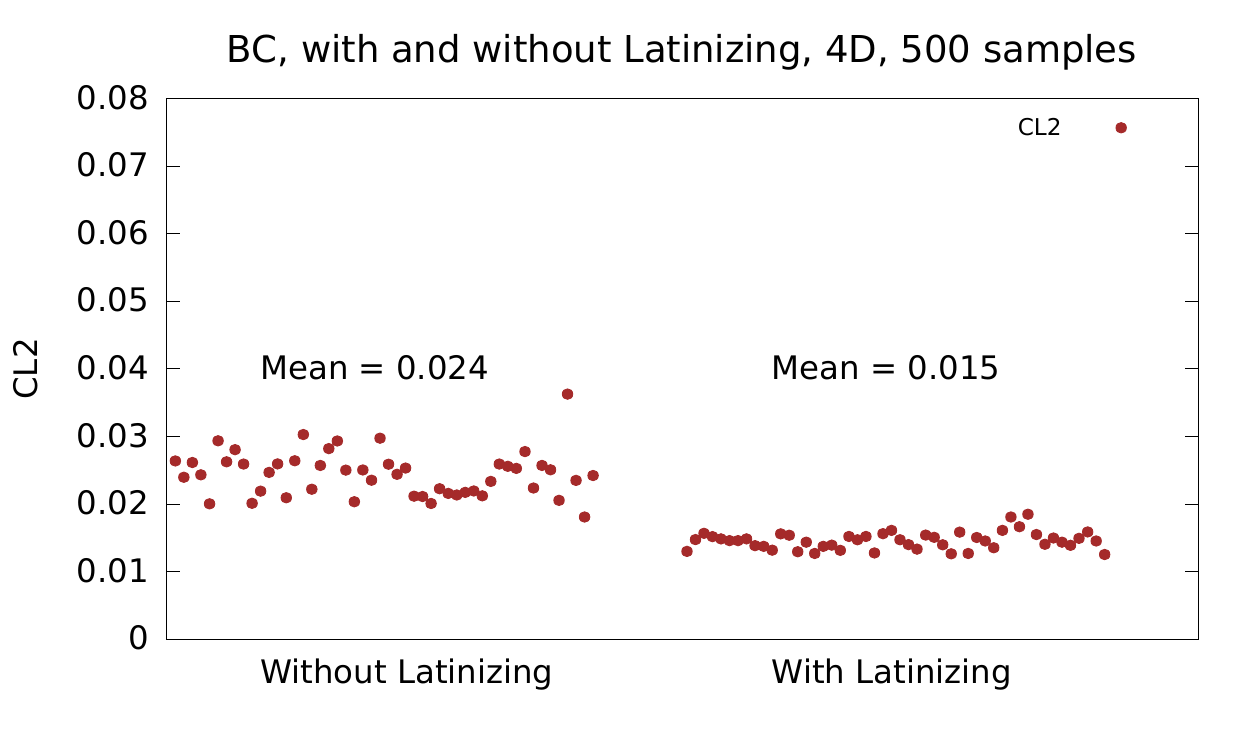} \\
\includegraphics[trim = 0.0cm 0cm 0.0cm 0cm, clip = true,width=0.31\textwidth]{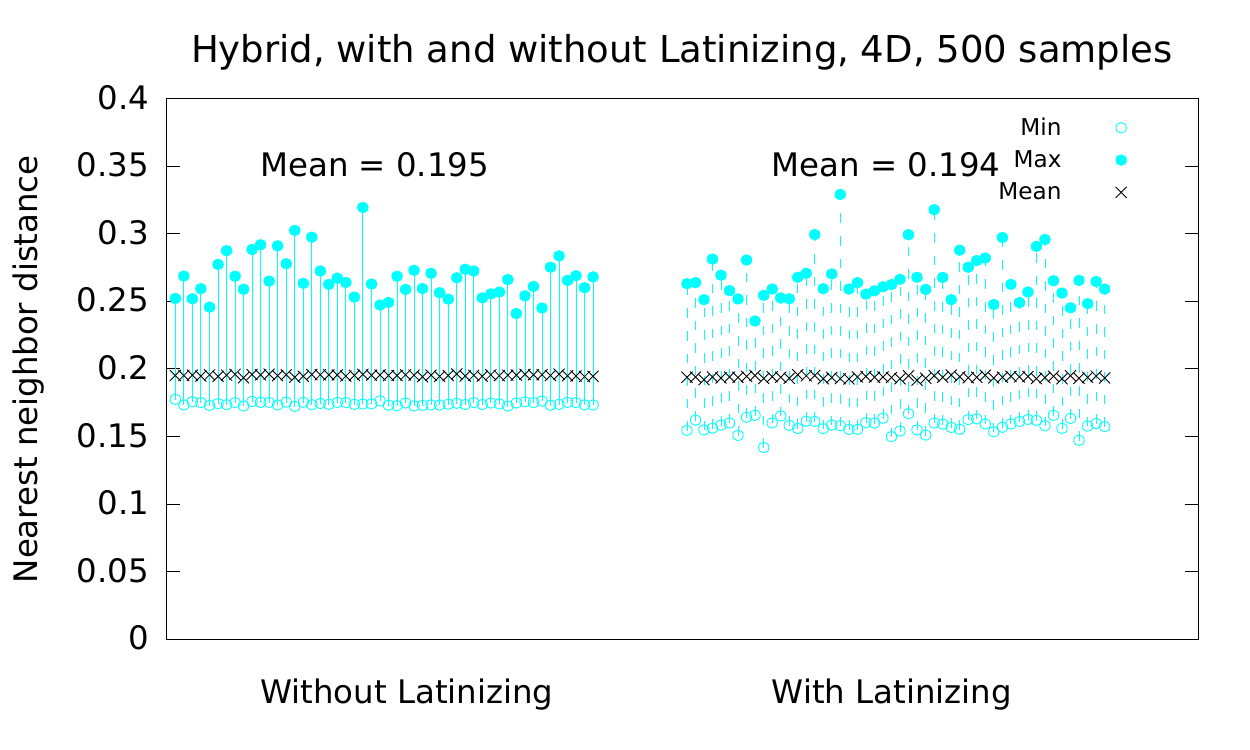} &
\includegraphics[trim = 0.0cm 0cm 0.0cm 0cm, clip = true,width=0.31\textwidth]{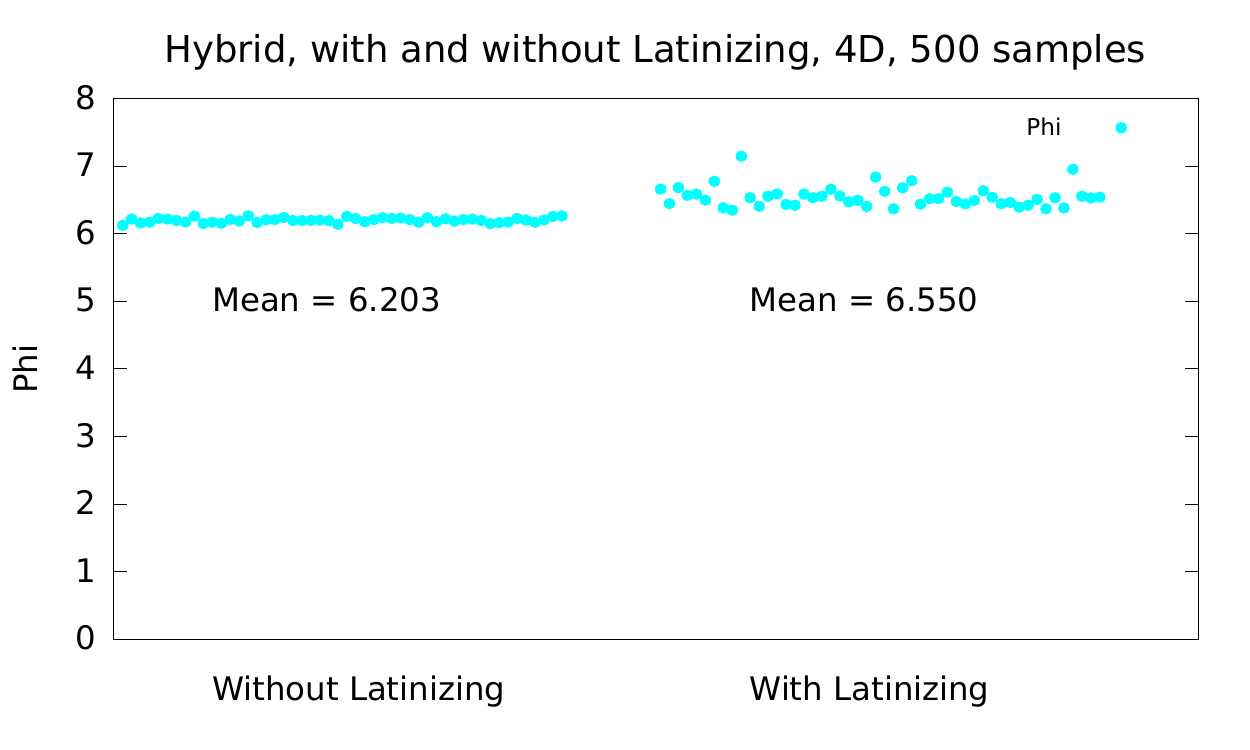} &
\includegraphics[trim = 0.0cm 0cm 0.0cm 0cm, clip = true,width=0.31\textwidth]{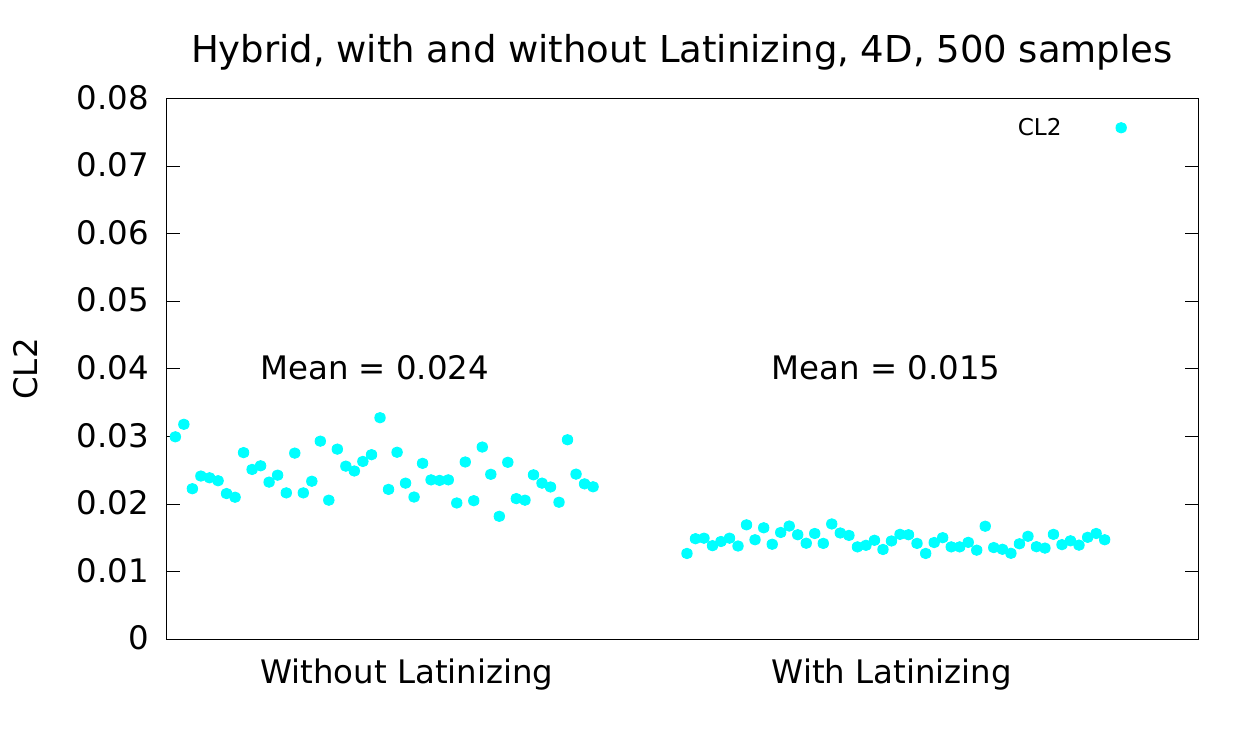} \\
(a) & (b) & (c) \\
\end{tabular}
\vspace{-0.2cm}
\caption{4D-500 experiment: Statistics on 50 repetitions of 500 samples in four dimensions
  generated using (top to bottom) random, LHS, GreedyFP, BC, and the
  hybrid methods. (a) distance to the nearest neighbor; (b)
  $\phi_{50}$ criterion; and (c) the CL2 discrepancy. Each plot shows
  the results without Latinizing (left) and with Latinizing (right);
  there are no Latinizing results for LHS.  The mean value of the metrics
  --- the average of the nearest neighbor distance in each sample set,
  the $\phi_{50}$ criterion and the CL2 discrepancy --- over the 50
  repetitions is also included. The range of values in the plots for
  $\phi_{50}$ are different for each algorithm. }
\label{fig:metrics_4d_500}
\end{figure}

\begin{figure}[!htb]
\centering
\begin{tabular}{ccc}
\includegraphics[trim = 0.0cm 0cm 0.0cm 0cm, clip = true,width=0.31\textwidth]{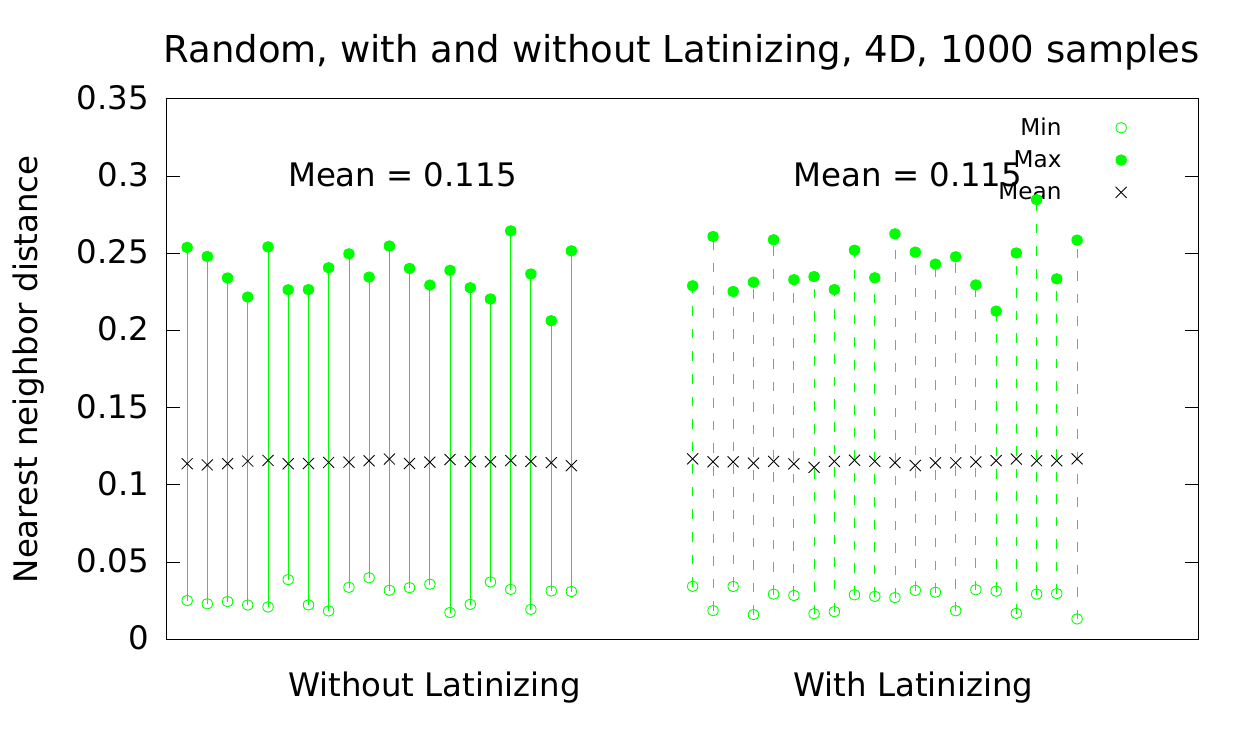} &
\includegraphics[trim = 0.0cm 0cm 0.0cm 0cm, clip = true,width=0.31\textwidth]{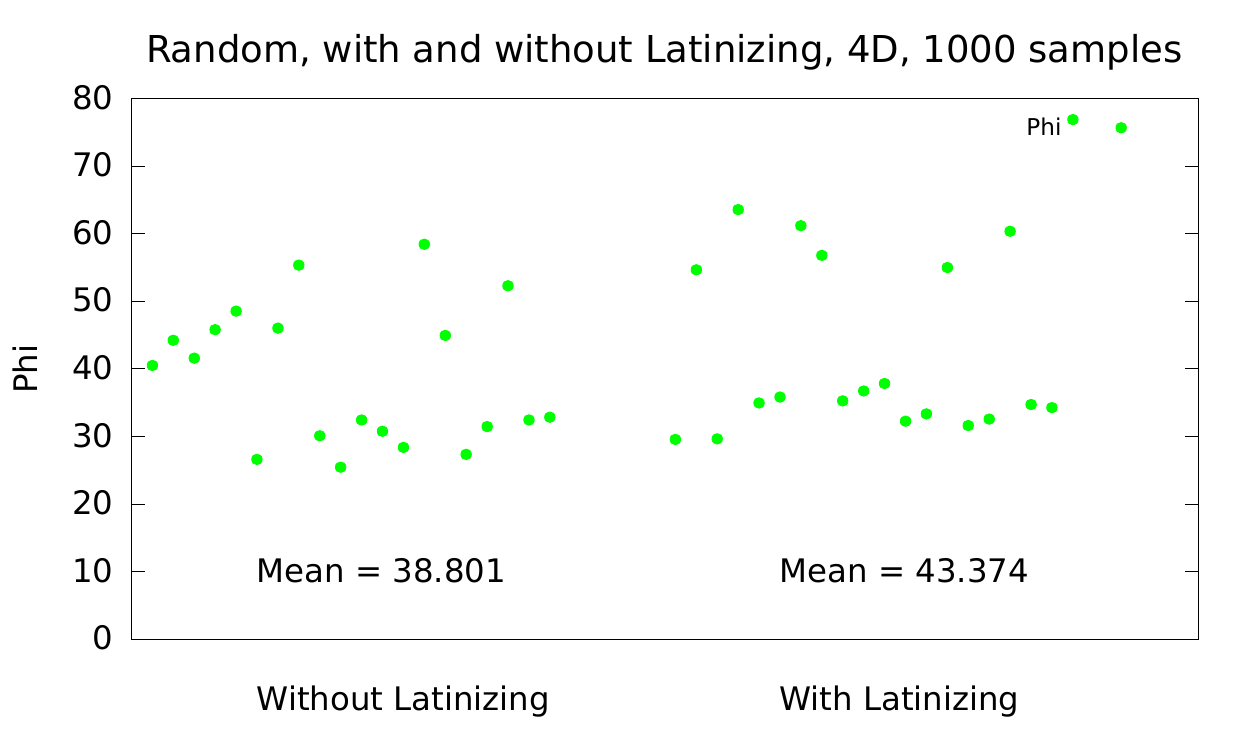} &
\includegraphics[trim = 0.0cm 0cm 0.0cm 0cm, clip = true,width=0.31\textwidth]{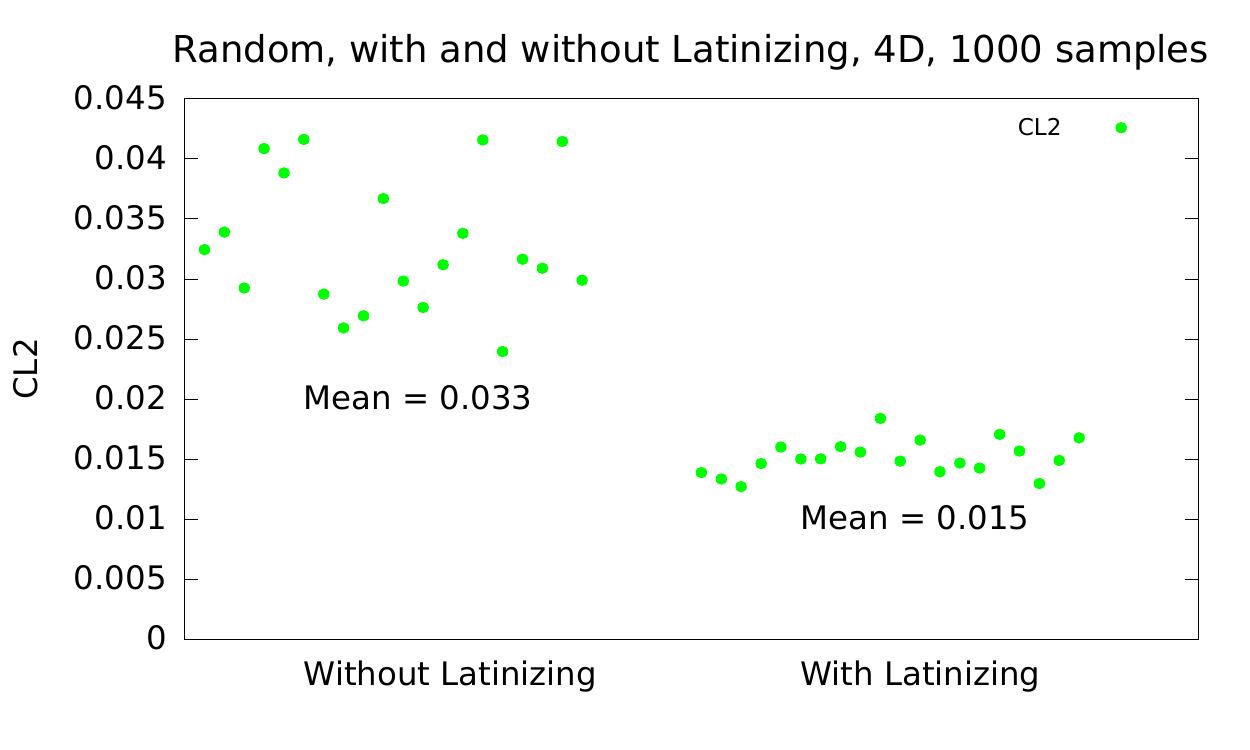} \\
\includegraphics[trim = 0.0cm 0cm 0.0cm 0cm, clip = true,width=0.31\textwidth]{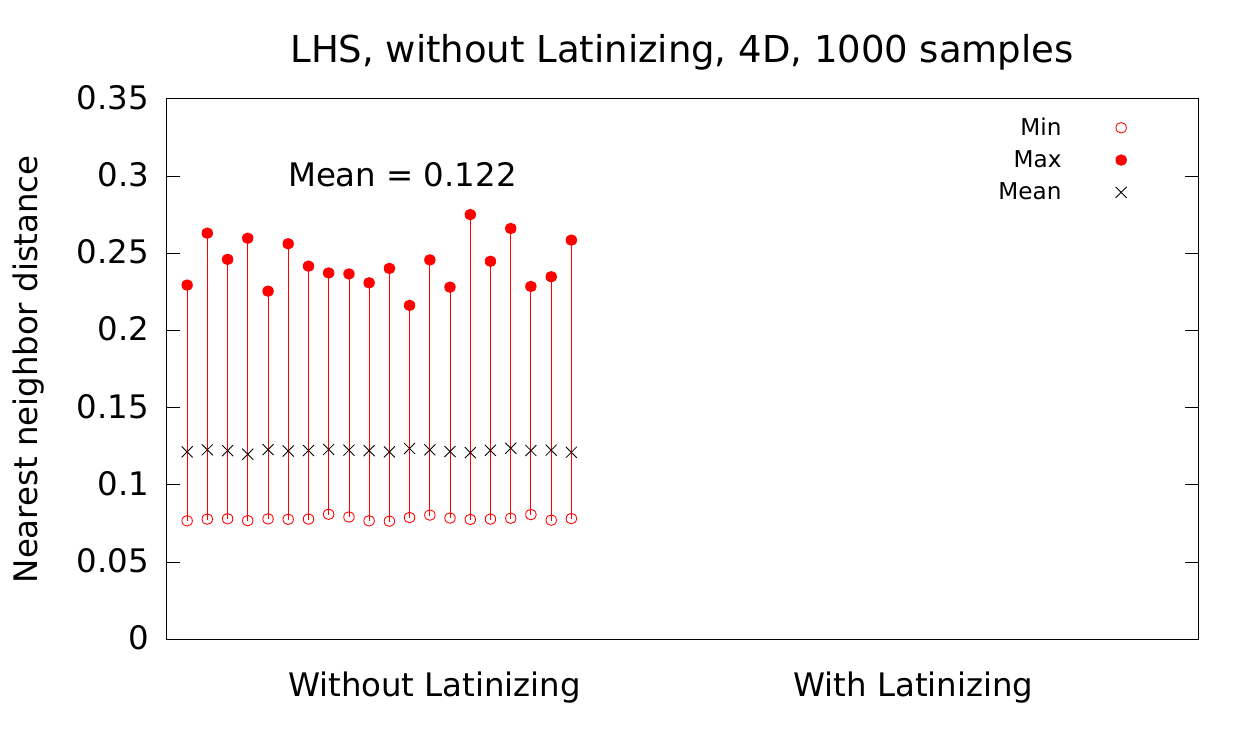} &
\includegraphics[trim = 0.0cm 0cm 0.0cm 0cm, clip = true,width=0.31\textwidth]{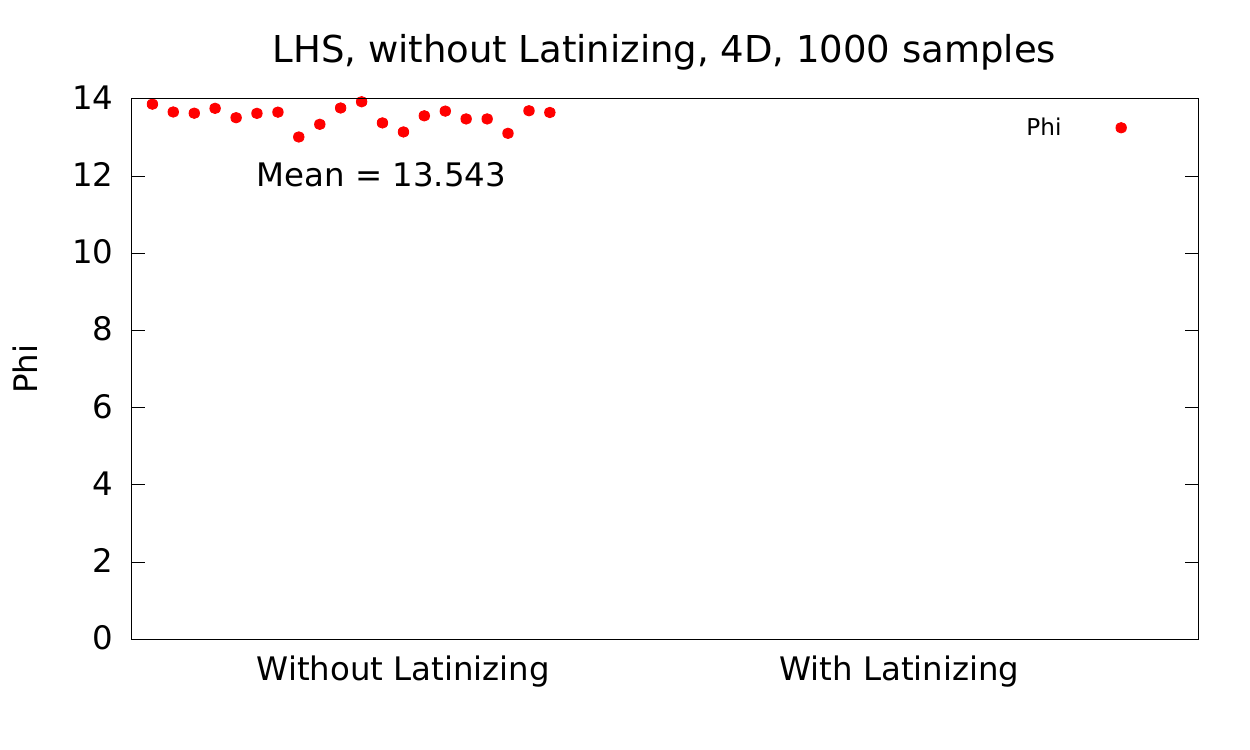} &
\includegraphics[trim = 0.0cm 0cm 0.0cm 0cm, clip = true,width=0.31\textwidth]{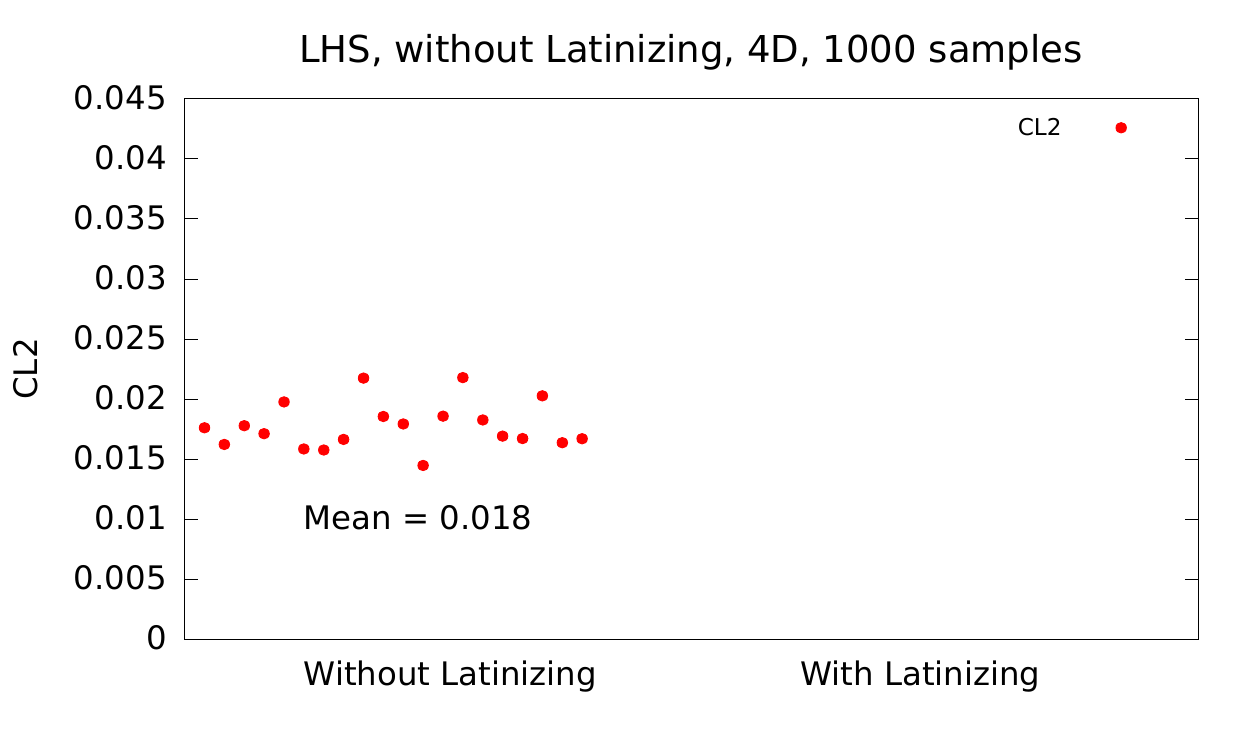} \\
\includegraphics[trim = 0.0cm 0cm 0.0cm 0cm, clip = true,width=0.31\textwidth]{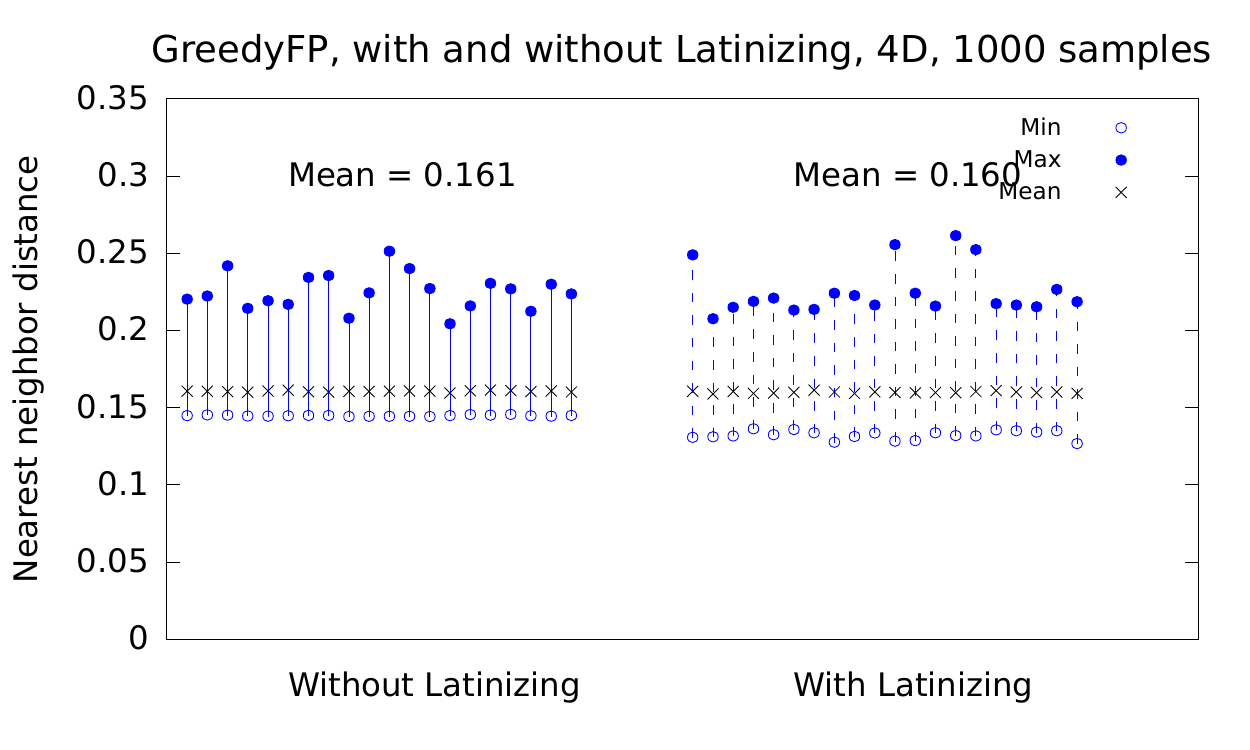} &
\includegraphics[trim = 0.0cm 0cm 0.0cm 0cm, clip = true,width=0.31\textwidth]{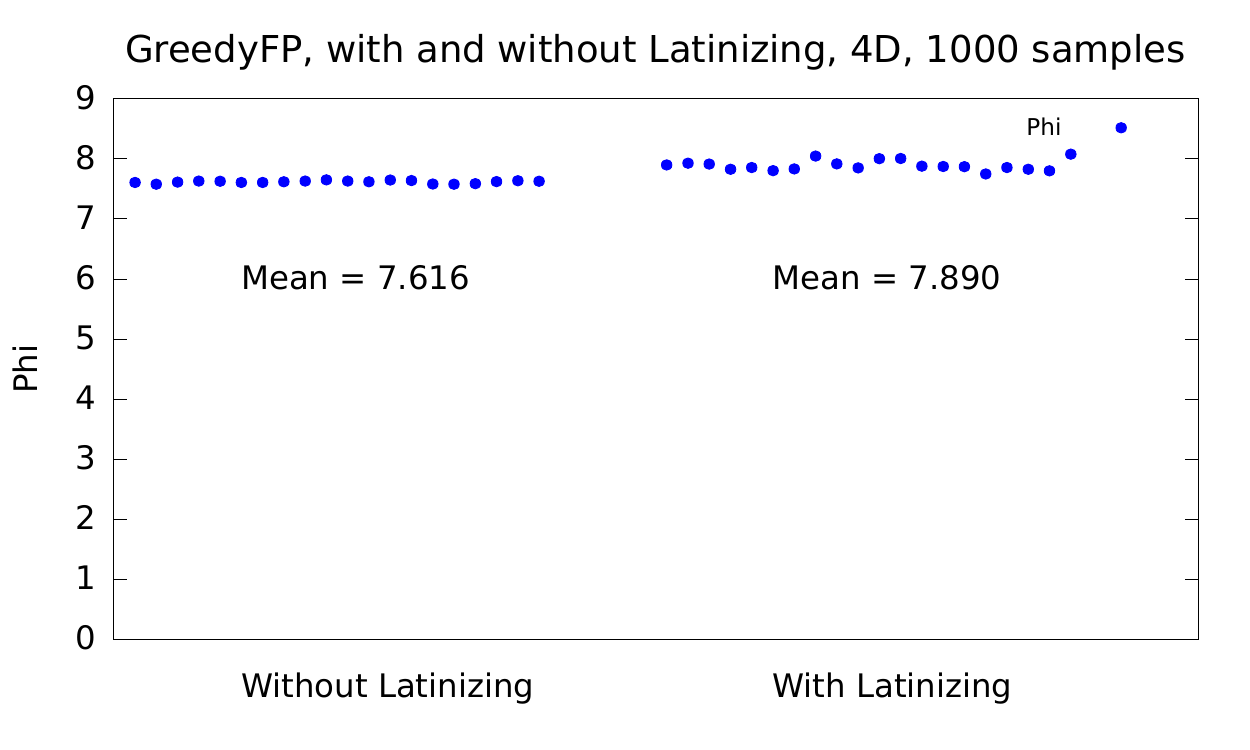} &
\includegraphics[trim = 0.0cm 0cm 0.0cm 0cm, clip = true,width=0.31\textwidth]{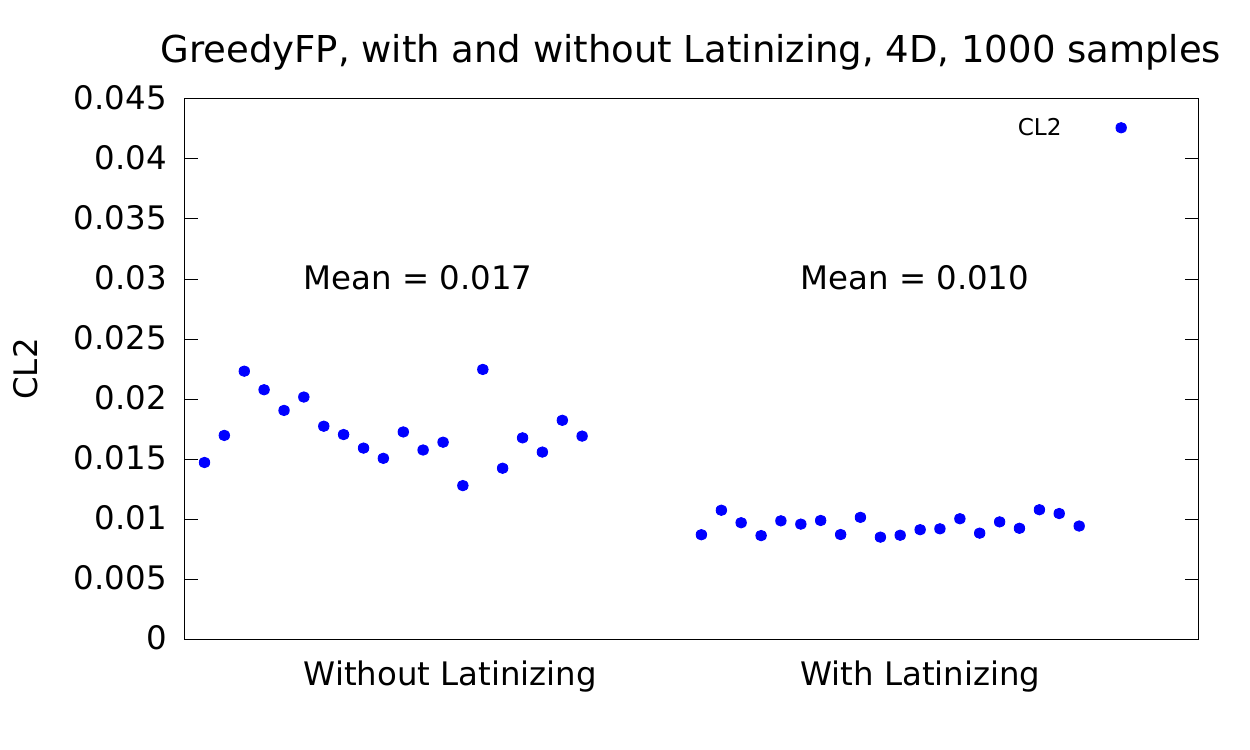} \\
\includegraphics[trim = 0.0cm 0cm 0.0cm 0cm, clip = true,width=0.31\textwidth]{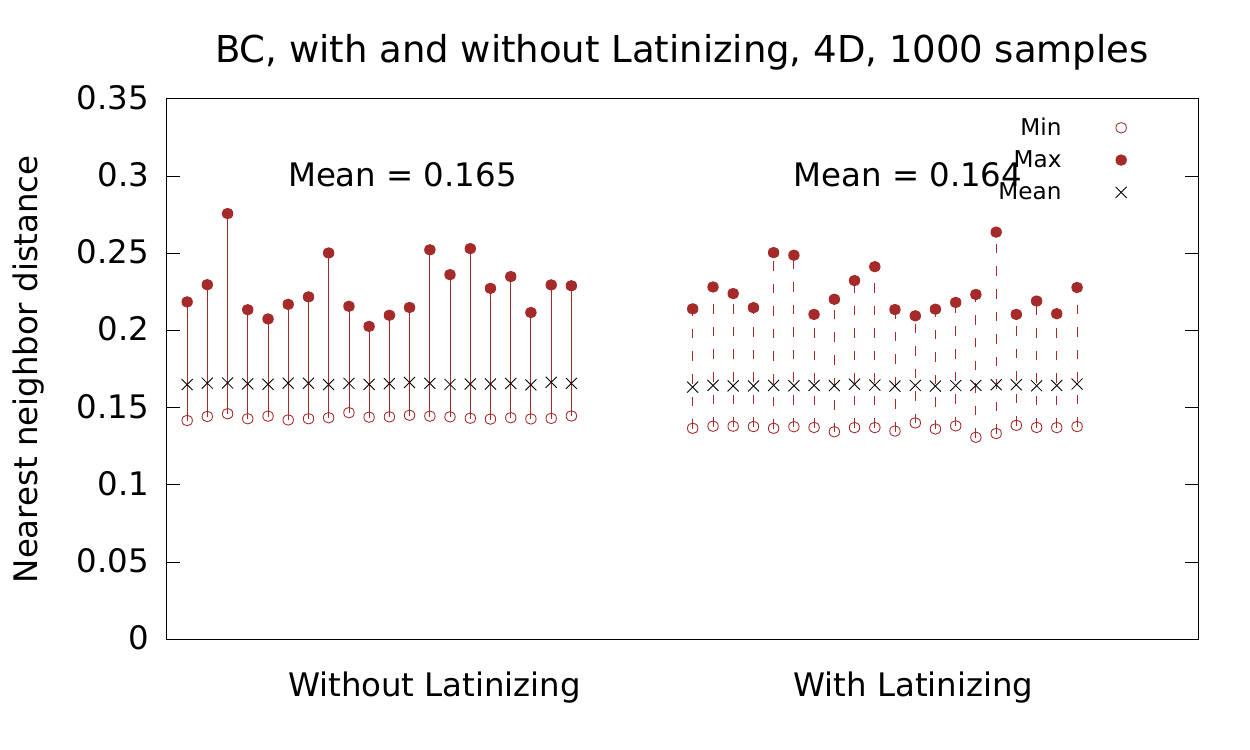} &
\includegraphics[trim = 0.0cm 0cm 0.0cm 0cm, clip = true,width=0.31\textwidth]{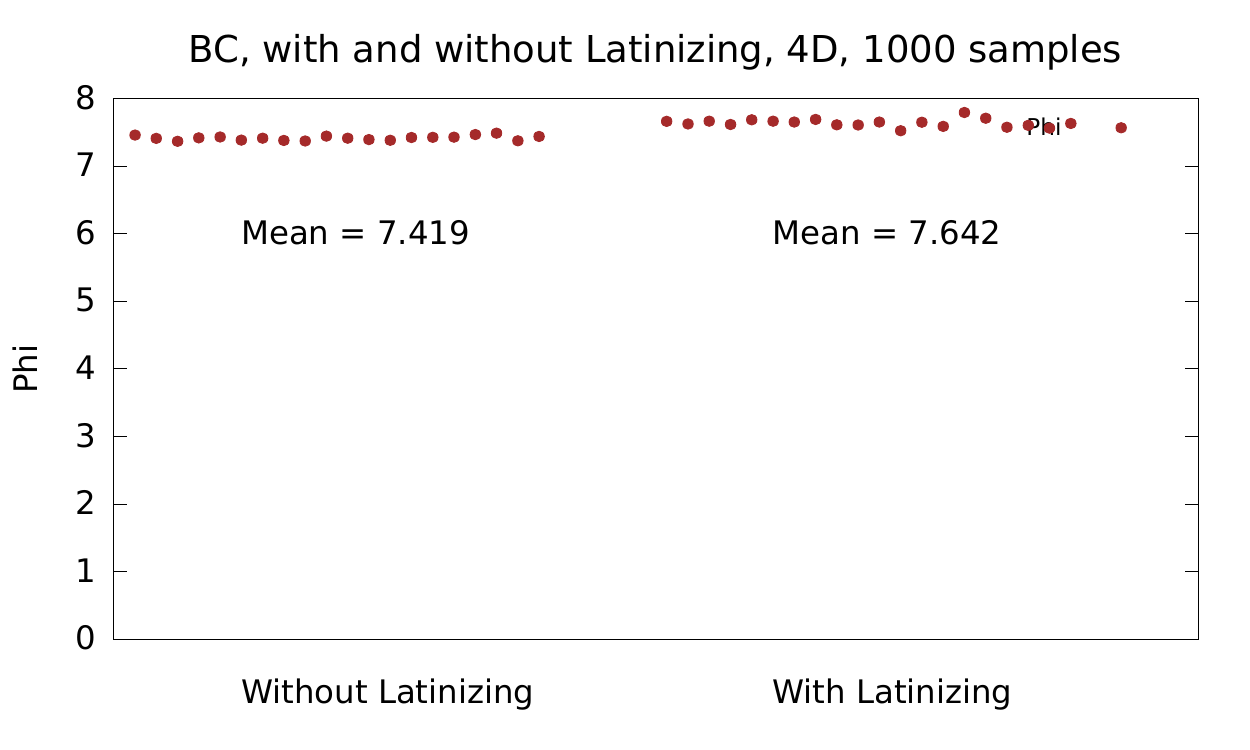} &
\includegraphics[trim = 0.0cm 0cm 0.0cm 0cm, clip = true,width=0.31\textwidth]{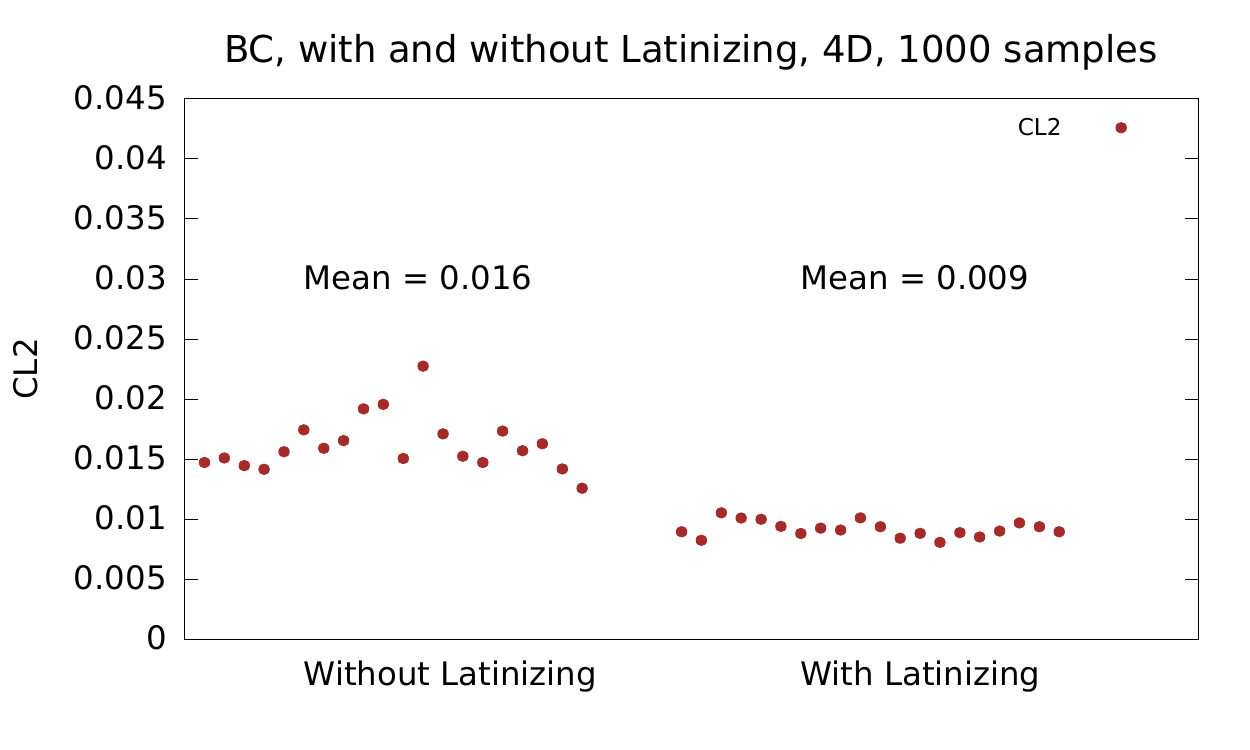} \\
\includegraphics[trim = 0.0cm 0cm 0.0cm 0cm, clip = true,width=0.31\textwidth]{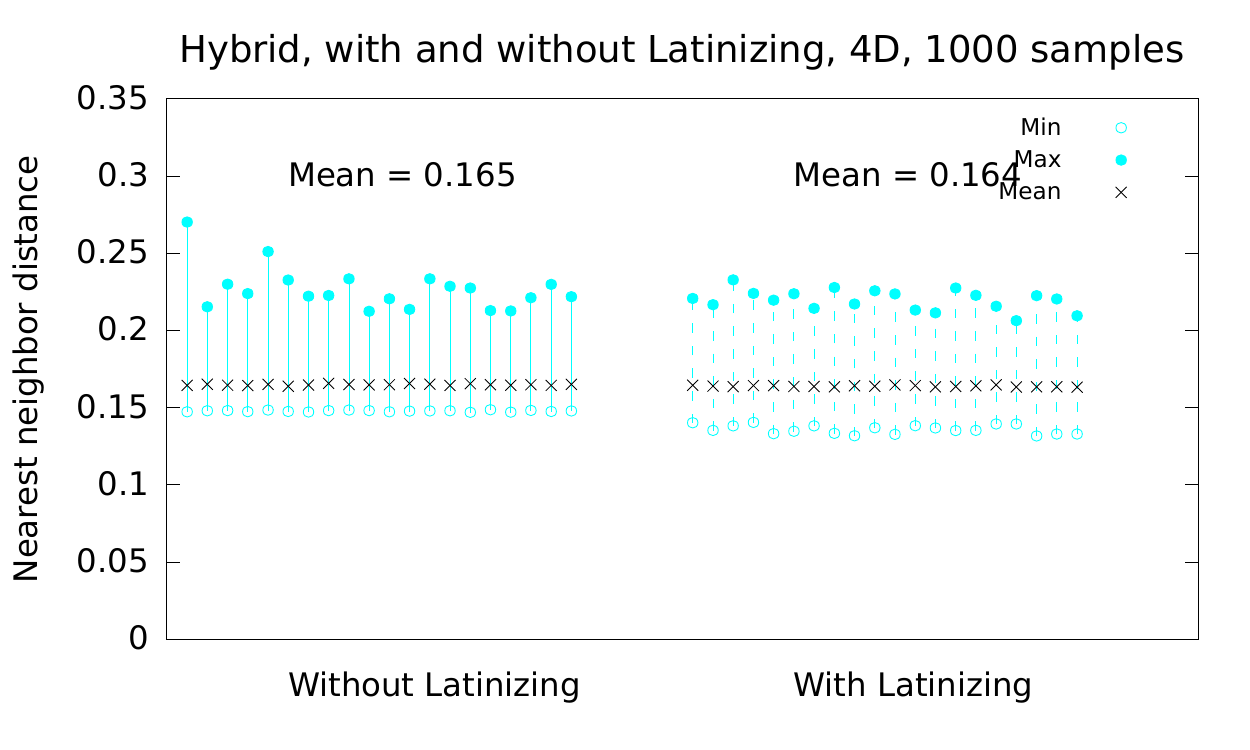} &
\includegraphics[trim = 0.0cm 0cm 0.0cm 0cm, clip = true,width=0.31\textwidth]{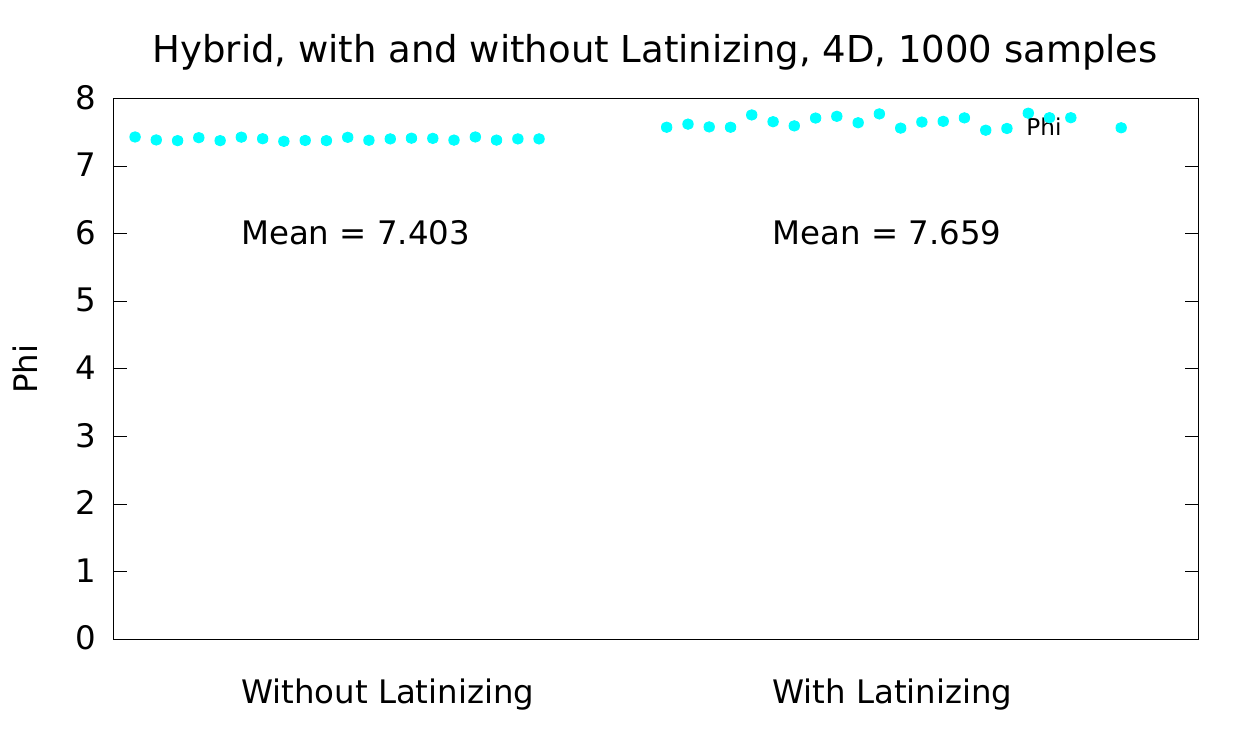} &
\includegraphics[trim = 0.0cm 0cm 0.0cm 0cm, clip = true,width=0.31\textwidth]{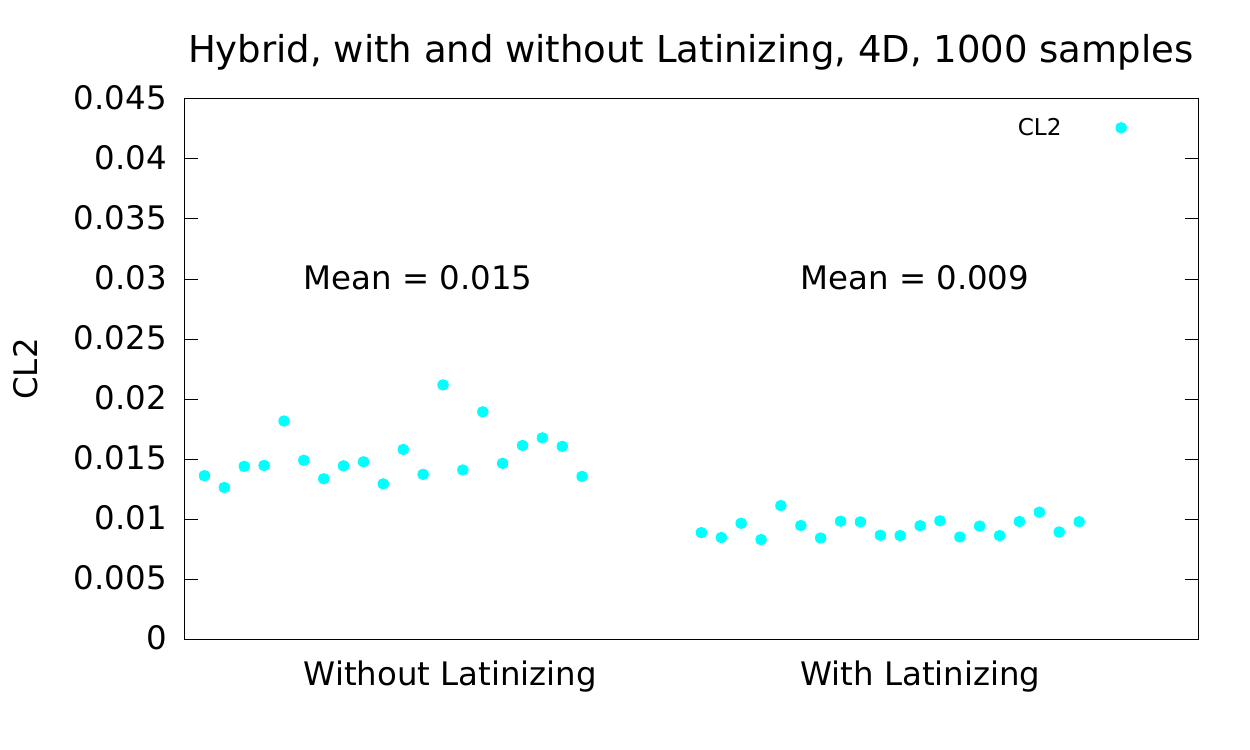} \\
(a) & (b) & (c) \\
\end{tabular}
\vspace{-0.2cm}
\caption{4D-1000 experiment: Statistics on 20 repetitions of 1000 samples in four dimensions
  generated using (top to bottom) random, LHS, GreedyFP, BC, and the
  hybrid methods. (a) distance to the nearest neighbor; (b)
  $\phi_{50}$ criterion; and (c) the CL2 discrepancy. Each plot shows
  the results without Latinizing (left) and with Latinizing (right);
  there are no Latinizing results for LHS.  The mean value of the metrics
  --- the average of the nearest neighbor distance in each sample set,
  the $\phi_{50}$ criterion and the CL2 discrepancy --- over the 20
  repetitions is also included. The range of values in the plots for
  $\phi_{50}$ are different for each algorithm. }
\label{fig:metrics_4d_1000}
\end{figure}

\begin{figure}[!htb]
\centering
\begin{tabular}{ccc}
\includegraphics[trim = 0.0cm 0cm 0.0cm 0cm, clip = true,width=0.31\textwidth]{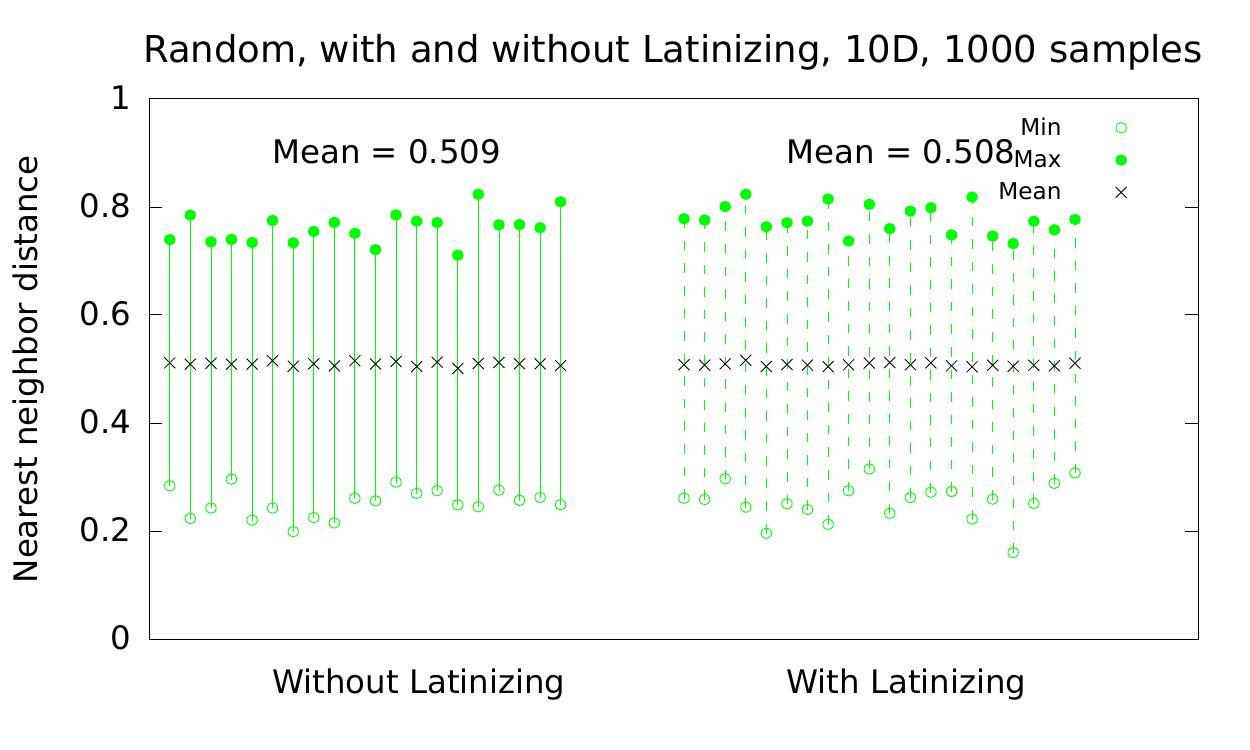} &
\includegraphics[trim = 0.0cm 0cm 0.0cm 0cm, clip = true,width=0.31\textwidth]{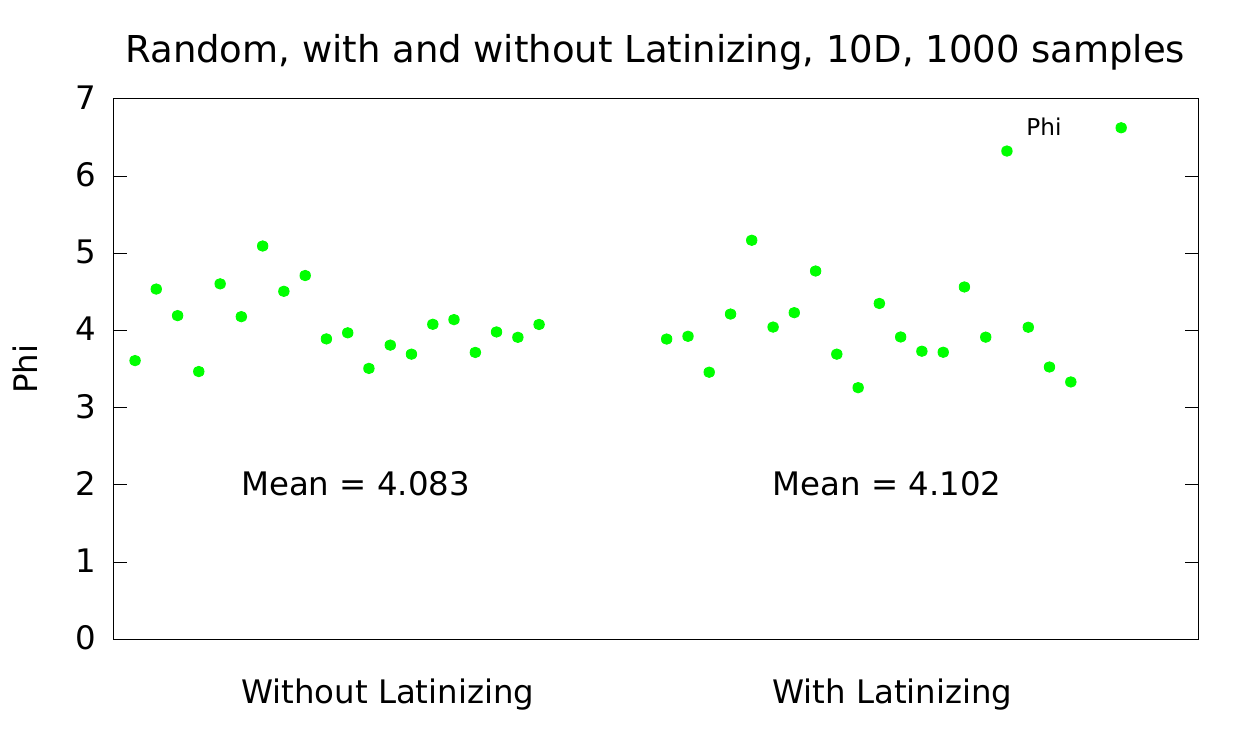} &
\includegraphics[trim = 0.0cm 0cm 0.0cm 0cm, clip = true,width=0.31\textwidth]{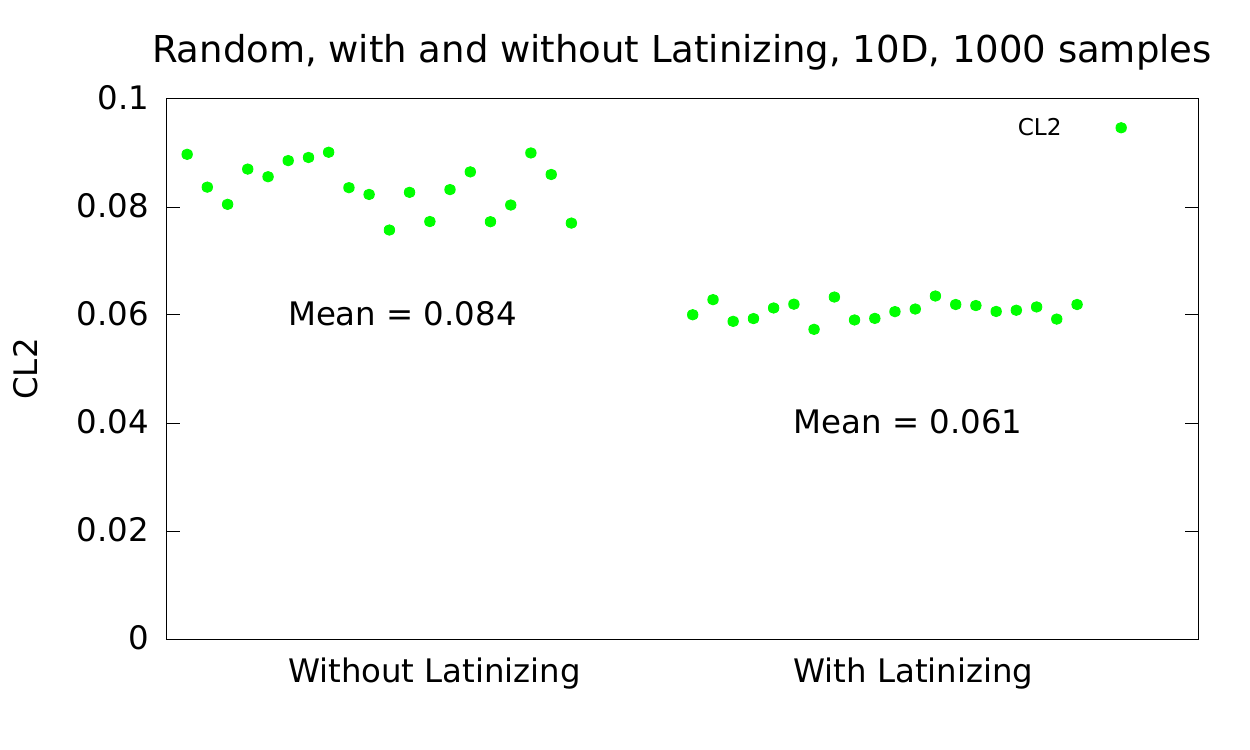} \\
\includegraphics[trim = 0.0cm 0cm 0.0cm 0cm, clip = true,width=0.31\textwidth]{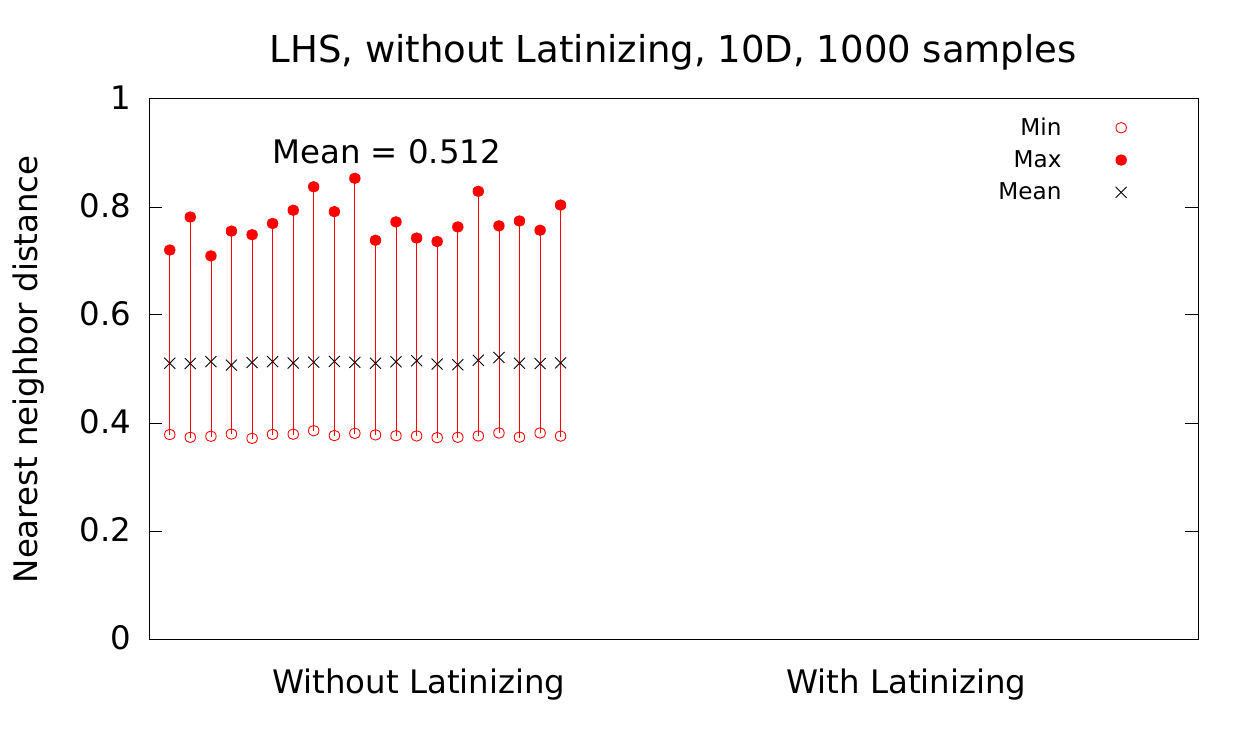} &
\includegraphics[trim = 0.0cm 0cm 0.0cm 0cm, clip = true,width=0.31\textwidth]{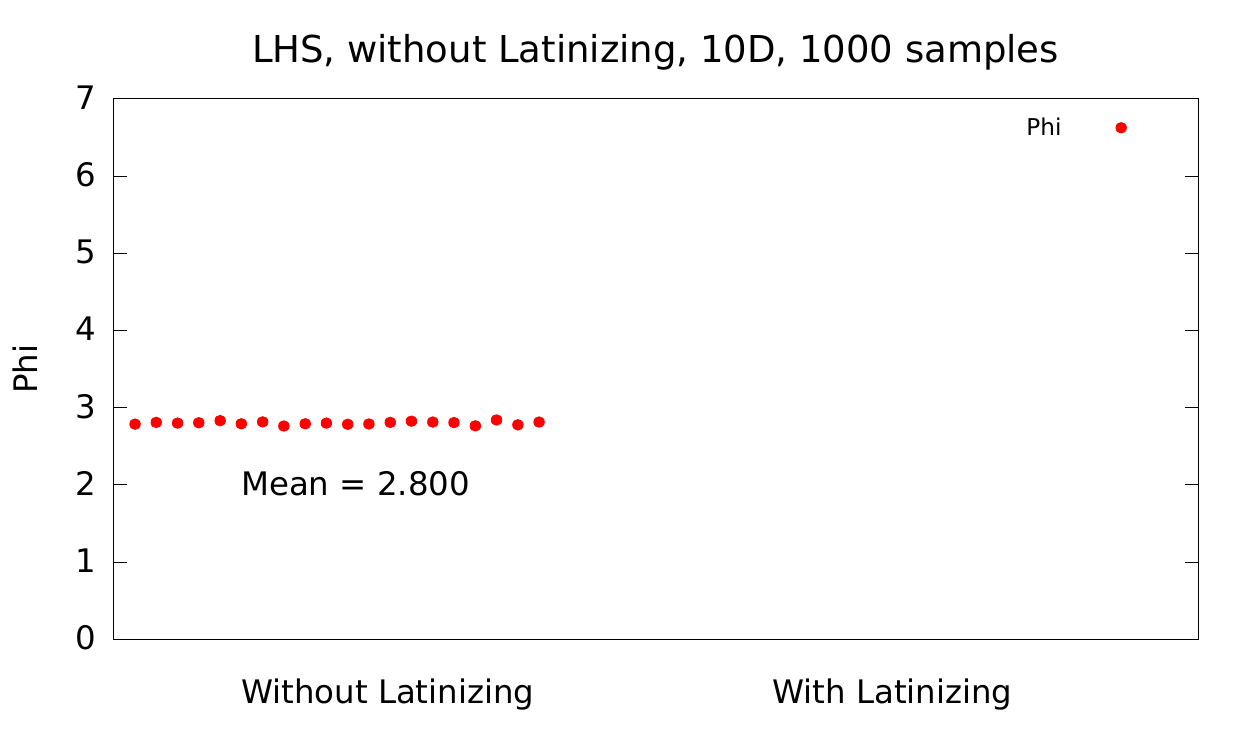} &
\includegraphics[trim = 0.0cm 0cm 0.0cm 0cm, clip = true,width=0.31\textwidth]{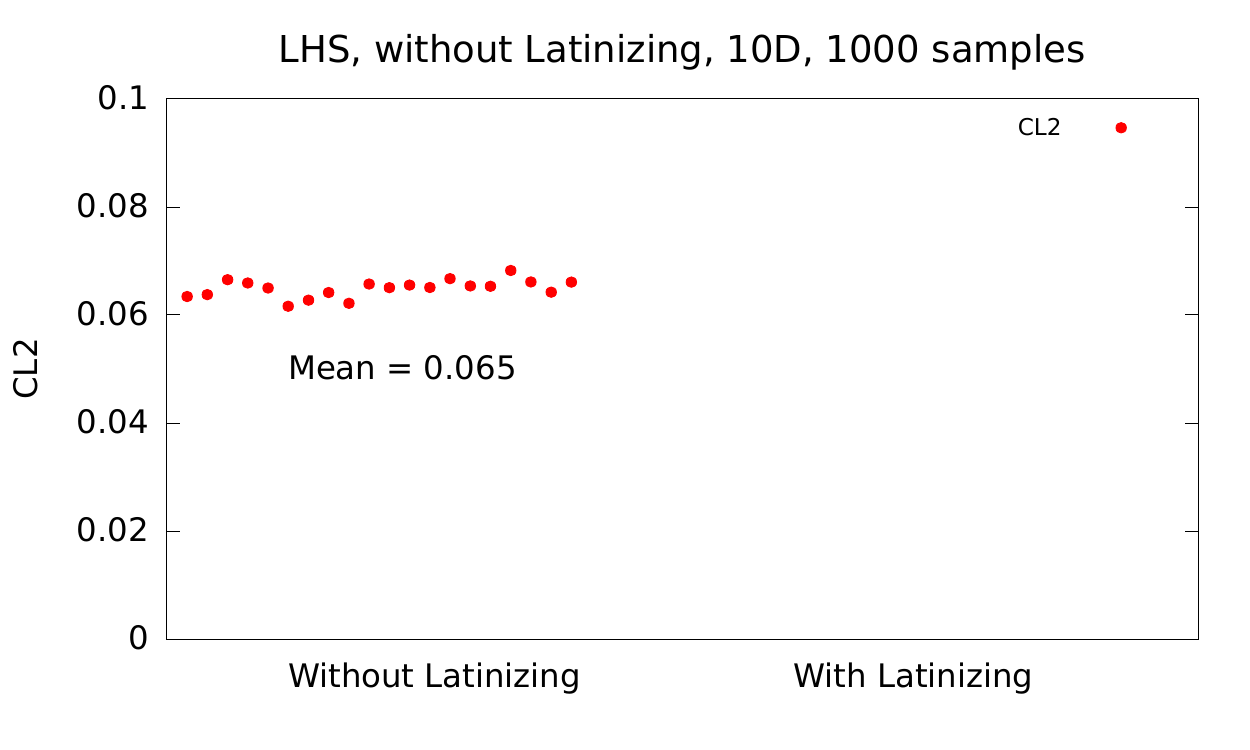} \\
\includegraphics[trim = 0.0cm 0cm 0.0cm 0cm, clip = true,width=0.31\textwidth]{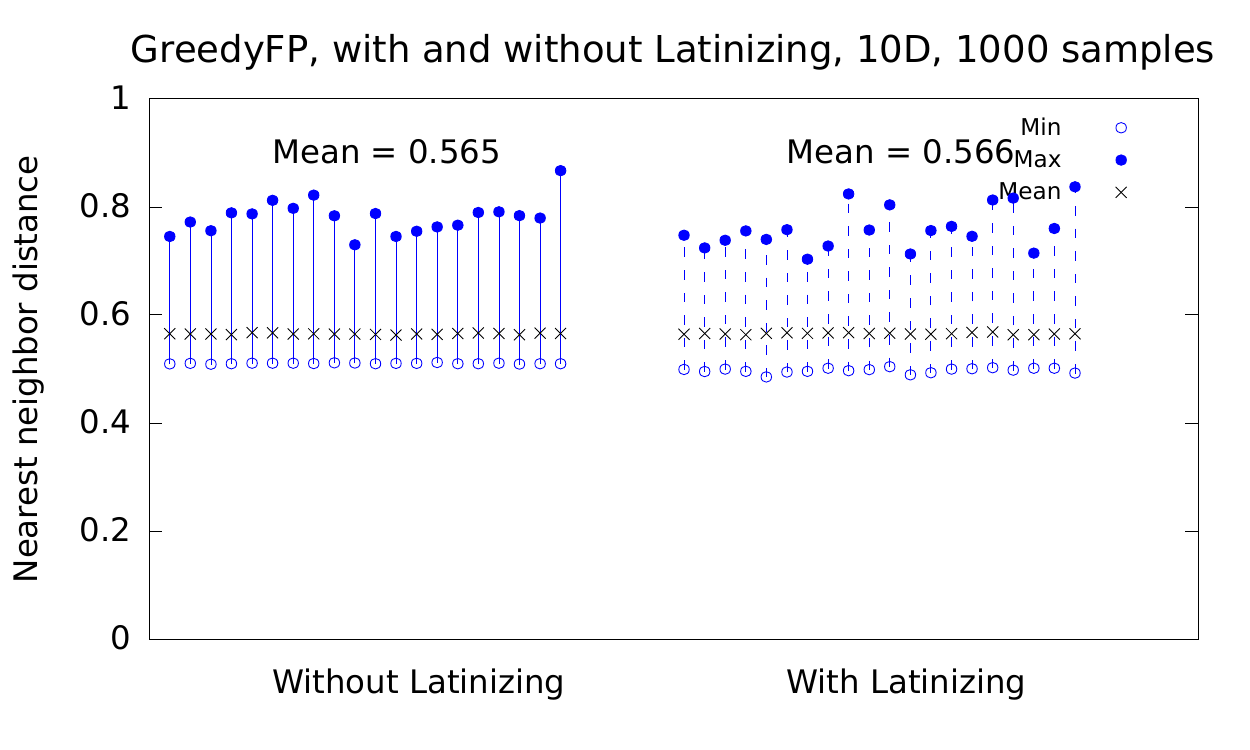} &
\includegraphics[trim = 0.0cm 0cm 0.0cm 0cm, clip = true,width=0.31\textwidth]{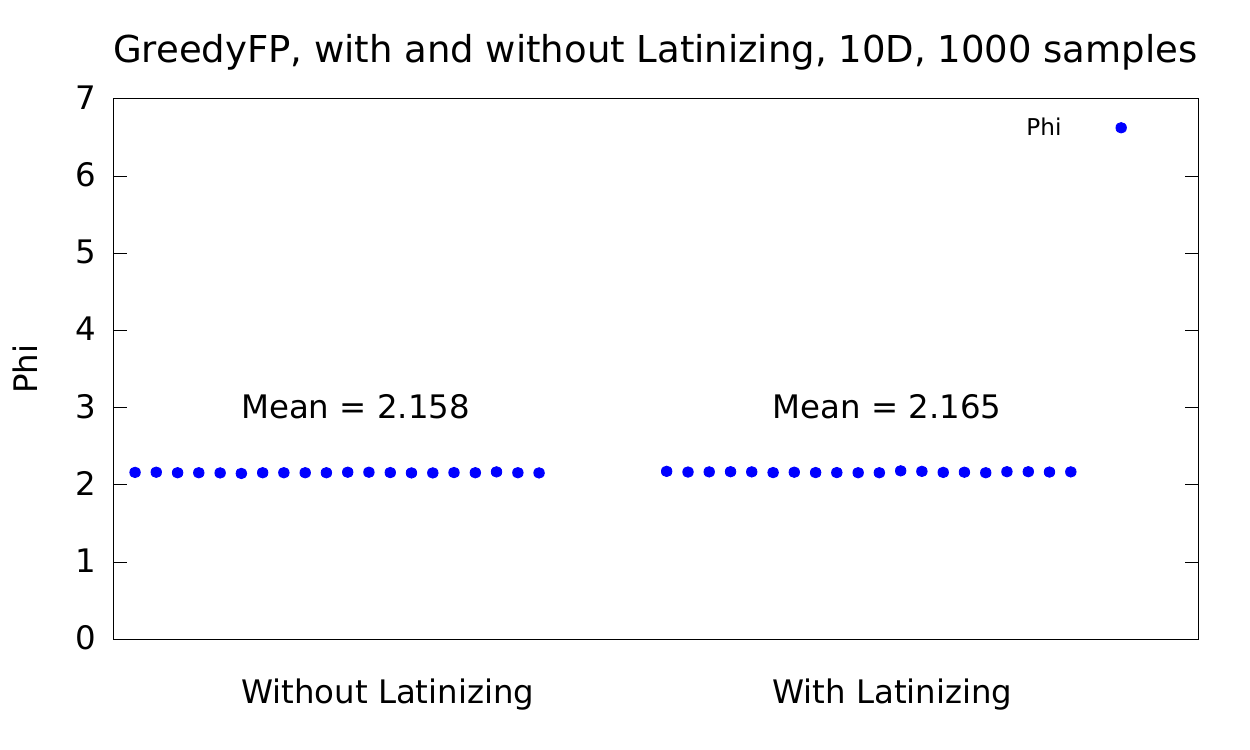} &
\includegraphics[trim = 0.0cm 0cm 0.0cm 0cm, clip = true,width=0.31\textwidth]{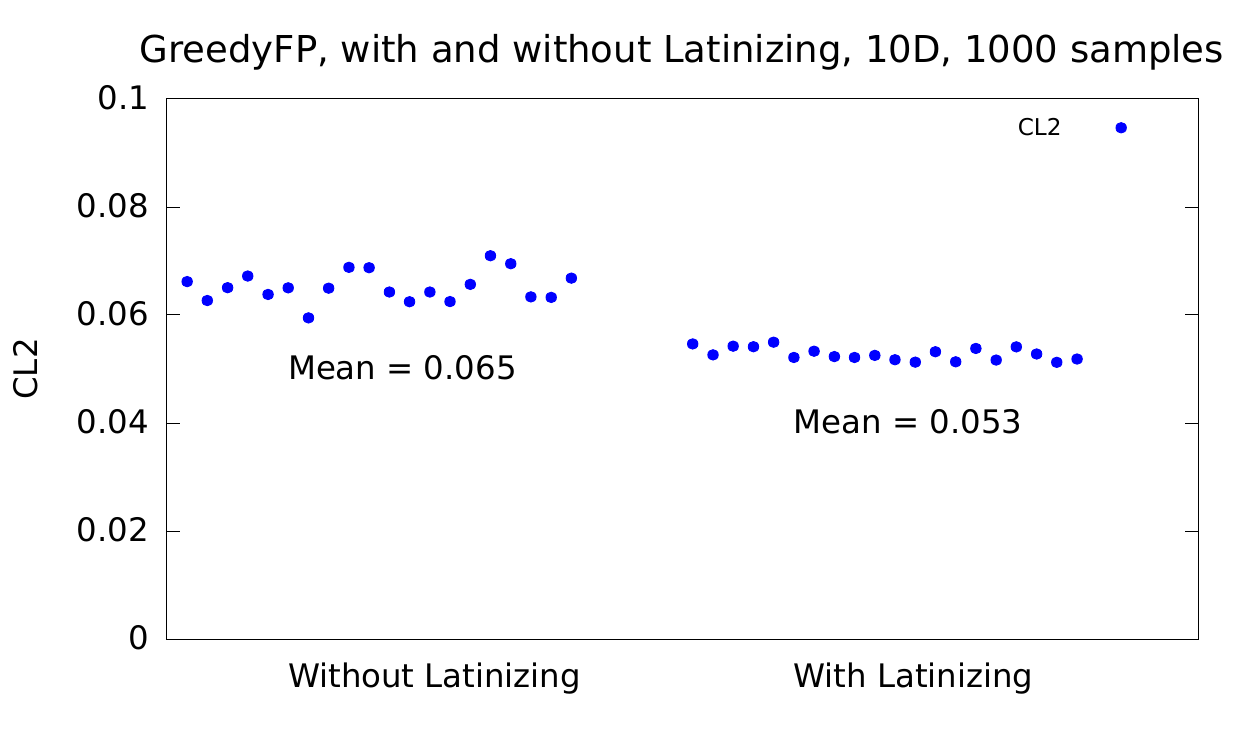} \\
\includegraphics[trim = 0.0cm 0cm 0.0cm 0cm, clip = true,width=0.31\textwidth]{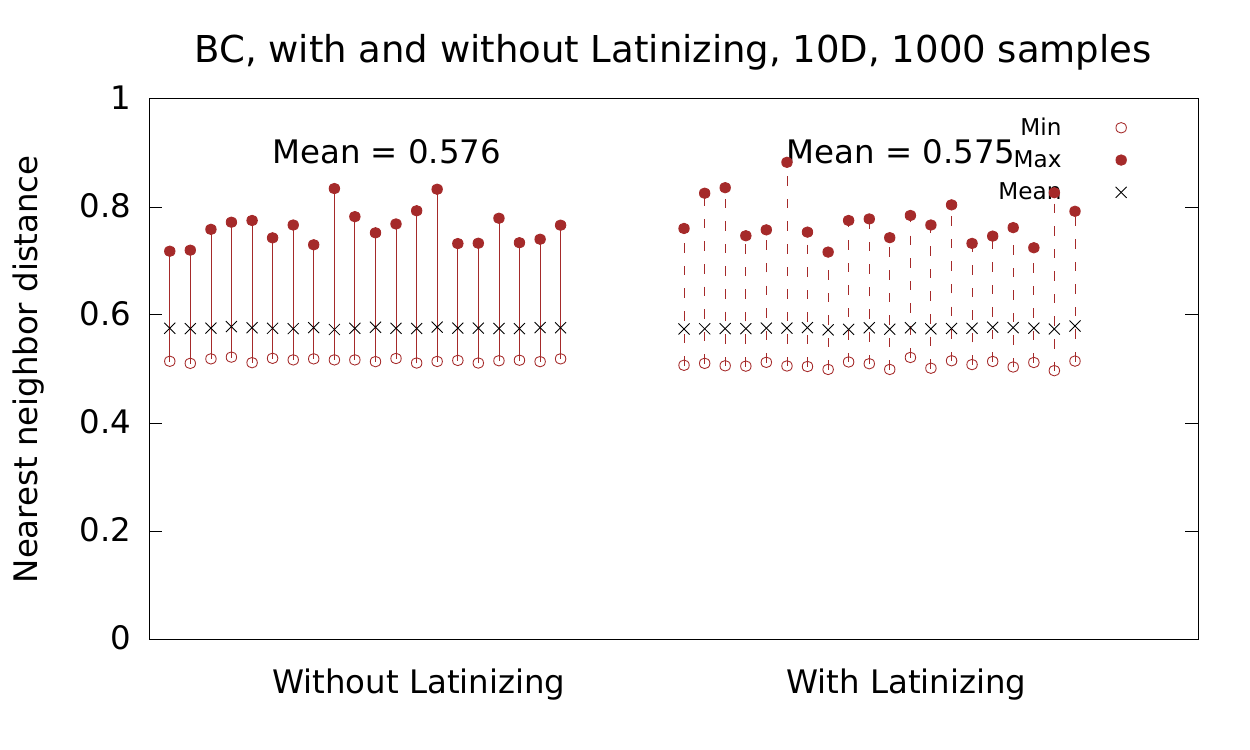} &
\includegraphics[trim = 0.0cm 0cm 0.0cm 0cm, clip = true,width=0.31\textwidth]{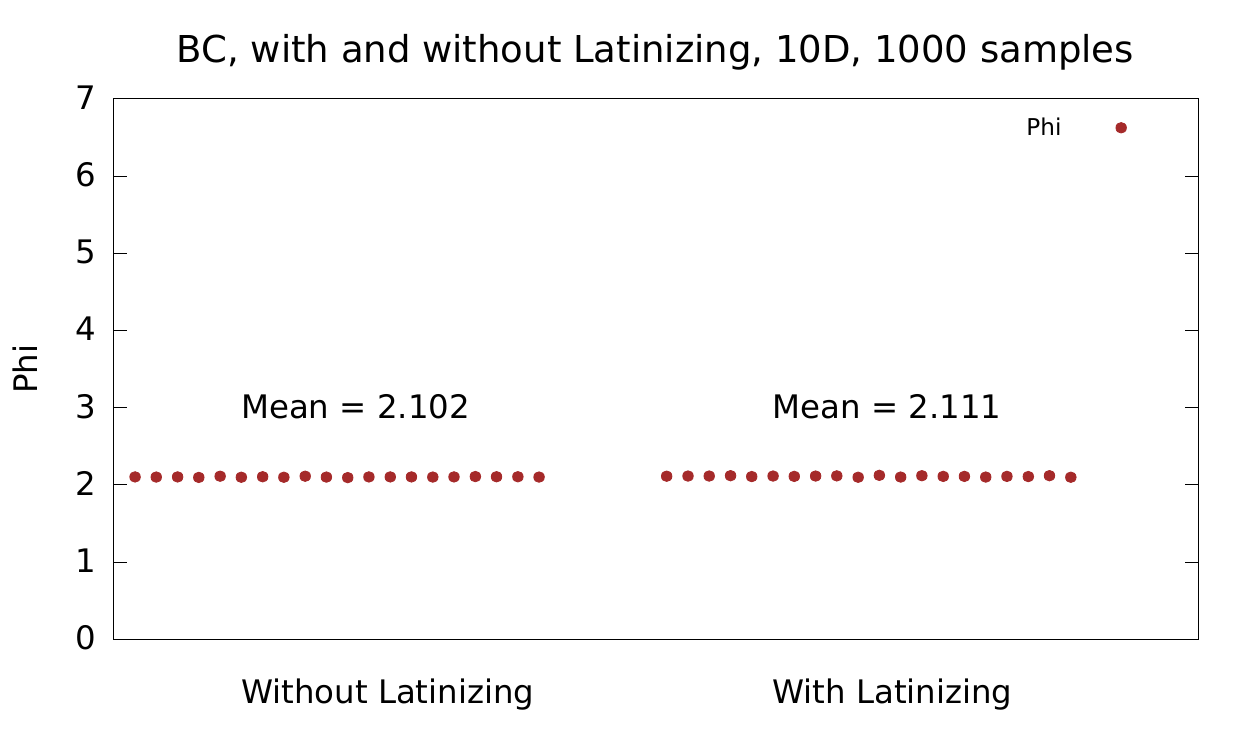} &
\includegraphics[trim = 0.0cm 0cm 0.0cm 0cm, clip = true,width=0.31\textwidth]{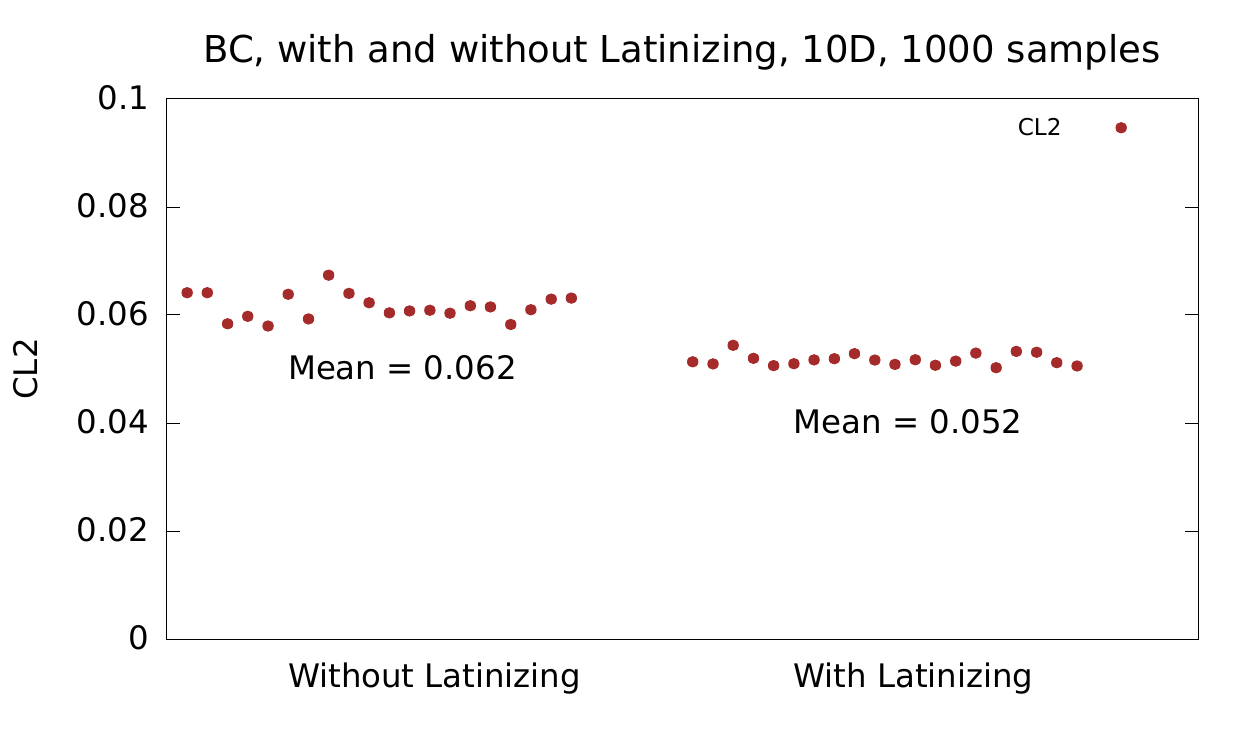} \\
\includegraphics[trim = 0.0cm 0cm 0.0cm 0cm, clip = true,width=0.31\textwidth]{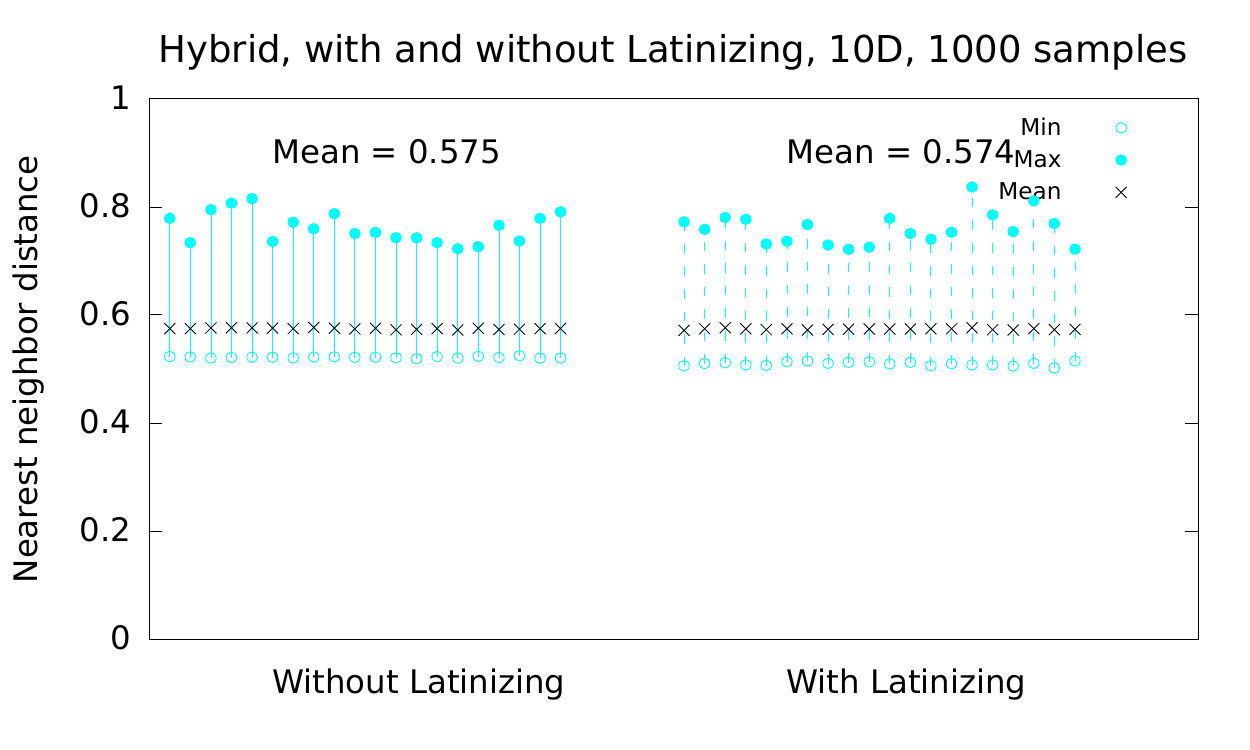} &
\includegraphics[trim = 0.0cm 0cm 0.0cm 0cm, clip = true,width=0.31\textwidth]{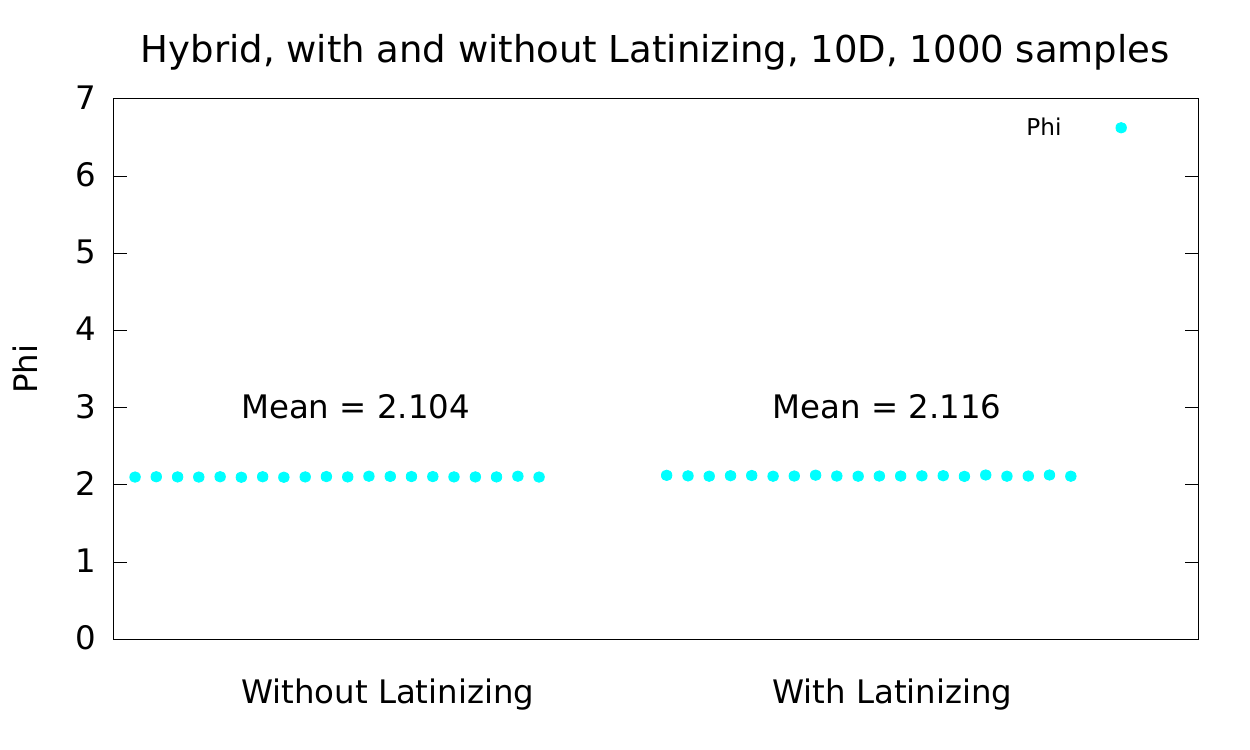} &
\includegraphics[trim = 0.0cm 0cm 0.0cm 0cm, clip = true,width=0.31\textwidth]{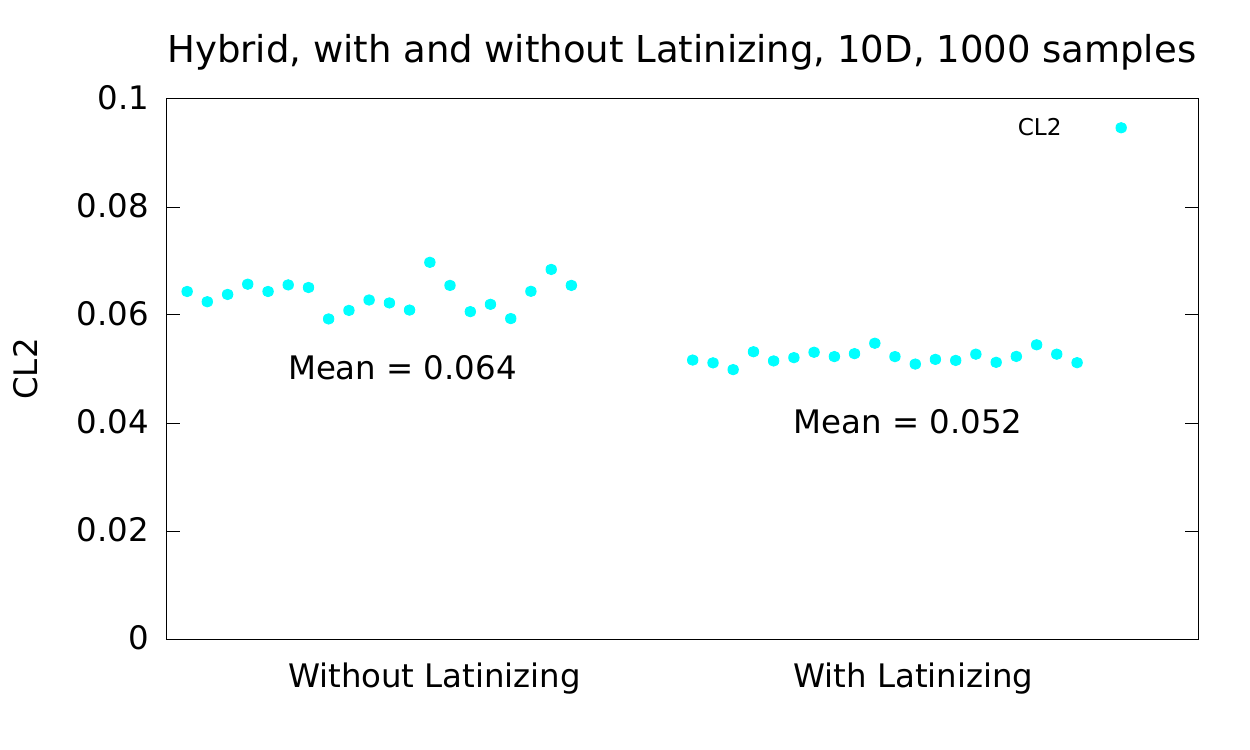} \\
(a) & (b) & (c) \\
\end{tabular}
\vspace{-0.2cm}
\caption{10D-1000 experiment: Statistics on 20 repetitions of 1000 samples in ten dimensions
  generated using (top to bottom) random, LHS, GreedyFP, BC, and the
  hybrid methods. (a) distance to the nearest neighbor; (b)
  $\phi_{50}$ criterion; and (c) the CL2 discrepancy. Each plot shows
  the results without Latinizing (left) and with Latinizing (right);
  there are no Latinizing results for LHS.  The mean value of the metrics
  --- the average of the nearest neighbor distance in each sample set,
  the $\phi_{50}$ criterion and the CL2 discrepancy --- over the 20
  repetitions is also included. The range of values in the plots for
  $\phi_{50}$ are different for each algorithm. }
\label{fig:metrics_10d_1000}
\end{figure}

We obtain the three metrics for four experiments defined by varying
the number of dimensions and the number of samples: 500 samples in
two- and four-dimensions shown in Figures~\ref{fig:metrics_2d_500} and
\ref{fig:metrics_4d_500}, respectively, that were repeated 50 times to
account for the randomness in the sample locations, as well as 1000
samples in four- and ten-dimensions shown in
Figures~\ref{fig:metrics_4d_1000} and~\ref{fig:metrics_10d_1000},
respectively, that were repeated 20 times.
For all algorithms except LHS, we include results both with and
without Latinization so we can understand the effects of Latinization.
The plots also include the mean of each metric taken over the number
of repetitions.
These mean values, along with the total time to generate the
repetitions for each algorithm, are summarized in
Table~\ref{tab:metrics_comparison}. All times reported in this paper
are for codes run on an Intel Xeon W-2155 workstation with ten cores,
two threads per core, and 62GB of memory.  The codes are run on a
single thread.

\renewcommand{\arraystretch}{1.2}
\begin{table}[tb]
  \begin{center}
    \begin{tabular}{|l|l|l|l|l|l|l|l|l|l|} \hline
      \multirow{3}{2cm}{Dimensions-samples} & \#repeats & Method & \multicolumn{6}{c|}{Mean across repetitions} & Time \\
      \cline{4-9}
      & & & \multicolumn{2}{c|}{Avg. N-N dist} & \multicolumn{2}{c|}{$\phi_{50}$} & \multicolumn{2}{c|}{CL2} & \\
      \cline{4-9}
      & & & no Lat & Lat & no Lat & Lat & no Lat & Lat & \\
      \hline
      \multirow{5}{2cm}{2D-500} & 50 & Random & 0.023 & 0.024 & 818.871 & 425.866 & 0.027 & 0.008 & 4s \\
      & 50 & LHS & 0.027 & N/A & 67.021 & N/A & 0.007 & N/A & 205s \\
      & 50 & GreedyFP & 0.039 & 0.038 & 32.581 & 43.626 & 0.009 & 0.004 & 8s \\
      & 50 & BC & 0.039 & 0.038 & 32.654 & 40.723 & 0.009 & 0.004 & 55s \\
      & 50 & Hybrid & 0.040 & 0.038 & 32.199 & 40.516 & 0.009 & 0.005 & 20s \\
      \hline
      \multirow{5}{2cm}{4D-500} & 50 & Random & 0.138 & 0.139 & 34.532 & 34.210 & 0.047 & 0.022 & 6s \\
      & 50 & LHS & 0.151 & N/A & 9.899 & N/A & 0.022 & N/A & 401s \\
      & 50 & GreedyFP &  0.192 & 0.191 & 6.292 & 6.656 & 0.026 & 0.015 & 15s \\
      & 50 & BC & 0.198 & 0.196 & 6.139 & 6.402 & 0.024 & 0.015 & 110s \\
      & 50 & Hybrid & 0.195 & 0.194 & 6.203 & 6.550 & 0.024 & 0.015 & 40s \\
      \hline
      \multirow{5}{2cm}{4D-1000} & 20 & Random & 0.115 & 0.115 & 38.801 & 43.374 & 0.033 & 0.015 & 10s \\
      & 20 & LHS & 0.122 & N/A & 13.543 & N/A & 0.018 & N/A & 641s \\
      & 20 & GreedyFP & 0.161 & 0.160 & 7.616 & 7.890 & 0.017 & 0.010 & 24s \\
      & 20 & BC & 0.165 & 0.164 & 7.419 & 7.642 & 0.016 & 0.009 & 175s \\
      & 20 & Hybrid &  0.165 & 0.164 & 7.403 & 7.659 & 0.015 & 0.009 & 97s \\
      \hline
      \multirow{5}{2cm}{10D-1000} & 20 & Random & 0.509 & 0.508 & 4.083 & 4.102 & 0.084 & 0.061 & 24s \\
      & 20 & LHS & 0.512 & N/A & 2.800 & N/A & 0.065 & N/A & 1641s \\
      & 20 & GreedyFP & 0.565 & 0.566 & 2.158 & 2.165 & 0.065 & 0.053 & 58s \\
      & 20 & BC & 0.576 &0.575 & 2.102 & 2.111 & 0.062 & 0.052 & 432s \\
      & 20 & Hybrid &  0.575 & 0.574 & 2.104 & 2.116 & 0.064 & 0.052 & 241s \\
      \hline
    \end{tabular}
  \end{center}
  \vspace{-0.2cm}
  \caption{Summary of the results in Figures~\ref{fig:metrics_2d_500} -\ref{fig:metrics_10d_1000}. The four experiments (column 1) are repeated a number of times (column 2) to account for the randomness in the samples generated. The three metrics listed are the mean values of the average nearest neighbor distance, $\phi_{50}$, and CL2 for a sample set, with the mean taken across the number of repetitions. Results with and without Latinization are listed for each method with the exception of LHS, where no Latinization is needed. The time reported in the last column is for all repetitions with Latinization and is roughly the same as the time without Latinization. Figures~\ref{fig:metrics_2d_500} - \ref{fig:metrics_10d_1000} show the variation across the repetitions.}
  \label{tab:metrics_comparison}
\end{table}

Based on the summary in Table~\ref{tab:metrics_comparison} and the
details in Figures~\ref{fig:metrics_2d_500}
-\ref{fig:metrics_10d_1000}, we make the following observations on the
four experiments that are identified by number of dimensions and
number of samples:

\begin{itemize}

\item {\bf Distance to nearest neighbor:} First, focusing on the
  nearest neighbor distances within each sample set (panel (a) in
  Figures~\ref{fig:metrics_2d_500} -\ref{fig:metrics_10d_1000}), we
  find that across all four experiments, random sampling has the
  largest range of values, which reduces somewhat for LHS, and more so
  for the three farthest point algorithms. Further, the minimum
  (maximum) nearest neighbor distance is smallest (largest) for random
  sampling, then LHS, followed by the three farthest-point algorithms.
  This is expected as random sampling creates under-and over-sampled
  regions, leading to a larger range of nearest neighbor distances,
  while for LHS, we perform only limited number of interchanges for
  optimization as calculating the optimal sampling is very expensive.
  We also find that the average nearest neighbor distance within a
  sample set is relatively unchanged across repetitions for all
  algorithms, except for random sampling, where it varies a little
  across repetitions.

  Focusing on the effect of Latinization, we find that for the 2D-500
  sample experiment, Latinizing reduces the minimum nearest neighbor
  distance, a behavior we also observe in
  Figure~\ref{fig:samples_latinize} for samples in two dimensions.
  This effect is less pronounced in the 4D-500 and 4D-1000
  experiments, and barely noticeable in the 10D-1000 experiment. We
  suspect this is due to a lower density of samples in the latter
  three experiments. We also observe that the average nearest neighbor
  distance within each sample set remains roughly the same with and
  without Latinization.

  Next, using the results in Table~\ref{tab:metrics_comparison}, we
  compare the mean, across repetitions, of the average nearest
  neighbor distance within a sample set; we want this mean value to be
  large. We observe very little difference between the columns with
  and without Latinization. Within each of the four experiments, the
  mean across repetitions is smallest for random sampling, then LHS,
  and largest for the three farthest point algorithms. Among these
  three algorithms, BC has slightly larger mean value than GreedyFP,
  while the hybrid method has a value in between the two, as expected.
  This reflects the slightly better statistical properties we expect to
  see with BC compared to Greedy-FP resulting from the regeneration of
  candidates for each sample.
 
\item {\bf $\phi_{50}$ criterion:} Using this metric, a better
  algorithm would have a smaller value of $\phi_{50}$, as it reflects
  the inverse distance between samples. The plots for $\phi_{50}$ are
  shown in Figures~\ref{fig:metrics_2d_500}
  -~\ref{fig:metrics_10d_1000}, panel (b), with different ranges of y
  values across algorithms. As would be expected, the values of
  $\phi_{50}$ are much higher for random sampling than for the other
  four methods, with LHS having smaller values, and the three farthest
  point algorithms having the smallest values. The differences between
  methods, as well as the values of $\phi_{50}$, shrink as we move to
  higher dimensions.

  Latinizing tends to increase $\phi_{50}$ for most experiments,
  except 2D-500 with random sampling. We suspect that this is may be
  due to Latinization reducing the distance between some samples and
  the metric being dominated by the smallest distances.

  The $\phi_{50}$ metric has a large range of values across
  repetitions, especially for random sampling, though the range
  reduces for LHS and is almost constant for the three farthest point
  algorithms. In the latter three algorithms, and to a lesser extent
  in LHS, we control the distances between samples, so each term in
  Equation~\ref{eqn:phi50} is roughly the same. But in random
  sampling, we do not have this explicit control, and the calculation
  of $\phi_{50}$ is dominated by the smallest distance between
  samples, which could vary substantially across repetitions.

  As with the nearest neighbor distance metric,
  Table~\ref{tab:metrics_comparison} indicates that the three farthest
  point algorithms have very similar performance, with BC having the
  smallest value of $\phi_{50}$ and the hybrid algorithm having a
  value between BC and Greedy-FP.

\item {\bf CL2 discrepancy criterion:} For the CL2 criterion (panel (c) in
  Figures~\ref{fig:metrics_2d_500} -\ref{fig:metrics_10d_1000}), a smaller
  value indicates a better algorithm.  This metric shows some
  interesting behavior, especially with regard to Latinization. First
  we observe that Latinizing always reduces the metric for all
  methods. While random sampling always has the largest value among
  the methods, LHS can sometimes have smaller values than the three FP
  algorithms without Latinization, though with Latinization, these
  algorithms give a lower CL2 value than LHS (which does not require
  Latinization). Latinization also tends to reduce the variation
  across repetitions of an algorithm, for all algorithms.

  These results suggest that the CL2 criterion could be biased towards
  sample sets that have been Latinized. This metric is different from
  the other two as it focuses on the coordinates of the samples
  instead of distances between them. It also involves products of
  terms, so a small term could have a disproportionate effect.  In
  particular, the last component, $|x_{ik} - x_{jk} |$, in the third
  term of Equation~\ref{eqn:cl2}, could be small before Latinization
  if two coordinate values along a dimension are close to each other.
  By making these coordinates more equispaced, Latinization would
  increase the value of this component, which appears with a negative
  sign. This would reduce the value of the quantity within square
  brackets, making the last term in the equation smaller than before
  Latinization. This would explain why the metric always reduces with
  Latinization.

  Table~\ref{tab:metrics_comparison} also indicates that the three
  farthest point algorithms have very similar performance.

\end{itemize}

Finally, we make some observations based on the time required for
multiple repetitions of the five algorithms for different experiments
as listed in the last column of Table~\ref{tab:metrics_comparison}.
Random sampling is the fastest and LHS the slowest. The three fastest
point algorithms are moderately fast, with GreedyFP the fastest, BC
somewhat slower, but a lot faster than LHS, and the hybrid somewhere
in between BC and GreedyFP.

\afterpage{\clearpage}

\afterpage{\clearpage}

\afterpage{\clearpage}

\afterpage{\clearpage}

\clearpage

%
\section{Conclusions}
\label{sec:conclusions}
%

In this paper, we considered the role of sampling in data analysis
tasks, including surrogate modeling and the closely-related task of
hyperparameter optimization. We showed that the increasing use of
these tasks to solve real problems in practical applications has
resulted in new requirements being placed on the sampling algorithms
used in these tasks. To address these new requirements, the typical
approach is to modify the algorithm already in use, but this often
results in a less than satisfactory solution. We therefore proposed
the concept of {\it intelligent sampling}, where we showed how we can
devise algorithms specifically tailored to meet our requirements,
either by creating new algorithms or by modifying suitable algorithms
from other fields.  Both qualitative and quantitative analyses
indicated that three farthest-point algorithms - Greedy-FP, best
candidate, and a hybrid we proposed - all meet our needs and perform
quite well, with Greedy-FP being the fastest. Latin hypercube, despite
its popularity in surrogate modeling, does not meet many of the
capabilities we need and, due to an optimization step, can be quite
expensive to compute.  Centroidal Voronoi tesselation produced well
separated samples, but does not support incremental or progressive
sampling, while Poisson disk sampling does not support a
user-specified number of samples.  Random sampling, despite its known
drawback of creating under- and over-sampled regions, is very fast,
simple to implement, and an excellent algorithm to generate candidate
samples required by all three farthest-point algorithms.

%
\section{Acknowledgment}
\label{sec:ack}
%

LLNL-TR-829837 This work performed under the auspices of the U.S.
Department of Energy by Lawrence Livermore National Laboratory under
Contract DE-AC52-07NA27344. This work is part of the IDEALS project
funded by the ASCR Program (late Dr.~Lucille Nowell, Program Manager)
at the Office of Science, US Department of Energy.

\clearpage

\stepcounter{section} 
\addcontentsline{toc}{section}{\thesection \quad References} 

\bibliographystyle{acm}
\bibliography{ms_arxiv}

\begin{thebibliography}{10}

\bibitem{arthur2007:kmeanspp}
{\sc Arthur, D., and Vassilvitskii, S.}
\newblock K-means++: The advantages of careful seeding.
\newblock In {\em Proceedings of the Eighteenth Annual ACM-SIAM Symposium on
  Discrete Algorithms\/} (USA, 2007), SODA '07, Society for Industrial and
  Applied Mathematics, pp.~1027--–1035.

\bibitem{aurenhammer1991:voronoi}
{\sc Aurenhammer, F.}
\newblock Voronoi diagrams - {A} survey of a fundamental geometric data
  structure.
\newblock {\em ACM Comput. Surv. 23\/} (1991), 345--405.

\bibitem{bergstra2012:hyperparam}
{\sc Bergstra, J., and Bengio, Y.}
\newblock Random search for hyper-parameter optimization.
\newblock {\em Journal of Machine Learning Research 13}, 10 (2012), 281--305.

\bibitem{bhosekar2018:surrogate}
{\sc Bhosekar, A., and Ierapetritou, M.}
\newblock Advances in surrogate based modeling, feasibility analysis, and
  optimization: A review.
\newblock {\em Computers \& Chemical Engineering 108\/} (2018), 250--267.

\bibitem{brandt2019:gpufp}
{\sc Brandt, S., J\"ahn, C., Fischer, M., and auf~der Heide, F.~M.}
\newblock Visibility-aware progressive farthest point sampling on the {GPU}.
\newblock {\em Computer Graphics Forum 38}, 7 (2019), 413--424.

\bibitem{bridson07:sampling}
{\sc Bridson, R.}
\newblock {Fast Poisson disk sampling in arbitrary dimensions}.
\newblock In {\em ACM SIGGRAPH 2007 Sketches\/} (New York, NY, USA, 2007),
  SIGGRAPH '07, ACM.

\bibitem{cheng07:compress}
{\sc Cheng, L., and Vishwanathan, S. V.~N.}
\newblock Learning to compress images and videos.
\newblock In {\em Proceedings of the 24th International Conference on Machine
  Learni ng\/} (New York, NY, USA, 2007), ICML '07, ACM, pp.~161--168.

\bibitem{chi2005:halton}
{\sc Chi, H., Mascagni, M., and Warnock, T.}
\newblock {On the optimal Halton sequence}.
\newblock {\em Math. Comput. Simul. 70}, 1 (Sept. 2005), 9–21.

\bibitem{cho2017:samplingcomparison}
{\sc Cho, I., Lee, Y., Ryu, D., and Choi, D.-H.}
\newblock Comparison study of sampling methods for computer experiments using
  various performance measures.
\newblock {\em Struct. Multidiscip. Optim. 55}, 1 (Jan. 2017), 221–235.

\bibitem{cook1986:stochastic}
{\sc Cook, R.~L.}
\newblock Stochastic sampling in computer graphics.
\newblock {\em ACM Trans. Graph. 5\/} (1986), 51--72.

\bibitem{crombecq2011:hybrid}
{\sc Crombecq, K., Gorissen, D., Deschrijver, D., and Dhaene, T.}
\newblock A novel hybrid sequential design strategy for global surrogate
  modeling of computer experiments.
\newblock {\em SIAM Journal on Scientific Computing 33}, 4 (2011), 1948--1974.

\bibitem{crombecq2011:spacefilling}
{\sc Crombecq, K., Laermans, E., and Dhaene, T.}
\newblock Efficient space-filling and non-collapsing sequential design
  strategies for simulation-based modeling.
\newblock {\em European Journal of Operational Research 214}, 3 (2011), 683 --
  696.

\bibitem{crombecq2011:phdthesis}
{\sc {Crombecq, Karel}}.
\newblock {\em {Surrogate modeling of computer experiments with sequential
  experimental design}}.
\newblock PhD thesis, {Ghent University}, {2011}.

\bibitem{du1999:cvt}
{\sc Du, Q., Faber, V., and Gunzburger, M.}
\newblock {Centroidal Voronoi tessellations: Applications and algorithms}.
\newblock {\em SIAM Review 41}, 4 (1999), 637--676.

\bibitem{eason2014:adaptive}
{\sc Eason, J., and Cremaschi, S.}
\newblock Adaptive sequential sampling for surrogate model generation with
  artificial neural networks.
\newblock {\em Computers \& Chemical Engineering 68\/} (2014), 220 -- 232.

\bibitem{eldar1997:farthest}
{\sc {Eldar}, Y., {Lindenbaum}, M., {Porat}, M., and {Zeevi}, Y.~Y.}
\newblock The farthest point strategy for progressive image sampling.
\newblock {\em IEEE Transactions on Image Processing 6}, 9 (1997), 1305--1315.

\bibitem{frazier2018:bopttutorial}
{\sc Frazier, P.~I.}
\newblock {A tutorial on Bayesian optimization}, 2018.
\newblock arXiv:1807.02811 [stat.ML].

\bibitem{garud2018:smartsampling}
{\sc Garud, S.~S., Karimi, I.~A., Brownbridge, G., and Kraft, M.}
\newblock Evaluating smart sampling for constructing multidimensional surrogate
  models.
\newblock {\em Comput. Chem. Eng. 108\/} (2018), 276--288.

\bibitem{garud2017:doereview}
{\sc Garud, S.~S., Karimi, I.~A., and Kraft, M.}
\newblock Design of computer experiments: A review.
\newblock {\em Computers \& Chemical Engineering 106\/} (2017), 71--95.
\newblock ESCAPE-26.

\bibitem{gonzalez1985:clustering}
{\sc Gonzalez, T.}
\newblock Clustering to minimize the maximum intercluster distance.
\newblock {\em Theor. Comput. Sci. 38\/} (1985), 293--306.

\bibitem{grundland2009:quasicrystal}
{\sc Grundland, M., Patera, J., Masáková, Z., and Dodgson, N.~A.}
\newblock Image sampling with quasicrystals.
\newblock {\em SIGMA. Symmetry, Integrability and Geometry: Methods and
  Applications [electronic only] 5\/} (2009), Paper 075, 23 p., electronic
  only--Paper 075, 23 p., electronic only.

\bibitem{hickernell1998:cl2}
{\sc Hickernell, F.~J.}
\newblock A generalized discrepancy and quadrature error bound.
\newblock {\em Mathematics of Computation 67}, 221 (1998), 299--322.

\bibitem{husslage2011:lhs}
{\sc Husslage, B. G.~M., Rennen, G., van Dam, E.~R., and den Hertog, D.}
\newblock Space-filling latin hypercube designs for computer experiments.
\newblock {\em Optimization and Engineering 12\/} (2011), 611--630.

\bibitem{husslage2006:phdthesis}
{\sc {Husslage B. G. M.}}
\newblock {\em {Maximin designs for computer experiments}}.
\newblock PhD thesis, {Tilburg University}, {2006}.

\bibitem{iman2006:lhs}
{\sc Iman, R.~L.}
\newblock Latin hypercube sampling.
\newblock In {\em Encyclopedia of Statistical Sciences}. American Cancer
  Society, 2006.

\bibitem{jin2005:optimallhs}
{\sc Jin, R., Chen, W., and Sudjianto, A.}
\newblock An efficient algorithm for constructing optimal design of computer
  experiments.
\newblock {\em Journal of Statistical Planning and Inference 134}, 1 (Sept.
  2005), 268--287.

\bibitem{johnson1990:maximin}
{\sc Johnson, M., Moore, L., and Ylvisaker, D.}
\newblock Minimax and maximin distance designs.
\newblock {\em Journal of Statistical Planning and Inference 26}, 2 (1990),
  131--148.

\bibitem{joseph2016:spacefilling}
{\sc Joseph, V.~R.}
\newblock Space-filling designs for computer experiments: A review.
\newblock {\em Quality Engineering 28}, 1 (2016), 28--35.

\bibitem{joseph2015:maxpro}
{\sc Joseph, V.~R., Gul, E., and Ba, S.}
\newblock {Maximum projection designs for computer experiments}.
\newblock {\em Biometrika 102}, 2 (03 2015), 371--380.

\bibitem{ju2002:parallelcvt}
{\sc Ju, L., Du, Q., and Gunzburger, M.}
\newblock {Probabilistic methods for centroidal Voronoi tessellations and their
  parallel implementations}.
\newblock {\em Parallel Comput. 28}, 10 (Oct. 2002), 1477–1500.

\bibitem{kamath17:compression}
{\sc Kamath, C.}
\newblock Learning to compress unstructured mesh data from simulations.
\newblock In {\em 2017 {IEEE} International Conference on Data Science and
  Advanced Analytics, {DSAA} 2017, Tokyo, Japan, October 19-21, 2017\/} (2017),
  pp.~621--630.

\bibitem{kamath19:bigdata}
{\sc {Kamath}, C.}
\newblock Intelligent exploration of large-scale data: What can we learn in two
  passes?
\newblock In {\em 2019 IEEE International Conference on Big Data\/} (Dec 2019),
  pp.~1831--1840.

\bibitem{kamath18:compression}
{\sc Kamath, C., and Fan, Y.}
\newblock Compressing unstructured mesh data using spline fits, compressed
  sensing, and regression methods.
\newblock In {\em 2018 {IEEE} Global Conference on Signal and Information
  Processing, GlobalSIP 2018, Anaheim, CA, USA, November 26-29, 2018\/} (2018),
  pp.~316--320.

\bibitem{kamath2018:small}
{\sc Kamath, C., and Fan, Y.-J.}
\newblock Regression with small data sets: a case study using code surrogates
  in additive manufacturing.
\newblock {\em Knowledge and Information Systems 57\/} (2018), 475--493.

\bibitem{kamath2021:inverse}
{\sc Kamath, C., Franzman, J., and Ponmalai, R.}
\newblock Data mining for faster, interpretable solutions to inverse problems:
  A case study using additive manufacturing.
\newblock {\em Machine Learning with Applications 6\/} (2021), 100122.

\bibitem{kopf2006:recursivewang}
{\sc Kopf, J., Cohen-Or, D., Deussen, O., and Lischinski, D.}
\newblock {Recursive Wang tiles for real-time blue noise}.
\newblock {\em ACM Trans. Graph. 25}, 3 (July 2006), 509–518.

\bibitem{lagae2008:comparisonpd}
{\sc Lagae, A., and Dutr\'e, P.}
\newblock { A comparison of methods for generating Poisson disk distributions}.
\newblock {\em Computer Graphics Forum 27}, 1 (March 2008), 114--129.

\bibitem{larson81:book}
{\sc Larson, R.~C., and Odoni, A.~R.}
\newblock {\em Urban Operations Research}.
\newblock Prentice Hall, Hoboken, NJ, 1981.
\newblock Available online at
  \url{https://web.mit.edu/urban_or_book/www/book/}.

\bibitem{loyola2016:smartsampling}
{\sc {Loyola R}, D.~G., Pedergnana, M., and {Gimeno García}, S.}
\newblock Smart sampling and incremental function learning for very large high
  dimensional data.
\newblock {\em Neural Networks 78\/} (2016), 75 -- 87.
\newblock Special Issue on Neural Network Learning in Big Data.

\bibitem{mccool1992:poisson}
{\sc McCool, M., and Fiume, E.}
\newblock Hierarchical {Poisson} disk sampling distributions.
\newblock In {\em Proceedings of the Conference on Graphics Interface '92\/}
  (San Francisco, CA, USA, 1992), Morgan Kaufmann Publishers Inc., p.~94–105.

\bibitem{mckay1979:lhs}
{\sc McKay, M.~D., Beckman, R.~J., and Conover, W.~J.}
\newblock A comparison of three methods for selecting values of input variables
  in the analysis of output from a computer code.
\newblock {\em Technometrics 21}, 2 (1979), 239--245.

\bibitem{mitchell1987:antialiased}
{\sc Mitchell, D.~P.}
\newblock Generating antialiased images at low sampling densities.
\newblock In {\em SIGGRAPH '87\/} (1987).

\bibitem{mitchell91:sampling}
{\sc Mitchell, D.~P.}
\newblock Spectrally optimal sampling for distribution ray tracing.
\newblock {\em Computer Graphics 25}, 4 (1991), 157--164.

\bibitem{mitchell2018:spokedart}
{\sc Mitchell, S.~A., Ebeida, M.~S., Awad, M.~A., Park, C., Patney, A., Rushdi,
  A.~A., Swiler, L.~P., Manocha, D., and Wei, L.-Y.}
\newblock Spoke-darts for high-dimensional blue-noise sampling.
\newblock {\em ACM Trans. Graph. 37}, 2 (May 2018).

\bibitem{mitry2014:surrogatethesis}
{\sc Mitry, M.}
\newblock {Reduced Order Surrogate Modelling (ROSM) of high dimensional
  deterministic simulations}.
\newblock Master's thesis, Department of Aerospace Engineering, University of
  Toronto, 2014.
\newblock Available from \url{http://hdl.handle.net/1807/68047}.

\bibitem{montgomery12:book}
{\sc Montgomery, D.~C.}
\newblock {\em Design and Analysis of Experiments}.
\newblock John Wiley, Hoboken, NJ, 2012.

\bibitem{morissette2013:kmeans}
{\sc Morissette, L., and Chartier, S.}
\newblock The k-means clustering technique: General considerations and
  implementation in {Mathematica}.
\newblock {\em Tutorials in Quantitative Methods for Psychology 9}, 1 (2013),
  15--24.

\bibitem{morokoff1994:quasirandom}
{\sc Morokoff, W.~J., and Caflisch, R.~E.}
\newblock Quasi-random sequences and their discrepancies.
\newblock {\em SIAM Journal on Scientific Computing 15}, 6 (1994), 1251--1279.

\bibitem{morris95:design}
{\sc Morris, M.~D., and Mitchell, T.~J.}
\newblock Exploratory designs for computational experiments.
\newblock {\em Journal of Statistical Planning and Inference 43}, 3 (1995), 381
  -- 402.

\bibitem{nuchit2013:samplesize}
{\sc Nuchitprasittichai, A., and Cremaschi, S.}
\newblock An algorithm to determine sample sizes for optimization with
  artificial neural networks.
\newblock {\em AIChE Journal 59}, 3 (2013), 805--812.

\bibitem{patel2018:fastpoissondisk}
{\sc Patel, B.}
\newblock Faster poisson disk sampling, December 2018.
\newblock Available from
  \url{https://medium.com/basudevpatel/faster-poisson-sampling-a76cb9a99825}.
  Accessed 9 September 2021.

\bibitem{roberts2021:fastpoissondisk}
{\sc Roberts, M.}
\newblock {Maximal Poisson disk sampling: An improved version of Bridson's
  algorithm}.
\newblock Available from
  \url{http://extremelearning.com.au/an-improved-version-of-bridsons-algorithm-n-for-poisson-disc-sampling/}.
  Accessed 9 September 2021.

\bibitem{romero2006:latinized}
{\sc Romero, V.~J., Burkardt, J.~V., Gunzburger, M.~D., and Peterson, J.~S.}
\newblock {Comparison of pure and Latinized centroidal Voronoi tessellation
  against various other statistical sampling methods}.
\newblock {\em Reliability Engineering and System Safety 91}, 10 (2006), 1266
  -- 1280.
\newblock Special issue: The Fourth International Conference on Sensitivity
  Analysis of Model Output (SAMO 2004).

\bibitem{sacks1989:dace}
{\sc Sacks, J., Welch, W.~J., Mitchell, T.~J., and Wynn, H.~P.}
\newblock {Design and Analysis of Computer Experiments}.
\newblock {\em Statistical Science 4}, 4 (1989), 409 -- 423.

\bibitem{saka2007:latinized}
{\sc Saka, Y., Gunzburger, M., and Burkardt, J.}
\newblock Latinized, improved {LHS}, and {CVT} point sets in hypercubes.
\newblock {\em International Journal of Numerical Analysis and Modeling 4}, 3-4
  (2007), 729--743.

\bibitem{santner2018:book}
{\sc Santner, T.~J., Williams, B., and Notz, W.}
\newblock {\em The Design and Analysis of Computer Experiments, Second
  Edition}.
\newblock Springer-Verlag, 2018.

\bibitem{schubert2021:kmediods}
{\sc Schubert, E., and Rousseeuw, P.~J.}
\newblock Fast and eager k-medoids clustering: O(k) runtime improvement of the
  pam, clara, and clarans algorithms.
\newblock {\em Information Systems 101\/} (2021), 101804.

\bibitem{secord2002:stippling}
{\sc Secord, A.}
\newblock Weighted {Voronoi} stippling.
\newblock In {\em Proceedings of the 2nd International Symposium on
  Non-Photorealistic Animation and Rendering\/} (New York, NY, USA, 2002), NPAR
  '02, Association for Computing Machinery, p.~37–43.

\bibitem{settles2009:active}
{\sc Settles, B.}
\newblock Active learning literature survey.
\newblock Tech. rep., University of Wisconsin-Madison Computer Sciences 1648,
  January 2009.
\newblock Available at \url{https://minds.wisconsin.edu/handle/1793/60660}.

\bibitem{shewchuk2012:delaunaynotes}
{\sc Shewchuk, J.~R.}
\newblock {Lecture Notes on Delaunay Mesh Generation}.
\newblock Tech. rep., University of California, Berkeley, February 2012.
\newblock Available at
  \url{https://people.eecs.berkeley.edu/~jrs/meshpapers/delnotes.pdf}.

\bibitem{viana2016:lhstutorial}
{\sc Viana, F. A.~C.}
\newblock {A tutorial on Latin hypercube design of experiments}.
\newblock {\em Quality and Reliability Engineering International 32}, 5 (2016),
  1975--1985.

\bibitem{wei2008:parallelpd}
{\sc Wei, L.-Y.}
\newblock Parallel {Poisson} disk sampling.
\newblock In {\em ACM SIGGRAPH 2008 Papers\/} (New York, NY, USA, 2008),
  SIGGRAPH '08, Association for Computing Machinery.

\bibitem{yan2011:inclhs}
{\sc Yan, S., and Minsker, B.}
\newblock Applying dynamic surrogate models in noisy genetic algorithms to
  optimize groundwater remediation designs.
\newblock {\em Journal of Water Resources Planning and Management 137}, 3
  (2011), 284--292.

\bibitem{ying2013:parallelpd}
{\sc Ying, X., Xin, S., Sun, Q., and He, Y.}
\newblock An intrinsic algorithm for parallel poisson disk sampling on
  arbitrary surfaces.
\newblock {\em IEEE Transactions on Visualization and Computer Graphics 19\/}
  (2013), 1425--1437.

\bibitem{zack2010:esa}
{\sc Zack, J., Natenberg, E., Young, S., Manobianco, J., and Kamath, C.}
\newblock Application of ensemble sensitivity analysis to observation targeting
  for short-term wind speed forecasting.
\newblock Tech. rep., Lawrence Livermore National Laboratory LLNL-TR-424442,
  February 2010.
\newblock Available at \url{https://www.osti.gov/biblio/972845}.

\end{thebibliography}

\end{document}